\PassOptionsToPackage{bookmarks=true}{hyperref}
\RequirePackage{float}
\documentclass{agibot-report}

\definecolor{agiblue}{HTML}{9273B0}
\definecolor{agiurl}{HTML}{76558F}
\definecolor{agicite}{HTML}{76558F}
\definecolor{RVLinkBlue}{HTML}{3478F6}

\usepackage{bookmark}

\hypersetup{
  linktoc=all,
  bookmarksopen=true,
  bookmarksnumbered=true
}

\setcitestyle{authoryear,round,citesep={;},aysep={,},yysep={;}}
\makeatletter
\def\NAT@sort{0}
\makeatother

\usepackage{amsmath,amsfonts,bm}

\def\eqref#1{equation~\ref{#1}}

\def\1{\bm{1}}

\DeclareMathAlphabet{\mathsfit}{\encodingdefault}{\sfdefault}{m}{sl}
\SetMathAlphabet{\mathsfit}{bold}{\encodingdefault}{\sfdefault}{bx}{n}

\usepackage{graphicx}
\usepackage{wrapfig}
\usepackage{capt-of}
\usepackage{pdfpages}
\usepackage[table]{xcolor}
\usepackage{booktabs}
\usepackage{placeins}
\usepackage{needspace}
\usepackage{multirow}
\usepackage{array}
\usepackage{makecell}
\usepackage{adjustbox}
\usepackage{tabularx}

\usepackage{RoboValue_Real_Overleaf4/RoboValue_Real_Compact_iclr}
\usepackage{RoboValue_Sim_Overleaf/RoboValue_Sim_Compact_iclr}

\usepackage{url}

\newcolumntype{Q}[1]{>{\raggedright\arraybackslash}p{#1}}
\newcolumntype{Y}{>{\centering\arraybackslash}X}

\definecolor{RVZeroBand}{RGB}{232,240,249}
\definecolor{RVOneBand}{RGB}{246,239,249}
\definecolor{RVBestHighlight}{HTML}{EFE8F6}

\definecolor{RVUnderstandingInk}{HTML}{366C9A}
\definecolor{RVProgressInk}{HTML}{447D64}
\definecolor{RVRecoveryInk}{HTML}{A56F2D}
\definecolor{RVConsistencyInk}{HTML}{A45368}

\DeclareRobustCommand{\RVbest}[1]{%
  \begingroup
  \setlength{\fboxsep}{0.12em}%
  \leavevmode\kern-0.27em%
  \colorbox{RVBestHighlight}{%
    \kern0.15em\textbf{#1}\kern0.15em%
  }%
  \kern-0.27em%
  \endgroup
}

\DeclareRobustCommand{\RVsecond}[1]{%
  \underline{\textbf{#1}}%
}

\definecolor{RVSensSimInk}{HTML}{695184}
\definecolor{RVSensRealInk}{HTML}{3D7185}
\colorlet{RVSensRefInk}{agiurl}
\definecolor{RVSensRule}{HTML}{8B8591}


\newcommand{\modeband}[2]{%
  \midrule[0.6pt]
  \noalign{\vskip -0.75pt}%
  \rowcolor{#1}%
  \multicolumn{12}{c}{\textit{#2}}\\[-1.5pt]
  \noalign{\vskip -0.75pt}%
  \midrule[0.6pt]
}

\title{RoboValue: A Fine-Grained Sim-and-Real Benchmark for Unified Evaluation of Robotic Value Models}

\author{%
Shengbang Liu\textsuperscript{1,2,*} \quad
Zhengye Du\textsuperscript{1,2,*} \quad
Zhilong Wan\textsuperscript{1,2,*} \quad
Honghao Su\textsuperscript{3,*}\\
Chenxiang Xia\textsuperscript{1} \quad
Chang Ge\textsuperscript{4} \quad
Jinyang Xiao\textsuperscript{4} \quad
Nan Wang\textsuperscript{2,\textdagger} \quad
Zhipeng Hu\textsuperscript{1}\\
Huining Yuan\textsuperscript{1} \quad
Shu'ang Yu\textsuperscript{1} \quad
Yue Chen\textsuperscript{3,5} \quad
Tianxing Chen\textsuperscript{3,6} \quad
Chuankang Li\textsuperscript{2}\\
Maoqing Yao\textsuperscript{2} \quad
Wenbo Ding\textsuperscript{1,3} \quad
Yu Wang\textsuperscript{1} \quad
Guanghui Ren\textsuperscript{2,\textdaggerdbl} \quad
Chao Yu\textsuperscript{1,\textdaggerdbl}
}

\affiliation{%
\textsuperscript{1}Tsinghua University \quad
\textsuperscript{2}AgiBot \quad
\textsuperscript{3}Xspark AI\\[2pt]
\textsuperscript{4}Sun Yat-sen University \quad
\textsuperscript{5}Peking University \quad
\textsuperscript{6}The University of Hong Kong\\[2pt]
{\footnotesize
\textsuperscript{*}Equal contribution. \quad
\textsuperscript{\textdagger}Project leader. \quad
\textsuperscript{\textdaggerdbl}Corresponding authors.
}
}

\authornote{}
\date{}

\headerlogo{%
  \includegraphics[height=17pt]{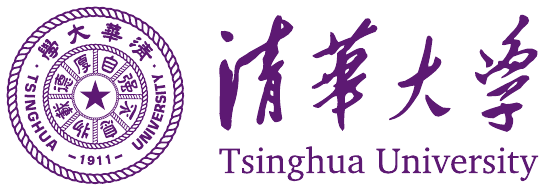}%
  \hspace{9pt}%
  \raisebox{2pt}{%
    \includegraphics[height=14pt]{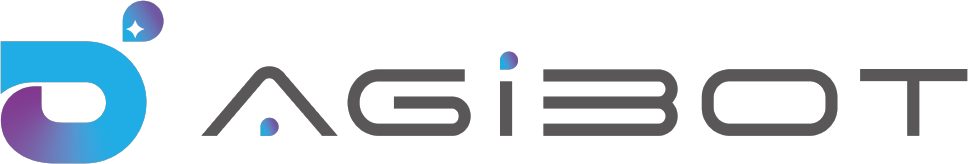}%
  }%
  \hspace{8pt}%
  \raisebox{2pt}{%
    \includegraphics[height=11pt]{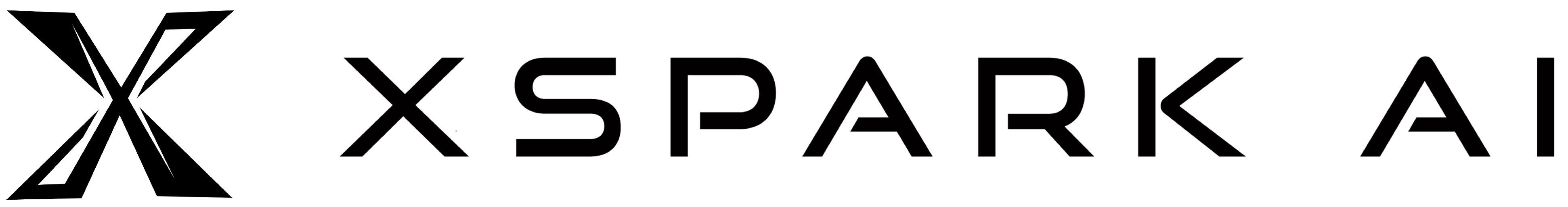}%
  }%
  \hspace{8pt}%
  \includegraphics[height=16pt]{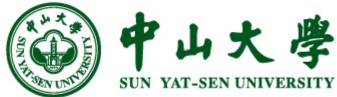}%
  \hspace{9pt}%
  \includegraphics[height=16pt]{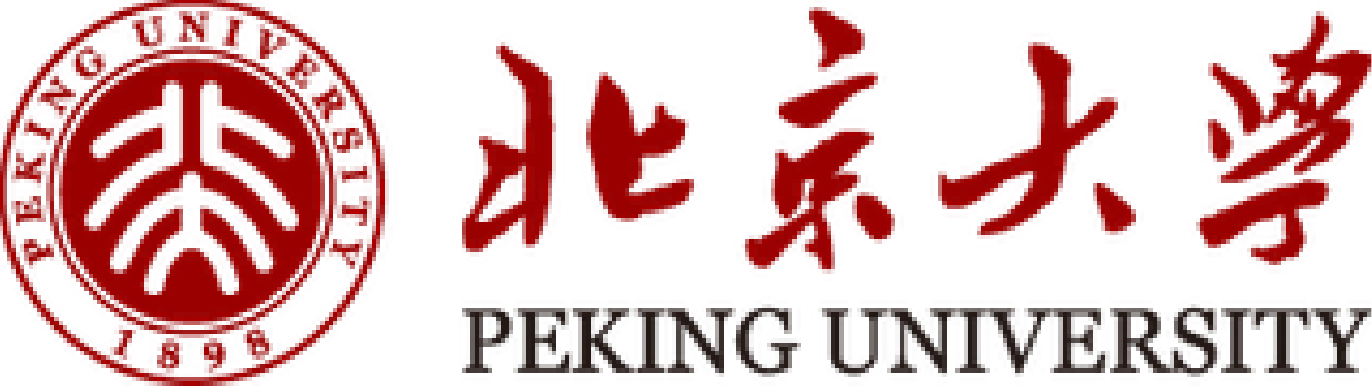}%
  \hspace{9pt}%
  \includegraphics[height=16pt]{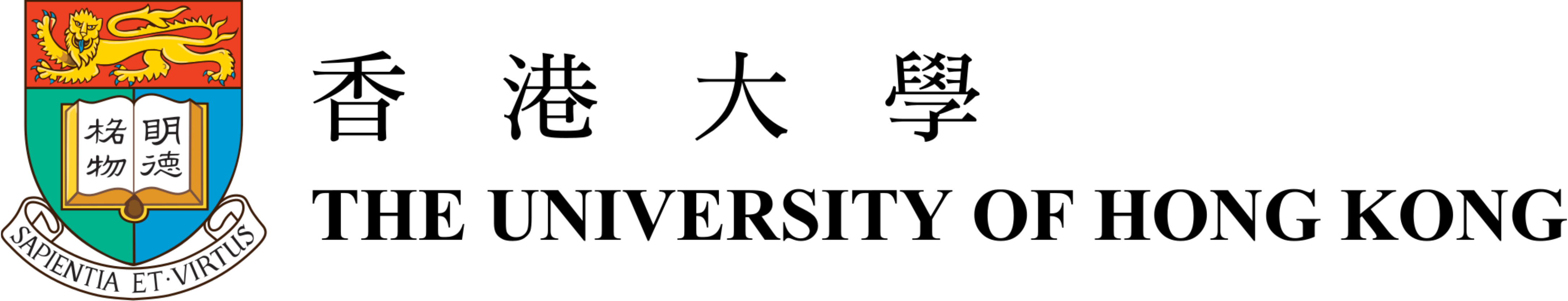}%
}

\usepackage{etoolbox}

\renewenvironment{abstract}{%
  \vskip 0.075in
  \centerline{\large\bfseries\color{agiblue}Abstract}%
  \vspace{0.5ex}%
  \begin{quote}%
}{%
  \par
  \end{quote}%
  \vskip 1ex
}

\usepackage{tocloft}

\cftsetindents{section}{0em}{2em}
\cftsetindents{subsection}{2em}{2.8em}
\cftsetindents{subsubsection}{4.8em}{3.2em}

\usepackage{fontawesome5}
\usetikzlibrary{shadows}

\newlength{\RVcoverlinkwidth}
\newcommand{\RVcoverlink}[3]{%
  \begingroup
  \def\RVcoverlinkcontent{%
    \tikz[baseline=(resource.base)]{%
      \node[
        fill=agiblue!7, draw=none, rounded corners=4pt,
        minimum height=21pt, inner sep=0pt, outer sep=0pt,
        text=agiurl, font=\normalsize,
        drop shadow={opacity=0.08, shadow xshift=0pt, shadow yshift=-0.8pt}
      ] (resource) {\makebox[\RVcoverlinkwidth][c]{#1\hspace{4pt}#2}};
    }%
  }%
  \ifstrempty{#3}
    {\RVcoverlinkcontent}%
    {\href{#3}{\RVcoverlinkcontent}}%
  \endgroup
}

\begin{document}

\enlargethispage{16pt}
\begingroup
  \makeatletter
  \patchcmd{\agb@toptitlebar}{0.12in}{0.04in}{}{}
  \patchcmd{\agb@bottomtitlebar}{0.12in}{0.04in}{}{}
  \patchcmd{\agb@titleitem}{24\p@}{10\p@}{}{}
  \patchcmd{\agb@authorblock}{0.8em}{0.5em}{}{}
  \patchcmd{\@maketitle}{\vskip 0.3in \@minus 0.1in}{\vskip 4pt}{}{}
  \makeatother
  \fontsize{11}{13.2}\selectfont
  \maketitle
\endgroup
\vspace{4pt}

\begingroup
  \normalsize
  \setlength{\parskip}{0pt}
  \setlength{\RVcoverlinkwidth}{\dimexpr(\linewidth-28pt)/5\relax}
  \noindent\makebox[\linewidth][s]{%
    \RVcoverlink{\faGlobe}{Project Page}{https://robovalue-benchmark.github.io/}%
    \hfill
    \RVcoverlink{\faGithub}{Code}{https://github.com/RoboValue-Benchmark/RoboValue}%
    \hfill
    \RVcoverlink{\faDatabase}{Dataset}{https://huggingface.co/datasets/wzlllong/value-model-benchmark-training-data}%
    \hfill
    \RVcoverlink{\faBookOpen}{Document}{https://robovalue-benchmark.github.io/doc/}%
    \hfill
    \RVcoverlink{\faTrophy}{Leaderboard}{https://robovalue-benchmark.github.io/leaderboard/}%
  }\par
\endgroup
\vspace{6pt}

\noindent
\begin{minipage}{\textwidth}
  \centering

  \includegraphics[width=\linewidth]{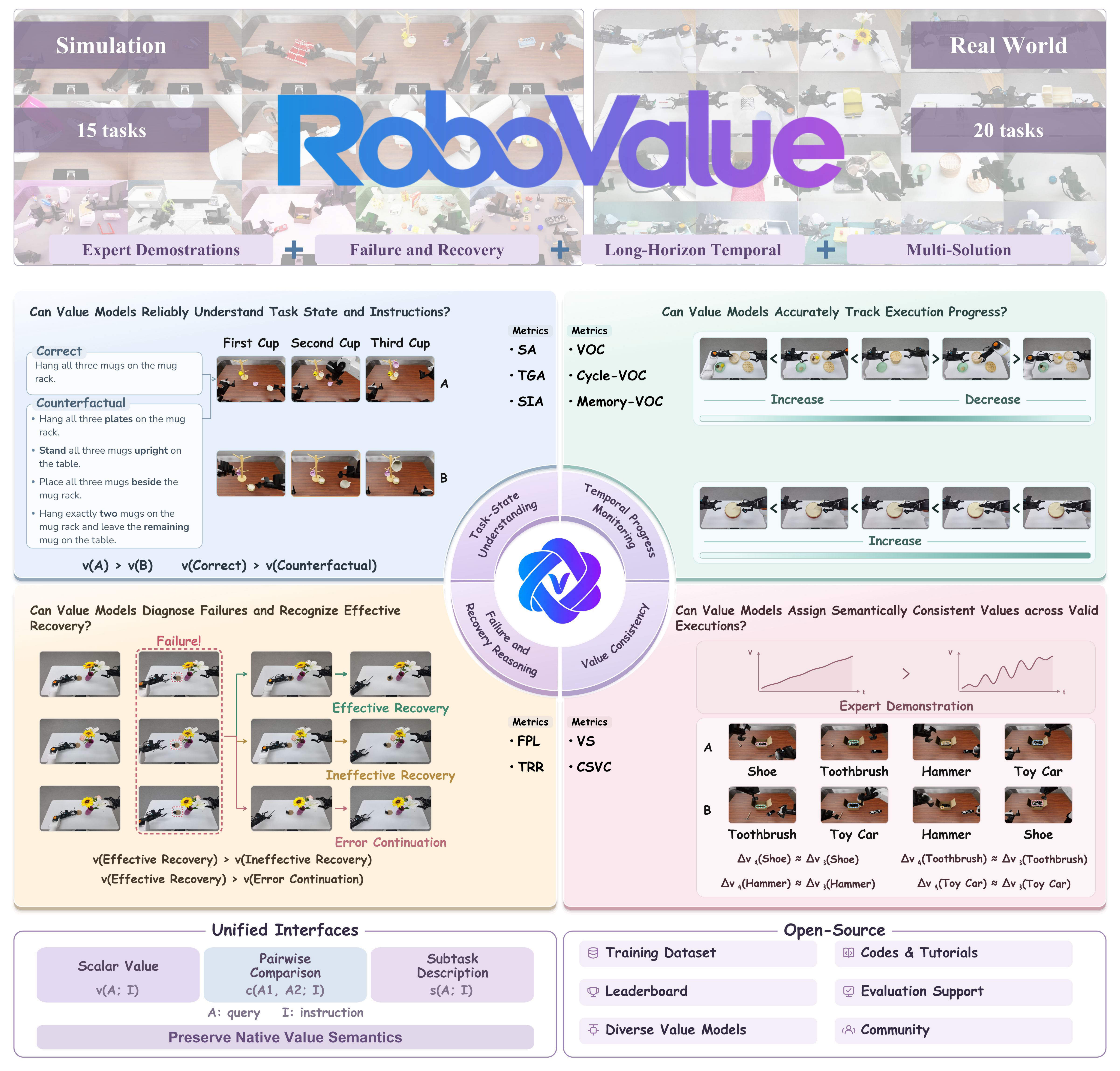}

  \setlength{\abovecaptionskip}{2pt}
  \captionof{figure}{%
    \textbf{Overview of RoboValue.}
    RoboValue evaluates the reliability of robotic value models across
    15 simulation and 20 real-world tasks. Expert demonstrations, diagnostic
    trajectories, and instruction variations probe four complementary
    dimensions: task-state understanding, temporal progress monitoring,
    failure and recovery reasoning, and value consistency. Joint evaluation
    reveals limitations that outcome and forward-progress scores can miss.
    The dataset and leaderboard support the development of reliable value models
    for policy improvement.
  }
  \label{fig:overview}
\end{minipage}

\vfill
\clearpage

\begingroup
  \normalsize
  \color{black}

  \hypersetup{
    linkcolor=black,
    linktoc=all
  }

  \setlength{\parskip}{0pt}
  \setcounter{tocdepth}{3}
  \tableofcontents
\endgroup

\clearpage

\begingroup
\raggedbottom
\pretocmd{\section}{\Needspace{9\baselineskip}}{}{%
  \PackageError{RoboValue}{Could not set section page spacing}{}%
}
\pretocmd{\subsection}{\Needspace{5\baselineskip}}{}{%
  \PackageError{RoboValue}{Could not set subsection page spacing}{}%
}
\pretocmd{\subsubsection}{\Needspace{4\baselineskip}}{}{%
  \PackageError{RoboValue}{Could not set subsubsection page spacing}{}%
}
\setlength{\textfloatsep}{14pt plus 2pt minus 2pt}
\setlength{\floatsep}{12pt plus 2pt minus 2pt}
\setlength{\intextsep}{12pt plus 2pt minus 2pt}
\phantomsection
\addcontentsline{toc}{section}{Abstract}
\begin{abstract}
Robotic value models assess task execution from visual observations and instructions, providing feedback for policy learning and execution monitoring. 
However, existing evaluations cover only selected aspects of execution understanding, leaving the reliability of value feedback insufficiently tested. 
We introduce \textbf{RoboValue}, a fine-grained sim-and-real benchmark for unified evaluation of robotic value models. 
Its framework evaluates four complementary capability dimensions through shared interfaces: task-state understanding, temporal progress monitoring, failure and recovery reasoning, and value consistency. 
Diagnostic trajectories and instruction variations test whether value judgments accurately reflect task requirements and changes throughout execution. 
RoboValue-Dataset covers 15 simulation and 20 real-world manipulation tasks, providing expert demonstrations for training and annotated trajectories for fine-grained evaluation. 
We evaluate representative robotic value models in zero-shot and one-shot settings under standard conditions and separate shifts in robot embodiment and environment. 
Our results reveal errors in fine-grained instruction grounding, history-dependent progress tracking, and recovery assessment, even in models with strong outcome and forward-progress scores. 
The same semantic subtask can also receive substantially different value gains across valid execution orders. 
These findings identify concrete targets for value supervision, while the dataset and public leaderboard support continued assessment of feedback reliability. 
The website is available at \url{https://robovalue-benchmark.github.io/}.
\end{abstract}

\section{Introduction}
\label{sec:introduction}
General-purpose robotic value models assess task execution from visual observations and task instructions~\citep{alakuijala2024video}, providing learned feedback that reduces reliance on task-specific reward design~\citep{mu2024drs,kim2025subtask}.
Their predictions support training data selection and reweighting~\citep{budzianowski2025opengvl,liang2026robometer,ma2025vision,yu2026warp} and supply reward or advantage signals for policy improvement~\citep{tan2025robo,huang2026rynnvalue,tsao2026learning,intelligence2025pi,fei2026srpo,liu2026taco,yong2026generalizable}.
At inference time, value estimates guide planning over predicted future states~\citep{ma2023liv,kim2026cosmospolicy}, while action-conditioned evaluators rank and refine candidate actions~\citep{dai2025rover}.
They also enable execution monitoring and failure detection~\citep{liang2026robometer,tan2025interactive,ye2025robofac,park2026hide,zhai2025vision} and process-level policy diagnosis~\citep{ji2026prm,liu2026prm15}.
These assessments influence which behaviors are selected, reinforced, or corrected, so their errors can propagate into both policy learning and execution.
Systematic evaluation is therefore essential to determine when value feedback accurately reflects task execution and can reliably guide downstream policies.

\begin{wrapfigure}{R}{0.46\textwidth}
  \vspace{-10pt}
  \centering
  \includegraphics[width=\linewidth]{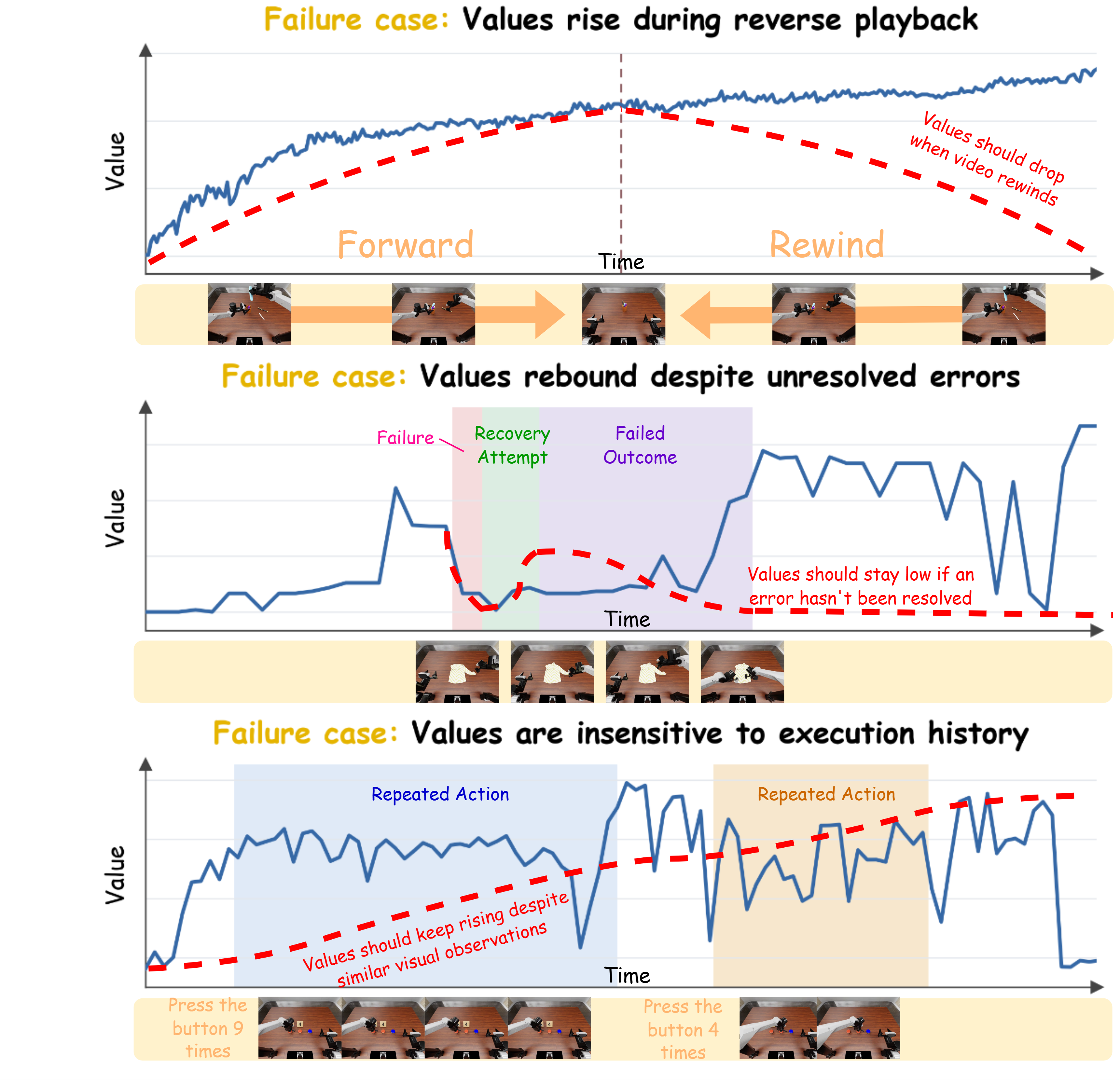}
  \caption{\textbf{Failure Cases of Value Models.} Blue curves show model predictions. Red dashed curves illustrate expected qualitative trends during reverse playback, unsuccessful recovery, and repeated actions.
  }
  \label{fig:teaser}
\end{wrapfigure}

Existing evaluations assess task completion and trajectory preferences~\citep{lee2026roboreward,liang2026robometer}, as well as process understanding through diagnostic questions and progress comparisons~\citep{xia2026roboprocessbench,ji2026prm,liu2026prm15}.
These tests capture important aspects of execution, but strong outcome accuracy or forward-progress correlation can still conceal errors in intermediate value judgments.
Differences in model interfaces and value semantics further hinder consistent diagnosis across models~\citep{zhang2026progress}.
Such blind spots matter for policy learning: optimizing imperfect feedback can increase predicted reward while degrading actual task performance~\citep{pan2022effects}.
Recent evidence in robotics likewise shows that embodied reward models can over-reward poor execution and shortcut behaviors, motivating broader coverage of negative behavior in training and evaluation~\citep{tian2026badbehavior}.
Figure~\ref{fig:teaser} illustrates values rising during reverse playback, rebounding after unsuccessful recovery, or failing to reflect accumulated progress when similar states recur.
If used as learning signals, such errors could reinforce unproductive continuation, ineffective correction, or omission of required subtasks.
These cases motivate joint assessment of whether value changes correspond to changes in task attainment across diverse execution conditions.

We introduce \textbf{RoboValue}, a fine-grained sim-and-real benchmark for unified evaluation of robotic value models.
RoboValue is organized around a central principle: value feedback should accurately reflect how well task requirements are being fulfilled throughout execution.
This requires grounding assessments in the specified goal, tracking advancement and regression using relevant history, and distinguishing unresolved errors from successful recovery.
Reliability also depends on stable, informative feedback within trajectories and consistent subtask credit across valid solutions.
As illustrated in Figure~\ref{fig:overview}, these requirements motivate four complementary capability dimensions: \emph{task-state understanding}, \emph{temporal progress monitoring}, \emph{failure and recovery reasoning}, and \emph{value consistency}.

\Needspace{24\baselineskip}
We make these requirements testable through diagnostic trajectories and instruction variations that probe specific ambiguities in execution assessment.
Counterfactual instructions hold the observed execution fixed while changing the requested goal, and recurring visual states require history to disambiguate accumulated progress.
Contrasting error continuations with effective and ineffective recoveries tests whether models distinguish corrective activity from its outcome.
Alternative valid subtask orders test whether the same semantic subtask receives consistent value gains across solutions.
RoboValue-Dataset covers 15 simulation and 20 real-world manipulation tasks (Section~\ref{sec:construction}).
Simulation provides diverse object assets and varied scene configurations under controlled conditions, while real-world data captures visual sensing and contact dynamics in complex physical environments.
Shared interfaces and model-specific adapters enable comparison across heterogeneous models while preserving their native value semantics (Section~\ref{sec:robovaluebenchmark}).

We evaluate representative robotic value models in zero-shot and one-shot settings under standard conditions and separate shifts in robot embodiment and environment (Sections~\ref{sec:leaderboard} and~\ref{sec:experiment}).
Our analyses reveal uneven capability profiles across model scales and input designs, as well as broader degradation under the tested environment shifts than embodiment shifts.
These findings motivate supervision that captures task semantics, relevant execution history, and recovery outcomes (Section~\ref{sec:analysis}).
Our main contributions are:
\par
\WFclear
\begin{itemize}
    \item We construct \textbf{RoboValue-Dataset}, with expert training demonstrations and annotated test trajectories in simulation and the real world. Contrasts in task instructions, execution history, recovery outcomes, and valid subtask orders expose unreliable value judgments, while separate embodiment and environment shifts test generalization.
    \item We introduce \textbf{RoboValue-Benchmark}, a unified framework for evaluating robotic value models across four complementary capability dimensions. Joint assessment under a shared protocol reveals limitations that outcome discrimination and forward-progress correlation alone can overlook.
    \item We establish \textbf{RoboValue-Leaderboard} with overall rankings and capability profiles for each evaluation track. Our analyses distinguish instruction insensitivity from incorrect semantic grounding and reveal weaknesses in history awareness, recovery reasoning, and subtask credit, informing value supervision for policy learning, action planning, and execution monitoring.
\end{itemize}

\section{Related Work}
\label{sec:related_work}

\subsection{Value Models for Robotic Manipulation}
Robotic value models differ in both the meaning of their predictions and the evidence used to produce them.
Some derive scores from ordinal completion ratings or the likelihood of a completion response~\citep{lee2026roboreward,chen2026topreward}, while others estimate absolute task progress~\citep{ma2025vision,zhang2025rewind,yan2026progressvla}.
Stage-aware models further structure progress around procedures and subtasks~\citep{feng2026procvlm,chen2026sarm,chen2026sarm2,hung2025victor}.
Other methods estimate temporal distances between observations or to task completion~\citep{huang2026rynnvalue,liu2025timerewarder}, or predict relative progress and local advantages between states~\citep{tan2025robo,zhai2025vision,mao2026arm,liu2026steam}.
Large Reward Models expose progress, completion, and temporal-comparison interfaces within a single framework for robot policy refinement~\citep{wu2026largereward}.

Recent work also expands supervision and temporal context to address imperfect execution.
RoboMeter combines frame-level progress and trajectory-preference supervision~\citep{liang2026robometer}; ReTVL models degradation and recovery around retries~\citep{qin2026beyond}, and DenseReward incorporates synthesized physical failures~\citep{fang2026densereward}.
ViVa and World Value Models draw on video-generation priors for value estimation~\citep{lv2026viva,wang2026world}, while Robo-Dopamine~2.0 explicitly addresses execution history and out-of-distribution states~\citep{xu2026robo}.
These developments motivate evaluation that tests shared behavioral requirements while respecting differences in value semantics.

\subsection{Evaluation of Robotic Value Models}
Fine-grained policy evaluation calls for assessing execution behavior alongside task outcomes, as exemplified by RoboEval's structured behavioral measures~\citep{wang2025roboeval}.
StepEval proposes using VLMs to automate assessment of intermediate subgoal outcomes~\citep{elmallah2025score}.
Using learned models as evaluators requires checking whether their judgments faithfully reflect task execution.
Eval-Actions addresses this need by comparing predicted execution-quality scores with expert grading and ranking annotations~\citep{liu2026evalactions}.

Value-model evaluations examine complementary aspects of feedback quality.
RoboRewardBench evaluates completion ratings against human-verified labels~\citep{lee2026roboreward}, while RBM-EVAL measures task alignment, progress correlation, and trajectory ranking~\citep{liang2026robometer}.
GVL and OpenGVL assess whether predicted progress respects temporal ordering~\citep{ma2025vision,budzianowski2025opengvl}.
ManiRewardBench adds subtask-level progress annotations and success-detection tests~\citep{chen2026topreward}, whereas Progress-Bench probes demonstration-conditioned progress estimation under viewpoint changes and queries whose progress cannot be determined from the reference demonstration~\citep{zhang2026progresslm}.
ContextProgress-Bench, introduced with ProgressCompass, evaluates context-dependent progress estimation through state recall, sequence tracking, and disambiguation of recurring observations~\citep{zhang2026progresscompass}.

Process-level diagnostics examine reasoning during execution. 
RoboProcessBench uses static and temporal questions about manipulation~\citep{xia2026roboprocessbench}.
PRIMO Bench assesses progress estimation from execution context, accompanying a model trained to reason explicitly about progress~\citep{liu2026passive}.
RoboPulse evaluates pairwise micro-progress judgments~\citep{ji2026prm}, and RoboPulse++ extends evaluation to interval-level progress and regression~\citep{liu2026prm15}.
Suboptimal-Value-Bench directly tests value stability during hesitation and sensitivity to progress loss during retries~\citep{wang2026world}.

Failure-oriented systems study complementary forms of execution understanding. AHA detects and explains manipulation failures~\citep{duan2025aha}, while REFLECT uses execution histories to explain failures and guide correction~\citep{liu2023reflect}. FailSafe additionally pairs generated failures with executable recovery actions~\citep{lin2025failsafe}. These systems motivate examining whether value feedback distinguishes corrective activity from a successful recovery outcome.

Robustness is another emerging focus. 
FailBench tests success/failure judgments across diverse data sources~\citep{navasardyan2026failbench}, while RoboRMBench examines reward invariance under semantically equivalent instruction paraphrases~\citep{jeung2026robormbench}.
Paraphrase invariance complements RoboValue's tests of sensitivity to instructions that change task requirements.
The ARS semantic-mismatch suite directly probes this sensitivity by testing whether wrong-object manipulation receives spurious progress credit~\citep{hu2026ars}.

RoboValue connects these evaluation concerns through four complementary capabilities: task-state understanding, temporal progress monitoring, failure and recovery reasoning, and value consistency.
Using diagnostic trajectories under a shared sim-and-real protocol, it examines progress judgments at recurring visual states, responses to effective and ineffective recovery, and the consistency of subtask value gains across valid execution orders.
This joint diagnosis identifies which aspects of value feedback remain unreliable despite strong aggregate outcome or progress scores.

\section{RoboValue Dataset}
\label{sec:construction}

\begin{figure}[t]
  \centering
  \includegraphics[width=\textwidth]{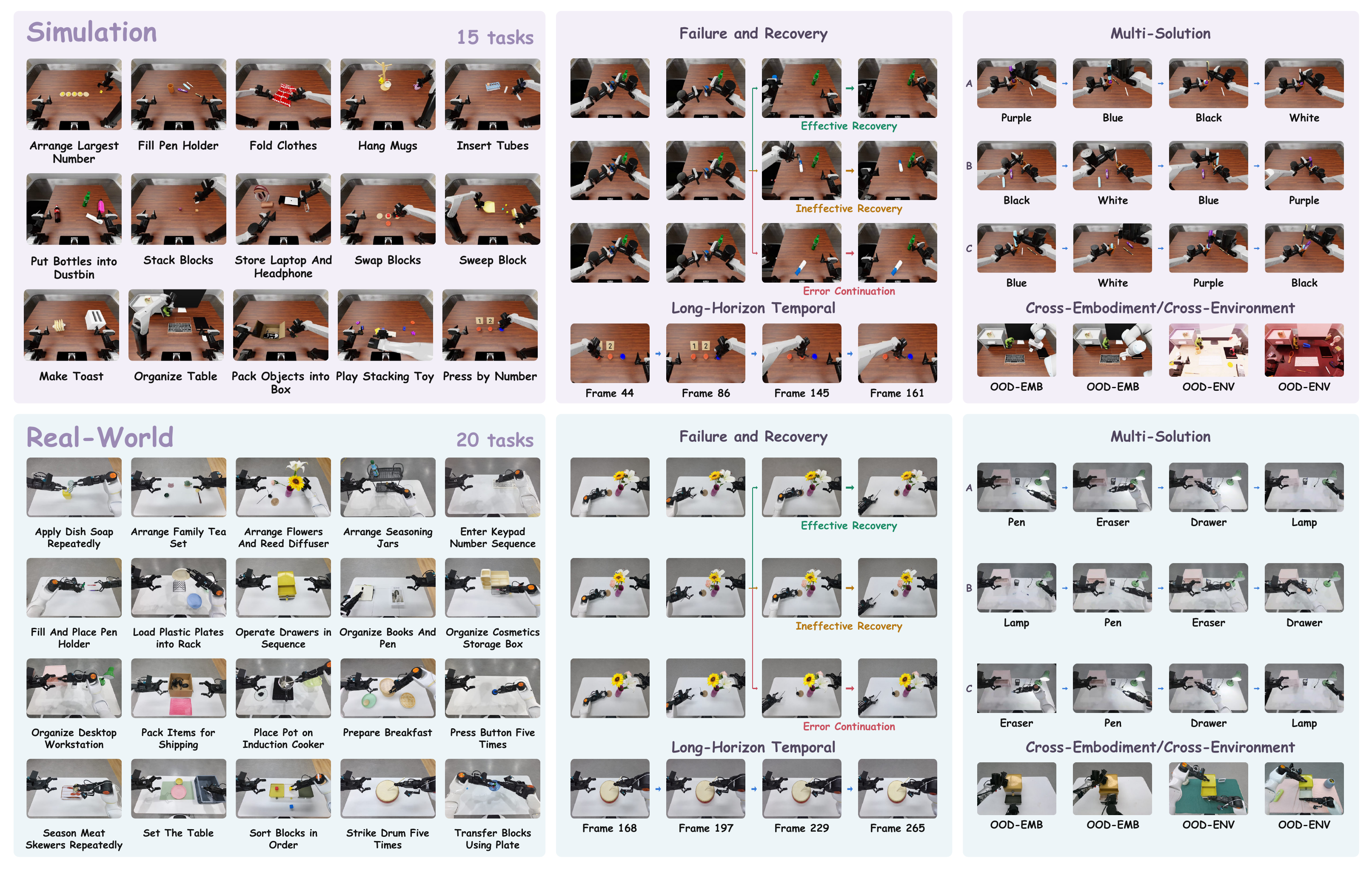}
  \caption{
    \textbf{RoboValue Dataset.} Tasks and diagnostic trajectories in simulation (top) and the real world (bottom). Examples show contrasting failure and recovery outcomes, recurring visual states, alternative valid solutions, and separate embodiment and environment shifts.
  }
  \label{fig:Dataset}
\end{figure}

\subsection{Dataset Composition} 
RoboValue-Dataset comprises training and test sets covering 35 dual-arm manipulation tasks: 15 simulation tasks adapted from RoboDojo~\citep{chen2026robodojo} and 20 real-world tasks.
These tasks span diverse manipulation skills and vary in procedural complexity and physical interaction requirements.
Simulation provides diverse object assets and controllable scene configurations, while real-world data captures realistic visual conditions and physical interactions.
The training set contains 3,500 expert demonstrations, with 100 per task, all collected under the standard in-domain (ID) setting.
The One-Shot track uses one demonstration per task from this split; the Full-Shot track will allow training or fine-tuning on the complete split.
The test set is disjoint from the training set and contains 2,792 trajectories spanning the standard setting and two out-of-domain (OOD) settings: cross-embodiment and cross-environment.
Cross-embodiment evaluation changes the robot platform, while cross-environment evaluation varies scene appearance, lighting, and clutter.
Each OOD setting varies either the embodiment or the environment relative to the standard setting, with a shared evaluation protocol across all three settings.
Figure~\ref{fig:Dataset} summarizes the task coverage, evaluation settings, and representative diagnostic trajectories.
Task specifications, execution visualizations, and OOD configurations are detailed in Appendices~\ref{app:simulation_tasks} and~\ref{app:realworld_tasks}.

\subsection{Trajectory Design}
\label{sec:trajectory_design}
We organize six trajectory types into four categories to probe how value judgments respond to execution quality, history, and subtask order.

\noindent (1) \textbf{Expert Demonstrations:} The robot carries out all required subtasks fluently and without errors, providing complete demonstrations of successful task execution.

\noindent (2) \textbf{Failure and Recovery Trajectories:} The three types share an initial failure scenario but differ in subsequent actions and outcomes.
These contrasting outcomes test whether value predictions distinguish corrective activity from actual recovery.

\begin{itemize}
  \item \emph{Error Continuation:} After an execution error, the robot continues with the remaining task without correcting it.
  \item \emph{Effective Recovery:} The robot takes corrective actions that resolve the error.
  \item \emph{Ineffective Recovery:} The robot attempts to recover but fails to correct the error before continuing with the remaining task.
\end{itemize}

\noindent (3) \textbf{Long-Horizon Temporal Trajectories:} The robot performs repeated actions and revisits visually similar states with different execution histories.
These trajectories test whether models use execution history to track accumulated task progress when the current observation alone is ambiguous.

\noindent (4) \textbf{Multi-Solution Trajectories:} The robot completes the task using alternative valid subtask orders that respect task dependencies.
For selected tasks, valid solutions also vary the operating arm or the division of work between arms while preserving the task requirements.
These trajectories test whether the same semantic subtask receives consistent value gains across valid solutions.

\begin{figure}[t]
  \centering
  \includegraphics[width=\textwidth]{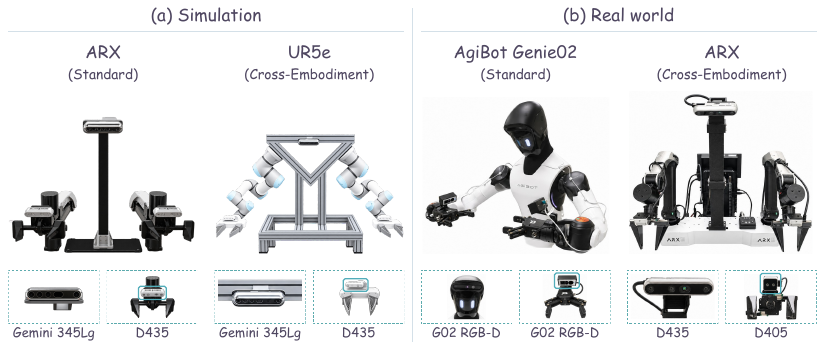}
  \caption{\textbf{Robot Platforms and Camera Configurations.}
    Standard and cross-embodiment platforms in (a) simulation and (b) the real world.
    Insets show each platform's head (left) and wrist (right) cameras; solid outlines identify wrist cameras.}
  \label{fig:robots}
\end{figure}

\subsection{Data Collection}
\label{sec:data_collection}
We collect all trajectory types through VR-based teleoperation using handheld controllers, with controller motions mapped to target end-effector poses. Each trajectory includes synchronized RGB-D observations from one head-mounted camera and two wrist-mounted cameras, together with robot states and action targets. Test-trajectory collection covers the standard, cross-embodiment, and cross-environment settings in both simulation and the real world; training demonstrations use only the standard setting. Figure~\ref{fig:robots} illustrates the robot platforms and their camera and gripper configurations.

\noindent\textbf{Simulation Platform and Trajectory Collection.}
We collect trajectories for 15 manipulation tasks using RoboDojo, a simulation and data-collection framework built on NVIDIA Isaac Sim~\citep{nvidia_isaac_sim}. ARX serves as the standard embodiment, while a dual-arm UR5e setup is used for cross-embodiment evaluation. Simulation provides diverse object assets and varied scene configurations under controlled conditions.

\noindent\textbf{Real-World Platform and Trajectory Collection.}
We collect real-world trajectories for 20 manipulation tasks. AgiBot Genie02 serves as the standard embodiment, while a dual-arm ARX setup is used for cross-embodiment evaluation. Real-world data capture visual sensing and contact dynamics in complex physical environments.

\subsection{Trajectory Annotation}
We manually annotate the test trajectories to support evaluation across different capability dimensions.
For \textbf{successful trajectories}, we record the start and end timestamps of each subtask.
The resulting intervals support subtask identification, while their boundaries enable comparison of value gains for the same semantic subtask across valid solutions.
The units used to enumerate valid execution orders may group several annotated subtasks. This grouping leaves the subtask definitions and temporal boundaries used for evaluation unchanged.
For \textbf{failure and recovery trajectories}, we mark the beginning and end of the initial failure stage, together with the endpoint of the subtask in which the failure occurs.
For both Effective Recovery and Ineffective Recovery trajectories, we additionally mark the moment when the recovery outcome becomes observable.
This timestamp separates the recovery attempt from the stage in which its success or failure can be assessed.
Together, these annotations support failure-point localization and separate evaluation of value trends during failure, error continuation, recovery attempts, and their outcomes.
The training split does not include subtask boundary annotations; annotations constructed for a model's adaptation are separate from these test-set labels.

\Needspace{7\baselineskip}
\section{RoboValue Benchmark}
\label{sec:robovaluebenchmark}
\subsection{Preliminaries}
\label{sec:formulation}
In RoboValue, a \emph{robotic value model} is a learned evaluator that uses visual observations and task instructions to assess how well a robot's execution satisfies, or advances toward, a specified goal.
This definition encompasses reward and progress models with different output semantics and numerical scales (Section~\ref{sec:related_work}).

Consider an execution with task instruction $l$.
A query $A_t$ at timestamp $t$ combines the current visual observation, which may include multiple camera views, with model-specific context such as execution history, a reference demonstration, or robot states. For a query $A$ and instruction $l$, we write the scalar execution score as
\[
v(A;l)=f_\theta(A,l)\in\mathbb{R},
\]
where $\theta$ denotes the model parameters and $f_\theta$ includes the model-specific adaptation described in Section~\ref{sec:protocol}.
Scores are oriented so that higher values indicate more favorable task assessments, while their semantics and numerical scales remain model-dependent.
We additionally use a signed pairwise assessment $c(A_1,A_2;l)$, where a positive value favors $A_2$ over $A_1$, and a textual subtask description $s(A;l)$ for models supporting this output.
Along an execution, we write $v(t)=v(A_t;l)$ and use $v_i=v(t_i)$ for predictions sampled at timestamps $t_i$.
We use $\mathbf{1}[\cdot]$ for the indicator function; metric-specific quantities are defined where they are introduced.

%




\subsection{Evaluation Protocol}
\label{sec:protocol}

\noindent\textbf{Shared Evaluation Interfaces.}
We compare execution judgments through the interfaces defined above, using model-specific adapters to accommodate differences in observation requirements and output semantics.
Scalar scores $\boldsymbol{v(A;l)}$ are obtained by transforming native predictions or aggregating relative progress estimates.
For pairwise assessments $\boldsymbol{c(A_1,A_2;l)}$, we use native comparison outputs when available; otherwise, we compute $c(A_1,A_2;l)=v(A_2;l)-v(A_1;l)$.
The subtask interface $\boldsymbol{s(A;l)}$ extracts descriptions of the current subtask from supported textual outputs.
These shared operations enable evaluation with common metrics while retaining model-dependent score interpretations.
Appendix~\ref{app:adapter_implementations} details the input construction, output mappings, and inference settings for each model.

\noindent\textbf{Evaluation Tracks.}
We define three tracks according to access to task-specific demonstrations.
The Zero-Shot track uses released checkpoints without task-specific adaptation or reference demonstrations.
The One-Shot track provides a single demonstration per task, selected from that task's 100 standard-scenario training demonstrations and kept separate from the test trajectories.
Depending on the model, this demonstration is used for inference-time conditioning or task-specific adaptation.
The Full-Shot track will permit training or fine-tuning on the complete RoboValue training split.
This report evaluates the Zero-Shot and One-Shot tracks, with results reported separately.

\noindent\textbf{Generalization Settings.}
We report in-domain (ID) performance under the standard setting and out-of-domain (OOD) performance under the cross-embodiment and cross-environment settings defined in Section~\ref{sec:construction}.
Embodiment and environment shifts are assessed separately under the same evaluation protocol, without additional adaptation to the shifted conditions.
These comparisons assess the robustness of value judgments to changes in robot platform and visual context.

\noindent\textbf{Score Aggregation.}
For each metric and evaluation setting, we first average task scores separately within simulation and the real world.
The main tables report the arithmetic mean of these two domain averages, giving each domain equal weight regardless of its number of evaluated tasks.

\subsection{Capability Dimensions and Metrics}
\label{sec:metrics}
RoboValue evaluates what an execution achieves, how progress evolves, how errors and recovery affect value, and whether the resulting feedback remains consistent.
Table~\ref{tab:metric_overview} summarizes the specific capabilities targeted by individual metrics.

\begin{table}[t]
  \centering
  \caption{\textbf{Overview of RoboValue Evaluation Metrics.} Fine-grained capabilities assessed within the four complementary evaluation dimensions.}
  \label{tab:metric_overview}
  \begingroup
  \small
  \setlength{\tabcolsep}{5.5pt}
  \setlength{\aboverulesep}{4.5pt}
  \setlength{\belowrulesep}{4.5pt}
  \renewcommand{\arraystretch}{1.10}
  \renewcommand{\tabularxcolumn}[1]{m{#1}}
  \arrayrulecolor{black!55}
  \newlength{\RVMetricDimensionWidth}
  \newlength{\RVMetricBodyWidth}
  \newlength{\RVMetricNameWidth}
  \setlength{\RVMetricDimensionWidth}{0.175\linewidth}
  \setlength{\RVMetricBodyWidth}{\dimexpr\linewidth-\RVMetricDimensionWidth-2\tabcolsep\relax}
  \settowidth{\RVMetricNameWidth}{\textbf{Memory-VOC}}
  \newcommand{\RVMetricBlock}[2]{%
    \begingroup
    \newcommand{\RVMetricName}[1]{\textcolor{#1}{\textbf{##1}}}%
    \begin{tabularx}{\RVMetricBodyWidth}[c]{@{}
      >{\centering\arraybackslash}m{\RVMetricNameWidth}
      >{\raggedright\arraybackslash}X@{}}
      #2%
    \end{tabularx}%
    \endgroup
  }
  \newcommand{\RVMetricRule}{%
    \arrayrulecolor{black!18}\midrule[0.3pt]%
    \arrayrulecolor{black!55}%
  }
  \begin{tabular}{@{}cc@{}}
    \toprule
    \makebox[\RVMetricDimensionWidth][c]{\textbf{Dimension}}
    & \RVMetricBlock{black}{%
      \textbf{Metric}
      & \centering\arraybackslash\textbf{Capability} \\
    } \\
    \midrule
    \textcolor{RVUnderstandingInk}{\textbf{\makecell[c]{Task-State\\Understanding}}}
    & \RVMetricBlock{RVUnderstandingInk}{%
      \RVMetricName{SA}
      & Discriminating overall goal satisfaction from partial or failed execution. \\
      \RVMetricRule
      \RVMetricName{TGA-CT}
      & Grounding value judgments in language-specified goals across different tasks. \\
      \RVMetricRule
      \RVMetricName{TGA-CF}
      & Fine-grained grounding of object, action, placement, and constraint requirements. \\
      \RVMetricRule
      \RVMetricName{SIA}
      & Recognizing the active semantic subtask within a multi-stage execution. \\
    } \\
    \midrule
    \textcolor{RVProgressInk}{\textbf{\makecell[c]{Temporal\\Progress\\Monitoring}}}
    & \RVMetricBlock{RVProgressInk}{%
      \RVMetricName{VOC}
      & Ordering intermediate states by their progress toward task completion. \\
      \RVMetricRule
      \RVMetricName{Cycle-VOC}
      & Distinguishing the direction of task progress from elapsed-time cues. \\
      \RVMetricRule
      \RVMetricName{Memory-VOC}
      & Using execution history to resolve progress ambiguity at recurring visual states. \\
    } \\
    \midrule
    \textcolor{RVRecoveryInk}{\textbf{\makecell[c]{Failure and\\Recovery\\Reasoning}}}
    & \RVMetricBlock{RVRecoveryInk}{%
      \RVMetricName{FPL}
      & Temporally localizing the onset of task-relevant execution errors. \\
      \RVMetricRule
      \RVMetricName{TRR}
      & Distinguishing unresolved errors, corrective progress, and recovery outcomes. \\
    } \\
    \midrule
    \textcolor{RVConsistencyInk}{\textbf{\makecell[c]{Value\\Consistency}}}
    & \RVMetricBlock{RVConsistencyInk}{%
      \RVMetricName{VS}
      & Maintaining stable local value feedback while retaining informative updates. \\
      \RVMetricRule
      \RVMetricName{CSVC}
      & Consistent valuation of semantic subtasks across valid execution orders. \\
    } \\
    \bottomrule
  \end{tabular}
  \arrayrulecolor{black}
  \endgroup
\end{table}
\FloatBarrier

\subsubsection{Task-State Understanding}
Understanding an execution involves relating observed behavior to the intended goal and the subtasks required to achieve it.
We evaluate outcome discrimination (SA), instruction grounding (TGA), and identification of the active subtask (SIA).

\noindent\textbf{Success Accuracy (SA, $\uparrow$).}
Using value predictions to supervise policy learning requires distinguishing executions that fulfill the task requirements from those that do not.
Reliable outcome discrimination supports selecting successful demonstrations for imitation learning and deriving outcome-based rewards for reinforcement learning.
However, partially completed executions may still make substantial task-relevant progress and visually resemble successful ones, making this distinction challenging.
SA evaluates outcome discrimination by testing whether successful episodes receive higher terminal scores than unsuccessful ones, without requiring a shared absolute threshold.

For each episode $e$, we average predictions over the final $2\%$ of its duration into a terminal score $z_e$ to reduce sensitivity to local fluctuations. Let $\mathcal E^+$ and $\mathcal E^-$ denote the successful and unsuccessful episodes of the same task and evaluation setting, respectively.
SA is the fraction of episode pairs in which the successful episode receives a higher score:
\[
\mathrm{SA} = \frac{\sum_{e\in\mathcal E^+}\sum_{e'\in\mathcal E^-}\mathbf{1}[z_e>z_{e'}]}{|\mathcal E^+|\,|\mathcal E^-|}.
\]




\noindent\textbf{Task Grounding Accuracy (TGA, $\uparrow$).}
Correct outcome discrimination does not establish instruction grounding: a model may distinguish successful from unsuccessful executions while assigning similar value gains under instructions that specify different goals.
The same sequence of actions may fulfill one instruction while leaving another unmet, so value judgments should reflect the requested task.
This grounding supports policy learning with a shared value model across tasks and the retrieval of demonstrations that match a requested goal.

For each successful episode $e$, we sample query pairs $(A_0,A_T)$ from the first and last $2\%$ of the episode, respectively. Its mean predicted value gain under instruction $l'$ is
\[
\Delta_e(l')
=
\operatorname{mean}_{(A_0,A_T)}
c(A_{0},A_{T};l').
\]
TGA compares this gain under the correct and negative instructions while keeping the visual execution fixed.
The \textbf{Cross-Task (CT)} variant uses instructions from other tasks to test discrimination between distinct goals.
The \textbf{Counterfactual (CF)} variant modifies the target object, action, placement, or constraint to test sensitivity to specific task requirements.

\Needspace{6\baselineskip}
Let $\mathcal N$ denote the negative instructions for the variant being evaluated, with CT and CF scored separately.
Over the successful episodes $\mathcal E^+$, TGA is the fraction for which the correct instruction $l$ yields a larger gain than every negative instruction:
\[
\mathrm{TGA} = \frac{1}{|\mathcal E^+|}\sum_{e\in\mathcal E^+}\mathbf{1}\left[\Delta_e(l)>\max_{l'\in\mathcal N}\Delta_e(l')\right].
\]




\noindent\textbf{Subtask Identification Accuracy (SIA, $\uparrow$).}
Outcome discrimination and instruction grounding do not establish whether a model can identify the subtask being performed at a particular point in execution.
Recognizing the active subtask connects observed behavior to the current intermediate objective, supporting stage-specific progress estimation and reward assignment.
It also helps associate execution errors with the corresponding step of a task plan, providing context for failure diagnosis and corrective guidance.
SIA evaluates subtask recognition through the model's description of the current subtask.

We sample queries within annotated subtask intervals, away from stage boundaries to reduce ambiguity.
Models supporting the textual subtask interface produce a free-form description $s(A;l)$ without being given the candidate labels.
We use DeepSeek-V4-Pro~\citep{deepseekai2026deepseekv4} as a separate LLM judge to evaluate each description against the task's candidate subtasks, assigning probability $q_i$ to the ground-truth subtask for query $i$.
SIA is the geometric mean of these probabilities over $N$ queries:
\[
\mathrm{SIA} = \exp\left(\frac{1}{N}\sum_{i=1}^{N}\log q_i\right).
\]





\subsubsection{Temporal Progress Monitoring}
Beyond assessing task states, value models need to track how task progress changes over an execution.
We examine whether values follow forward progress (VOC), respond to regression (Cycle-VOC), and reflect accumulated progress when visually similar states recur under different histories (Memory-VOC).


\noindent\textbf{Value-Order Correlation (VOC, $\uparrow$).}
Intermediate value estimates can provide supervision throughout an execution by indicating how successive states advance the task.
This feedback supports dense reward construction and temporal credit assignment, helping policy learning associate intermediate transitions with progress toward the goal.
VOC~\citep{ma2025vision} assesses whether value predictions preserve the order of progress along successful trajectories, using temporal position as a proxy.
By comparing ranks, it accommodates variation in the rate of progress without requiring calibrated value magnitudes or a linear relationship between predicted value and time.

For sampled timestamps $0\leq t_1<\cdots<t_N<T$ within a successful trajectory of duration $T$, we compute the Spearman rank correlation $\rho_{\mathrm S}$ between predicted values and normalized time, assigning zero to constant predictions:
\[ \mathrm{VOC} = \rho_{\mathrm S}\left((v_i)_{i=1}^{N},(t_i/T)_{i=1}^{N}\right). \]




\noindent\textbf{Cycle-VOC ($\uparrow$).}
Along a successful trajectory, progress and elapsed time usually increase together, so a model that raises its predictions with time can obtain a high VOC without responding to changes in the task state.
Cycle-VOC probes this elapsed-time shortcut by reversing the order of recorded states while playback time continues to advance.
For the successful trajectories used here, agreement with the recorded progress order corresponds to increasing predictions in the forward half and decreasing predictions in the reverse half.
Reverse playback provides a controlled visual diagnostic and need not represent a physically executable trajectory.

We replay a successful trajectory forward and then backward as one continuous sequence, with a single shared turning point.
For models that use execution history, the reverse half retains the preceding forward context.
Cycle-VOC averages the Spearman correlations with increasing progress order in the forward half and decreasing progress order in the reverse half; both scored halves include the turning point, and a constant half receives a score of zero.


\noindent\textbf{Memory-VOC ($\uparrow$).}
A robot may return to a visually similar configuration after completing another subtask or repeating an operation. The current observation can therefore correspond to different amounts of completed work, making execution history important for interpreting progress. Memory-VOC targets the ability to retain and use this task-relevant history. This capability supports progress feedback for long-horizon policy learning and execution monitoring.

Memory-VOC applies the same Spearman scoring rule as VOC to successful long-horizon trajectories containing these recurring visual states, using the execution history available under each model's evaluation interface. Since temporal position remains correlated with progress in these trajectories, a high score alone does not establish effective use of history. Cycle-VOC provides a complementary test of whether the predictions also respond appropriately when progress reverses.


\subsubsection{Failure and Recovery Reasoning}
We extend evaluation to trajectories containing execution errors, where subsequent actions may restore progress or leave the underlying problem unresolved.
FPL evaluates failure localization, while TRR assesses value responses to initial failure, continued error, recovery attempts, and their outcomes.


\noindent\textbf{Failure-Point Localization (FPL, $\downarrow$).}
A value model may recognize that an error has occurred without identifying when it began.
Accurate failure localization supports policy learning from partially successful trajectories by retaining correctly executed segments as useful supervision and directing negative feedback toward the relevant errors.
FPL assesses whether a model's most pronounced value decline is temporally aligned with the onset of an error.
A delayed response or a larger decline during unrelated activity can lead to an incorrect localization even when the model reacts to the actual error.

We evaluate Error Continuation and Effective Recovery trajectories, each containing a single execution error, so that the target failure onset is uniquely identifiable.
We sample value predictions at $5$ Hz, including the final frame. 
A ZigZag filter extracts declines using a reversal threshold of $\alpha=0.2$ times the largest drop between an earlier and a later value in the trajectory, suppressing minor reversals while retaining unfinished terminal declines. 
The peak preceding the largest detected decline gives the predicted failure onset $\hat t$. 
With annotated onset $t^{\mathrm{GT}}$ and final-frame timestamp $T$, FPL measures the normalized localization error, assigning the maximum possible error within the trajectory when no decline is detected:

\[ \mathrm{FPL} = \frac{1}{T}\begin{cases}
\left|\hat t-t^{\mathrm{GT}}\right|, & \text{if a decline is detected},\\[4pt]
\max\!\left(t^{\mathrm{GT}},\,T-t^{\mathrm{GT}}\right), & \text{otherwise}.
\end{cases} \]




\noindent\textbf{Trajectory Recovery Reasoning (TRR, $\uparrow$).}
Corrective actions can improve an execution before their outcome becomes observable, and an attempt that initially makes progress may still fail. Distinguishing these stages supports reward supervision for recovery policies and execution monitoring, where corrective activity can otherwise be mistaken for a resolved error. TRR assesses whether values reflect the initial failure, the progression of a corrective attempt, and its eventual outcome, as well as execution that continues without correction.

TRR uses matched trajectories that share an initial failure scenario and follow three branches: continued error (CE), successful recovery (SR), and failed recovery (FR). Each branch is divided into the annotated stages it contains: initial failure ($f$), continued error ($c$), recovery attempt ($r$), and outcome ($o$).

We expect a decrease during initial failure and no overall increase during continued error. Recovery-attempt stages are expected to show an increase as corrective actions work toward restoring progress. Once the recovery outcome becomes observable, values are expected to continue increasing for SR and decrease for FR. In FR, the outcome interval ends before the next subtask begins, so the assessment reflects the unresolved failure without mixing in later task progress.

\begin{center}
\begin{tabular}{c|cccc}
\hline
Branch & Failure & Continued Error & Recovery Attempt & Outcome \\
\hline
CE & $\downarrow$ & $\not\uparrow$ & -- & -- \\
SR & $\downarrow$ & -- & $\uparrow$ & $\uparrow$ \\
FR & $\downarrow$ & -- & $\uparrow$ & $\downarrow$ \\
\hline
\end{tabular}
\end{center}

Within each annotated stage, we retain samples from the $10$ Hz base grid and add native frames when necessary to reach $K=10$ samples; stages containing fewer than $K$ native frames use all available frames. No interpolation is applied.

For temporally ordered samples $i<j$ in stage $s$, let $v_i$ denote the predicted value and $w_i$ the duration represented by sample $i$. We summarize the stage trend with $D_s$, a normalized sum of value changes over all ordered pairs. The duration weights account for unequal spacing between samples; positive and negative $D_s$ indicate increasing and decreasing trends, respectively, and $D_s=0$ when the denominator is zero.

Each branch is evaluated on its own trajectory and receives credit only when all required stage trends are satisfied. Branch scores are averaged within each matched group, then across groups within each task and evaluation setting. The stage direction and binary branch score are

\[
D_s = \frac{\sum_{i<j}w_iw_j\bigl(v_j-v_i\bigr)}
{\sum_{i<j}w_iw_j\left|v_j-v_i\right|},
\qquad
\mathrm{TRR} = \mathbf{1}[D_f<0]
\begin{cases}
\mathbf{1}[D_c\leq0], & \mathrm{CE}\\[2pt]
\mathbf{1}[D_r>0\land D_o>0], & \mathrm{SR}\\[2pt]
\mathbf{1}[D_r>0\land D_o<0], & \mathrm{FR}
\end{cases}
\]

\subsubsection{Value Consistency}
Reliable feedback should also remain stable and informative within an execution and assign comparable credit to equivalent task progress across valid solutions.
VS assesses unnecessary fluctuations and prolonged flat regions along expert trajectories, while CSVC compares the gains assigned to the same semantic subtask across valid solutions.


\noindent\textbf{Value Stability (VS, $\uparrow$).}
When value differences are used as local rewards, spurious fluctuations can alternate between positive and negative feedback along an otherwise successful execution, obscuring which transitions advance the task.
However, stability alone is insufficient: prolonged flat regions can appear stable while leaving intermediate changes in execution undifferentiated.
VS therefore assesses whether predictions maintain a consistent trend along expert demonstrations while providing updates throughout execution.

We combine Kaufman's Efficiency Ratio (ER)~\citep{kaufman1995smarter}, which compares the absolute net value change with the total variation, with a non-flat time coverage factor $C$, which measures the fraction of sampled duration covered by intervals with a value change. Together, they penalize reversals and prolonged intervals without updates. Since a stable trend can still evolve in the wrong direction, VS is interpreted alongside progress-oriented metrics such as VOC and Cycle-VOC.

We sample each expert trajectory at $5$ Hz, obtaining $N$ predictions $v_i$ at timestamps $t_i$. The sequence $\tilde v_i$ replaces the first and last three predictions with their respective means and leaves the other predictions unchanged. ER uses this sequence to reduce sensitivity to endpoint fluctuations, whereas $C$ uses the raw values and timestamps. We set $\mathrm{ER}=1$ when its denominator is zero; a constant raw sequence therefore has $C=0$ and $\mathrm{VS}=0$. The per-trajectory score is

\[ \mathrm{ER} = \frac{|\tilde v_N-\tilde v_1|}{\sum_{i=1}^{N-1}|\tilde v_{i+1}-\tilde v_i|},\qquad C = \frac{\sum_{i=1}^{N-1}(t_{i+1}-t_i)\mathbf{1}[v_{i+1}\neq v_i]}{t_N-t_1},\qquad \mathrm{VS} = \mathrm{ER}\cdot C. \]



\noindent\textbf{Cross-Solution Value Consistency (CSVC, $\uparrow$).}
Many manipulation tasks admit several valid subtask orders that respect the same task dependencies.
CSVC evaluates whether the same semantic subtask receives comparable value gains across these solutions.
When these gains are used to select or weight training segments, variation associated with execution position can assign different learning weights to semantically matched subtasks with otherwise comparable contributions to the task goal.
If valid orders are unevenly represented in training data, such differences may reinforce associations between subtask identity and execution order and reduce the transferability of local feedback to less frequently observed but equally valid solutions.

We align subtasks by their semantic identities and compare their local gains, allowing different solutions to follow different overall value curves. For subtask $a$ in solution $q$, let $(A_j^-,A_j^+)$ denote the $j$th matched query pair near its annotated start and end boundaries. We use $m=4$ pairs, allowing repeated pairs for very short subtasks, and define the gain $c_{q,a}$ as the median pairwise assessment.

Let $\mathcal A$ denote the evaluated semantic subtasks and $\mathcal S_a$ the valid solutions containing subtask $a$. 
The median gain across these solutions is $\widetilde c_a$. 
We compare the root-mean-square magnitude $G$ of the median gains with the root-mean-square deviation $D$ from these medians, weighting semantic subtasks equally. 
A larger $G$ relative to $D$ indicates greater consistency across solutions. 
This agreement alone does not establish that the gains correctly reflect task progress, so CSVC is interpreted together with progress-oriented metrics. 
Differences in local gains must also be interpreted in light of the model's value semantics and execution context, since semantically matched subtasks need not have identical value contributions. 
When $G+D=0$, the ratio is undefined and we assign $\mathrm{CSVC}=-1$.

\[ c_{q,a} = \operatorname{median} c(A_j^-,A_j^+;l),\qquad \widetilde c_a = \operatorname{median}_{q\in\mathcal S_a}c_{q,a}. \]

\[
G = \sqrt{\frac{1}{|\mathcal A|}\sum_{a\in\mathcal A}\widetilde c_a^2},\qquad
D = \sqrt{\frac{1}{|\mathcal A|}\sum_{a\in\mathcal A}\frac{1}{|\mathcal S_a|}\sum_{q\in\mathcal S_a}\left(c_{q,a}-\widetilde c_a\right)^2},\qquad
\mathrm{CSVC} = \frac{G-D}{G+D}.
\]

\FloatBarrier
\section{RoboValue Leaderboard}
\label{sec:leaderboard}

\subsection{Evaluated Value Models}
\label{sec:models}
Our evaluation covers the main value formulations discussed in Section~\ref{sec:related_work}. These include completion assessments based on ordinal ratings or response likelihoods~\citep{lee2026roboreward,chen2026topreward}, absolute or stage-aware progress estimates~\citep{liang2026robometer,feng2026procvlm}, temporal distance predictions~\citep{huang2026rynnvalue}, and relative progress judgments between states~\citep{tan2025robo,xu2026robo,zhai2025vision}. We further include LIV~\citep{ma2023liv}, which derives scores from image--text similarity, and RoboFAC~\citep{ye2025robofac}, whose execution assessments are mapped to scalar values.

TOPReward uses 8B and 32B Qwen3-VL backbones. Models use the shared interfaces and zero-shot or one-shot settings defined in Section~\ref{sec:protocol}. ProcVLM and VLAC are evaluated in both settings; their one-shot variants use task-specific adaptation and demonstration conditioning, respectively. Model profiles and adapter details are provided in Appendices~\ref{app:model_profiles} and~\ref{app:adapter_implementations}.

\subsection{Leaderboard}
\label{sec:leaderboard_scores}
We establish the RoboValue leaderboard to compare robotic value models across simulation and real-world tasks. Table~\ref{tab:robovalue_leaderboard} reports scores for the four capability dimensions and ranks models by overall performance, separately for zero-shot and one-shot evaluation.

The leaderboard summarizes the underlying metric-level evaluations: we aggregate normalized scores across in-domain (ID) and out-of-domain (OOD) settings where available, then average the resulting capability scores. Detailed scoring rules are provided in Appendix~\ref{app:leaderboard_scoring}.

\begin{table}[htbp]
\centering
\caption{\textbf{RoboValue Leaderboard.} Capability scores and their overall mean, with models ranked separately for zero-shot and one-shot evaluation. The \RVbest{Best} and \RVsecond{Second Best} scores are marked in each column within each setting.}
\label{tab:robovalue_leaderboard}
\begingroup
\small
\setlength{\tabcolsep}{3pt}
\renewcommand{\arraystretch}{1.11}
\newcommand{\RVLeaderboardHeader}[1]{{\fontsize{8}{9.5}\selectfont\bfseries\makecell[c]{#1}}}
\begin{tabularx}{\linewidth}{@{}l*{5}{>{\centering\arraybackslash}X}@{}}
\toprule
\textbf{Model}
& \textcolor{RVUnderstandingInk}{\RVLeaderboardHeader{Task-State\\Understanding}}
& \textcolor{RVProgressInk}{\RVLeaderboardHeader{Temporal Progress\\Monitoring}}
& \textcolor{RVRecoveryInk}{\RVLeaderboardHeader{Failure and\\Recovery Reasoning}}
& \textcolor{RVConsistencyInk}{\RVLeaderboardHeader{Value\\Consistency}}
& \RVLeaderboardHeader{Overall} \\
\midrule
\rowcolor{RVZeroBand}[0pt][0pt]
\multicolumn{6}{@{}c@{}}{\textit{Zero-Shot Evaluation}} \\
\midrule
RoboMeter-4B & 68.53 & \RVbest{83.33} & \RVsecond{46.78} & \RVsecond{41.57} & \RVbest{60.05} \\
RynnValue-4B & \RVbest{75.54} & 78.64 & 40.29 & 36.00 & \RVsecond{57.62} \\
RynnValue-8B & \RVsecond{70.98} & \RVsecond{78.78} & 42.39 & 32.59 & 56.19 \\
TOPReward-32B$^{\ddagger}$ & 46.96 & 65.81 & 41.30 & 38.04 & 48.03 \\
TOPReward-8B$^{\ddagger}$ & 48.09 & 57.02 & 39.64 & \RVbest{43.65} & 47.10 \\
ProcVLM-2B & 40.31 & 67.65 & 42.63 & 28.46 & 44.76 \\
\bottomrule
\end{tabularx}
\par\smallskip
\begin{minipage}{\linewidth}
\scriptsize\raggedright
$^{\dagger}$ 2.0 Preview checkpoints.\quad
$^{\ddagger}$ Qwen3-VL backbones.\quad
$^{*}$ RoboReward is excluded from VS and CSVC due to its discrete outputs; both count as zero.
\end{minipage}
\endgroup
\end{table}

RoboMeter-4B leads the zero-shot leaderboard, while Robo-Dopamine 2.0-8B Preview leads the one-shot leaderboard. The capability profiles reveal uneven performance: RynnValue-4B achieves the strongest task-state understanding, whereas Robo-Dopamine 2.0-8B Preview leads the other three dimensions. Failure and recovery reasoning and value consistency remain challenging even for these leading models.

\Needspace{7\baselineskip}
\noindent\textbf{Evaluation Tracks.}
RoboValue comprises three tracks: \emph{Zero-Shot}, \emph{One-Shot}, and \emph{Full-Shot}.
The current Zero-Shot and One-Shot evaluations do not use the full RoboValue training set for model training.
Opening with the dataset release, the Full-Shot track will allow training or fine-tuning on all 3,500 demonstrations, supporting methods that learn reward and value functions from task-domain trajectories~\citep{lv2026viva,chen2026sarm,chen2026sarm2,mao2026arm,liu2025timerewarder,liu2026steam,yu2026warp}.
All three tracks share the held-out test set and evaluation metrics, with separate rankings.

\FloatBarrier

\section{Experiments}
\label{sec:experiment}

\subsection{Main Results}
\label{sec:idresults}
Table~\ref{tab:robovalue_main} reports in-domain (ID) performance under the standard robot embodiment and environment. The metric-level results reveal capability differences that the aggregate leaderboard alone cannot resolve.

\begin{table}[htbp]
    \caption{\textbf{Main Results.} In-domain (ID) scores across the four capability dimensions, multiplied by 100. Lower FPL and higher scores elsewhere are better. Cyc and Mem denote Cycle-VOC and Memory-VOC. The \RVbest{Best} and \RVsecond{Second Best} scores are marked per metric within each zero-shot or one-shot setting.}
    \label{tab:robovalue_main}
    \centering
    \begingroup

    \small
    \rmfamily
    \setlength{\tabcolsep}{1.2pt}
    \renewcommand{\arraystretch}{1.03}

    \settowidth{\dimen0}{TGA-CT}
    \edef\rvtgawidth{\the\dimen0}
    \settowidth{\dimen0}{TGA-CF}
    \ifdim\dimen0>\rvtgawidth\relax
        \edef\rvtgawidth{\the\dimen0}
    \fi
    \edef\rvtgawidth{\the\dimexpr\rvtgawidth+2pt\relax}

    \def\modeband#1#2{%
        \specialrule{\lightrulewidth}{\aboverulesep}{2pt}
        \multicolumn{11}{@{}c@{}}{%
            \makebox[\textwidth][c]{%
                \rlap{\smash{\color{#1}%
                    \rule[-\dp\csname @arstrutbox\endcsname]
                        {\textwidth}
                        {\dimexpr\ht\csname @arstrutbox\endcsname
                            +\dp\csname @arstrutbox\endcsname\relax}}}%
                \makebox[\textwidth][c]{\textit{#2}}%
            }%
        }\\%
        \specialrule{\lightrulewidth}{2pt}{\belowrulesep}
    }

\begin{tabularx}{\textwidth}{
        @{}
        l
        >{\centering\arraybackslash}X
        *{2}{>{\centering\arraybackslash}p{\rvtgawidth}}
        *{7}{>{\centering\arraybackslash}X}
        @{}
    }

        \toprule

        \multirow{2}{*}{\textbf{Model}}
        & \multicolumn{3}{c}{\textcolor{RVUnderstandingInk}{\textbf{Understanding}}}
        & \multicolumn{3}{c}{\textcolor{RVProgressInk}{\textbf{Tracking}}}
        & \multicolumn{2}{c}{\textcolor{RVRecoveryInk}{\textbf{Diagnosis}}}
        & \multicolumn{2}{c}{\textcolor{RVConsistencyInk}{\textbf{Consistency}}} \\

        \cmidrule(lr){2-4}
        \cmidrule(lr){5-7}
        \cmidrule(lr){8-9}
        \cmidrule(lr){10-11}

        & \textcolor{RVUnderstandingInk}{SA}
        & \textcolor{RVUnderstandingInk}{\mbox{TGA-CT}}
        & \textcolor{RVUnderstandingInk}{\mbox{TGA-CF}}
        & \textcolor{RVProgressInk}{VOC}
        & \textcolor{RVProgressInk}{Cyc}
        & \textcolor{RVProgressInk}{Mem}
        & \textcolor{RVRecoveryInk}{FPL}
        & \textcolor{RVRecoveryInk}{TRR}
        & \textcolor{RVConsistencyInk}{VS}
        & \textcolor{RVConsistencyInk}{CSVC} \\

        \modeband{RVZeroBand}{Zero-Shot Evaluation}

        LIV
        & 54.44 & 9.10 & 6.17 & 8.33 & 8.33 & 1.50 & 31.19 & 24.25 & 3.36 & -8.47 \\

        ProcVLM-2B
        & 62.20 & 20.04 & 14.36 & 51.10 & 47.41 & 9.73 & 35.93 & 21.55 & 5.25 & 4.60 \\

        RoboFAC-7B
        & 72.63 & 4.82 & 4.34 & 20.44 & 36.95 & -25.23 & 35.66 & \RVbest{33.80} & 3.34 & -0.30 \\

        RoboMeter-4B
        & \RVsecond{81.29} & \RVsecond{82.00} & 29.29 & \RVbest{86.05} & \RVbest{89.43} & 2.36 & \RVbest{29.76} & 24.83 & \RVbest{18.17} & \RVsecond{32.41} \\

        RoboReward-4B
        & 28.10 & 35.97 & 9.12 & 82.76 & 44.23 & \RVsecond{81.46} & 32.61 & 3.07 & - & - \\

        RoboReward-8B
        & 31.31 & 28.89 & 12.81 & \RVsecond{83.40} & 43.00 & \RVbest{81.77} & 31.90 & 2.60 & - & - \\
\bottomrule

    \end{tabularx}

    \par\smallskip
    \parbox{\textwidth}{\scriptsize\raggedright
        $^{\dagger}$ 2.0 Preview checkpoints.\quad
        $^{\ddagger}$ Qwen3-VL backbones.\quad
        $^{\S}$ High forward correlation coexists with weak reversal tracking; these entries are excluded from best/second-best marking.
    }
    \endgroup
\end{table}

\noindent\textbf{Can Value Models Reliably Understand Task State and Instructions?} Strong outcome discrimination does not guarantee reliable instruction grounding. As shown in Table~\ref{tab:robovalue_main}, RynnValue-4B and RoboMeter combine strong outcome discrimination with task grounding, consistent with instruction-verification training and mismatched-task trajectory comparisons, respectively.
Both are less accurate on counterfactual instructions, indicating that they can distinguish broadly different task goals but still struggle with fine-grained instruction changes involving the specified object, action, placement, or constraints.
Robo-Dopamine shows strong outcome discrimination but weak instruction grounding, suggesting that its value estimates are anchored more strongly to the demonstrated goal than to the instruction.
One-shot ProcVLM improves both outcome discrimination and instruction grounding, suggesting that task-specific adaptation better aligns procedural knowledge with visual evidence. Fine-grained grounding nevertheless remains difficult.

\noindent\textbf{Can Value Models Accurately Track Execution Progress?} Our results show that high forward-progress correlation does not imply a reliable understanding of progress regression and execution history.
As shown in Table~\ref{tab:robovalue_main}, both TOPReward variants achieve high zero-shot VOC and Memory-VOC but perform poorly on Cycle-VOC. This discrepancy suggests a bias toward temporal progression: their predicted values increase as the video advances even when the visual evidence indicates task regression. Their Qwen3-VL backbones use explicit textual timestamps for video alignment~\citep{bai2025qwen3vl}, providing a possible cue for such a shortcut. These observations are consistent with timestamp dependence, although isolating its contribution requires a controlled timestamp ablation.
Conversely, RoboMeter and RynnValue handle reversals more reliably, consistent with rewind augmentation and signed temporal supervision, respectively. 
Robo-Dopamine 2.0 and adapted ProcVLM also track both execution directions well, in line with their supervision on signed relative progress and procedural stages. 
Yet these models struggle when visually similar states recur, indicating that they can accurately track progress changes without reliably remembering which subgoals have been completed.

\noindent\textbf{Can Value Models Diagnose Failures and Recognize Effective Recovery?}
Reliable feedback requires value models to localize failures, track unresolved errors, and update values according to recovery outcomes. Yet current models struggle to meet these requirements. As shown in Table~\ref{tab:robovalue_main}, mean FPL error is approximately 34\% of the trajectory length, slightly above the 32\% random baseline, which uniformly samples a failure-onset timestamp within each trajectory under the same scoring protocol. TRR scores are also generally low. Among zero-shot models, RoboMeter localizes failures most accurately, whereas RoboFAC achieves the highest overall TRR, with stronger responses on continued-error and failed-recovery branches. These differences show that locating failure onset and tracking its subsequent consequences require distinct abilities.
Robo-Dopamine 2.0 Preview performs comparatively well on both metrics, consistent with its supervision across failure and recovery transitions. However, trajectory-level analyses reveal persistent errors: responses to non-failure events can obscure the actual failure onset, and values may rebound while an error remains unresolved or after an unsuccessful recovery.

\noindent\textbf{Can Value Models Assign Semantically Consistent Values Across Valid Executions?}
Our results reveal that providing stable and informative values and assigning consistent gains to the same semantic subtasks across valid solutions remain challenging.
As shown in Table~\ref{tab:robovalue_main}, TOPReward-8B and RoboMeter provide comparatively stable local feedback and consistent subtask gains across solutions among zero-shot models. Robo-Dopamine 2.0 achieves relatively strong performance in both metrics under one-shot evaluation.
Trajectory-level comparisons further reveal more consistent gains for repeated placements than for semantic subtasks performed in different valid orders. The same subtask can receive different credit depending on its position in the execution sequence. Thus, strong global progress tracking does not guarantee consistent local value assignment across valid solutions.

\FloatBarrier
\subsection{Generalization Results}
\label{sec:oodresults}
Table~\ref{tab:robovalue_ood} reports out-of-domain (OOD) performance, separating cross-embodiment and cross-environment shifts. Both settings preserve the task goals and use the shared evaluation protocol without additional adaptation to the shifted conditions.

\begin{table}[htbp]
    \caption{\textbf{Generalization Results.} Each cell reports cross-embodiment (left) and cross-environment (right) OOD scores. The \RVbest{Best} and \RVsecond{Second Best} scores are marked separately for each shift within each zero-shot or one-shot setting. Metric scales, directions, and abbreviations follow Table~\ref{tab:robovalue_main}.}
    \label{tab:robovalue_ood}
    \centering
    \begingroup

    \footnotesize
    \rmfamily
    \setlength{\tabcolsep}{3pt}
    \renewcommand{\arraystretch}{1.14}

    \settowidth{\dimen0}{\textbf{-00.00}}
    \edef\oodhalfwidth{\the\dimen0}
    \edef\oodslashwidth{\the\dimexpr0.7em\relax}
    \edef\oodmetricwidth{\the\dimexpr2\dimen0+\oodslashwidth\relax}
    \settowidth{\dimen0}{\textbf{Consistency}}
    \ifdim\dimen0>\oodmetricwidth\relax
        \edef\oodmetricwidth{\the\dimen0}
    \fi
    \edef\oodhalfwidth{\the\dimexpr(\oodmetricwidth-\oodslashwidth)/2\relax}

    \def\modeband#1#2{%
        \specialrule{\lightrulewidth}{\aboverulesep}{2pt}
        \noalign{%
            \begingroup
            \color{#1}%
            \hrule height\dimexpr\ht\csname @arstrutbox\endcsname
                +\dp\csname @arstrutbox\endcsname\relax
            \kern-\dimexpr\ht\csname @arstrutbox\endcsname
                +\dp\csname @arstrutbox\endcsname\relax
            \endgroup
        }%
        \multicolumn{9}{@{}c@{}}{\textit{#2}}\\%
        \specialrule{\lightrulewidth}{2pt}{\belowrulesep}
    }

    \def\oodpair#1#2{%
        \makebox[\oodmetricwidth][c]{%
            \makebox[\oodhalfwidth][r]{#1}%
            \makebox[\oodslashwidth][c]{/}%
            \makebox[\oodhalfwidth][l]{#2}%
        }%
    }

    \resizebox{\linewidth}{!}{%
\begin{tabular}{
        @{}
        l
        *{8}{>{\centering\arraybackslash}p{\oodmetricwidth}}
        @{}
    }

        \toprule

        \multirow{2}{*}{\textbf{Model}}
        & \multicolumn{3}{c}{\textcolor{RVUnderstandingInk}{\textbf{Understanding}}}
        & \multicolumn{3}{c}{\textcolor{RVProgressInk}{\textbf{Tracking}}}
        & \textcolor{RVRecoveryInk}{\textbf{Diagnosis}}
        & \textcolor{RVConsistencyInk}{\textbf{Consistency}} \\

        \cmidrule(lr){2-4}
        \cmidrule(lr){5-7}
        \cmidrule(lr){8-8}
        \cmidrule(lr){9-9}

        & \textcolor{RVUnderstandingInk}{SA}
        & \textcolor{RVUnderstandingInk}{TGA-CT}
        & \textcolor{RVUnderstandingInk}{TGA-CF}
        & \textcolor{RVProgressInk}{VOC}
        & \textcolor{RVProgressInk}{Cyc}
        & \textcolor{RVProgressInk}{Mem}
        & \textcolor{RVRecoveryInk}{FPL}
        & \textcolor{RVConsistencyInk}{VS} \\

        \modeband{RVZeroBand}{Zero-Shot Evaluation}

        LIV
        & \oodpair{55.90}{52.61} & \oodpair{17.34}{8.36} & \oodpair{9.40}{10.36} & \oodpair{13.65}{5.52} & \oodpair{13.65}{5.52} & \oodpair{-2.51}{2.73} & \oodpair{\RVsecond{32.13}}{\RVsecond{32.79}} & \oodpair{2.48}{2.84} \\

        ProcVLM-2B
        & \oodpair{66.75}{55.71} & \oodpair{31.26}{17.50} & \oodpair{17.69}{16.05} & \oodpair{53.71}{45.50} & \oodpair{47.06}{40.66} & \oodpair{8.84}{15.31} & \oodpair{37.61}{35.72} & \oodpair{4.31}{3.64} \\

        RoboFAC-7B
        & \oodpair{78.22}{73.15} & \oodpair{6.16}{3.85} & \oodpair{7.20}{3.99} & \oodpair{27.12}{5.99} & \oodpair{36.74}{26.38} & \oodpair{-16.75}{-42.24} & \oodpair{39.28}{39.42} & \oodpair{3.41}{5.55} \\

        RoboMeter-4B
        & \oodpair{\RVbest{85.42}}{\RVsecond{79.14}} & \oodpair{\RVsecond{79.53}}{\RVbest{91.13}} & \oodpair{28.63}{20.33} & \oodpair{\RVbest{86.79}}{81.38} & \oodpair{\RVbest{90.04}}{\RVbest{85.50}} & \oodpair{-7.57}{-6.22} & \oodpair{35.14}{\RVbest{30.38}} & \oodpair{\RVbest{15.48}}{\RVbest{15.91}} \\
\bottomrule

    \end{tabular}%
    }

    \par\smallskip
    \parbox{\linewidth}{\scriptsize\raggedright
        $^{\dagger}$ 2.0 Preview checkpoints.\quad
        $^{\ddagger}$ Qwen3-VL backbones.\quad
        $^{\S}$ High forward correlation coexists with weak reversal tracking; these entries are excluded from best/second-best marking.
    }
    \endgroup
    
\end{table}

\noindent\textbf{Cross-Embodiment Generalization.} Cross-embodiment generalization requires value models to judge task execution from object states and interaction outcomes relative to the task goal, rather than relying on the appearance or motion patterns of a particular robot. As shown in Table~\ref{tab:robovalue_ood}, models largely retain their in-domain strengths, suggesting that task understanding can transfer across robot morphologies and execution patterns. RoboMeter retains strong outcome discrimination, while RynnValue-4B maintains strong instruction grounding. Their transfer is consistent with training across diverse embodiments, coupled with trajectory-preference supervision for RoboMeter and instruction verification for RynnValue. ProcVLM's adapted checkpoint retains strong outcome discrimination but declines in grounding and progress tracking, suggesting that calibration from a single demonstration transfers only partially.

\noindent\textbf{Cross-Environment Generalization.} Cross-environment generalization requires value models to distinguish actual changes in task state from irrelevant visual variation while maintaining accurate recognition of target objects and actions. As shown in Table~\ref{tab:robovalue_ood}, environmental shifts cause broader degradation across models than embodiment shifts, revealing limited robustness to visual variation. ProcVLM and VLAC-8B decline in outcome discrimination and progress tracking despite access to a task demonstration. VLAC-8B's advantage over ProcVLM on history-dependent executions also narrows. These declines suggest that task adaptation in ProcVLM and reference conditioning in VLAC leave judgments tied to the demonstration's visual context. Robo-Dopamine 2.0 better preserves progress tracking and retains the highest value stability, with its tracking advantage over version 1.0 widening after the shift. This resilience is consistent with its explicit training to preserve progress targets under visual perturbations that leave task semantics unchanged, suggesting that robustness benefits from learning which visual changes should not affect value estimates.

\FloatBarrier
\subsection{Fine-Grained Evaluation}
\label{sec:diagnostics}

\subsubsection{Counterfactual Instruction Grounding}
\label{sec:tga_cf_diagnostics}

Table~\ref{tab:tga_cf_breakdown} splits TGA-CF results by which instruction component is changed: object, action, placement, or constraint. For each category, we report how often the original instruction receives strictly higher trajectory gains than the counterfactual version, together with the frequency of tie situations. These statistics distinguish two sources of grounding error: failing to differentiate the instructions and assigning greater gain to the mismatched goal.

\begin{table}[htbp]\centering\caption{\textbf{Fine-Grained TGA-CF Results.}
In-domain strict win rates (\%), with category-level tie rates in parentheses.
Overall requires a win against all four counterfactual variants.
The \RVbest{Best} and \RVsecond{Second Best} win rates are marked per column within each zero-shot or one-shot setting.}
\label{tab:tga_cf_breakdown}
\begingroup
\small
\rmfamily
\setlength{\tabcolsep}{3pt}
\renewcommand{\arraystretch}{1.08}

\begin{tabularx}{\textwidth}{
    @{}
    >{\raggedright\arraybackslash}p{0.25\textwidth}
    *{5}{>{\centering\arraybackslash}X}
    @{}
}
\toprule
\textbf{Model}
& \textbf{Objects}
& \textbf{Actions}
& \textbf{Placement}
& \textbf{Constraints}
& \textbf{Overall} \\
& Win (Tie) & Win (Tie) & Win (Tie) & Win (Tie) & \\
\midrule

\rowcolor{RVZeroBand}
\multicolumn{6}{c}{\textit{Zero-Shot Evaluation}} \\
\midrule

LIV
& 41.94 (11.01) & 41.10 (0.24)
& 46.01 (4.35) & 44.86 (4.09) & 6.17 \\

ProcVLM-2B
& 48.50 (12.02) & 57.11 (8.83)
& 48.30 (12.57) & 48.37 (8.87) & 14.36 \\

RoboFAC-7B
& 36.06 (0.00) & 28.17 (0.00)
& 26.05 (0.00) & 42.09 (0.00) & 4.34 \\

RoboMeter-4B
& 54.31 (0.00) & \RVsecond{84.77} (0.00)
& 71.01 (0.00) & 72.18 (0.00) & 29.29 \\

RoboReward-4B
& 40.22 (55.17) & 30.99 (55.04)
& 33.28 (56.25) & 24.97 (58.55) & 9.12 \\
\bottomrule
\end{tabularx}

\par\smallskip
\parbox{\textwidth}{\small\raggedright
$^{\dagger}$ 2.0 Preview checkpoints.\quad
$^{\ddagger}$ Qwen3-VL backbones.
}
\endgroup
\end{table}

The first pattern is \emph{instruction insensitivity}. Every evaluated Robo-Dopamine checkpoint has tie rates above 50\% in all four categories, indicating that substantial changes to task requirements often leave the predicted trajectory gain unchanged. The observed execution and reference demonstration remain fixed in these comparisons; only the queried instruction changes. This pattern is consistent with value estimates being anchored strongly to the demonstrated trajectory without adequately resolving the goal specified by the instruction, providing a plausible explanation for strong outcome discrimination despite weak instruction grounding.

The second pattern is \emph{responsive but semantically incorrect grounding}. RoboMeter, both TOPReward variants, and one-shot VLAC produce far fewer ties, yet still often assign greater gains to counterfactual instructions. Their predictions do respond to instruction changes, but they do not consistently favor the goal actually satisfied by the execution. Changes to the specified object are particularly challenging: among zero-shot models, the mean win rate is 46.53\% for object changes, compared with 52.52\% for actions, 53.88\% for placement, and 54.48\% for constraints. RoboMeter achieves 84.77\% on action changes but 54.31\% on object changes; TOPReward-8B achieves 79.58\% on constraint changes but 40.15\% on object changes. TOPReward-32B also has its lowest win rate on object changes (53.72\%), despite improving over the 8B variant in this category. RynnValue-8B likewise performs better on placement and constraint changes than on object changes. These results suggest that a recurring source of grounding error is the failure to connect the object specified in the instruction with the object involved in the observed interaction.

Task-specific adaptation improves ProcVLM-2B's discrimination across all four categories. Its overall TGA-CF score rises from 14.36\% to 51.16\%, and the one-shot version produces no ties in these comparisons. The improvement extends from outcome discrimination to recognizing individual task requirements. Nevertheless, the remaining errors show that access to a task demonstration does not fully resolve fine-grained instruction grounding.

\subsubsection{Subtask Identification}
\label{sec:sia_diagnostics}

\begin{table}[htbp]
    \caption{\textbf{Subtask Identification Results.} SIA ($\times100$) under the standard, cross-embodiment, and cross-environment settings; Overall averages the three. The \RVbest{Best} and \RVsecond{Second Best} scores are marked per column within each zero-shot or one-shot setting; ties share the same mark.}
    \label{tab:sia_results}
    \centering
    \begingroup

    \small
    \rmfamily
    \setlength{\tabcolsep}{3pt}
    \renewcommand{\arraystretch}{1.14}

    \begin{tabularx}{\linewidth}{
        @{}l
        *{4}{>{\centering\arraybackslash}X}
        @{}
    }
        \toprule

        \textbf{Model}
        & \textbf{Standard Scenario}
        & \textbf{Cross-Embodiment}
        & \textbf{Cross-Environment}
        & \textbf{Overall} \\

        \midrule
        \rowcolor{RVZeroBand}
        \multicolumn{5}{c}{\textit{Zero-Shot Evaluation}} \\
        \midrule

        ProcVLM-2B
        & \RVbest{23.75}
        & \RVbest{21.34}
        & \RVbest{24.68}
        & \RVbest{23.26} \\

        RoboFAC-7B
        & 9.97 & \RVsecond{10.36} & \RVsecond{9.94} & \RVsecond{10.09} \\

        FailSafe~\citep{lin2025failsafe}
        & \RVsecond{10.01} & 9.62 & \RVsecond{9.94} & 9.85 \\

        \midrule
        \rowcolor{RVOneBand}
        \multicolumn{5}{c}{\textit{One-Shot Evaluation}} \\
        \midrule

        ProcVLM-2B
        & \RVbest{24.66}
        & \RVbest{19.74}
        & \RVbest{19.15}
        & \RVbest{21.19} \\

        \bottomrule
    \end{tabularx}

    \endgroup
\end{table}

SIA examines whether models can associate the observed execution with the semantic subtask currently being performed. Alongside ProcVLM and RoboFAC, we evaluate the subtask descriptions generated by FailSafe, a model designed for failure reasoning and recovery~\citep{lin2025failsafe}. Table~\ref{tab:sia_results} shows that zero-shot ProcVLM outperforms RoboFAC and FailSafe under all three evaluation settings. However, the generally low scores indicate that reliable identification of the active subtask remains difficult for the evaluated textual interfaces.

For ProcVLM, one-shot adaptation slightly improves in-domain SIA but reduces performance under both out-of-domain shifts. This contrasts with its gains in other capabilities: improved scalar value estimates do not necessarily translate into more accurate descriptions of the current subtask.

\subsubsection{Failure and Recovery Reasoning}
\label{sec:failure_diagnostics}

Figure~\ref{fig:trr_heatmaps} separates TRR into stage-level responses and complete-branch judgments. In both simulation and real-world evaluations, RoboFAC more consistently predicts value declines during failures than RoboMeter and achieves higher pass rates on the continued-error (CE) and failed-recovery (FR) branches. RoboMeter, however, more consistently predicts increases when recovery succeeds. The two models thus differ in their ability to recognize adverse events and recovery progress.

For several models, the main difficulty in failed recovery is assessing its outcome. TOPReward-8B and Robo-Dopamine 2.0 Preview often increase their values during corrective attempts while failing to decrease them when those attempts fail. This gap indicates that recognizing activity directed toward recovery does not ensure recognition of whether the error has been resolved. Complete-branch pass rates are low because a branch passes only when all required phase-wise trends are correct, revealing how rarely models maintain a consistent response pattern across the full branch.

\begin{figure}[htbp]
    \centering
    \includegraphics[
        width=0.98\linewidth,
        trim=24bp 18bp 14bp 54bp,
        clip,
        keepaspectratio
    ]{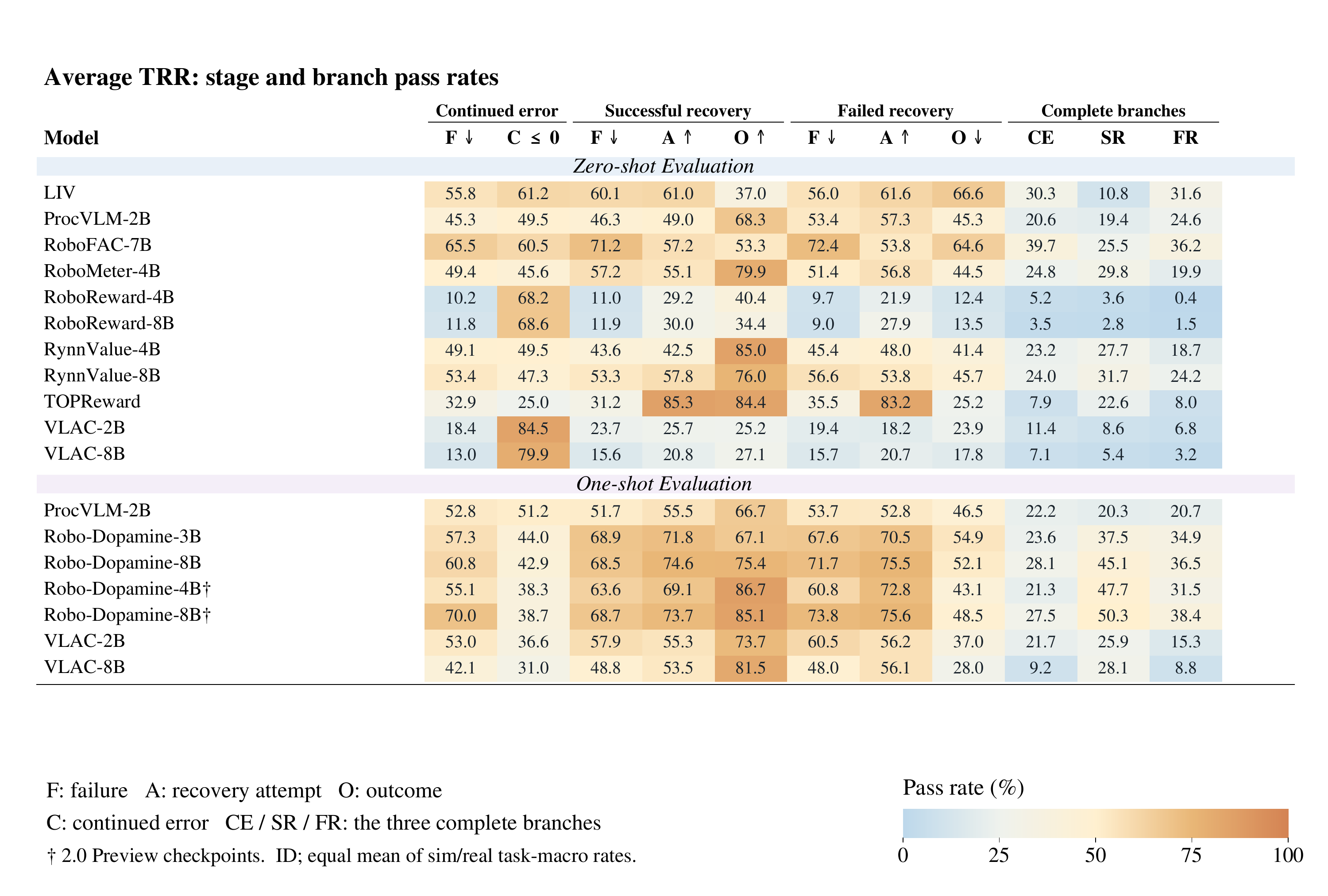}

    \caption{\textbf{Stage- and Branch-Level TRR Pass Rates.}
    Pass rates averaged across simulation and real-world evaluations.
    Stage columns test individual value trends; a complete branch passes only if all required stage trends are correct. TOPReward denotes the 8B variant.}
    \label{fig:trr_heatmaps}
\end{figure}

The FPL example in Figure~\ref{fig:fpl_trr_cases} illustrates why responding to failure is insufficient for precise localization. RoboFAC's value decreases near the annotated onset, but a larger decline occurs earlier during otherwise normal execution. Selecting the largest detected ZigZag decline therefore identifies that earlier event, giving a normalized error of 45.0\%. RoboMeter and Robo-Dopamine 2.0-8B Preview place their largest declines closer to the true onset, with errors of 9.9\% and 2.5\%, respectively. This example shows how unrelated value fluctuations can obscure detectable failure responses.

The lower panel shows a separate failed-recovery execution. RoboFAC follows the required decrease–increase–decrease sequence across the trajectory. RoboMeter, instead, has positive phase trends during both the initial failure and the failed outcome. Robo-Dopamine 2.0-8B Preview responds correctly to the initial failure and corrective attempt, but its value rebounds after a brief decline within the failed-outcome phase, yielding an overall positive trend for that stage. These examples reveal two distinct failure modes: some models respond in the wrong direction when an error occurs, while others respond appropriately only transiently before their values rebound despite the error remaining unresolved. The latter further shows that recovery assessment must extend beyond the onset of a correction to its eventual outcome.

\begin{figure}[tp]
    \centering
    \includegraphics[
        width=0.98\linewidth,
        keepaspectratio
    ]{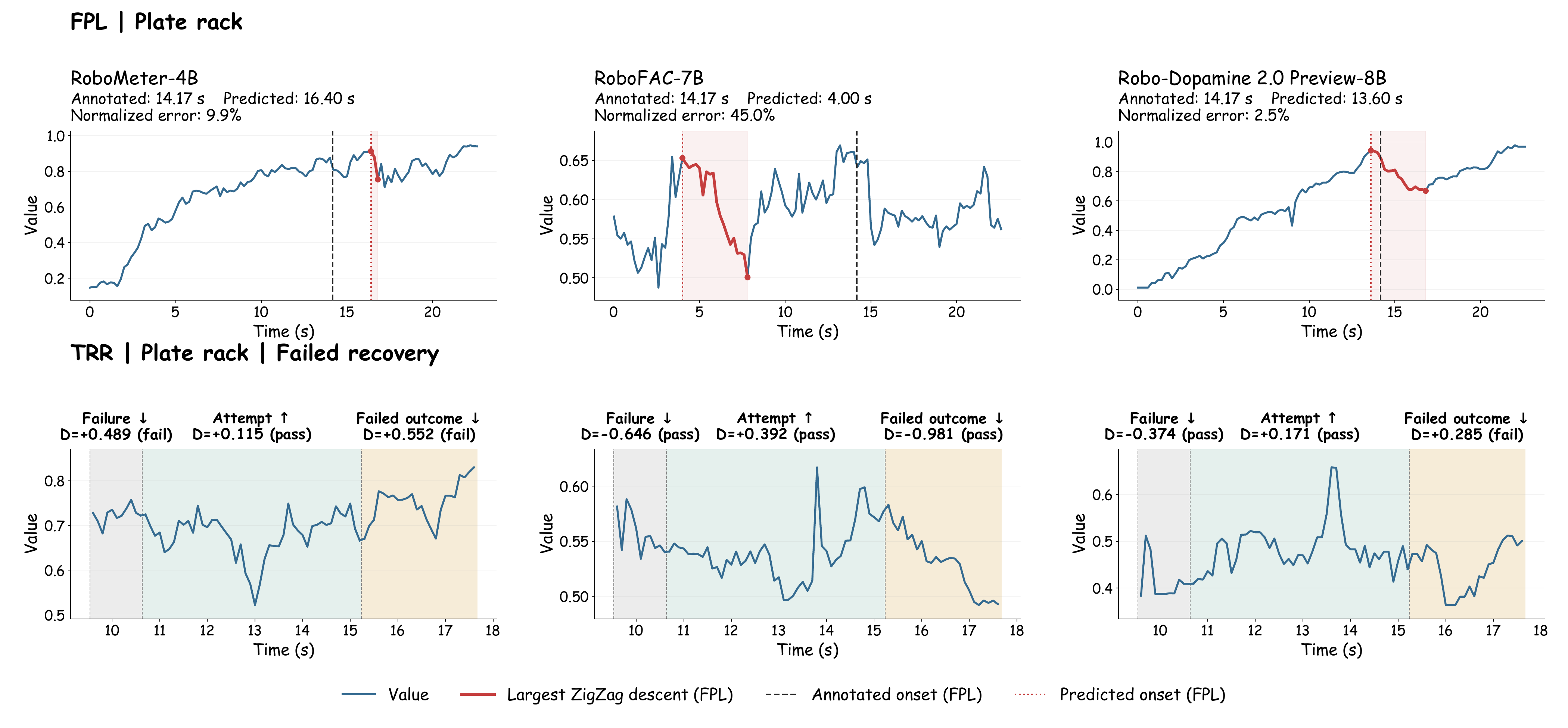}

    \caption{\textbf{Failure Localization and Recovery Assessment.}
    Top: FPL on an effective-recovery execution containing an annotated failure. Red segments mark the largest ZigZag declines; black dashed and red dotted lines mark annotated and predicted onsets.
    Bottom: TRR on a separate failed-recovery execution, with required trend directions and measured $D$ values shown for each shaded stage.
    Values retain each model's native scale.}
    \label{fig:fpl_trr_cases}
\end{figure}

\subsubsection{Value Consistency Across Valid Solutions}
\label{sec:consistency_diagnostics}

\begin{figure}[tp]
    \centering
    \includegraphics[
        width=0.98\linewidth,
        keepaspectratio
    ]{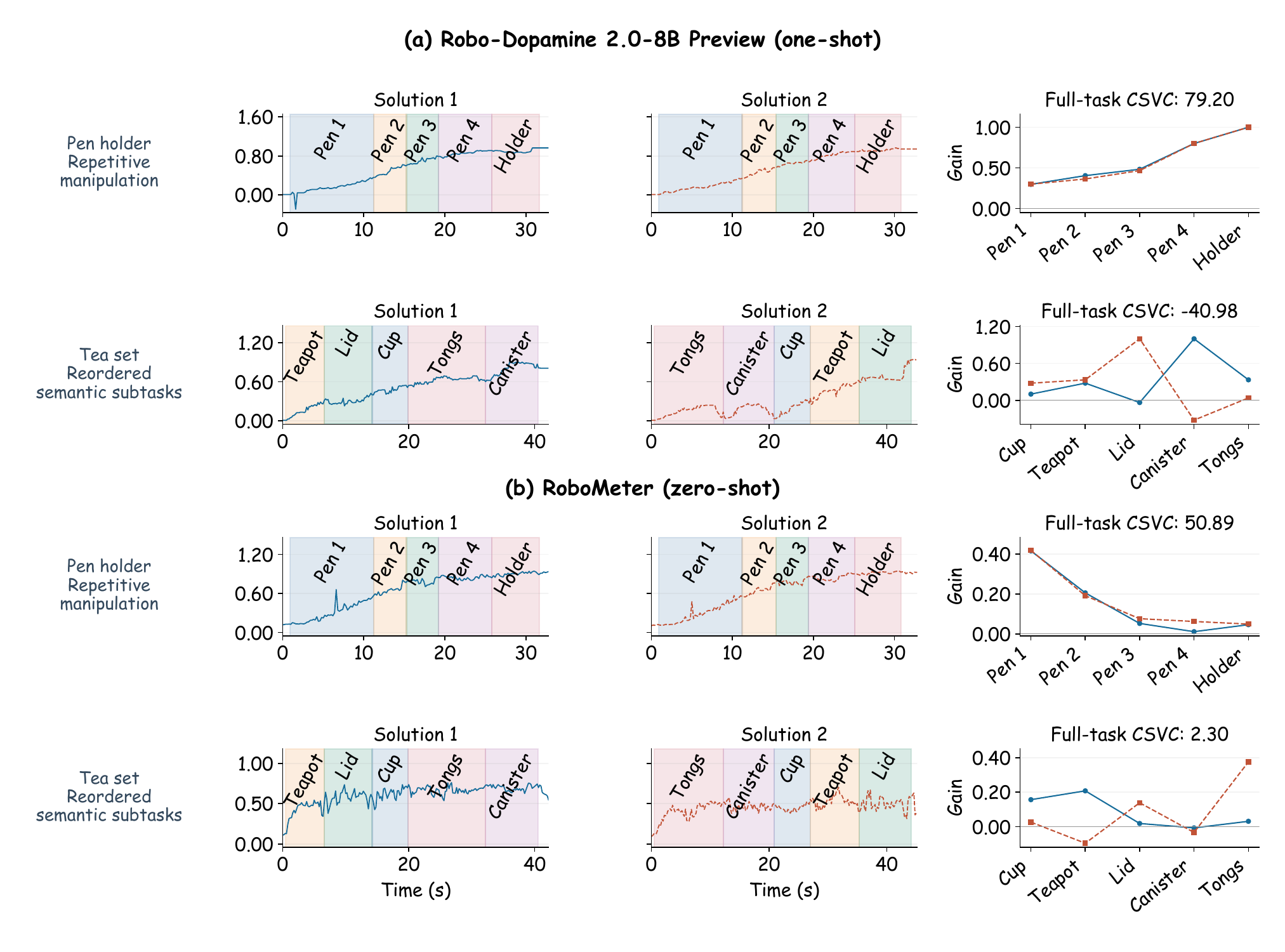}

    \caption{\textbf{Value Consistency Across Valid Solutions.}
    Each row compares two valid executions. Left and middle: value curves with semantic subtasks shaded. Right: gains aligned by subtask identity.
    CSVC scores are computed over all eligible executions of each task.}
    \label{fig:csvc_cases}
\end{figure}

Figure~\ref{fig:csvc_cases} contrasts value consistency in a repetitive pen-holder task and a tea-set task with alternative valid subtask orders. In the illustrated pen-holder executions, Robo-Dopamine 2.0-8B Preview assigns larger gains to later placements, whereas RoboMeter assigns larger gains to earlier ones, revealing different temporal credit-allocation profiles across models. Despite these differences, each model produces similar gain profiles across solutions that vary which object is placed at each step and where it is placed, while preserving the same underlying structure of repeated placement operations. The tea-set task provides a stronger test by varying the order of semantically distinct subtasks.

The tea-set executions expose a different pattern. For Robo-Dopamine 2.0-8B Preview, moving cup placement from the first to the last position substantially increases its assigned gain, while moving canister placement in the opposite direction reduces it. RoboMeter shows different shifts, but it also changes the credit assigned to the same subtasks across valid orders. Their full-task CSVC scores fall to $-40.98$ and $2.30$, respectively.

Together, these examples reveal strong order dependence in local value gains: the same semantic subtask may receive substantially different credit when executed earlier or later within another valid solution. Such variation matters when value gains are used to select or weight training segments, because semantically matched segments may be treated differently solely due to their execution context and position.

\FloatBarrier
\section{Analysis}
\label{sec:analysis}

We examine the capability gaps in Section~\ref{sec:experiment} through input interfaces, supervision, and model capacity.
The comparisons identify design priorities for reliable value feedback, while differences in training data and model configurations limit attribution to individual components.

\subsection{How Do Input Interfaces Relate to Value Reliability?}
\label{sec:interfaces}

\noindent\textbf{Temporal Context.}
When similar observations recur, estimating progress requires evidence of earlier task-relevant events~\citep{zhang2026recurrent,chen2026rmbench,zhang2026progresscompass}.
LIV's single-frame interface provides no direct record of those events.
RoboMeter and RynnValue receive sampled observation histories and achieve strong VOC and Cycle-VOC, yet their near-zero or negative Memory-VOC reveals difficulty tracking accumulated progress (Table~\ref{tab:robovalue_main}).
Sampled context can omit brief interactions, and models may fail to integrate events that remain visible into their current estimates.
These results motivate context construction and memory mechanisms that preserve evidence of completed subtasks and make it available for later judgments.

\noindent\textbf{Camera Views.}
Robo-Dopamine combines front and wrist views, providing complementary observations of the scene and local object interactions.
Its strong progress scores nevertheless coexist with weak instruction grounding and limited history-dependent tracking (Table~\ref{tab:robovalue_main}).
This capability profile highlights the need to connect visible interactions with their task-specific meaning: broader visual coverage alone does not establish which requirements have been fulfilled.

\noindent\textbf{Reference Demonstrations.}
Task demonstrations enter the evaluated models through distinct mechanisms: ProcVLM uses task-specific parameter adaptation, whereas VLAC conditions its predictions on reference observations (Appendix~\ref{app:adapter_implementations}).
Both achieve higher SA and VOC in the one-shot setting (Table~\ref{tab:robovalue_main}).
However, ProcVLM's gains in scalar progress estimation are not accompanied by consistent improvements in subtask identification (Table~\ref{tab:sia_results}).
Scalar progress and subtask recognition thus provide complementary checks on the benefit of task-specific demonstrations.

\subsection{Which Distinctions Should Supervision Capture?}
\label{sec:semantics}

The results motivate supervision that specifies when value judgments should change and when their task-level meaning should remain consistent.

\noindent\textbf{Instruction Grounding.}
RoboMeter and RynnValue use instruction-mismatch supervision~\citep{liang2026robometer,huang2026rynnvalue} and distinguish broadly different tasks more reliably than fine-grained counterfactual instructions (Table~\ref{tab:robovalue_main}).
The tie-rate analysis in Table~\ref{tab:tga_cf_breakdown} further separates failing to react to an instruction change from reacting in the wrong semantic direction.
Training examples that hold execution and reference evidence fixed while varying a specific task requirement can directly supervise this distinction.
Supervision can then reward the instruction that matches the observed behavior, discouraging reliance on visual similarity to a demonstration when the requested goal differs.

\noindent\textbf{Execution History and Recovery.}
RoboMeter's rewind augmentation and RynnValue's temporal-order perturbations with signed temporal targets provide supervision for changes in temporal direction~\citep{liang2026robometer,huang2026rynnvalue}.
Their strong Cycle-VOC scores indicate sensitivity to the direction of change in our constructed forward--reverse sequences.
Low Memory-VOC and the recovery-stage errors in Figure~\ref{fig:trr_heatmaps} expose further distinctions that temporal reversal alone does not capture.
Pairing visually similar states with different completed subgoals would directly target ambiguity in accumulated progress.
Failure-and-recovery examples could additionally supervise which completed conditions remain valid and whether corrective actions actually restore the task state.
The rebounds after failed corrections in Figure~\ref{fig:fpl_trr_cases} illustrate why these targets need to remain tied to the unresolved task conditions.

\noindent\textbf{Semantic Consistency.}
Robo-Dopamine 2.0 Preview better preserves progress tracking under environmental shifts than the original models (Table~\ref{tab:robovalue_ood}).
The 2.0 formulation assigns unchanged progress targets to visual perturbations that preserve task semantics~\citep{xu2026robo}.
Learning this invariance offers a plausible explanation for the observed robustness.
Consistency across valid execution paths requires a related form of alignment at the subtask level.
In Figure~\ref{fig:csvc_cases}, the gain assigned to a semantic subtask changes with its position in an otherwise valid execution.
Cross-solution supervision could align gains for equivalent subtask completions across admissible orders, directly addressing this dependence of local credit on execution position.

\subsection{What Do Model Size and Backbone Choice Explain?}
\label{sec:scaling}

\noindent\textbf{Model Size.}
The benefits of larger checkpoints vary across capabilities.
In zero-shot evaluation, VLAC-8B improves instruction grounding and progress tracking over VLAC-2B, but its TRR decreases (Table~\ref{tab:robovalue_main}).
Increasing RynnValue from 4B to 8B yields only a small improvement in Cycle-VOC, while both TGA variants and CSVC decline.
TOPReward-32B improves Cycle-VOC and TRR over its 8B counterpart, but achieves lower SA, VS, and CSVC.
Model size is therefore an incomplete predictor of which execution judgments will improve.

\noindent\textbf{Shared Backbone.}
RoboMeter-4B and RoboReward-4B share a Qwen3-VL-4B backbone, yet RoboMeter performs substantially better on SA and Cycle-VOC (Table~\ref{tab:robovalue_main}).
RoboMeter combines frame-level progress supervision with trajectory preferences, allowing failed executions to contribute training signals without dense progress annotations~\citep{liang2026robometer}.
RoboReward instead learns five-level completion ratings for video clips~\citep{lee2026roboreward}.
These objectives provide different constraints on intermediate predictions; the models also differ in training data, temporal inputs, and score construction.
Their contrasting results show that backbone identity alone does not determine the execution properties captured by the learned values.
These differences suggest that supervision coverage and prediction granularity deserve attention alongside model capacity.

\section{Conclusion}
\label{sec:conclusion}
We presented \textbf{RoboValue}, a fine-grained sim-and-real benchmark for unified evaluation of robotic value models. 
Strong outcome discrimination and forward-progress correlation do not ensure accurate instruction grounding or progress tracking when similar states recur. 
Failure localization and recovery assessment also remain challenging. 
Models struggle to provide stable, informative feedback within trajectories and consistent subtask gains across valid execution orders. 
Models generally retain their in-domain strengths under embodiment shifts, whereas environment shifts lead to broader performance degradation.
RoboValue-Dataset and the public leaderboard provide a common basis for diagnosing model limitations and assessing future advances. 
An important next step is to develop more general and reliable value models that provide informative supervision for downstream policy improvement.

\endgroup

\begingroup
\setlength{\bibsep}{6pt plus 2pt minus 1pt}
\bibliography{iclr2027_conference}
\bibliographystyle{iclr2027_conference}
\endgroup

\clearpage

\phantomsection
\addcontentsline{toc}{section}{Appendix}

\addtocontents{toc}{\protect\setcounter{tocdepth}{2}}

\let\RoboValueOriginalAddContentsLine\addcontentsline

\renewcommand{\addcontentsline}[3]{%
  \ifstrequal{#1}{toc}{%
    \RoboValueOriginalAddContentsLine{apc}{#2}{#3}%
    \ifstrequal{#2}{section}{%
      \RoboValueOriginalAddContentsLine{toc}{subsection}{#3}%
    }{%
      \ifstrequal{#2}{subsection}{%
        \RoboValueOriginalAddContentsLine{toc}{subsubsection}{#3}%
      }{%
        \RoboValueOriginalAddContentsLine{#1}{#2}{#3}%
      }%
    }%
  }{%
    \RoboValueOriginalAddContentsLine{#1}{#2}{#3}%
  }%
}


\definecolor{RVStatsSTBand}{RGB}{234,241,248}
\definecolor{RVStatsFRTBand}{RGB}{240,244,237}
\providecolor{RVZeroBand}{RGB}{232,240,248}
\providecolor{RVOneBand}{RGB}{248,238,225}
\colorlet{RVCFNew}{agiurl}
\definecolor{RVCFOld}{HTML}{454545}
\definecolor{RVCFRule}{HTML}{D2D2D2}
\definecolor{RVCFHeader}{HTML}{F1EEF5}
\newcommand{\RVCFDelta}[2]{%
  {\color{RVCFOld}#1}\allowbreak\ {\color{RVCFOld}$\rightarrow$}\allowbreak\
  {\color{RVCFNew}#2}%
}
\newcommand{\RVCFRow}[6]{%
  \textbf{#1}\par #2 &
  \textbf{Objects:} #3\par
  \textbf{Actions:} #4\par
  \textbf{Placement:} #5\par
  \textbf{Constraints:} #6\\
}
\newcommand{\RVCFSeparator}{%
  \arrayrulecolor{RVCFRule}\midrule\arrayrulecolor{black}%
}

\colorlet{RVCompareStripe}{agiblue!6}
\colorlet{RVTrajectoryYes}{agiblue}
\newcommand{\RVTrajectoryCheck}{\textcolor{RVTrajectoryYes}{\faCheck}}

\clearpage
\appendix
\raggedbottom
\setlength{\parskip}{4pt}

\newcommand{\RVSkillPanel}[2]{%
  \includegraphics[width=0.26\linewidth,viewport=#2,clip]{#1}%
}
\newcommand{\RVSkillGrid}[1]{%
  \includegraphics[width=\dimexpr0.78\linewidth+10pt\relax,
    viewport=50 844 1750 913,clip]{#1}\par\vspace{5pt}%
  \begingroup
  \setlength{\tabcolsep}{0pt}%
  \renewcommand{\arraystretch}{0}%
  \begin{tabular}{@{}c@{\hspace{5pt}}c@{\hspace{5pt}}c@{}}
    \RVSkillPanel{#1}{52 600 382 831} & \RVSkillPanel{#1}{393 600 723 831} & \RVSkillPanel{#1}{734 600 1064 831} \\[4pt]
    \RVSkillPanel{#1}{1076 600 1406 831} & \RVSkillPanel{#1}{1417 600 1747 831} & \RVSkillPanel{#1}{52 340 382 571} \\[4pt]
    \RVSkillPanel{#1}{393 340 723 571} & \RVSkillPanel{#1}{734 340 1064 571} & \RVSkillPanel{#1}{1076 340 1406 571} \\[4pt]
    \RVSkillPanel{#1}{1417 340 1747 571} & \RVSkillPanel{#1}{52 81 382 312} & \RVSkillPanel{#1}{393 81 723 312} \\[4pt]
    \RVSkillPanel{#1}{734 81 1064 312} & \RVSkillPanel{#1}{1076 81 1406 312} & \RVSkillPanel{#1}{1417 81 1747 312}
  \end{tabular}%
  \endgroup
}

\makeatletter
\let\ftype@figure\ftype@table
\let\@fb@topbarrier\@empty
\setlength{\@fptop}{0pt}
\setlength{\@dblfptop}{0pt}
\setlength{\@fpsep}{12pt plus 2pt minus 2pt}
\setlength{\@dblfpsep}{12pt plus 2pt minus 2pt}
\setlength{\@fpbot}{0pt plus 1fil}
\setlength{\@dblfpbot}{0pt plus 1fil}
\makeatother
\setcounter{topnumber}{6}
\setcounter{dbltopnumber}{6}
\setcounter{bottomnumber}{6}
\setcounter{totalnumber}{8}
\setlength{\floatsep}{12pt plus 2pt minus 2pt}
\setlength{\dblfloatsep}{12pt plus 2pt minus 2pt}
\setlength{\textfloatsep}{14pt plus 2pt minus 2pt}
\setlength{\intextsep}{7pt plus 1pt minus 1pt}
\renewcommand{\topfraction}{0.97}
\renewcommand{\dbltopfraction}{0.97}
\renewcommand{\bottomfraction}{0.97}
\renewcommand{\floatpagefraction}{0.75}
\renewcommand{\dblfloatpagefraction}{0.75}
\renewcommand{\textfraction}{0.03}
\suppressfloats[t]

\begingroup
\normalsize\color{black}
\hypersetup{linkcolor=black,linktoc=all}
\setlength{\parskip}{0pt}
\setcounter{tocdepth}{3}
\renewcommand{\contentsname}{Appendix Contents}
\makeatletter
\@cftmaketoctitle
\@starttoc{apc}
\makeatother
\endgroup

\clearpage

\section{Trajectory Types for Metric Evaluation}
\label{app:metric_trajectory_types}

Table~\ref{tab:metric-trajectory-design} links the evaluation metrics to the trajectory categories introduced in Section~\ref{sec:trajectory_design}. Expert demonstrations provide complete successful executions for assessing task understanding and progress. Failure and recovery trajectories supply contrasting outcomes for SA and distinct error and recovery stages for TRR. FPL uses Error Continuation and Effective Recovery trajectories, each containing a single execution error. Long-horizon temporal trajectories expose ambiguity that requires execution history. Multi-solution trajectories contribute to SA and SIA and support comparisons of subtask value gains across valid execution orders for CSVC.

\begin{table}[htbp]
  \centering
  \caption{\textbf{Trajectory Types Used by Each Evaluation Metric.} A purple check (\RVTrajectoryCheck) marks a trajectory type used by the metric; unused types are left blank.}
  \label{tab:metric-trajectory-design}
  \begingroup
  \normalsize
  \setlength{\tabcolsep}{2.5pt}
  \renewcommand{\arraystretch}{1.45}
  \newcommand{\RVTypeHeader}[1]{{\fontsize{9}{10.5}\selectfont\bfseries\makecell[c]{#1}}}
  \begin{tabularx}{\linewidth}{@{}l*{6}{>{\centering\arraybackslash}X}@{}}
    \toprule
    \multirow{3}{*}{\textbf{Metric}}
    & \multirow{3}{*}{\RVTypeHeader{Expert\\Demonstrations}}
    & \multicolumn{3}{c}{\RVTypeHeader{Failure and Recovery Trajectories}}
    & \multirow{3}{*}{\RVTypeHeader{Long-Horizon\\Temporal\\Trajectories}}
    & \multirow{3}{*}{\RVTypeHeader{Multi-Solution\\Trajectories}} \\
    \cmidrule(lr){3-5}
    & & \RVTypeHeader{Error\\Continuation}
      & \RVTypeHeader{Effective\\Recovery}
      & \RVTypeHeader{Ineffective\\Recovery} & & \\
    \midrule
    \rowcolor{RVCompareStripe}
    SA         & \RVTrajectoryCheck & \RVTrajectoryCheck & \RVTrajectoryCheck & \RVTrajectoryCheck & & \RVTrajectoryCheck \\
    TGA-CT     & \RVTrajectoryCheck & & & & & \\
    \rowcolor{RVCompareStripe}
    TGA-CF     & \RVTrajectoryCheck & & & & & \\
    SIA        & \RVTrajectoryCheck & & & & & \RVTrajectoryCheck \\
    \midrule
    \rowcolor{RVCompareStripe}
    VOC        & \RVTrajectoryCheck & & & & & \\
    Cycle-VOC  & \RVTrajectoryCheck & & & & & \\
    \rowcolor{RVCompareStripe}
    Memory-VOC & & & & & \RVTrajectoryCheck & \\
    \midrule
    FPL        & & \RVTrajectoryCheck & \RVTrajectoryCheck & & & \\
    \rowcolor{RVCompareStripe}
    TRR        & & \RVTrajectoryCheck & \RVTrajectoryCheck & \RVTrajectoryCheck & & \\
    \midrule
    VS         & \RVTrajectoryCheck & & & & & \\
    \rowcolor{RVCompareStripe}
    CSVC       & \RVTrajectoryCheck & & & & & \RVTrajectoryCheck \\
    \bottomrule
  \end{tabularx}
  \endgroup
\end{table}

\section{Comparison with Existing Benchmarks}
\label{app:benchmark_comparison}

We compare RoboValue with existing robotic value-model benchmarks and evaluation suites in Tables~\ref{tab:existing_value_benchmarks} and~\ref{tab:benchmark_test_design}. The comparison highlights RoboValue's \textbf{controlled execution contrasts}: trajectories with the same task goal and failure scenario branch into error continuation, effective recovery, and ineffective recovery. Its \textbf{cross-solution comparisons} test whether the same semantic subtask receives consistent value gains across valid execution orders. These designs distinguish responses to changes in task attainment from variation caused by the choice of a valid solution. Alongside instruction interventions, history-dependent tasks, and progress reversals, they support fine-grained diagnosis under a shared sim-and-real protocol. RoboValue further evaluates \textbf{embodiment and environment shifts separately} while retaining task goals, allowing performance changes to be traced to each shifted factor. Together, these contrasts expose why value feedback can become unreliable despite strong aggregate outcome or forward-progress scores.

\begin{table}[htbp]
  \centering
  \caption{\textbf{Data Coverage and Diagnostic Test Design.} Purple checks mark data domains and diagnostic tests reported in each benchmark; blank cells indicate unreported coverage.}
  \label{tab:existing_value_benchmarks}
  \begingroup
  \fontsize{8.5}{10}\selectfont
  \setlength{\tabcolsep}{2.5pt}
  \renewcommand{\arraystretch}{1.05}
  \newcommand{\RVCompareHead}[1]{{\fontsize{8}{9.2}\selectfont\bfseries\makecell[c]{#1}}}
\begin{tabularx}{\linewidth}{@{}>{\RaggedRight\arraybackslash}p{163pt}*{7}{>{\centering\arraybackslash}X}@{}}
    \toprule
    \multirow{2}{*}{\textbf{Benchmark}}
    & \multicolumn{2}{c}{\textbf{Data Domain}}
    & \multicolumn{5}{c}{\textbf{Diagnostic Tests}} \\
    \cmidrule(lr){2-3}\cmidrule(lr){4-8}
    & \RVCompareHead{Sim.} & \RVCompareHead{Real}
    & \RVCompareHead{Goal\\Changes}
    & \RVCompareHead{History}
    & \RVCompareHead{Progress\\Decline}
    & \RVCompareHead{Recovery\\Contrasts}
    & \RVCompareHead{Valid\\Orders} \\
    \midrule
    ARS suite~\citep{hu2026ars} & \RVTrajectoryCheck &  & \RVTrajectoryCheck &  &  &  &  \\
    \rowcolor{RVCompareStripe}
    ContextProgress-Bench~\citep{zhang2026progresscompass} & \RVTrajectoryCheck &  & \RVTrajectoryCheck & \RVTrajectoryCheck &  &  &  \\
    Eval-Actions~\citep{liu2026evalactions} &  & \RVTrajectoryCheck &  &  &  &  &  \\
    \rowcolor{RVCompareStripe}
    FailBench~\citep{navasardyan2026failbench} & \RVTrajectoryCheck & \RVTrajectoryCheck &  &  &  &  &  \\
    ManiRewardBench~\citep{chen2026topreward} &  & \RVTrajectoryCheck & \RVTrajectoryCheck &  &  &  &  \\
    \rowcolor{RVCompareStripe}
    OpenGVL~\citep{budzianowski2025opengvl} &  & \RVTrajectoryCheck &  &  &  &  &  \\
\bottomrule
  \end{tabularx}
  \par\smallskip
  \begin{minipage}{\linewidth}
    \fontsize{8}{9.5}\selectfont\raggedright
    Goal Changes pair fixed visual evidence with different task requirements. History covers history-dependent progress tests or temporal-context ablations for progress estimation. Progress Decline denotes dedicated reversed-state or declining-progress tests. Recovery Contrasts compare error continuation, effective recovery, and ineffective recovery for the same failure scenario. Valid Orders compare subtask value gains across admissible execution orders. Blank cells indicate that the specified evaluator test is not reported; training augmentations and downstream policy trials do not qualify.\par
    \textsuperscript{*} Related process and failure question-answering benchmarks.
  \end{minipage}
  \endgroup
\end{table}

\begin{table}[htbp]
  \centering
  \caption{\textbf{Comparison of Generalization Protocols.} Purple checks mark separate tests of embodiment or environment changes that retain task goals.}
  \label{tab:benchmark_test_design}
  \begingroup
  \fontsize{8.5}{10}\selectfont
  \setlength{\tabcolsep}{4pt}
  \renewcommand{\arraystretch}{1.08}
  \begin{tabularx}{\linewidth}{@{}>{\RaggedRight\arraybackslash}p{116pt}*{2}{>{\centering\arraybackslash}p{61pt}}>{\RaggedRight\arraybackslash}X@{}}
    \toprule
    \multirow{2}{*}{\textbf{Benchmark}}
    & \multicolumn{2}{c}{\textbf{Separate Shift Tests}}
    & \multirow{2}{*}{\textbf{Evaluation Protocol}} \\
    \cmidrule(lr){2-3}
    & \textbf{Embodiment} & \textbf{Environment} & \\
    \midrule
    Eval-Actions & \RVTrajectoryCheck &
    & Unseen-arm-only and unseen-task-only splits are evaluated separately. \\
    \rowcolor{RVCompareStripe}
    FailBench & &
    & Cross-source evaluation spans different tasks, robots, and scenes. \\
    PRIMO Bench & &
    & Held-out tasks; real-humanoid transfer changes the robot, tasks, and environment together. \\
    \rowcolor{RVCompareStripe}
    ProcVQA & &
    & Unseen RoboTwin tasks form the OOD split; ID data come from real-robot datasets. \\
    Progress-Bench & &
    & Query observations use the same or a different camera view from the demonstration. \\
    \rowcolor{RVCompareStripe}
    RBM-EVAL & &
    & New collection sites, embodiments, and viewpoints occur in a combined OOD set. \\
    ReWiND suite & &
    & Unseen tasks are evaluated in simulation and the real world. \\
    \rowcolor{RVCompareStripe}
    Robo-Dopamine 2.0 suite & & \RVTrajectoryCheck\textsuperscript{$\dagger$}
    & Background and distractor edits preserve task semantics; additional OOD tests alter execution validity. \\
    RoboFAC\textsuperscript{*} & &
    & A simulation-trained evaluator is tested on simulation and real-robot videos. \\
    \midrule
    \rowcolor{RVCompareStripe}
    \textbf{RoboValue (ours)} & \RVTrajectoryCheck & \RVTrajectoryCheck
    & Task goals are retained while embodiment and environment vary separately in simulation and the real world. \\
    \bottomrule
  \end{tabularx}
  \par\smallskip
  \begin{minipage}{\linewidth}
    \fontsize{8.5}{10}\selectfont\raggedright
    Combined domain shifts and viewpoint-only comparisons are described in the final column rather than marked as separate embodiment or environment tests.\par
    \textsuperscript{$\dagger$} Environment variation is introduced through synthetic scene edits.
    \textsuperscript{*} Related process and failure question-answering benchmark.
  \end{minipage}
  \endgroup
\end{table}

\FloatBarrier
\section{Simulation Task Details}
\label{app:simulation_tasks}

\subsection{Manipulation Skill Coverage}
\label{app:simulation_skill_coverage}

\begin{figure}[htbp]
  \centering
  \RVSkillGrid{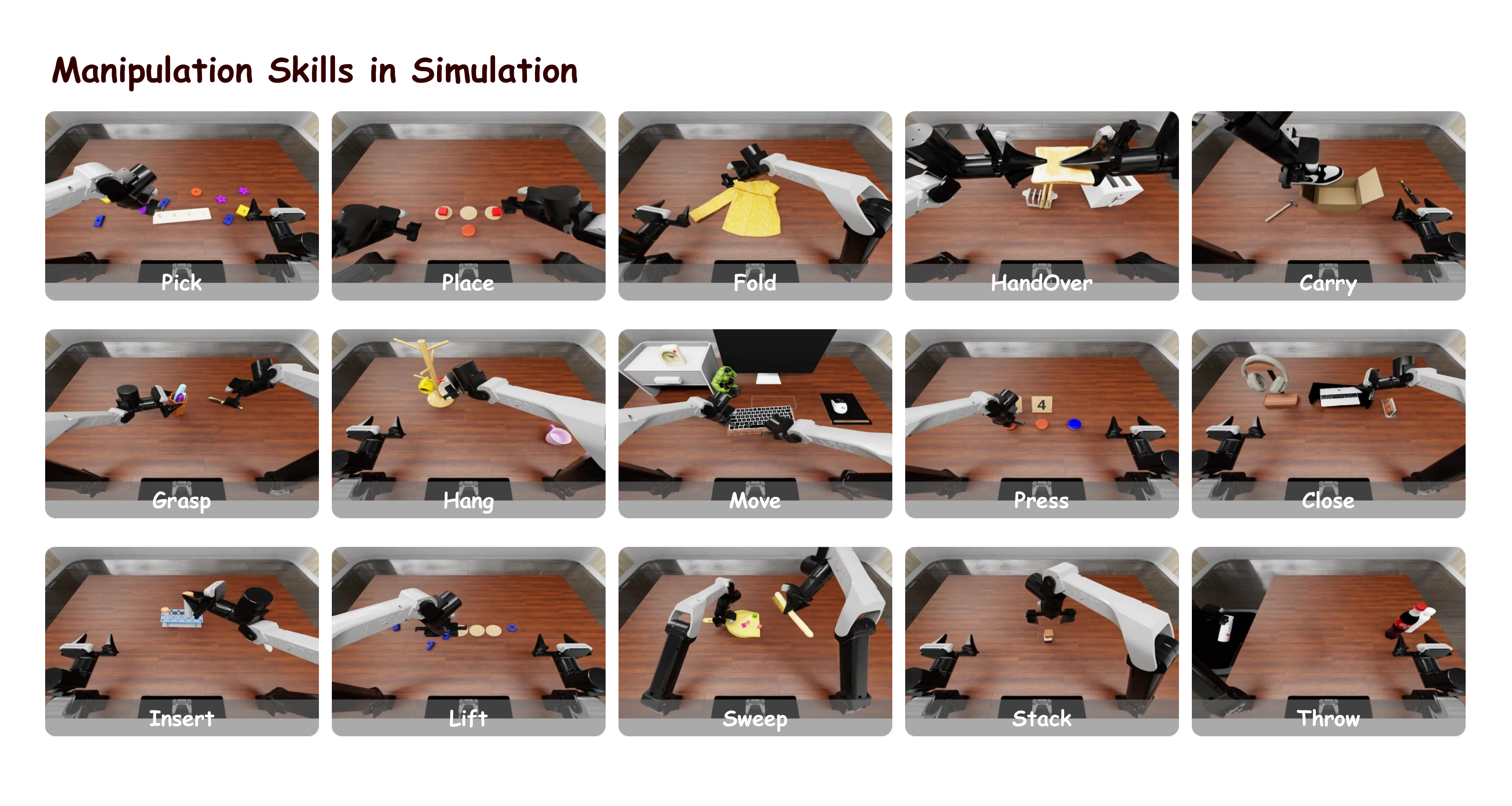}
  \caption{\textbf{Manipulation Skill Coverage in Simulation.} Fifteen representative skills, including object transport, placement, insertion, stacking, folding, and tool use.}
  \label{fig:sim_skill}
\end{figure}

The simulation tasks combine 15 representative manipulation skills, shown in Figure~\ref{fig:sim_skill}. They range from object transport and arrangement to insertion, cloth folding, and sweeping with tools; handovers and coordinated holding also require cooperation between the two arms. Composing these skills into multi-stage tasks introduces varied spatial relations, contact conditions, and ordering constraints. Value models must therefore distinguish meaningful changes toward the task goal across different forms of interaction.

\FloatBarrier
\subsection{Task Descriptions and Subtask Decomposition}
\label{app:simulation_task_descriptions}

The task cards describe all 15 simulation tasks. Each card pairs the original instruction with the initial scene, an intended execution procedure, and a decomposition into semantic subtasks. Three representative frames show the standard in-domain, cross-embodiment, and cross-environment settings, making the changes in robot platform and visual context explicit. The subtask descriptions provide a reference for temporal annotation and subtask-level evaluation. For tasks with multiple valid solutions, the illustrated procedure represents one admissible ordering rather than a unique required sequence.

\RoboValuePkgDEEHDUEHDIDTEEENDXDRDFEBEFDRCPEHEFEIDTDVEMAppendixLayoutBegin
\def\RoboValuePkgDEEHDUEHDIDTEEENDXDRDFEBEFDRCPEHEFEIDTDVEMIncludeRoot#1{RoboValue_Sim_Overleaf/#1}
\RoboValuePkgDEEHDUEHDIDTEEENDXDRDFEBEFDRCPEHEFEIDTDVEMSetFigureDir{\RoboValuePkgDEEHDUEHDIDTEEENDXDRDFEBEFDRCPEHEFEIDTDVEMIncludeRoot{figures}}

\RoboValuePkgDEEHDUEHDIDTEEENDXDRDFEBEFDRCPEHEFEIDTDVEMTaskCard
  {Arrange Largest Number}
  {arrange_largest_number}
  {Arrange the numbers from left to right to form the largest possible number, and place them on the pad.}
  {\textbf{Scene:} Four number tiles lie separately on the table, with four target pads arranged in a row. \textbf{Procedure:} Determine the descending numerical order that forms the largest possible number and place the tiles on the pads. \textbf{Order and variations:} Place the tiles from left to right in descending numerical order.}
  {\textbf{S1} Place the largest number on the leftmost mat. \textbf{S2} Place the largest number outside the mats on the second mat from the left. \textbf{S3} Place the largest number outside the mats on the third mat from the left. \textbf{S4} Place the remaining number on the fourth mat from the left.}

\RoboValuePkgDEEHDUEHDIDTEEENDXDRDFEBEFDRCPEHEFEIDTDVEMTaskCard
  {Fill Pen Holder}
  {fill_pen_holder}
  {Pick up the pen holder, place all the pens into it one by one, then put the filled holder back on the table.}
  {\textbf{Scene:} An empty pen holder stands on the table, with four pens lying separately nearby. \textbf{Procedure:} Pick up the holder with one hand and use the other hand to insert all four pens individually, then return the filled holder to the table. \textbf{Order and variations:} The pens may be inserted in any order. Picking up the holder precedes insertion, and putting it down follows all four insertions.}
  {\textbf{S1} Hold the empty pen holder. \textbf{S2} Insert the first selected pen into the pen holder. \textbf{S3} Insert the second selected pen into the pen holder. \textbf{S4} Insert the third selected pen into the pen holder. \textbf{S5} Insert the remaining pen into the pen holder. \textbf{S6} Place the filled pen holder upright on the table.}

\RoboValuePkgDEEHDUEHDIDTEEENDXDRDFEBEFDRCPEHEFEIDTDVEMTaskCard
  {Fold Clothes}
  {fold_clothes}
  {Fold the left sleeve inward toward the opposite chest point, fold the right sleeve inward toward the opposite chest point, then fold the lower hem upward until it is aligned with the shoulders.}
  {\textbf{Scene:} An unfolded clothes lies flat on the table, with both sleeves extended. \textbf{Procedure:} Fold both sleeves across the body of the clothes, then fold the lower section upward over them. \textbf{Order and variations:} Fold the left sleeve first, the right sleeve second, and the lower hem last.}
  {\textbf{S1} Fold the left sleeve inward. \textbf{S2} Fold the right sleeve inward. \textbf{S3} Fold the lower half of the garment over the upper half.}


\RoboValuePkgDEEHDUEHDIDTEEENDXDRDFEBEFDRCPEHEFEIDTDVEMTaskCard
  {Hang Mugs}
  {hang_mugs}
  {Hang all three mugs on the mug rack.}
  {\textbf{Scene:} Three mugs are placed separately on the table beside an empty mug rack. \textbf{Procedure:} Pick up and hang all three mugs on the rack, using an empty hook for each mug. \textbf{Order and variations:} The mugs may be hung in any order.}
  {\textbf{S1} Hang the first cup on an empty hook. \textbf{S2} Hang the second cup on another empty hook. \textbf{S3} Hang the third cup on the remaining empty hook.}

\RoboValuePkgDEEHDUEHDIDTEEENDXDRDFEBEFDRCPEHEFEIDTDVEMTaskCard
  {Insert Tubes}
  {insert_tubes}
  {Insert the three tubes into the rack one by one.}
  {\textbf{Scene:} Three tubes lie separately on the table in front of an empty tube rack. \textbf{Procedure:} Pick up and insert each tube into an empty rack slot until all three tubes are in the rack. \textbf{Order and variations:} Insert the tubes one by one in any order.}
  {\textbf{S1} Insert the first tube into an empty slot. \textbf{S2} Insert the second tube into another empty slot. \textbf{S3} Insert the third tube into the remaining empty slot.}

\RoboValuePkgDEEHDUEHDIDTEEENDXDRDFEBEFDRCPEHEFEIDTDVEMTaskCard
  {Make Toast}
  {make_toast}
  {Pick up two bread slices one by one, place them into the toaster, and press the lever down.}
  {\textbf{Scene:} Bread slices are placed beside a toaster with two empty slots on the table. \textbf{Procedure:} Pick up two slices one by one, insert them into the toaster's two slots, and press the lever down. \textbf{Order and variations:} Both slices must be inserted before the lever is pressed.}
  {\textbf{S1} Insert the first bread slice into one toaster slot. \textbf{S2} Insert the second bread slice into the remaining toaster slot. \textbf{S3} Press the toaster lever down.}


\RoboValuePkgDEEHDUEHDIDTEEENDXDRDFEBEFDRCPEHEFEIDTDVEMTaskCard
  {Organize Table}
  {organize_table}
  {Place the alarm clock on the drawer, put the figurine on its stand, place the mouse on the mouse pad, and push the keyboard into the frame.}
  {\textbf{Scene:} An alarm clock, a figurine, a mouse, and a keyboard are placed on the table, outside their respective destinations: a drawer top, a figurine stand, a mouse pad, and a keyboard frame. \textbf{Procedure:} Place the alarm clock on the drawer top, the figurine on its stand, and the mouse on its pad, and push the keyboard into its frame. \textbf{Order and variations:} The four placements have no required global order.}
  {\textbf{S1} Place the alarm clock on top of the drawer. \textbf{S2} Place the figurine on the stand. \textbf{S3} Place the mouse on the mouse pad. \textbf{S4} Insert the keyboard into the frame.}

\RoboValuePkgDEEHDUEHDIDTEEENDXDRDFEBEFDRCPEHEFEIDTDVEMTaskCard
  {Press By Number}
  {press_by_number}
  {Press the two red buttons the required number of times according to the number cards, then press the blue button to confirm.}
  {\textbf{Scene:} Two number cards lie above two corresponding red buttons on the table, with a blue confirmation button beside them. \textbf{Procedure:} Read the two cards and press each red button the corresponding number of times, then press the blue button to confirm. \textbf{Order and variations:} The required press counts vary with the cards. Complete the red-button presses before pressing the blue confirmation button.}
  {\textbf{S1} Press the first button the required number of times. \textbf{S2} Press the blue button. \textbf{S3} Press the second button the required number of times. \textbf{S4} Press the blue button again.}

\RoboValuePkgDEEHDUEHDIDTEEENDXDRDFEBEFDRCPEHEFEIDTDVEMTaskCard
  {Put Bottles Into Dustbin}
  {put_bottles_into_dustbin}
  {Pick up all four bottles one by one and throw them into the dustbin.}
  {\textbf{Scene:} Four bottles are placed on the table, with a dustbin beside the table. \textbf{Procedure:} Pick up and deposit all four bottles in the dustbin. \textbf{Order and variations:} Handle the bottles one by one, order is not specified.}
  {\textbf{S1} Place the first bottle in the dustbin. \textbf{S2} Place the second bottle in the dustbin. \textbf{S3} Place the third bottle in the dustbin. \textbf{S4} Place the last bottle in the dustbin.}


\RoboValuePkgDEEHDUEHDIDTEEENDXDRDFEBEFDRCPEHEFEIDTDVEMTaskCard
  {Pack Objects Into Box}
  {pack_objects_into_box}
  {Move the box to the center of the table, then place all the objects into the box.}
  {\textbf{Scene:} An open box and four separate objects are placed on the table, with the objects initially outside the box. \textbf{Procedure:} Clear the center of the table and move the box there, then pick up and place all four objects inside. \textbf{Order and variations:} The objects may be packed in any order after the box is positioned.}
  {\textbf{S1} Clear the center of the table, then place the box in the center of the table. \textbf{S2} Place the shoe into the box. \textbf{S3} Place the hammer into the box. \textbf{S4} Place the toothbrush into the box. \textbf{S5} Place the toy car into the box.}

\RoboValuePkgDEEHDUEHDIDTEEENDXDRDFEBEFDRCPEHEFEIDTDVEMTaskCard
  {Play Stacking Toy}
  {play_stacking_toy}
  {Place the four matching pieces onto the longest peg one by one, place the three matching pieces onto the second-longest peg, place the two matching pieces onto the second-shortest peg, then place the remaining piece onto the shortest peg.}
  {\textbf{Scene:} A stacking-toy base with four pegs of different lengths lies on the table. Ten matching pieces lie separately around it, with four, three, two, and one piece for the respective pegs. \textbf{Procedure:} Load the longest peg with its four matching pieces, the second-longest with three, the second-shortest with two, and the shortest with the remaining piece. \textbf{Order and variations:} Follow the peg order from longest to shortest and place the pieces individually.}
  {\textbf{S1} Place the four matching pieces on the longest peg. \textbf{S2} Place the three matching pieces on the second-longest peg. \textbf{S3} Place the two matching pieces on the second-shortest peg. \textbf{S4} Place the remaining piece on the shortest peg.}

\RoboValuePkgDEEHDUEHDIDTEEENDXDRDFEBEFDRCPEHEFEIDTDVEMTaskCard
  {Stack Blocks}
  {stack_blocks}
  {Stack the three blocks with different textures into one stable pile.}
  {\textbf{Scene:} Three blocks with different textures are placed separately on the table. \textbf{Procedure:} Place a base block at the table center, stack a second block on it, and place the third on top to form one stable pile. \textbf{Order and variations:} The blocks may be selected in any order. Each supporting block must be in place before the next block is stacked on it.}
  {\textbf{S1} Place the first block in the center of the table. \textbf{S2} Place the second block on the first block. \textbf{S3} Place the third block on the second block.}


\RoboValuePkgDEEHDUEHDIDTEEENDXDRDFEBEFDRCPEHEFEIDTDVEMTaskCard
  {Store Laptop And Headphones}
  {store_laptop_and_headphone}
  {Hang the headphones on the headphone stand, close the laptop, then place the closed laptop into the vertical laptop stand.}
  {\textbf{Scene:} An open laptop rests on a laptop stand. Headphones, an empty headphone stand, and an empty vertical laptop stand are placed nearby. \textbf{Procedure:} Hang the headphones on their stand, close the laptop, lift it from its initial stand, and insert it into the vertical laptop stand. \textbf{Order and variations:} Hang the headphones first, close the laptop second, and place the closed laptop in the vertical stand last.}
  {\textbf{S1} Place the headphones on the headphone stand. \textbf{S2} Close the laptop. \textbf{S3} Place the closed laptop in the laptop stand.}

\RoboValuePkgDEEHDUEHDIDTEEENDXDRDFEBEFDRCPEHEFEIDTDVEMTaskCard
  {Swap Blocks}
  {swap_blocks}
  {Swap the two blocks using the empty mat, pressing the button after each move.}
  {\textbf{Scene:} Three mats are arranged in a row on the table. Two hold one block each, the third is empty, and a button lies nearby. \textbf{Procedure:} Move one block to the empty mat, move the other block to the newly vacant mat, and move the first block to the second block's original mat, so that the two blocks exchange positions. \textbf{Order and variations:} Press the button after every block move.}
  {\textbf{S1} Put the first block on the empty mat. \textbf{S2} Press the button to confirm the first move. \textbf{S3} Transfer the second block onto the original first block's position. \textbf{S4} Press the button to confirm the second move. \textbf{S5} Put the first block on the original second block's mat. \textbf{S6} Press the button to confirm the last move.}

\RoboValuePkgDEEHDUEHDIDTEEENDXDRDFEBEFDRCPEHEFEIDTDVEMTaskCard
  {Sweep Blocks}
  {sweep_blocks}
  {Pick up the broom, pick up and place the dustpan upright on the table, then use the broom to sweep all the objects into the dustpan.}
  {\textbf{Scene:} A broom and an empty dustpan lie on the table beside several scattered blocks. \textbf{Procedure:} Pick up the broom, pick up and position the dustpan upright on the table, and sweep all blocks into it. \textbf{Order and variations:} Pick up the broom first and position the dustpan before sweeping.}
  {\textbf{S1} Pick up the broom. \textbf{S2} Pick up the dustpan. \textbf{S3} Sweep all objects into the dustpan.}

\RoboValuePkgDEEHDUEHDIDTEEENDXDRDFEBEFDRCPEHEFEIDTDVEMAppendixLayoutEnd


\subsection{Original and Counterfactual Instructions}
\label{app:simulation_instruction}

For TGA-CF, each included simulation task has an original instruction and four counterfactual variants that modify its requirements. We evaluate all five instructions against the same successful execution, testing whether the model assigns greater value gain to the instruction that matches the observed behavior. Holding the visual input fixed isolates the model's sensitivity to these semantic changes. The instruction table covers 13 simulation tasks; the two tasks dedicated to Memory-VOC have no counterfactual variants and are excluded.

Table~\ref{tab:simulation_counterfactual_instructions} pairs each original instruction with four changes to objects, actions, placement, or constraints. Arrows show the original requirement and its replacement, with color highlighting the change. These labels identify the primary semantic change; evaluation uses the complete modified instruction.

\begin{table}[htbp]
  \centering
  \setlength{\belowcaptionskip}{4pt}
  \caption{\textbf{Original and Counterfactual Instructions for Simulation Tasks.} Gray text gives the original requirement; purple text after each arrow gives its counterfactual replacement.}
  \label{tab:simulation_counterfactual_instructions}
  \begingroup
  \fontsize{8}{9}\selectfont
  \setlength{\tabcolsep}{3pt}
  \setlength{\parskip}{0pt}
  \setlength{\aboverulesep}{1.3pt}
  \setlength{\belowrulesep}{1.3pt}
  \renewcommand{\arraystretch}{1}
  \begin{tabularx}{\linewidth}{@{}
    >{\hsize=0.88\hsize\linewidth=\hsize\RaggedRight\arraybackslash}X
    >{\hsize=1.12\hsize\linewidth=\hsize\RaggedRight\arraybackslash}X@{}}
    \toprule
    \rowcolor{RVCFHeader}
    \textbf{Task and Original Instruction} & \textbf{Counterfactual Changes} \\
    \midrule
    \RVCFRow
      {Arrange Largest Number}
      {Arrange the numbers from left to right to form the largest possible number, and place them on the pad.}
      {\RVCFDelta{numbers}{books}}
      {\RVCFDelta{arrange to form largest number}{turn face down and arrange left to right}}
      {\RVCFDelta{on pad}{beside pad}}
      {\RVCFDelta{largest}{smallest}}
    \RVCFSeparator
    \RVCFRow
      {Fill Pen Holder}
      {Pick up the pen holder, place all the pens into it one by one, then put the filled holder back on the table.}
      {\RVCFDelta{pens}{spoons}}
      {\RVCFDelta{return holder to table}{return holder to table upside down}}
      {\RVCFDelta{inside holder}{beside holder}}
      {\RVCFDelta{all pens}{exactly three pens}}
    \RVCFSeparator
    \RVCFRow
      {Fold Clothes}
      {Fold the left sleeve inward toward the opposite chest point, fold the right sleeve inward toward the opposite chest point, then fold the lower hem upward until it is aligned with the shoulders.}
      {\RVCFDelta{clothing: sleeves/chest/hem}{tablecloth: sides/center/edges}}
      {\RVCFDelta{three folds}{leave completely unfolded on table}}
      {\RVCFDelta{hem upward until aligned with shoulders}{hem downward away from shoulders}}
      {\RVCFDelta{both sleeves and hem folded}{exactly one sleeve folded; rest unfolded}}
    \RVCFSeparator
    \RVCFRow
      {Hang Mugs}
      {Hang all three mugs on the mug rack.}
      {\RVCFDelta{mugs}{plates}}
      {\RVCFDelta{hang on rack}{stand upright on table}}
      {\RVCFDelta{hang on rack}{place beside rack}}
      {\RVCFDelta{all three}{exactly two; one on table}}
    \RVCFSeparator
    \RVCFRow
      {Insert Tubes}
      {Insert three tubes into the rack one by one.}
      {\RVCFDelta{tubes}{blocks}}
      {\RVCFDelta{insert}{rotate each tube once, then insert}}
      {\RVCFDelta{insert into rack}{place beside rack}}
      {\RVCFDelta{three}{exactly two; one on table}}
    \RVCFSeparator
    \RVCFRow
      {Make Toast}
      {Pick up two bread slices one by one, place them into the toaster, and press the lever down.}
      {\RVCFDelta{bread slices}{wooden coasters}}
      {\RVCFDelta{press lever down}{leave lever up}}
      {\RVCFDelta{into toaster}{beside toaster}}
      {\RVCFDelta{two slices}{exactly one}}
    \RVCFSeparator
    \RVCFRow
      {Organize Table}
      {Place the alarm clock on the drawer, put the figurine on its stand, place the mouse on the mouse pad, and push the keyboard into the frame.}
      {\RVCFDelta{figurine on its stand; mouse on mouse pad}{figurine on mouse pad; mouse on stand}}
      {\RVCFDelta{push keyboard into frame}{leave keyboard outside frame}}
      {\RVCFDelta{mouse on pad}{beside pad}}
      {\RVCFDelta{all four}{exactly one; three stay initial}}
    \RVCFSeparator
    \RVCFRow
      {Pack Objects Into Box}
      {Move the box to the center of the table, then place all the objects into the box.}
      {\RVCFDelta{objects}{pillows}}
      {\RVCFDelta{place objects in box}{place in box, then remove to table}}
      {\RVCFDelta{inside box}{beside box}}
      {\RVCFDelta{all objects}{exactly three; one outside}}
    \RVCFSeparator
    \RVCFRow
      {Play Stacking Toy}
      {Place the four matching pieces onto the longest peg one by one, place the three matching pieces onto the second-longest peg, place the two matching pieces onto the second-shortest peg, then place the remaining piece onto the shortest peg.}
      {\RVCFDelta{matching pieces}{matching balls}}
      {\RVCFDelta{place on pegs}{rotate in place; leave on table}}
      {\RVCFDelta{longest to shortest pegs}{shortest to longest pegs}}
      {\RVCFDelta{per-peg counts 4/3/2/1}{per-peg counts 3/2/1/0; four remain on table}}
    \RVCFSeparator
    \RVCFRow
      {Put Bottles Into Dustbin}
      {Pick up all four bottles one by one and throw them into the dustbin.}
      {\RVCFDelta{bottles}{books}}
      {\RVCFDelta{throw into bin}{stand upright on table}}
      {\RVCFDelta{throw into dustbin}{place beside dustbin}}
      {\RVCFDelta{all four}{exactly three; one on table}}
    \RVCFSeparator
    \RVCFRow
      {Stack Blocks}
      {Stack the three blocks with different textures into one stable pile.}
      {\RVCFDelta{blocks}{plates}}
      {\RVCFDelta{stack in pile}{rotate in place; no stack}}
      {\RVCFDelta{stack in one pile}{arrange side by side on table}}
      {\RVCFDelta{one 3-block pile}{one 2-block stack + one single block}}
    \RVCFSeparator
    \RVCFRow
      {Store Laptop And Headphones}
      {Hang the headphones on the headphone stand, close the laptop, then place the closed laptop into the vertical laptop stand.}
      {\RVCFDelta{headphones}{scarf}}
      {\RVCFDelta{close laptop}{leave open}}
      {\RVCFDelta{into laptop stand}{beside laptop stand}}
      {\RVCFDelta{headphones/laptop on their own stands}{both on headphone stand; laptop stand empty}}
    \RVCFSeparator
    \RVCFRow
      {Sweep Blocks}
      {Pick up the broom, pick up and place the dustpan upright on the table, then use the broom to sweep all the objects into the dustpan.}
      {\RVCFDelta{objects}{balls}}
      {\RVCFDelta{sweep into dustpan}{scatter away from dustpan}}
      {\RVCFDelta{into dustpan}{beside dustpan}}
      {\RVCFDelta{sweep all objects into dustpan}{sweep exactly one object into dustpan}}
    \bottomrule
  \end{tabularx}
  \endgroup
\end{table}

\subsection{Out-of-Domain Scenario Construction}
\label{app:simulation_ood}

We construct two OOD settings by changing the robot embodiment or the environment separately while preserving the task goals. Both settings follow the shared evaluation protocol without additional adaptation to the shifted conditions.

\noindent\textbf{Cross-Embodiment Generalization.}
ARX serves as the standard embodiment, and a dual-arm UR5e setup provides the embodiment shift. Executing the same tasks with a different robot changes the visible morphology and execution patterns. This setting tests whether value judgments transfer across platforms while remaining grounded in task-relevant object interactions and outcomes.

\noindent\textbf{Cross-Environment Generalization.}
We vary surface appearance, clutter, and lighting while retaining the standard embodiment. Tabletop and floor appearances are sampled from a pool of 500 material variants, up to 20 task-irrelevant objects are added, and one of 13 lighting configurations is selected. Random combinations of these factors produce varied visual contexts for the same tasks. This setting tests whether models continue to track task-relevant state changes despite variations in background appearance and distractor objects.

\Needspace{6\baselineskip}
\subsection{Task Execution Visualizations}
  \label{app:simulation_visualizations}
  
\begin{figure}[htbp]
  \centering
  \includegraphics[width=0.80\linewidth]{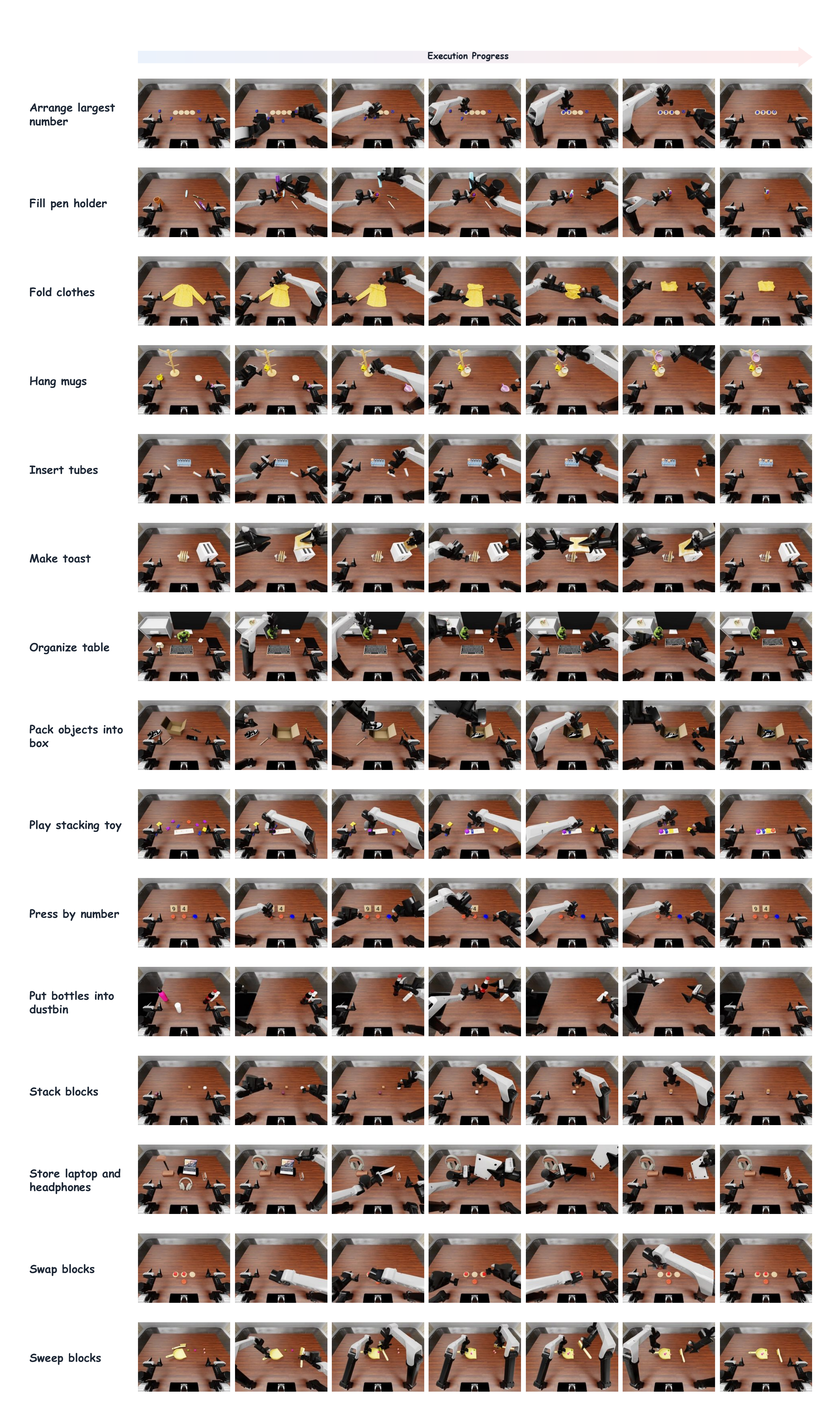}
  \caption{\textbf{Simulation Task Executions.} The simulation tasks are shown in seven chronologically ordered frames per row, from initial scene to successful outcome.}
  \label{fig:sim_7frames}
\end{figure}

  Figure~\ref{fig:sim_7frames} shows successful executions of all 15 simulation tasks, with seven chronologically ordered frames per task tracing the initial scene, intermediate manipulation stages, and completed outcome.

\FloatBarrier

\section{Real-World Task Details}
\label{app:realworld_tasks}

\subsection{Manipulation Skill Coverage}
\label{app:realworld_skill_coverage}

\begin{figure}[htbp]
  \centering
  \RVSkillGrid{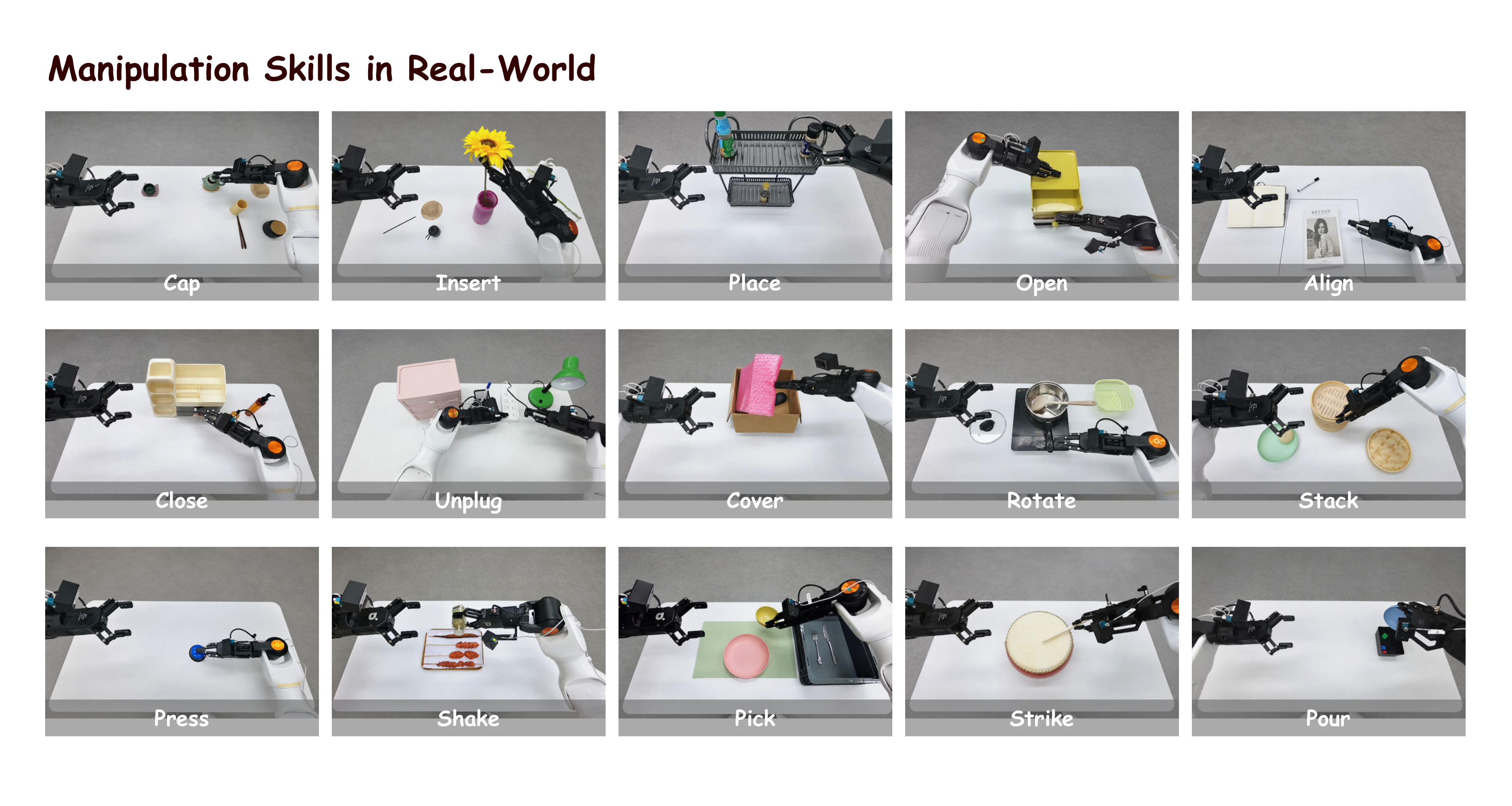}
  \caption{\textbf{Manipulation Skill Coverage in the Real World.} Fifteen representative skills, including object arrangement, container manipulation, pouring, and striking.}
  \label{fig:real_skill}
\end{figure}

Figure~\ref{fig:real_skill} illustrates 15 representative skills covered by the real-world tasks. Alongside picking and placing, the tasks involve precise alignment, capping, opening and closing containers, unplugging, and transferring objects by pouring. Repeated pressing, shaking, and striking introduce actions whose contribution to completion depends on the preceding execution. This range spans changes in object position, configuration, and accumulated task progress under physical sensing and contact conditions.

\FloatBarrier
\subsection{Task Descriptions and Subtask Decomposition}

\label{app:realworld_task_descriptions}
The task cards document all 20 real-world tasks through their original instructions, initial scenes, intended execution procedures, and semantic subtask decompositions. Each card includes a representative frame from the standard in-domain, cross-embodiment, and cross-environment settings. Together, the text and images specify the requested outcome and the intermediate interactions that contribute to it, providing a reference for trajectory annotation and subtask-level evaluation. For tasks that admit alternative valid orders, the listed procedure gives one admissible execution consistent with the task dependencies.

\RoboValuePkgDEEHDUEHDIDTEEENDXDRDEDXDTEEDRCPEHEFEIDTDVEMAppendixLayoutBegin
\def\RoboValuePkgDEEHDUEHDIDTEEENDXDRDEDXDTEEDRCPEHEFEIDTDVEMIncludeRoot#1{RoboValue_Real_Overleaf4/#1}
\RoboValuePkgDEEHDUEHDIDTEEENDXDRDEDXDTEEDRCPEHEFEIDTDVEMSetFigureDir{\RoboValuePkgDEEHDUEHDIDTEEENDXDRDEDXDTEEDRCPEHEFEIDTDVEMIncludeRoot{figures}}

\RoboValuePkgDEEHDUEHDIDTEEENDXDRDEDXDTEEDRCPEHEFEIDTDVEMTaskCard
  {Apply Dish Soap Repeatedly}
  {apply_dish_soap_repeatedly}
  {Press the dish-soap bottle 5 times continuously, dispensing soap onto the dishcloth.}
  {\textbf{Scene:} A dish-soap dispenser sits at the table center, with a dishcloth lying to its right. \textbf{Procedure:} Hold the cloth beneath the outlet with the right arm while the left arm presses and releases the pump, then return the cloth to the dispenser's right.}
  {\textbf{S1} Place the dishcloth below the pump outlet. \textbf{S2} Press the pump for the first time. \textbf{S3} Press the pump for the second time. \textbf{S4} Press the pump for the third time. \textbf{S5} Press the pump for the fourth time. \textbf{S6} Press the pump for the fifth time. \textbf{S7} Place the dishcloth on the table.}

\RoboValuePkgDEEHDUEHDIDTEEENDXDRDEDXDTEEDRCPEHEFEIDTDVEMTaskCard
  {Arrange Family Tea Set}
  {arrange_family_tea_set}
  {Place the teacup on the cat-head mat, the teapot on the yellow mat, the lid on the teapot, the tongs in their holder, and the canister on the round mat. Complete the four placement units in any order; within the teapot-and-lid unit, place the teapot first and seat its lid immediately afterward.}
  {\textbf{Scene:} A cup, a teapot, and a tea canister sit off their designated coasters on the table. The teapot lid lies beside the teapot, and tea tongs lie outside their holder. \textbf{Procedure:} Place the cup on the left-front coaster, the teapot on the right-front teapot coaster, the lid on the teapot, the tea canister on the tea-canister coaster, and the tea tongs in the bamboo holder tool position.}
  {\textbf{S1} Place the cup on the left-front coaster. \textbf{S2} Place the teapot on the right-front teapot coaster. \textbf{S3} Place the lid on the teapot. \textbf{S4} Place the tea canister on the tea-canister coaster. \textbf{S5} Place the tea tongs in the bamboo holder.}

\RoboValuePkgDEEHDUEHDIDTEEENDXDRDEDXDTEEDRCPEHEFEIDTDVEMTaskCard
  {Arrange Flowers and Reed Diffuser}
  {arrange_flowers_and_reed_diffuser}
  {Place the pink, white, and yellow flowers consecutively in the vase, in any order. Place the diffuser bottle on the coaster and insert the reed stick into it, in either order. Either diffuser action may occur before or after the flower sequence, but neither may occur between flower placements.}
  {\textbf{Scene:} A sunflower, a lily, and a rose lie outside a vase on the table. A reed diffuser bottle sits off its coaster, with a diffuser reed placed separately beside it. \textbf{Procedure:} Insert the three flowers into the vase one after another, place the reed diffuser bottle on the coaster, then insert the diffuser reed into the bottle.}
  {\textbf{S1} Insert the first flower into the vase. \textbf{S2} Insert the second flower into the vase. \textbf{S3} Insert the third flower into the vase. \textbf{S4} Place the reed diffuser bottle on the coaster. \textbf{S5} Insert the diffuser reed into the reed diffuser bottle.}

\RoboValuePkgDEEHDUEHDIDTEEENDXDRDEDXDTEEDRCPEHEFEIDTDVEMTaskCard
  {Arrange Seasoning Jars}
  {arrange_seasoning_jars}
  {Place the three seasoning jars on the two-tier rack in any order, on either tier and in any positions.}
  {\textbf{Scene:} A two-tier seasoning rack and three seasoning jars are placed on the table. \textbf{Procedure:} Pick up and place each of the three seasoning jars on the rack.}
  {\textbf{S1} Place the first seasoning jar on the two-tier seasoning rack. \textbf{S2} Place the second seasoning jar on the two-tier seasoning rack. \textbf{S3} Place the third seasoning jar on the two-tier seasoning rack.}

\RoboValuePkgDEEHDUEHDIDTEEENDXDRDEDXDTEEDRCPEHEFEIDTDVEMTaskCard
  {Enter Keypad Number Sequence}
  {enter_keypad_number_sequence}
  {Press the keys in the sequence 2-8-2-6-8, fully releasing each key and returning to the waiting position in front of the keypad between presses.}
  {\textbf{Scene:} A keyboard is placed on the right side of the table. The operating arm starts at its waiting position. \textbf{Procedure:} Press and fully release each required key, returning to the waiting position after every press, including the last.}
  {\textbf{S1} Press 2. \textbf{S2} Press 8. \textbf{S3} Press 2. \textbf{S4} Press 6. \textbf{S5} Press 8.}

\RoboValuePkgDEEHDUEHDIDTEEENDXDRDEDXDTEEDRCPEHEFEIDTDVEMTaskCard
  {Fill And Place Pen Holder}
  {fill_and_place_pen_holder}
  {Hold the pen holder with its opening upward, insert all four pens one by one in any order, then place the filled holder upright.}
  {\textbf{Scene:} Four pens lie on the table beside an upright, empty pen holder. \textbf{Procedure:} Pick up the pen holder, place the four pens inside it one by one, then place the filled holder upright on the table.}
  {\textbf{S1} Place the first pen in the pen holder. \textbf{S2} Place the second pen in the pen holder. \textbf{S3} Place the third pen in the pen holder. \textbf{S4} Place the fourth pen in the pen holder. \textbf{S5} Place the pen holder upright on the table.}

\RoboValuePkgDEEHDUEHDIDTEEENDXDRDEDXDTEEDRCPEHEFEIDTDVEMTaskCard
  {Load Plastic Plates Into Rack}
  {load_plastic_plates_into_rack}
  {Insert the three plastic plates into the rack in any order and in any slots.}
  {\textbf{Scene:} Three plastic plates lie outside a slotted plate rack on the table. The rack has 5 slots. \textbf{Procedure:} Insert all three plates upright into the rack's slots.}
  {\textbf{S1} Place the first plastic plate upright in an empty slot. \textbf{S2} Place the second plastic plate upright in an empty slot. \textbf{S3} Place the third plastic plate upright in an empty slot.}

\RoboValuePkgDEEHDUEHDIDTEEENDXDRDEDXDTEEDRCPEHEFEIDTDVEMTaskCard
  {Operate Drawers In Sequence}
  {operate_drawers_in_sequence}
  {Open the bottom, middle, and top drawers in that order, then close the top, middle, and bottom drawers in that order.}
  {\textbf{Scene:} A three-drawer storage unit stands on the table with all drawers closed. \textbf{Procedure:} Grasp the knobs to pull all drawers open, then grasp the knobs to push them closed and return to the initial pose.}
  {\textbf{S1} Open the bottom drawer. \textbf{S2} Open the middle drawer. \textbf{S3} Open the top drawer. \textbf{S4} Close the top drawer. \textbf{S5} Close the middle drawer. \textbf{S6} Close the bottom drawer.}

\RoboValuePkgDEEHDUEHDIDTEEENDXDRDEDXDTEEDRCPEHEFEIDTDVEMTaskCard
  {Organize Books And Pen}
  {organize_books_and_pen}
  {Center and align the closed book in the target rectangle. Place the pen along the spine of the open book, close the book, and stack it on the first book.}
  {\textbf{Scene:} A white book lies outside a marked area, a gray book is open on the table, and a pen lies nearby. \textbf{Procedure:} Move and align the white book in the marked area, place the pen in the gray book's spine, close the gray book, then place it on the white book.}
  {\textbf{S1} Place the white book in the marked area. \textbf{S2} Adjust the white book within the marked area. \textbf{S3} Place the pen in the spine of the gray book. \textbf{S4} Close the gray book. \textbf{S5} Place the gray book on the white book.}

\RoboValuePkgDEEHDUEHDIDTEEENDXDRDEDXDTEEDRCPEHEFEIDTDVEMTaskCard
  {Organize Cosmetics Storage Box}
  {organize_cosmetics_storage_box}
  {Open the upper drawer, cap the lipstick, and place the lipstick and eyebrow pencil inside. Close the drawer, then place the serum and cleansing oil in the storage area above, in either order.}
  {\textbf{Scene:} A storage cabinet stands on the table with its upper drawer closed. A lipstick and its separate cap, a pink eyebrow pencil, a yellow essence bottle, and cleansing oil are placed nearby. \textbf{Procedure:} Open the upper drawer, cap and store the lipstick, store the eyebrow pencil, and close the drawer. Place the yellow essence bottle in the upper open storage area with its base facing downward, then place the cleansing oil flat in the upper open storage area.}
  {\textbf{S1} Open the upper drawer. \textbf{S2} Place the lipstick cap on the lipstick. \textbf{S3} Place the lipstick in the drawer. \textbf{S4} Place the pink eyebrow pencil in the drawer. \textbf{S5} Close the drawer. \textbf{S6} Place the yellow essence bottle in the upper open storage area. \textbf{S7} Place the cleansing oil in the upper open storage area.}

\RoboValuePkgDEEHDUEHDIDTEEENDXDRDEDXDTEEDRCPEHEFEIDTDVEMTaskCard
  {Organize Desktop Workstation}
  {organize_desktop_workstation}
  {Place the pen in the holder, place the eraser in the open drawer and close it, switch off the lamp, and unplug it. These actions may be performed in any order, except that the lamp must be switched off before it is unplugged.}
  {\textbf{Scene:} An open drawer, an eraser, a desk lamp, a power strip, a pen holder, and a pen are placed on the table. The lamp is on and plugged into the power strip. \textbf{Procedure:} Put the pen in the pen holder, put the eraser in the drawer and close it, then turn off and unplug the lamp.}
  {\textbf{S1} Place the pen in the pen holder. \textbf{S2} Place the eraser in the drawer. \textbf{S3} Close the drawer. \textbf{S4} Turn off the desk lamp. \textbf{S5} Unplug the desk lamp.}

\RoboValuePkgDEEHDUEHDIDTEEENDXDRDEDXDTEEDRCPEHEFEIDTDVEMTaskCard
  {Pack Items For Shipping}
  {pack_items_for_shipping}
  {Place the wireless mouse, then the wired earphones, in the cardboard box. Cover both items with bubble wrap.}
  {\textbf{Scene:} A cardboard box sits at the table center, with wired earphones to its left, a wireless mouse to its right, and a sheet of bubble wrap in front. \textbf{Procedure:} Place the mouse and wired earphones inside the box and cover both with bubble wrap.}
  {\textbf{S1} Place the wireless mouse in the cardboard box. \textbf{S2} Place the wired earphones in the cardboard box. \textbf{S3} Place the bubble wrap over the packed items.}

\RoboValuePkgDEEHDUEHDIDTEEENDXDRDEDXDTEEDRCPEHEFEIDTDVEMTaskCard
  {Place Pot On Induction Cooker}
  {place_pot_on_induction_cooker}
  {Place the pot, with the spatula inside, on the induction cooker so that most of the pot is over the cooking surface. Turn the handle toward the camera, move the spatula to the basket, and cover the pot with the lid.}
  {\textbf{Scene:} An induction cooker sits at the table center, with a lid to its left and a pot containing a spatula to its right. A basket sits farther back and to the right of the pot. \textbf{Procedure:} Place the pot on the induction cooker, then adjust its position so that most of the pot is over the cooking surface and its handle points toward the camera. Move the spatula into the basket and cover the pot with the lid.
}  {\textbf{S1} Place the small pot on the induction cooker. \textbf{S2} Adjust the small pot on the induction cooker. \textbf{S3} Place the spatula in the basket. \textbf{S4} Cover the pot with the lid.}

\RoboValuePkgDEEHDUEHDIDTEEENDXDRDEDXDTEEDRCPEHEFEIDTDVEMTaskCard
  {Prepare Breakfast}
  {prepare_breakfast}
  {Place the pig bun on the left, the yellow steamed bun on the right, and the black steamed bun at the front of the upper tier of the two already assembled, in that order. Add the empty tier on top, place the large bun in its center, then cover the steamer with the lid.}
  {\textbf{Scene:} A plate holds a pig-shaped bun, a yellow steamed bun, a black steamed bun, and a large stuffed bun. Two steamer tiers are stacked at the table center, with an additional empty tier and a lid placed separately nearby. \textbf{Procedure:} Load the original upper steamer tier with the pig-shaped bun, the yellow bun, and the black bun. Add the separate tier, place the large bun at its center, and cover the resulting three-tier stack.}
  {\textbf{S1} Place the pig bun on the left side of the upper steamer tier. \textbf{S2} Place the yellow steamed bun on the right side of the upper steamer tier. \textbf{S3} Place the black steamed bun at the front of the upper steamer tier. \textbf{S4} Place the empty steamer tier on top. \textbf{S5} Place the large bun in the center of the top steamer tier. \textbf{S6} Cover the steamer with the lid.}

\RoboValuePkgDEEHDUEHDIDTEEENDXDRDEDXDTEEDRCPEHEFEIDTDVEMTaskCard
  {Press Button Five Times}
  {press_button_five_times}
  {Press the button 5 times, fully releasing it and returning to the waiting position after each press.}
  {\textbf{Scene:} A button sits near the table center. The operating arm starts at its waiting position. \textbf{Procedure:} Press and fully release the button, returning to the waiting position after each press.}
  {\textbf{S1} Press the button for the first time. \textbf{S2} Press the button for the second time. \textbf{S3} Press the button for the third time. \textbf{S4} Press the button for the fourth time. \textbf{S5} Press the button for the fifth time.}

\RoboValuePkgDEEHDUEHDIDTEEENDXDRDEDXDTEEDRCPEHEFEIDTDVEMTaskCard
  {Season Meat Skewers Repeatedly}
  {season_meat_skewers_repeatedly}
  {Sprinkle the meat skewers with the seasoning shaker 5 times continuously.}
  {\textbf{Scene:} Four meat skewers rest on a paper-lined baking tray, with a seasoning container to its right. \textbf{Procedure:} Pick up the container with the right arm, repeatedly tilt it downward over the skewers and return it upright, then place it back on the table.}
  {\textbf{S1} Sprinkle powder over the meat skewers for the first time. \textbf{S2} Sprinkle powder over the meat skewers for the second time. \textbf{S3} Sprinkle powder over the meat skewers for the third time. \textbf{S4} Sprinkle powder over the meat skewers for the fourth time. \textbf{S5} Sprinkle powder over the meat skewers for the fifth time. \textbf{S6} Place the seasoning bottle on the table.}

\RoboValuePkgDEEHDUEHDIDTEEENDXDRDEDXDTEEDRCPEHEFEIDTDVEMTaskCard
  {Set the Table}
  {set_the_table}
  {Take the plate, bowl, fork, and knife out of the crate. Place the plate face up in the lower center of the placemat and the bowl open side up in the upper center, in front of the plate. Place the fork on the left and the knife on the right, with the fork tines and knife tip pointing toward the front. Complete these actions in any order.}
  {\textbf{Scene:} A tablecloth lies on the table beside a tableware container holding a plate, a bowl, a fork, and a knife. \textbf{Procedure:} Take the plate, bowl, fork, and knife from the tableware container. Place the plate on the tablecloth, the bowl in front of the plate, the fork to the left of the plate, and the knife to the right of the plate.}
  {\textbf{S1} Place the plate on the tablecloth. \textbf{S2} Place the bowl in front of the plate. \textbf{S3} Place the fork to the left of the plate. \textbf{S4} Place the knife to the right of the plate.}

\RoboValuePkgDEEHDUEHDIDTEEENDXDRDEDXDTEEDRCPEHEFEIDTDVEMTaskCard
  {Sort Blocks in Order}
  {sort_blocks_in_order}
  {Place the red block on the yellow drawer, the yellow block on the white drawer, and the blue block on the dark-green drawer, in that order.}
  {\textbf{Scene:} Three drawer storage units are arranged in a row across the table. Red, yellow, and blue blocks form a separate row in front of them. \textbf{Procedure:} Place the red, yellow, and blue blocks on top of the yellow, white, and dark-green storage units, respectively.}
  {\textbf{S1} Place the red block on the yellow drawer. \textbf{S2} Place the yellow block on the white drawer. \textbf{S3} Place the blue block on the dark-green drawer.}

\RoboValuePkgDEEHDUEHDIDTEEENDXDRDEDXDTEEDRCPEHEFEIDTDVEMTaskCard
  {Strike Drum Five Times}
  {strike_drum_five_times}
  {Pick up the drumstick and strike the center of the drumhead 5 times continuously.}
  {\textbf{Scene:} A drum sits at the table center, with a drumstick lying to its right. \textbf{Procedure:} Pick up the drumstick with the right arm, strike the drumhead center repeatedly, lifting after each strike, and return the stick to the drum's right.}
  {\textbf{S1} Strike the center of the drum for the first time. \textbf{S2} Strike the center of the drum for the second time. \textbf{S3} Strike the center of the drum for the third time. \textbf{S4} Strike the center of the drum for the fourth time. \textbf{S5} Strike the center of the drum for the fifth time. \textbf{S6} Place the drumstick on the table.}

\RoboValuePkgDEEHDUEHDIDTEEENDXDRDEDXDTEEDRCPEHEFEIDTDVEMTaskCard
  {Transfer Blocks Using Plate}
  {transfer_blocks_using_plate}
  {Place the three colored blocks on the plate one at a time in the order red, green, then blue. Tilt the loaded plate to pour all three blocks into the box.}
  {\textbf{Scene:} Three blocks (red, green, and blue) are placed on the table beside an empty plate and a black box. \textbf{Procedure:} Load all three blocks onto the plate, align it with the edge of the black box, and tilt it until all blocks enter the black box.}
  {\textbf{S1} Place the red block onto the plate. \textbf{S2} Place the green block onto the plate. \textbf{S3} Place the blue block onto the plate. \textbf{S4} Pour all three colored blocks into the black box.}

\RoboValuePkgDEEHDUEHDIDTEEENDXDRDEDXDTEEDRCPEHEFEIDTDVEMAppendixLayoutEnd

\FloatBarrier

  \subsection{Original and Counterfactual Instructions}
  \label{app:realworld_instructions}
  Table~\ref{tab:realworld_counterfactual_instructions} covers the 14 real-world tasks in TGA-CF; the six Memory-VOC tasks are excluded. Each original instruction and its four counterfactual variants are evaluated on the same successful execution, using the complete modified instructions.

\begin{table}[htbp]
  \centering
  \setlength{\belowcaptionskip}{3pt}
  \caption{\textbf{Original and Counterfactual Instructions for Real-World Tasks.} Gray text gives the original requirement; purple text after each arrow gives its counterfactual replacement.}
  \label{tab:realworld_counterfactual_instructions}
  \begingroup
  \fontsize{7.5}{7.8}\selectfont
  \setlength{\tabcolsep}{3pt}
  \setlength{\parskip}{0pt}
  \setlength{\aboverulesep}{1.2pt}
  \setlength{\belowrulesep}{1.2pt}
  \renewcommand{\arraystretch}{1}
  \begin{tabularx}{\linewidth}{@{}
    >{\hsize=0.88\hsize\linewidth=\hsize\RaggedRight\arraybackslash}X
    >{\hsize=1.12\hsize\linewidth=\hsize\RaggedRight\arraybackslash}X@{}}
    \toprule
    \rowcolor{RVCFHeader}
    \textbf{Task and Original Instruction} & \textbf{Counterfactual Changes} \\
    \midrule
    \RVCFRow
      {Arrange Family Tea Set}
      {Place the teacup on the cat-head mat, the teapot on the yellow mat, the lid on the teapot, the tongs in their holder, and the canister on the round mat. Complete the four placement units in any order; within the teapot-and-lid unit, place the teapot first and seat its lid immediately afterward.}
      {\RVCFDelta{teacup on cat-head mat; teapot on yellow mat}{teacup on yellow mat; teapot on cat-head mat}}
      {\RVCFDelta{lid on teapot}{leave lid off teapot}}
      {\RVCFDelta{tongs in holder}{tongs beside holder}}
      {\RVCFDelta{place teapot immediately before lid}{place lid immediately before teapot}}
    \RVCFSeparator
    \RVCFRow
      {Arrange Flowers and Reed Diffuser}
      {Place the pink, white, and yellow flowers consecutively in the vase, in any order. Place the diffuser bottle on the coaster and insert the reed stick into it, in either order. Either diffuser action may occur before or after the flower sequence, but neither may occur between flower placements.}
      {\RVCFDelta{pink, white, yellow flowers}{pink, white, yellow paintbrushes}}
      {\RVCFDelta{insert reed into bottle}{leave reed beside bottle}}
      {\RVCFDelta{diffuser on coaster}{diffuser beside coaster}}
      {\RVCFDelta{all three consecutive in vase}{pink and white consecutive; yellow on table}}
    \RVCFSeparator
    \RVCFRow
      {Arrange Seasoning Jars}
      {Place the three seasoning jars on the two-tier rack in any order, on either tier and in any positions.}
      {\RVCFDelta{seasoning jars}{small ceramic cups}}
      {\RVCFDelta{place all three on rack}{place all three, then rotate each in place}}
      {\RVCFDelta{all three on rack}{all three beside rack}}
      {\RVCFDelta{all three}{exactly two; one beside rack}}
    \RVCFSeparator
    \RVCFRow
      {Fill And Place Pen Holder}
      {Hold the pen holder with its opening upward, insert all four pens one by one in any order, then place the filled holder upright.}
      {\RVCFDelta{pen holder}{mug}}
      {\RVCFDelta{leave holder filled}{remove all pens; leave holder empty}}
      {\RVCFDelta{filled holder upright}{filled holder horizontal on table}}
      {\RVCFDelta{all four pens}{exactly three pens}}
    \RVCFSeparator
    \RVCFRow
      {Load Plastic Plates Into Rack}
      {Insert the three plastic plates into the rack in any order and in any slots.}
      {\RVCFDelta{plastic plates}{thin plastic cutting boards}}
      {\RVCFDelta{insert all three into rack}{lay all three flat on table beside rack}}
      {\RVCFDelta{inside rack slots}{upright in front of rack}}
      {\RVCFDelta{all three}{exactly two; one beside rack}}
    \RVCFSeparator
    \RVCFRow
      {Organize Books And Pen}
      {Center and align the closed book in the target rectangle. Place the pen along the spine of the open book, close the book, and stack it on the first book.}
      {\RVCFDelta{books}{binders}}
      {\RVCFDelta{stack newly closed book}{leave it unstacked}}
      {\RVCFDelta{pen along open book's spine}{pen beside open book}}
      {\RVCFDelta{center/align first book, then place pen, close and stack second book}{place pen, close and stack second book, then center/align first book}}
    \RVCFSeparator
    \RVCFRow
      {Organize Cosmetics Storage Box}
      {Open the upper drawer, cap the lipstick, and place the lipstick and eyebrow pencil inside. Close the drawer, then place the serum and cleansing oil in the storage area above, in either order.}
      {\RVCFDelta{lipstick}{mascara}}
      {\RVCFDelta{cap lipstick}{leave lipstick uncapped}}
      {\RVCFDelta{pencil ends in drawer}{pencil ends in upper storage}}
      {\RVCFDelta{serum and oil in either order}{cleansing oil before serum}}
    \RVCFSeparator
    \RVCFRow
      {Organize Desktop Workstation}
      {Place the pen in the holder, place the eraser in the open drawer and close it, switch off the lamp, and unplug it. These actions may be performed in any order, except that the lamp must be switched off before it is unplugged.}
      {\RVCFDelta{pen to holder; eraser to drawer}{pen to drawer; eraser to holder}}
      {\RVCFDelta{close drawer}{leave drawer open}}
      {\RVCFDelta{pen in holder}{pen beside holder}}
      {\RVCFDelta{lamp off before unplug; object placements otherwise unordered}{lamp off, then unplug, before both object placements}}
    \RVCFSeparator
    \RVCFRow
      {Pack Items For Shipping}
      {Place the wireless mouse, then the wired earphones, in the cardboard box. Cover both items with bubble wrap.}
      {\RVCFDelta{wireless mouse}{wireless game controller}}
      {\RVCFDelta{cover both items}{leave both uncovered}}
      {\RVCFDelta{mouse and earphones both in box}{mouse in box; earphones beside box}}
      {\RVCFDelta{cover both items}{cover mouse only}}
    \RVCFSeparator
    \RVCFRow
      {Place Pot On Induction Cooker}
      {Place the pot, with the spatula inside, on the induction cooker so that most of the pot is over the cooking surface. Turn the handle toward the camera, move the spatula to the basket, and cover the pot with the lid.}
      {\RVCFDelta{spatula}{wooden spoon}}
      {\RVCFDelta{move spatula to basket}{leave spatula inside pot}}
      {\RVCFDelta{pot mostly over cooker}{pot beside cooker}}
      {\RVCFDelta{move spatula to basket, then cover pot}{cover pot, then move spatula to basket}}
    \RVCFSeparator
    \RVCFRow
      {Prepare Breakfast}
      {Place the pig bun on the left, the yellow steamed bun on the right, and the black steamed bun at the front of the upper tier of the two already assembled, in that order. Add the empty tier on top, place the large bun in its center, then cover the steamer with the lid.}
      {\RVCFDelta{black at current upper-tier front; large at added-tier center}{black at added-tier center; large at current upper-tier front}}
      {\RVCFDelta{cover with lid}{leave steamer uncovered}}
      {\RVCFDelta{large bun centered on top tier}{large bun beside third tier}}
      {\RVCFDelta{three small buns at left/right/front, in fixed order}{any two small buns at left/right; third left on table}}
    \RVCFSeparator
    \RVCFRow
      {Set the Table}
      {Take the plate, bowl, fork, and knife out of the crate. Place the plate face up in the lower center of the placemat and the bowl open side up in the upper center, in front of the plate. Place the fork on the left and the knife on the right, with the fork tines and knife tip pointing toward the front. Complete these actions in any order.}
      {\RVCFDelta{plate lower; bowl upper/in front of plate}{bowl lower; plate upper/in front of bowl}}
      {\RVCFDelta{bowl open side up}{bowl open side down}}
      {\RVCFDelta{bowl in front of plate}{bowl behind plate}}
      {\RVCFDelta{any order}{fixed order: plate, bowl, fork, knife}}
    \RVCFSeparator
    \RVCFRow
      {Sort Blocks in Order}
      {Place the red block on the yellow drawer, the yellow block on the white drawer, and the blue block on the dark-green drawer, in that order.}
      {\RVCFDelta{red on yellow; yellow on white; order red-yellow-blue}{red on white; yellow on yellow; order yellow-red-blue}}
      {\RVCFDelta{finish with blocks on drawers}{after placement, remove all blocks to table}}
      {\RVCFDelta{blue on dark-green drawer}{blue on yellow drawer}}
      {\RVCFDelta{red, yellow, blue order}{blue, red, yellow order}}
    \RVCFSeparator
    \RVCFRow
      {Transfer Blocks Using Plate}
      {Place the three colored blocks on the plate one at a time in the order red, green, then blue. Tilt the loaded plate to pour all three blocks into the box.}
      {\RVCFDelta{load onto plate; pour into box}{load into box; pour onto plate}}
      {\RVCFDelta{tilt plate to pour}{push blocks into box without tilting}}
      {\RVCFDelta{pour into box}{pour onto table beside box}}
      {\RVCFDelta{all three, red-green-blue}{exactly two, red then green; blue on table}}
    \bottomrule
  \end{tabularx}
  \endgroup
\end{table}



\subsection{Out-of-Domain Scenario Construction}
\label{app:realworld_ood}

Real-world OOD evaluation introduces separate changes in embodiment and environment while retaining the task goals. As in simulation, the shifted conditions use the shared evaluation protocol without additional adaptation.

\noindent\textbf{Cross-Embodiment Generalization.}
AgiBot Genie02 serves as the standard embodiment, and a dual-arm ARX setup provides the embodiment shift. The two platforms perform the same tasks with different visible morphologies and execution patterns. This setting examines whether value judgments remain informative about task progress and outcomes when transferred to a different physical robot.

\noindent\textbf{Cross-Environment Generalization.}
We retain the standard embodiment and construct two environment configurations that vary tablecloth color, lighting, and tabletop clutter. The first combines one tablecloth color with cool-toned frontal lighting. The second uses a different tablecloth color and mixed warm- and cool-toned lighting from an oblique angle on the right. Each trajectory includes two to four task-irrelevant objects, with positions and orientations varied across trajectories. These changes test whether models distinguish task-relevant interactions from variations in surface appearance, illumination, and clutter under real-world conditions.

\Needspace{6\baselineskip}
\subsection{Task Execution Visualizations}
  \label{app:realworld_visualizations}
  
\begin{figure}[htbp]
  \centering
  \includegraphics[width=0.80\linewidth,trim=0 44bp 0 62bp,clip]{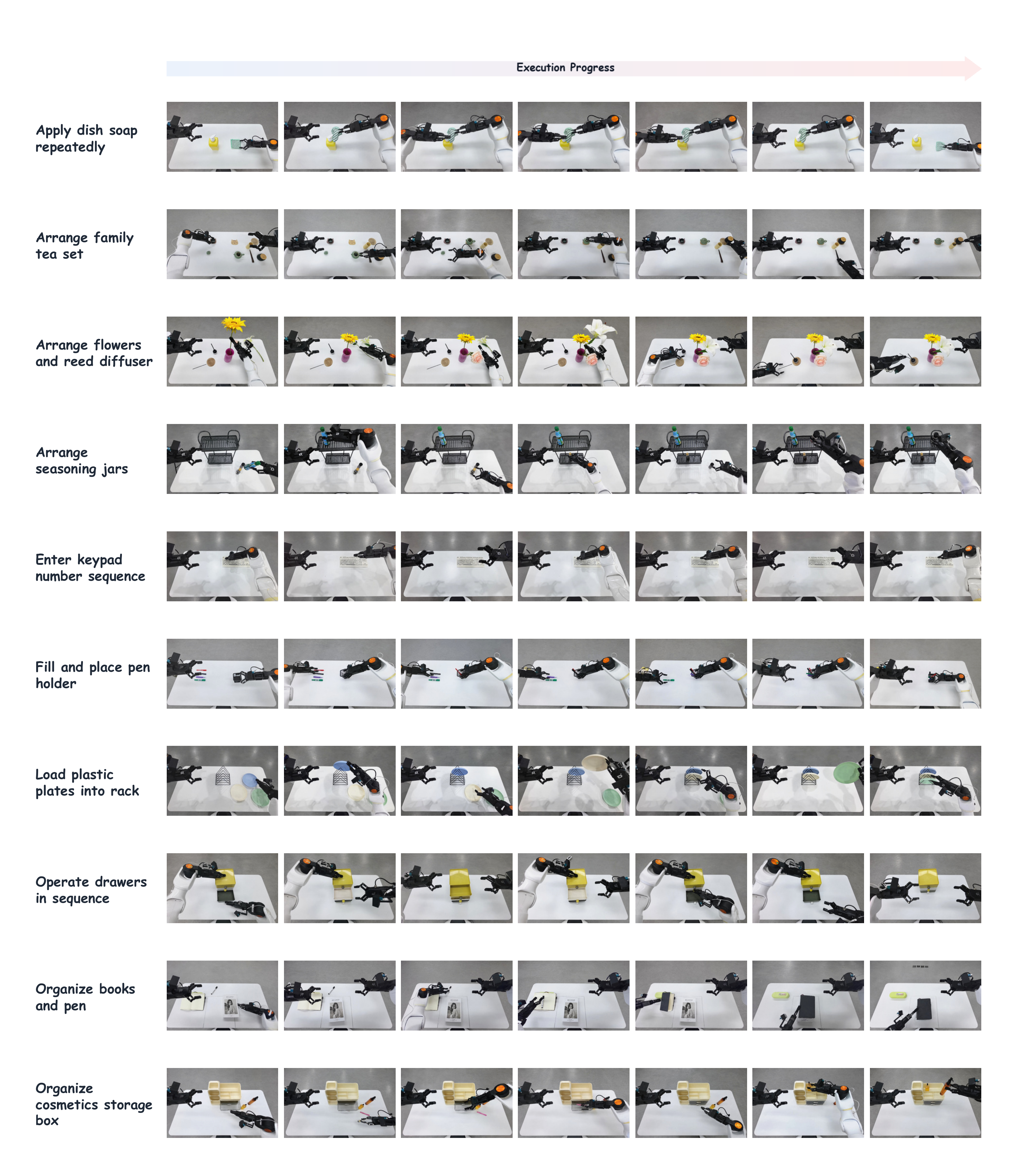}
  \caption{\textbf{Real-World Task Executions (Part 1 of 2).} Seven chronologically ordered frames per row show successful executions of the first ten real-world tasks.}
  \label{fig:real_7frames_1}
\end{figure}

\begin{figure}[htbp]
  \centering
  \includegraphics[width=0.95\linewidth,trim=0 44bp 0 62bp,clip]{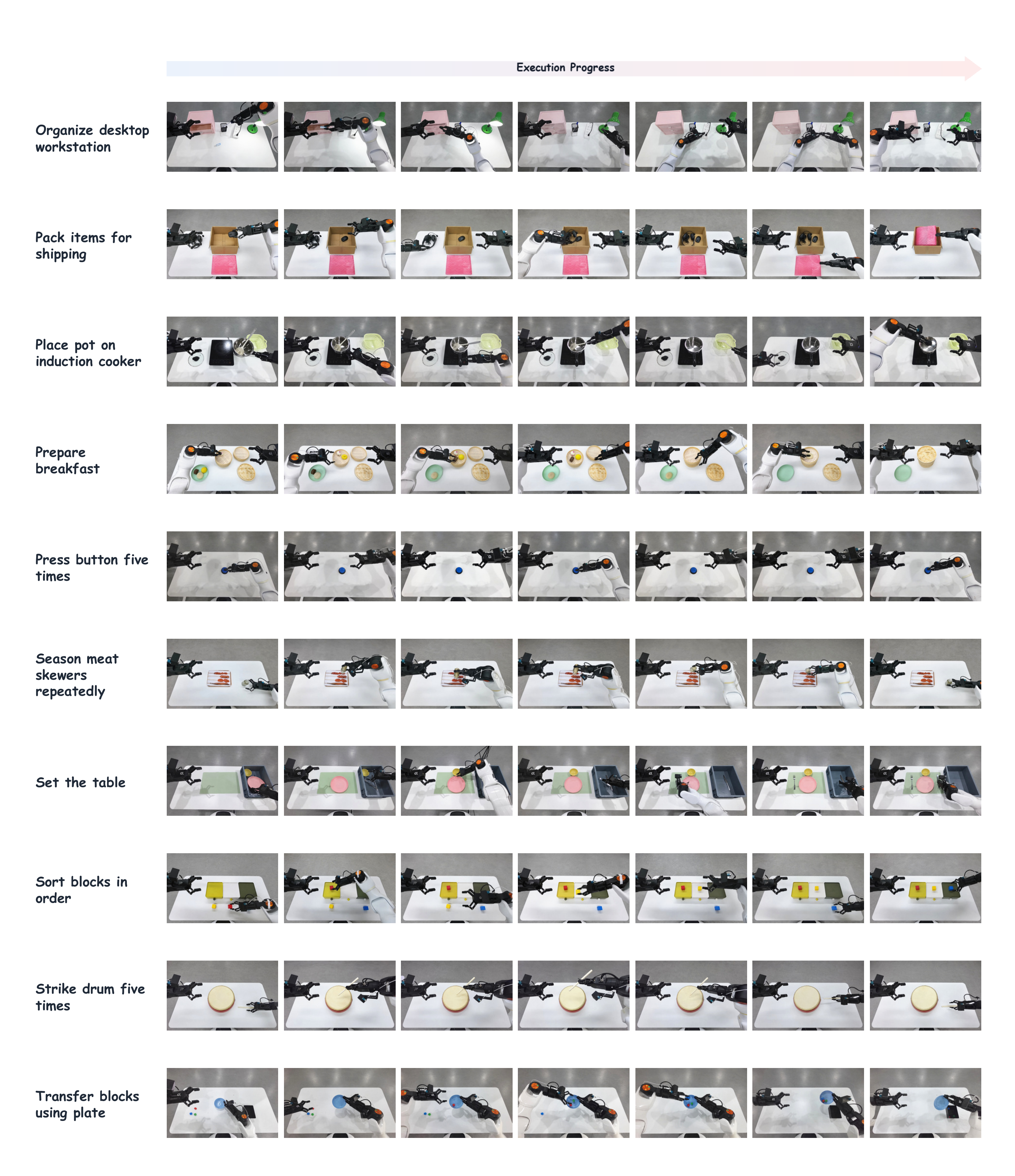}
  \caption{\textbf{Real-World Task Executions (Part 2 of 2).} Seven chronologically ordered frames per row show successful executions of the remaining ten real-world tasks.}
  \label{fig:real_7frames_2}
\end{figure}

  Figures~\ref{fig:real_7frames_1} and~\ref{fig:real_7frames_2} show successful executions of all 20 real-world tasks, with seven chronologically ordered frames per task tracing the initial scene, intermediate object interactions, and final configuration.

\FloatBarrier

\definecolor{RVProtocolHeader}{HTML}{F1EEF5}
\definecolor{RVProtocolStripe}{HTML}{F9F7FB}
\section{Data Collection Protocol}
\label{app:data_collection_protocol}

The following examples describe test-set collection, specifying how task constraints determine successful executions, alternative solutions, and controlled failure and recovery scenarios.

\subsection{Simulation Collection Example}
\label{app:pen_holder_sim_sop}

The \emph{Fill Pen Holder} task illustrates the collection of successful executions, alternative valid solutions, and contrasting responses to the same failure scenario.

\noindent\textbf{Platform and Initial Scene.}
We collect trajectories through RoboDojo on NVIDIA Isaac Sim, using ARX as the standard embodiment. An empty orange pen holder stands on the left side of the table, with four pens---purple, blue, white, and black-and-gold---placed separately on the right. Initial positions may vary within reach of both arms, provided that the holder remains upright and empty and each pen is accessible. Cross-embodiment collection replaces ARX with a dual-arm UR5e setup while retaining the background and task objects. Cross-environment collection retains ARX and varies surface appearance, clutter, and lighting as specified in Section~\ref{app:simulation_ood}. Both shifts preserve the task-relevant initial state and success criteria.

\noindent\textbf{Task Goal and Success Criteria.}
One arm picks up and holds the pen holder while the other inserts all four pens individually. The filled holder is then placed upright on the table. Success requires all four pens to remain inside the holder and the holder to remain upright after release. Holder pickup precedes the insertions, and final placement follows them; the pen insertion order is unrestricted.

\noindent\textbf{Alternative Valid Solutions.}
The canonical execution uses the left arm to hold the holder and the right arm to insert the purple, blue, white, and black-and-gold pens in that order. The left arm then places the filled holder upright at the center of the table. The four insertions admit $4!=24$ orders. The multi-solution set covers the 20 orders in Table~\ref{tab:pen-holder-sim-orders}, with two trajectories per order under the standard ARX setting. These trajectories also reverse the arm roles: the right arm holds the holder and the left arm inserts the pens. Thus, execution order and arm assignment vary while the task requirements remain fixed.

\begin{table}[htbp]
  \centering
  \caption{\textbf{Selected Insertion Orders for Fill Pen Holder.} Each order is collected twice, with the right arm holding the holder and the left arm inserting the pens.}
  \label{tab:pen-holder-sim-orders}
  \begingroup
  \small
  \setlength{\tabcolsep}{7pt}
  \renewcommand{\arraystretch}{1.16}
  \rowcolors{2}{RVProtocolStripe}{white}
  \begin{tabularx}{\linewidth}{@{}c>{\centering\arraybackslash}Xc>{\centering\arraybackslash}X@{}}
    \toprule
    \rowcolor{RVProtocolHeader}
    \textbf{Order} & \textbf{Insertion Sequence} & \textbf{Order} & \textbf{Insertion Sequence} \\
    \midrule
     1 & $P\rightarrow B\rightarrow K\rightarrow W$ & 11 & $B\rightarrow W\rightarrow K\rightarrow P$ \\
     2 & $P\rightarrow K\rightarrow B\rightarrow W$ & 12 & $K\rightarrow P\rightarrow W\rightarrow B$ \\
     3 & $P\rightarrow K\rightarrow W\rightarrow B$ & 13 & $K\rightarrow P\rightarrow B\rightarrow W$ \\
     4 & $P\rightarrow W\rightarrow B\rightarrow K$ & 14 & $K\rightarrow B\rightarrow W\rightarrow P$ \\
     5 & $P\rightarrow W\rightarrow K\rightarrow B$ & 15 & $K\rightarrow B\rightarrow P\rightarrow W$ \\
     6 & $B\rightarrow P\rightarrow W\rightarrow K$ & 16 & $K\rightarrow W\rightarrow P\rightarrow B$ \\
     7 & $B\rightarrow P\rightarrow K\rightarrow W$ & 17 & $K\rightarrow W\rightarrow B\rightarrow P$ \\
     8 & $B\rightarrow K\rightarrow P\rightarrow W$ & 18 & $W\rightarrow P\rightarrow B\rightarrow K$ \\
     9 & $B\rightarrow K\rightarrow W\rightarrow P$ & 19 & $W\rightarrow P\rightarrow K\rightarrow B$ \\
    10 & $B\rightarrow W\rightarrow P\rightarrow K$ & 20 & $W\rightarrow K\rightarrow P\rightarrow B$ \\
    \bottomrule
  \end{tabularx}
  \par\smallskip
  \raggedright\footnotesize $P$: purple; $B$: blue; $W$: white; $K$: black-and-gold. Holder pickup and final placement enclose every insertion sequence.
  \endgroup
\end{table}

\noindent\textbf{Trajectory Allocation.}
The collection comprises 75 successful trajectories and 65 failure/recovery trajectories. The canonical execution accounts for 35 successful trajectories: 15 in-domain, 10 cross-embodiment, and 10 cross-environment. The remaining 40 successful trajectories cover the selected alternative solutions. Each of the five failure scenarios contributes seven error continuations (three in-domain, two cross-embodiment, and two cross-environment), three effective recoveries, and three ineffective recoveries. Recovery trajectories use ARX in the standard setting. In total, the collection contains 100 in-domain trajectories and 20 trajectories for each OOD setting.

\noindent\textbf{Failure and Recovery Trajectories.}
Each scenario introduces one primary error into the canonical execution, with initial and error states matched across branches wherever possible. Error Continuation (EC) proceeds without correction; Effective Recovery (ER) resolves the error before proceeding; Ineffective Recovery (IR) attempts a correction but leaves the error unresolved. All branches continue with the remaining subtasks. EC and IR end unsuccessfully, whereas ER satisfies the task goal. The five scenarios are:

\begin{enumerate}
\item \textbf{FRT-1: Dropped Holder.} The holder slips during pickup. In EC, the robot picks up the pens in turn and performs insertion motions against the fallen holder's outer wall. In ER, it regrasps the holder, inserts all four pens, and places the filled holder upright. In IR, it fails to regrasp the holder and continues the insertion motions against its outer wall. Subsequent task motions continue despite the unresolved error in EC and IR.

\item \textbf{FRT-2: Dropped Second Pen.} The blue pen slips before insertion. EC performs the insertion motion with an empty gripper. ER retrieves and inserts the pen. IR fails to retrieve it and repeats the empty insertion motion. Each branch then attempts the remaining insertions and final holder placement.

\item \textbf{FRT-3: Misaligned Second-Pen Insertion.} The blue pen misses the opening and falls onto the table. EC proceeds without retrieving it. ER regrasps, aligns, and inserts the pen. IR fails to regrasp it and repeats the off-target insertion motion with an empty gripper. Each branch continues with the remaining pens and final holder placement.

\item \textbf{FRT-4: Dropped Fourth Pen.} The black-and-gold pen slips before insertion. EC performs an empty insertion motion. ER retrieves and inserts the pen. IR fails to retrieve it and repeats the empty insertion motion. All branches then place the holder upright on the table.

\item \textbf{FRT-5: Misaligned Fourth-Pen Insertion.} The black-and-gold pen misses the opening and falls onto the table. EC leaves it there. ER regrasps, aligns, and inserts it. IR fails to regrasp it and repeats the off-target insertion motion with an empty gripper. All branches proceed to final holder placement.
\end{enumerate}

\subsection{Real-World Collection Example}
\label{app:arrange_family_tea_set}

The \emph{Arrange Family Tea Set} task illustrates how placement constraints and subtask dependencies define valid solutions and the outcomes of recovery attempts.

\noindent\textbf{Platform and Initial Scene.}
The standard AgiBot Genie02 setup uses two Luoshi arms, each equipped with a CTAG2F90-D gripper, and software version \texttt{genie\_feat-rlinf\_rebase\_2.3.5}. The scene contains cup A, teapot B, its separate lid, a closed tea canister, tea tongs, and a fixed bamboo utensil holder. The cup and teapot begin upright at separate reachable locations in the rear material area. The lid begins in the left material area with a clear transfer path to the teapot. The canister remains closed throughout the task. The tongs lie flat near the material area or holder; the holder's upward-facing opening permits vertical insertion. Cross-embodiment collection uses a dual-arm ARX setup, while cross-environment collection retains Genie02 and varies the scene conditions described in Section~\ref{app:realworld_ood}.

\noindent\textbf{Task Goal and Success Criteria.}
The robot places cup A on the left-front cat-head mat and teapot B on the right-front yellow mat, both upright with their entire bases inside the designated support regions. The lid must sit stably on the teapot. The canister must stand upright on the round mat with its lid closed and aligned. The tongs must be inserted vertically into the bamboo holder with their gripping tips upward, without catching on the rim or extending outside the tool target region. Success requires all five conditions to hold after release, together with the ordering constraint below.

\noindent\textbf{Alternative Valid Solutions.}
We enumerate four placement units: cup placement ($A$), teapot placement immediately followed by lid placement ($P$), canister placement ($C$), and tong insertion ($T$). The two actions in $P$ remain consecutive and retain their internal order. Permuting the four units therefore yields $4!=24$ valid solutions (Table~\ref{tab:family_tea_set_orders}), all with the same target configuration. Teapot and lid placement retain separate annotated subtask boundaries even though they form one unit for order enumeration.

\begin{table}[htbp]
  \centering
  \caption{\textbf{Valid Execution Orders for Arrange Family Tea Set.} The four units may appear in any order; teapot placement and lid placement remain consecutive within unit $P$.}
  \label{tab:family_tea_set_orders}
  \begingroup
  \small
  \setlength{\tabcolsep}{7pt}
  \renewcommand{\arraystretch}{1.16}
  \rowcolors{2}{RVProtocolStripe}{white}
  \begin{tabularx}{\linewidth}{@{}c>{\centering\arraybackslash}Xc>{\centering\arraybackslash}X@{}}
    \toprule
    \rowcolor{RVProtocolHeader}
    \textbf{Order} & \textbf{Unit Sequence} & \textbf{Order} & \textbf{Unit Sequence} \\
    \midrule
1  & \(A\rightarrow P\rightarrow C\rightarrow T\) &
13 & \(C\rightarrow A\rightarrow P\rightarrow T\) \\
2  & \(A\rightarrow P\rightarrow T\rightarrow C\) &
14 & \(C\rightarrow A\rightarrow T\rightarrow P\) \\
3  & \(A\rightarrow C\rightarrow P\rightarrow T\) &
15 & \(C\rightarrow P\rightarrow A\rightarrow T\) \\
4  & \(A\rightarrow C\rightarrow T\rightarrow P\) &
16 & \(C\rightarrow P\rightarrow T\rightarrow A\) \\
5  & \(A\rightarrow T\rightarrow P\rightarrow C\) &
17 & \(C\rightarrow T\rightarrow A\rightarrow P\) \\
6  & \(A\rightarrow T\rightarrow C\rightarrow P\) &
18 & \(C\rightarrow T\rightarrow P\rightarrow A\) \\
7  & \(P\rightarrow A\rightarrow C\rightarrow T\) &
19 & \(T\rightarrow A\rightarrow P\rightarrow C\) \\
8  & \(P\rightarrow A\rightarrow T\rightarrow C\) &
20 & \(T\rightarrow A\rightarrow C\rightarrow P\) \\
9  & \(P\rightarrow C\rightarrow A\rightarrow T\) &
21 & \(T\rightarrow P\rightarrow A\rightarrow C\) \\
10 & \(P\rightarrow C\rightarrow T\rightarrow A\) &
22 & \(T\rightarrow P\rightarrow C\rightarrow A\) \\
11 & \(P\rightarrow T\rightarrow A\rightarrow C\) &
23 & \(T\rightarrow C\rightarrow A\rightarrow P\) \\
12 & \(P\rightarrow T\rightarrow C\rightarrow A\) &
24 & \(T\rightarrow C\rightarrow P\rightarrow A\) \\
    \bottomrule
  \end{tabularx}
  \par\smallskip
  \raggedright\footnotesize $A$: cup; $P$: teapot followed by lid; $C$: canister; $T$: tongs inserted vertically into the holder.
  \endgroup
\end{table}

\noindent\textbf{Trajectory Allocation.}
The collection comprises 58 successful trajectories and 52 failure/recovery trajectories. Among the successful trajectories, 35 follow the canonical order $A\rightarrow P\rightarrow C\rightarrow T$: 15 in-domain, 10 cross-embodiment, and 10 cross-environment. One in-domain trajectory provides solution 1 in Table~\ref{tab:family_tea_set_orders}; 23 additional trajectories cover solutions 2--24. Each of the four failure scenarios contributes seven error continuations (three in-domain, two cross-embodiment, and two cross-environment), three effective recoveries, and three ineffective recoveries.

\noindent\textbf{Failure and Recovery Trajectories.}
The failure trajectories use the canonical order and introduce one error scenario per execution. EC leaves the error uncorrected, ER corrects it, and IR attempts a correction that does not restore the required state. The robot continues with the remaining subtasks in every branch, and subtasks unrelated to the seeded error are completed correctly. EC and IR end unsuccessfully; ER satisfies all five success conditions. The four scenarios are:

\begin{enumerate}
\item \textbf{FRT-1: Dropped Lid.} The gripper releases the lid during transfer to the teapot. EC leaves it at the drop location. ER retrieves and seats it correctly. IR retries but drops it again before seating. Each branch then proceeds to the canister and tongs.

\item \textbf{FRT-2: Misoriented Tongs.} The robot places the tongs across the holder opening instead of inserting them vertically with the gripping tips upward. EC ends without adjustment. ER removes, reorients, and correctly inserts the tongs. IR attempts reinsertion but leaves them across the opening or outside the valid vertical configuration. As this is the final subtask, no further placement follows.

\item \textbf{FRT-3: Misplaced Teapot.} The teapot lands upright with part of its base outside the yellow mat's valid region. EC leaves it offset. ER repositions it fully within the mat. IR attempts repositioning but leaves it partly outside. Each branch then completes lid placement, canister placement, and tong insertion. The teapot remains stable and reachable for lid placement in all branches.

\item \textbf{FRT-4: Misplaced Canister.} The closed, upright canister is placed partly outside its target region. EC leaves it there. ER moves it fully into the region. IR attempts repositioning but leaves it partly outside. The canister remains closed and upright, and all branches proceed to the tongs.
\end{enumerate}

\FloatBarrier
\section{Evaluation Implementation}
\label{app:evaluation_setup}

\subsection{Baseline Model Profiles}
\label{app:model_profiles}

The following profiles summarize the evaluated models in alphabetical order, including the additional FailSafe baseline used for subtask identification.

\noindent\textbf{FailSafe.} FailSafe develops a simulation-based framework for generating robot failures and verifying executable recovery actions~\citep{lin2025failsafe}. Its FailSafe-VLM model fine-tunes LLaVA-OneVision-7B to describe the current subtask, identify execution failures, and suggest corrective actions from visual observations and a task instruction. The framework links failure reasoning to actions that can restore task execution. RoboValue evaluates its current-subtask descriptions through SIA; failure verdicts and recovery suggestions are not used as scalar value predictions.

\noindent\textbf{LIV.} Language-Image Value learning (LIV) jointly learns visual--language representations and goal-conditioned values from videos with text annotations, without action labels~\citep{ma2023liv}. Its objective couples value learning with image--text contrastive learning, allowing the learned embedding space to represent tasks specified by language or image goals. Pretraining on human videos provides representations that can transfer to robotic control and reward specification. In RoboValue, LIV supplies a single-image baseline through the similarity between the current observation and the task instruction.

\noindent\textbf{ProcVLM.} ProcVLM grounds progress estimation in task procedures, combining subtask semantics with visual change within each stage~\citep{feng2026procvlm}. Its annotation pipeline assigns progress budgets to subtasks and distributes them across their constituent frames. The model is trained jointly on progress estimation, action segmentation, and prediction of remaining actions. It uses a Qwen3-VL-2B backbone~\citep{bai2025qwen3vl} and supports both reasoning-formatted text generation and a dedicated continuous progress head. We evaluate ProcVLM-2B in zero-shot and task-adapted one-shot settings. The adapter used here reads progress from the generated response; this readout is distinguished from the continuous head described in the original paper.

\noindent\textbf{Robo-Dopamine.} Robo-Dopamine's General Reward Model predicts signed progress changes between observations conditioned on a task instruction and reference states~\citep{tan2025robo}. Step-aware supervision and multiple camera views support these comparisons, while its inference procedure combines incremental progress with estimates anchored to the initial and goal states. We evaluate the original 3B and 8B checkpoints and the subsequently released 4B and 8B \emph{2.0 Preview} checkpoints.\footnote{Official checkpoint releases: \url{https://github.com/FlagOpen/Robo-Dopamine}.} The Preview releases are marked with a dagger in Table~\ref{tab:robovalue_main}; we retain this designation to distinguish the evaluated weights from the later history-conditioned, OOD-aware formulation described in Robo-Dopamine 2.0~\citep{xu2026robo}.

\noindent\textbf{RoboFAC.} RoboFAC formulates task understanding, failure analysis, and corrective guidance as video question answering~\citep{ye2025robofac}. Its supervision covers task decomposition, failure detection and localization, explanations, and suggestions for recovery. We evaluate the 7B model, obtained by fine-tuning Qwen2.5-VL on 60K question--answer pairs from simulated robot executions; the original study reserves real-world data for evaluation. The model natively generates textual assessments. RoboValue uses these outputs for subtask identification and introduces a separate progress rubric to obtain scalar feedback, as specified in Appendix~\ref{app:adapter_implementations}.

\noindent\textbf{RoboMeter.} RoboMeter combines frame-level progress supervision with preference learning over trajectory pairs~\citep{liang2026robometer}. Its shared vision--language backbone has separate progress, preference, and success prediction heads, with success classification serving as an additional training objective. Trajectory comparisons allow successful, suboptimal, and failed executions to contribute supervision even when dense progress labels are unavailable. Training on the multi-embodiment RBM-1M corpus also uses temporal and task-mismatch augmentations. We evaluate RoboMeter-4B through its progress head; the preference and success heads are not substituted for the benchmark's shared comparison and outcome-discrimination protocols.

\noindent\textbf{RoboReward.} RoboReward predicts an ordinal task-completion rating from an execution video and a language instruction~\citep{lee2026roboreward}. Its training corpus draws on Open X-Embodiment and RoboArena. Counterfactual task relabeling and temporal clipping augment successful demonstrations with mismatched goals, near-misses, and partial executions. The resulting supervision uses five completion levels, so its native output is a discrete clip-level assessment. We evaluate the released 4B and 8B models, querying observed prefixes to obtain the scalar sequences required by RoboValue.

\noindent\textbf{RynnValue.} RynnValue learns language-conditioned temporal values from heterogeneous robot and human videos~\citep{huang2026rynnvalue}. Two distributional heads predict the remaining time to completion and signed temporal differences between observations, while a language branch learns video description, instruction matching, and success verification. Temporal targets are derived from timestamps and relabeled completion cutoffs. Random temporal sampling, order shuffling, and value-isolation attention are designed to reduce reliance on sampling regularities and other prediction shortcuts. We evaluate the 4B and 8B variants using their absolute remaining-time predictions, with the sign reversed so that larger values indicate greater task advancement.

\noindent\textbf{TOPReward.} TOPReward extracts a reward signal from a pretrained vision--language model without reward-specific training~\citep{chen2026topreward}. Given an observed video prefix and task instruction, it uses the log-probability of an affirmative completion token, such as \texttt{True}, as its score. We evaluate Qwen3-VL-8B-Instruct and Qwen3-VL-32B-Instruct backbones. The signal comes from the model's output distribution rather than a generated numerical estimate. RoboValue retains the raw prefix score, without the episode-level min--max normalization used for within-trajectory analysis in the original paper. This preserves a causal score at each queried state.

\noindent\textbf{VLAC.} VLAC combines action generation and execution assessment in a vision--language--action--critic model~\citep{zhai2025vision}. Its critic predicts signed relative progress from a pair of observations and provides a separate task-completion judgment from a single observation. Training combines vision--language data with human and robot trajectories, including negative and instruction-mismatched examples for irrelevant behavior, regression, and stagnation. We evaluate the critic outputs of the 2B and 8B variants in zero-shot and demonstration-conditioned one-shot settings. Their pairwise predictions are retained for comparisons and accumulated to produce scalar progress estimates.

\subsection{Input--Output Interfaces and Adapter Implementations}
\label{app:adapter_implementations}

Each adapter constructs the observation context required by a model and maps its predictions to the shared interfaces in Section~\ref{sec:protocol}: scalar value $v(A;l)$, directional comparison $c(A_1,A_2;l)$, and subtask description $s(A;l)$. Table~\ref{tab:adapter_routing} summarizes the supported interfaces and evaluated inputs. Scalar values retain their native meaning and scale, with their direction aligned so that larger values indicate greater task advancement. Robo-Dopamine and VLAC supply native pairwise predictions; for other models with a scalar interface, we use $c(A_1,A_2;l)=v(A_2;l)-v(A_1;l)$.

The scalar-sequence metrics query $v$, TGA and CSVC query $c$, and SIA queries $s$. For SIA, FailSafe, ProcVLM, and RoboFAC generate subtask descriptions without seeing the candidate labels. DeepSeek-V4-Pro~\citep{deepseekai2026deepseekv4} then receives the task instruction, the generated description, and the candidate subtasks, each assigned a single-digit label. The judge is prompted to select one candidate, and the next-token probability of the ground-truth label supplies $q_i$ in Section~\ref{sec:metrics}. These probabilities are aggregated geometrically within each task, followed by the shared task-level averaging protocol. The model descriptions below follow the order of scalar value, pairwise comparison, and subtask description wherever those interfaces are supported.

\begin{table}[htbp]
  \centering
  \caption{\textbf{Model-Specific Input and Output Mappings.} Frame counts and sampling rates describe the evaluated adapters. A dash marks an interface not evaluated; FailSafe contributes only to SIA.}
  \label{tab:adapter_routing}
  \begingroup
  \fontsize{8.5}{10.2}\selectfont
  \setlength{\tabcolsep}{3pt}
  \renewcommand{\arraystretch}{1.12}
  \begin{tabularx}{\linewidth}{@{}>{\RaggedRight\arraybackslash}p{64pt}>{\RaggedRight\arraybackslash}p{124pt}>{\RaggedRight\arraybackslash}p{92pt}>{\RaggedRight\arraybackslash}p{62pt}>{\RaggedRight\arraybackslash}X@{}}
    \toprule
    \multirow{2}{*}{\textbf{Model}} & \multirow{2}{*}{\textbf{Observation Context}}
    & \multicolumn{3}{c}{\textbf{Unified Output Interfaces}} \\
    \cmidrule(lr){3-5}
    & & \textbf{Scalar Value $v$} & \textbf{Comparison $c$} & \textbf{Subtask $s$} \\
    \midrule
    \rowcolor{RVCompareStripe}
    FailSafe & Latest ten prefix frames & -- & -- & Generated description \\
    LIV & Current image & Image--text cosine similarity & Value difference & -- \\
    \rowcolor{RVCompareStripe}
    ProcVLM & Latest eight points on a global grid; task LoRA in one-shot mode & Generated percentage divided by 100 & Value difference & First unfinished action \\
    Robo-Dopamine & Three-view state pairs; reference endpoints; ${\sim}3$-Hz increments & Fused relative progress & Native critic & -- \\
    \rowcolor{RVCompareStripe}
    RoboFAC & ${\sim}1$-Hz prefix for both value and SIA & Expected A--E rubric level & Value difference & Generated description \\
    RoboMeter & Eight uniformly sampled prefix frames & Final progress-head output & Value difference & -- \\
    \rowcolor{RVCompareStripe}
    RoboReward & Causal prefix; native frame selection & Five-level rating mapped to $[0,1]$ & Value difference & -- \\
    RynnValue & Eight uniformly sampled prefix frames & Negative remaining time & Value difference & -- \\
    \rowcolor{RVCompareStripe}
    TOPReward & Causal prefix sampled at ${\sim}2$ Hz & Final \texttt{True}-token log probability & Value difference & -- \\
    VLAC & Front-view pairs; six reference frames in one-shot mode & Accumulated relative progress & Native critic & -- \\
    \bottomrule
  \end{tabularx}
  \endgroup
\end{table}

\noindent\textbf{FailSafe.}
\emph{Subtask description.} FailSafe's structured response contains a current-subtask description followed by a yes/no failure verdict and, when a failure is predicted, its type and a corrective action. We retain the text preceding the failure verdict, remove an explicit \texttt{Subtask:} prefix when present, and pass the resulting description to the SIA judge.

\noindent\textbf{LIV.}
\emph{Scalar value.} LIV encodes the current image and task instruction separately, without observation history or a reference demonstration. We use their embedding cosine similarity, clipped to $[-1,1]$, as $v(A;l)$.
\emph{Pairwise comparison.} We take the difference between the two image--text similarities.

\noindent\textbf{ProcVLM.}
For one-shot adaptation, the selected training demonstration is converted into supervised question--answer examples using the model's subtask-structured format~\citep{feng2026procvlm}. Coarse subtask segments support an ordered list beginning with the current unfinished action, followed by later actions. Progress targets combine subtask structure with visual change within each stage. Each example pairs recent observations and the task instruction with the remaining-action list and a \texttt{<progress>} percentage; completed states receive a completion response. These adaptation targets are constructed from the selected training demonstration, independently of the benchmark's test annotations.

\emph{Scalar value.} The adapter places at most 512 uniformly spaced points on the full logical timeline and retains those up to the query state. It appends the query frame if it falls off this grid, then selects the latest eight points. Early windows are padded with their first point. One-shot adaptation uses a task-specific low-rank adaptation (LoRA) module~\citep{hu2022lora}; reference frames are not inserted into the query. A generated progress percentage $p$ gives $v(A;l)=p/100$ without clipping. A finished-task phrase maps to 1. If progress cannot be parsed, the query is retried once, then uses an earlier cached value from the same logical trajectory, or 0 if none exists.
\emph{Pairwise comparison.} Comparisons use the scalar-value difference defined above.

\emph{Subtask description.} We extract the first remaining action, which represents the current unfinished subtask. The parser also accepts recognized prose and direct action answers. A finished-task phrase or a parsed 100\% tag returns \texttt{All tasks are completed.} An unparseable answer is retried once and then reported as an error.

\noindent\textbf{Robo-Dopamine.}
The native critic compares three-view before/after states conditioned on the task instruction and the front-view endpoints of a reference demonstration~\citep{tan2025robo}. Its signed \texttt{<score>} percentage, divided by 100, defines $d(A_1,A_2;l)$. When the reference goal $G$ serves as a state endpoint, its front image fills all three view slots.

\emph{Scalar value.} We sample the logical prefix $A_0,\ldots,A_m=A$ at approximately 3 Hz, retaining the query frame. With $\delta_i=d(A_{i-1},A_i;l)$ and $r_0=0$, the released accumulation rule is
\begin{equation}
    r_i=\begin{cases}
        r_{i-1}+(1-r_{i-1})\delta_i, & \delta_i\geq0,\\
        r_{i-1}+r_{i-1}\delta_i, & \delta_i<0.
    \end{cases}
    \label{eq:robodopamine_update}
\end{equation}
The accumulated estimate is fused with initial- and goal-anchored predictions following Robo-Dopamine~\citep{tan2025robo}, without clipping:
\begin{equation}
    v(A;l)=\frac{d(A_0,A;l)+1+d(G,A;l)+r_m}{3}.
    \label{eq:robodopamine_value}
\end{equation}
\emph{Pairwise comparison.} We retain the native directional prediction, $c(A_1,A_2;l)=d(A_1,A_2;l)$.

\noindent\textbf{RoboFAC.}
For both scalar and subtask queries, the adapter samples the causal prefix at approximately 1 Hz, retains the initial and query frames, and resizes images to $512\times512$.
\emph{Scalar value.} A five-level A--E rubric asks how much task progress remains valid at the end of the observed prefix. For the five next-token logits $z$, we compute $v(A;l)=\sum_{i=0}^{4}(i/4)\,\mathrm{softmax}(z)_i$. The rubric and equal level spacing define the adapter's readout, rather than a calibrated native progress scale.
\emph{Pairwise comparison.} Comparisons use the scalar-value difference defined above.
\emph{Subtask description.} The generated description is passed unchanged to the SIA judge.

\noindent\textbf{RoboMeter.}
\emph{Scalar value.} The adapter selects eight uniformly spaced prefix indices, rounded to the nearest frame. Short prefixes retain all frames and repeat the query frame to fill the input. We take the final progress-head prediction, $v(A;l)=\hat p_8$. The preference and success heads are not used; reported SA is computed from these progress values under the shared protocol.
\emph{Pairwise comparison.} Comparisons use the scalar-value difference defined above.

\noindent\textbf{RoboReward.}
\emph{Scalar value.} The native video processor selects frames from the causal prefix, with a single-frame prefix duplicated. A generated answer \texttt{ANSWER: k}, where $k\in\{1,\ldots,5\}$, maps to $v(A;l)=(k-1)/4$. Predictions therefore retain the model's five discrete completion levels. Reported SA uses these prefix values under the shared protocol.
\emph{Pairwise comparison.} Comparisons use the scalar-value difference defined above.

\noindent\textbf{RynnValue.}
\emph{Scalar value.} Eight uniformly spaced prefix indices are rounded down, allowing repeated frames in short prefixes. The final absolute remaining-time prediction $\hat\tau(A;l)$ gives $v(A;l)=-\hat\tau(A;l)$, retaining seconds as its unit. The relative-displacement head and language outputs are not used.
\emph{Pairwise comparison.} We take the difference between the two sign-reversed remaining-time estimates.

\noindent\textbf{TOPReward.}
\emph{Scalar value.} The adapter samples the causal prefix at approximately 2 Hz, subject to the native processor's frame limits, and duplicates a single-frame prefix. For the resulting visual--language prompt $X(A,l)$, it returns $v(A;l)=\log P_\theta(\mathtt{True}\mid X(A,l))$, the log probability of the final supplied affirmative answer token. No textual verdict is generated, and no episode-level min--max normalization is applied.
\emph{Pairwise comparison.} Comparisons use the scalar-value difference defined above.

\noindent\textbf{VLAC.}
The native critic's signed percentage, divided by 100, defines $d(A_1,A_2;l)$. One-shot inputs include six uniformly selected reference frames.

\emph{Scalar value.} The adapter samples front-view frames on the released nominal-5-Hz integer-step grid and evaluates every fifth sampled position. A query frame outside these positions adds a final residual pair. Value queries use the episode's initial image as the start reference. For normalized pair outputs $\delta_i$, the released \texttt{simple=True} accumulation rule~\citep{zhai2025vision} is
\begin{equation}
    u_0=0,\qquad u_i=u_{i-1}+(1-u_{i-1})\delta_i,\qquad v(A;l)=u_m.
    \label{eq:vlac_value}
\end{equation}
The same update applies to positive and negative increments, without clipping. \emph{Pairwise comparison.} We use $c(A_1,A_2;l)=d(A_1,A_2;l)$ directly, with $A_1$ as the start image for a standalone comparison. Identical query states return 0.

\subsection{Leaderboard Scoring}
\label{app:leaderboard_scoring}

\noindent\textbf{Score Normalization and Domain Aggregation.}
We normalize metrics using their theoretical ranges, so scores do not depend on which other models enter the leaderboard. For the reported $\times100$ values, VOC, Memory-VOC, VROC, and CSVC are mapped by $(x+100)/2$, and FPL by $100-x$; the other metrics retain their percentage scores. For metrics evaluated in both ID and OOD settings in Tables~\ref{tab:robovalue_main} and~\ref{tab:robovalue_ood}, we first average the cross-embodiment and cross-environment scores to obtain OOD performance, then average ID and OOD. Their effective weights are therefore $1/2$, $1/4$, and $1/4$ for ID, cross-embodiment, and cross-environment, respectively. TRR and CSVC use their ID scores. Missing metrics contribute zero after normalization, without redistributing their weights. This applies to the unmeasured VS and CSVC of RoboReward-4B and RoboReward-8B.

\noindent\textbf{Dimension Scores.}
Task-State Understanding gives equal weight to SA and TGA, with TGA itself averaging TGA-CT and TGA-CF; SIA is reported separately. Temporal Progress Monitoring gives half its weight to the mean of VOC and Memory-VOC and half to VROC. Here, VROC is the reverse-half Spearman correlation from the continuous forward--reverse protocol used by Cycle-VOC, measured against decreasing progress order. VROC is averaged equally across ID, cross-embodiment, and cross-environment before correlation normalization. The forward context is retained during reverse playback, as specified in Section~\ref{sec:metrics}. Cycle-VOC is not added separately, avoiding repeated weighting of its forward component. Failure and Recovery Reasoning averages normalized FPL and TRR; Value Consistency averages normalized VS and CSVC. The overall score is the arithmetic mean of the four dimension scores, computed before rounding.

\FloatBarrier
\Needspace{7\baselineskip}
\section{Sensitivity to Evaluation Settings}
\label{app:sensitivity_analysis}

\noindent\textbf{Experimental Setup.}
We examine how the evaluation settings in Section~\ref{sec:metrics} affect metric scores and model rankings in simulation and real-world environments. RoboMeter-4B and RynnValue-4B are evaluated zero-shot, while ProcVLM-2B and Robo-Dopamine 2.0-8B Preview use their one-shot configurations. For each metric and environment, all models and parameter settings use the same fixed trajectory subset drawn from the tasks in Table~\ref{tab:sensitivity_tasks}.

The analysis covers eight metrics and eleven parameter sweeps (Table~\ref{tab:sensitivity_parameters}). Each sweep varies one parameter around its reference value while holding the remaining evaluation settings, model configurations, and instructions fixed. We compare both model rankings and absolute score changes, with simulation and real-world results analyzed separately.

Ranking stability is measured by the fraction of reference pairwise orderings preserved at the two alternative parameter values. Pairs tied at the reference setting are excluded; a strict ordering that becomes tied counts as not preserved. Table~\ref{tab:pairwise_preservation} reports preserved and eligible comparisons for each sweep.

\begin{table}[htbp]
\centering
\caption{\textbf{Task Subsets for Sensitivity Analysis.} CSVC uses a separate task subset. All models and parameter settings use the same trajectories within each metric and domain.}
\label{tab:sensitivity_tasks}
\begingroup
\normalsize
\setlength{\tabcolsep}{6pt}
\renewcommand{\arraystretch}{1.12}
\arrayrulecolor{RVSensRule}
\begin{tabularx}{\linewidth}{@{}>{\raggedright\arraybackslash}p{0.18\linewidth}
                              >{\raggedright\arraybackslash}X
                              >{\raggedright\arraybackslash}X@{}}
\toprule
\textbf{Metric Group}
& \textcolor{RVSensSimInk}{\textbf{Simulation}}
& \textcolor{RVSensRealInk}{\textbf{Real-world}} \\
\midrule
\makecell[tl]{Other tested\\metrics}
& \makecell[tl]{Fill Pen Holder\\Make Toast\\Organize Table\\Pack Objects into Box\\Stack Blocks}
& \makecell[tl]{Fill and Place Pen Holder\\Organize Desktop Workstation\\Pack Items for Shipping\\Prepare Breakfast\\Sort Blocks in Order} \\
\addlinespace[6pt]
\midrule
CSVC
& \makecell[tl]{Fill Pen Holder\\Hang Mugs\\Pack Objects into Box\\Stack Blocks}
& \makecell[tl]{Arrange Family Tea Set\\Fill and Place Pen Holder\\Load Plastic Plates into Rack\\Organize Desktop Workstation\\Set the Table} \\
\bottomrule
\end{tabularx}
\endgroup
\end{table}

\begin{table}[htbp]
\centering
\caption{\textbf{Parameter Grids for Sensitivity Analysis.}
Purple text marks the reference values. Parameters are varied one at a time, using the same grids in simulation and real-world evaluations.}
\label{tab:sensitivity_parameters}
\begingroup
\small
\setlength{\tabcolsep}{6pt}
\renewcommand{\arraystretch}{1.20}
\arrayrulecolor{RVSensRule}
\begin{tabularx}{\linewidth}{@{}l>{\raggedright\arraybackslash}X
                              c>{\color{RVSensRefInk}\bfseries\boldmath}cc@{}}
\toprule
\multirow{2}{*}{\textbf{Metric}}
& \multirow{2}{*}{\textbf{Evaluation parameter}}
& \multicolumn{3}{c@{}}{\textbf{Evaluated values}} \\
\cmidrule(l){3-5}
& & Lower & Reference & Higher \\
\midrule
SA
& Terminal-window proportion
& $1\%$ & $2\%$ & $4\%$ \\
TGA-CT
& Endpoint-window proportion
& $1\%$ & $2\%$ & $4\%$ \\
TGA-CF
& Endpoint-window proportion
& $1\%$ & $2\%$ & $4\%$ \\
\addlinespace[4pt]
\multirow{2}{*}{FPL}
& Temporal sampling rate
& $2.5$ Hz & $5$ Hz & $10$ Hz \\

& ZigZag reversal coefficient $\alpha$
& $0.1$ & $0.2$ & $0.3$ \\
\addlinespace[4pt]
\multirow{2}{*}{TRR}
& Temporal sampling rate
& $5$ Hz & $10$ Hz & $15$ Hz \\

& Minimum number of sampled points $K$
& $6$ & $10$ & $16$ \\
\addlinespace[4pt]
\multirow{2}{*}{VS}
& Temporal sampling rate
& $2.5$ Hz & $5$ Hz & $10$ Hz \\

& Number of points averaged at each endpoint
& $1$ & $3$ & $5$ \\
\addlinespace[4pt]
Cycle-VOC
& Sampling density relative to the reference
& $0.5\times$ & $1\times$ & $2\times$ \\
CSVC
& Number of boundary pairs
& $2$ & $4$ & $6$ \\
\bottomrule
\end{tabularx}
\endgroup
\end{table}

\begin{table}[htbp]
\centering
\caption{\textbf{Pairwise Ranking Stability.} Entries report preserved pairwise model orderings over eligible comparisons, pooled across both alternative parameter values. Summary rows combine all sweeps.}
\label{tab:pairwise_preservation}
\begingroup
\small
\setlength{\tabcolsep}{6pt}
\renewcommand{\arraystretch}{1.22}
\arrayrulecolor{RVSensRule}
\begin{tabularx}{\linewidth}{@{}l>{\raggedright\arraybackslash}X
                              >{\centering\arraybackslash}p{0.15\linewidth}
                              >{\centering\arraybackslash}p{0.15\linewidth}@{}}
\toprule
\multirow{2}{*}{\textbf{Metric}}
& \multirow{2}{*}{\textbf{Evaluation parameter}}
& \multicolumn{2}{c@{}}{\textbf{Preserved / eligible}} \\
\cmidrule{3-4}
& & \textcolor{RVSensSimInk}{\textbf{Simulation}}
& \textcolor{RVSensRealInk}{\textbf{Real-world}} \\
\midrule
SA & Terminal-window proportion & \textcolor{RVSensSimInk}{11/12} & \textcolor{RVSensRealInk}{9/10$^{*}$} \\
TGA-CT & Endpoint-window proportion & \textcolor{RVSensSimInk}{12/12} & \textcolor{RVSensRealInk}{11/12} \\
TGA-CF & Endpoint-window proportion & \textcolor{RVSensSimInk}{12/12} & \textcolor{RVSensRealInk}{12/12} \\
\addlinespace[4pt]
\multirow{2}{*}{FPL} & Temporal sampling rate & \textcolor{RVSensSimInk}{11/12} & \textcolor{RVSensRealInk}{12/12} \\
& Reversal coefficient $\alpha$ & \textcolor{RVSensSimInk}{11/12} & \textcolor{RVSensRealInk}{12/12} \\
\addlinespace[4pt]
\multirow{2}{*}{TRR} & Temporal sampling rate & \textcolor{RVSensSimInk}{12/12} & \textcolor{RVSensRealInk}{11/12} \\
\bottomrule
\end{tabularx}
\par\vspace{4pt}
\begin{minipage}{\linewidth}
\footnotesize\raggedright
$^{*}$ RoboMeter and Robo-Dopamine are tied at the reference setting for real-world SA. Excluding this pair leaves five eligible pairs at each alternative setting, giving ten comparisons in total.
\end{minipage}
\endgroup
\end{table}

\begin{figure}[tp]
    \centering
    \includegraphics[width=0.80\linewidth,height=0.45\textheight,keepaspectratio]{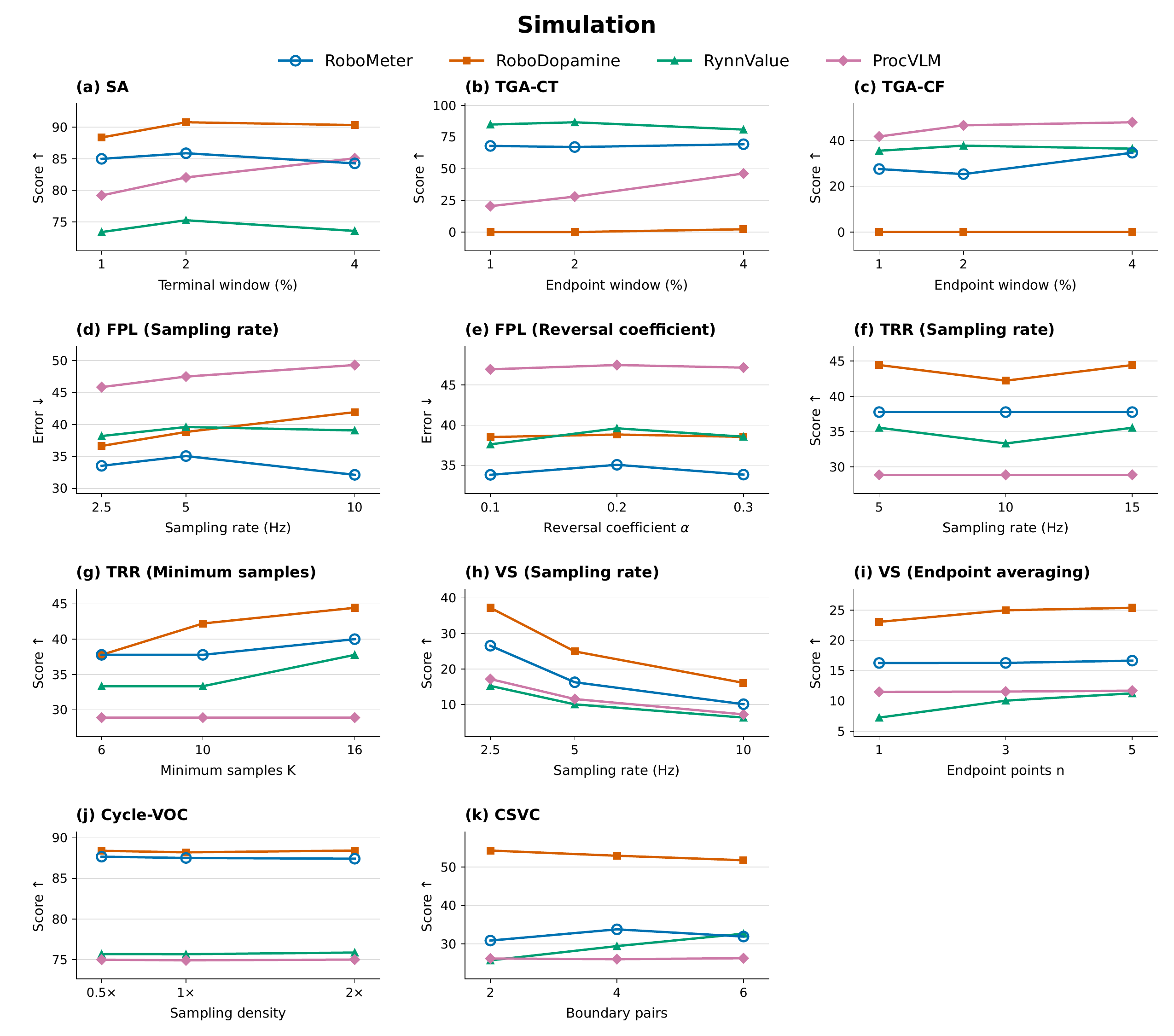}
    \par\vspace{6pt}
    \includegraphics[width=0.80\linewidth,height=0.45\textheight,keepaspectratio]{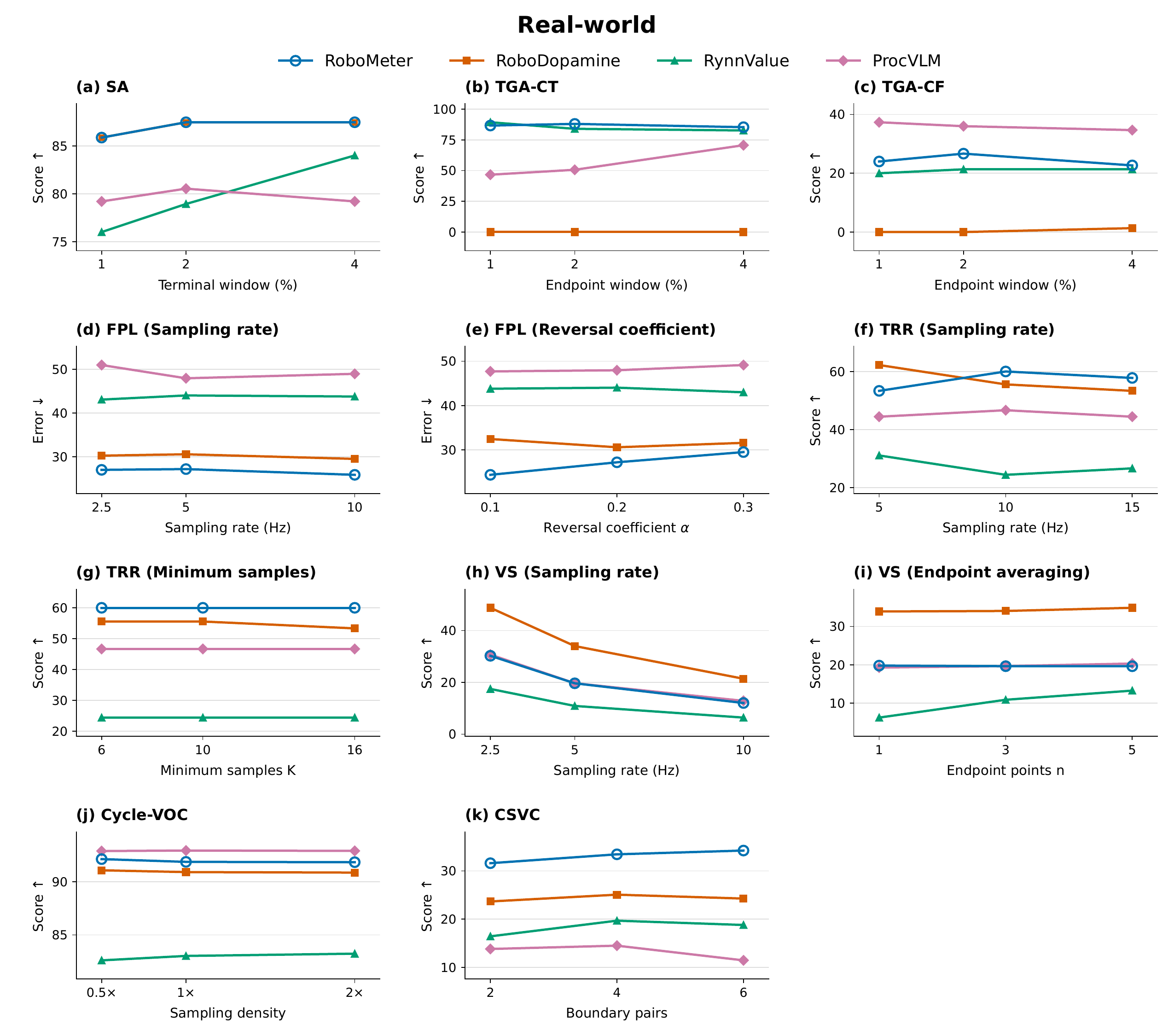}

    \caption{\textbf{Sensitivity to Evaluation Settings.}
    Results for eleven parameter sweeps across eight metrics in simulation (top) and the real world (bottom).
    Scores are multiplied by 100; lower FPL and higher scores elsewhere are better.}
    \label{fig:metric_robustness}
\end{figure}

\noindent\textbf{Overall Findings.}
Across the tested parameter values, 126 of 132 pairwise comparisons are preserved in simulation (95.45\%), and 126 of 130 are preserved in the real world (96.92\%). Figure~\ref{fig:metric_robustness} shows that model rankings remain largely stable despite changes in absolute scores. The exceptions involve a small number of model pairs, as detailed below.

\noindent\textbf{Endpoint Windows: SA and TGA.}
SA preserves most pairwise orderings across the tested terminal windows. At the 4\% window, RoboMeter and ProcVLM exchange positions in simulation, while RynnValue and ProcVLM exchange positions in the real-world setting. Both are adjacent pairs in their reference rankings, with score gaps of 3.82 and 1.60 percentage points, respectively. All remaining orderings are preserved.

TGA-CT preserves the full model ranking across endpoint windows of 1\%, 2\%, and 4\% in simulation. In the real-world setting, only RoboMeter and RynnValue exchange positions at 1\%; the rankings at 2\% and 4\% are identical in both environments. Examining ProcVLM's episode-level predictions shows that its marked score increase from 2\% to 4\% mainly reflects smaller maximum gains assigned to mismatched instructions, rather than a consistent increase in gains under the correct instruction. This change in absolute scores does not alter the aggregate model ordering. For TGA-CF, all four models retain their ordering across the three windows in both environments.

\noindent\textbf{Temporal Settings: FPL and TRR.}
For FPL, simulation ranking changes are confined to Robo-Dopamine and RynnValue, whose scores remain close across the tested settings. Their ordering reverses at 10 Hz in the sampling-rate sweep and at $\alpha=0.1$ in the reversal-coefficient sweep. The real-world ranking is unchanged in both sweeps. These results indicate that FPL preserves broad differences between models, while the ordering of closely matched models can depend on the evaluation parameters.

For TRR, the model rankings are identical at 10 and 15 Hz in both environments. At 5 Hz, the simulation ranking remains unchanged, whereas RoboMeter and Robo-Dopamine exchange positions in the real-world setting. This isolated reversal suggests sensitivity to coarse temporal sampling, although the sweep does not establish its cause. Varying the minimum sample count produces no strict ranking reversals; Robo-Dopamine and RoboMeter become tied at $K=6$ in simulation.

\noindent\textbf{Value Stability and Consistency: VS and CSVC.}
VS preserves model ordering across all tested sampling rates in both environments. Varying the endpoint-averaging window also preserves the simulation ranking. In the real-world setting, RoboMeter and ProcVLM exchange positions at $n=1$, with only a small score difference. Absolute VS scores decrease as sampling becomes denser for all four models. This is consistent with the efficiency-ratio term: additional samples can capture local fluctuations that increase total variation without a proportional increase in net endpoint change. Sampling density can also affect the non-flat-time component, and this sweep does not isolate the contribution of either term. The stable rankings support using a common sampling rate for model comparison, while absolute VS scores should be compared under matched sampling settings.

For CSVC, RynnValue and ProcVLM exchange positions at two boundary pairs in simulation, while RoboMeter and RynnValue exchange positions at six pairs. The score gaps after these reversals are 0.47 and 0.75 points, respectively. All other simulation orderings are preserved, and the full real-world ranking remains unchanged across the tested pair counts.

\noindent\textbf{Cycle Sampling.}
Cycle-VOC preserves model ordering across all tested sampling densities in both environments. The largest absolute score deviations from the reference are 0.22 points in simulation and 0.42 points in the real world.

These findings apply to the evaluated task subsets and parameter ranges.

\FloatBarrier

\makeatletter
\setlength{\@fptop}{0pt}
\setlength{\@fpsep}{8pt}
\setlength{\@fpbot}{0pt plus 1fil}
\setlength{\@dblfptop}{0pt}
\setlength{\@dblfpsep}{8pt}
\setlength{\@dblfpbot}{0pt plus 1fil}
\makeatother
\setlength{\floatsep}{8pt}
\setlength{\dblfloatsep}{8pt}
\setlength{\intextsep}{3pt}

\FloatBarrier
\section{Full Simulation Benchmark Results}\label{app:simulation_results}
\subsection{Main Results}\label{app:simulation_id_results}

The following tables report in-domain (ID) results on simulation trajectories collected with the standard ARX setup. These results provide the reference for the out-of-domain (OOD) comparisons below. Each table presents one metric, with task identifiers defined beneath it.


\begin{table}[H]
\renewcommand{\scriptsize}{\fontsize{7}{8}\selectfont}
\renewcommand{\footnotesize}{\fontsize{9}{9.5}\selectfont}
\setlength{\belowcaptionskip}{4pt}
\centering
\caption{\textbf{SA Results on In-Domain Simulation Trajectories.}}
\label{tab:id_sa_part1}
\label{tab:id_sa}
\begingroup
\footnotesize
\rmfamily
\setlength{\tabcolsep}{0.9pt}
\renewcommand{\arraystretch}{1.0}
\def\taskmodeband#1#2{%
  \specialrule{\lightrulewidth}{\aboverulesep}{2pt}
  \multicolumn{14}{@{}c@{}}{%
    \makebox[\textwidth][c]{%
      \rlap{\smash{\color{#1}%
        \rule[-\dp\csname @arstrutbox\endcsname]{\textwidth}%
        {\dimexpr\ht\csname @arstrutbox\endcsname
        +\dp\csname @arstrutbox\endcsname\relax}}}%
      \makebox[\textwidth][c]{\textit{#2}}}}\\%
  \specialrule{\lightrulewidth}{2pt}{\belowrulesep}
}
\begin{tabularx}{\textwidth}{
  @{} >{\raggedright\arraybackslash}p{0.24\textwidth}
  *{13}{>{\centering\arraybackslash}X} @{}
}
\toprule
\textbf{Model} & \textbf{T1} & \textbf{T2} & \textbf{T3} & \textbf{T4} & \textbf{T5} & \textbf{T6} & \textbf{T7} & \textbf{T8} & \textbf{T9} & \textbf{T10} & \textbf{T11} & \textbf{T12} & \textbf{T13} \\
\taskmodeband{RVZeroBand}{Zero-Shot Evaluation}
LIV & {\scriptsize\mbox{\textbf{55.09}}} & {\scriptsize\mbox{92.76}} & {\scriptsize\mbox{37.04}} & {\scriptsize\mbox{48.35}} & {\scriptsize\mbox{51.98}} & {\scriptsize\mbox{51.39}} & {\scriptsize\mbox{45.06}} & {\scriptsize\mbox{66.90}} & {\scriptsize\mbox{28.17}} & {\scriptsize\mbox{91.90}} & {\scriptsize\mbox{63.72}} & {\scriptsize\mbox{49.54}} & {\scriptsize\mbox{81.48}} \\
ProcVLM-2B & {\scriptsize\mbox{37.50}} & {\scriptsize\mbox{\textbf{100.00}}} & {\scriptsize\mbox{27.31}} & {\scriptsize\mbox{\textbf{92.36}}} & {\scriptsize\mbox{95.24}} & {\scriptsize\mbox{\textbf{77.08}}} & {\scriptsize\mbox{\textbf{72.84}}} & {\scriptsize\mbox{98.62}} & {\scriptsize\mbox{73.41}} & {\scriptsize\mbox{21.06}} & {\scriptsize\mbox{44.44}} & {\scriptsize\mbox{64.81}} & {\scriptsize\mbox{89.81}} \\
RoboFAC-7B & {\scriptsize\mbox{42.82}} & {\scriptsize\mbox{99.67}} & {\scriptsize\mbox{56.48}} & {\scriptsize\mbox{76.74}} & {\scriptsize\mbox{\textbf{100.00}}} & {\scriptsize\mbox{40.97}} & {\scriptsize\mbox{66.67}} & {\scriptsize\mbox{88.52}} & {\scriptsize\mbox{48.81}} & {\scriptsize\mbox{81.02}} & {\scriptsize\mbox{96.70}} & {\scriptsize\mbox{\textbf{87.27}}} & {\scriptsize\mbox{99.54}} \\
RoboMeter-4B & {\scriptsize\mbox{49.54}} & {\scriptsize\mbox{94.81}} & {\scriptsize\mbox{\textbf{98.38}}} & {\scriptsize\mbox{79.95}} & {\scriptsize\mbox{99.21}} & {\scriptsize\mbox{71.06}} & {\scriptsize\mbox{69.60}} & {\scriptsize\mbox{91.67}} & {\scriptsize\mbox{56.75}} & {\scriptsize\mbox{70.60}} & {\scriptsize\mbox{99.05}} & {\scriptsize\mbox{79.17}} & {\scriptsize\mbox{98.84}} \\
\bottomrule
\end{tabularx}
\renewcommand{\scriptsize}{\fontsize{6}{7}\selectfont}
\par\vspace{1pt}
\parbox{\textwidth}{\scriptsize\raggedright \textbf{T1}: Arrange largest number.\enspace \textbf{T2}: Fill pen holder.\enspace \textbf{T3}: Fold clothes.\enspace \textbf{T4}: Hang mugs.\enspace \textbf{T5}: Insert tubes.\enspace \textbf{T6}: Make toast.\enspace \textbf{T7}: Organize table.\enspace \textbf{T8}: Pack objects into box.\enspace \textbf{T9}: Play stacking toy.\enspace \textbf{T10}: Put bottles into dustbin.\enspace \textbf{T11}: Stack blocks.\enspace \textbf{T12}: Store laptop and headphone.\enspace \textbf{T13}: Sweep blocks.}
\par\vspace{1pt}
\parbox{\textwidth}{\scriptsize\raggedright
$^{\dagger}$ 2.0 Preview checkpoints.\quad
$^{\ddagger}$ Qwen3-VL backbones.
}
\endgroup
\end{table}

\begin{table}[H]
\renewcommand{\scriptsize}{\fontsize{7}{8}\selectfont}
\renewcommand{\footnotesize}{\fontsize{9}{9.5}\selectfont}
\setlength{\belowcaptionskip}{4pt}
\centering
\caption{\textbf{TGA-CT Results on In-Domain Simulation Trajectories.}}
\label{tab:id_tga_ct_part1}
\label{tab:id_tga_ct}
\label{tab:id_tga_easy_part1}
\label{tab:id_tga_easy}
\begingroup
\footnotesize
\rmfamily
\setlength{\tabcolsep}{0.9pt}
\renewcommand{\arraystretch}{1.0}
\def\taskmodeband#1#2{%
  \specialrule{\lightrulewidth}{\aboverulesep}{2pt}
  \multicolumn{14}{@{}c@{}}{%
    \makebox[\textwidth][c]{%
      \rlap{\smash{\color{#1}%
        \rule[-\dp\csname @arstrutbox\endcsname]{\textwidth}%
        {\dimexpr\ht\csname @arstrutbox\endcsname
        +\dp\csname @arstrutbox\endcsname\relax}}}%
      \makebox[\textwidth][c]{\textit{#2}}}}\\%
  \specialrule{\lightrulewidth}{2pt}{\belowrulesep}
}
\begin{tabularx}{\textwidth}{
  @{} >{\raggedright\arraybackslash}p{0.24\textwidth}
  *{13}{>{\centering\arraybackslash}X} @{}
}
\toprule
\textbf{Model} & \textbf{T1} & \textbf{T2} & \textbf{T3} & \textbf{T4} & \textbf{T5} & \textbf{T6} & \textbf{T7} & \textbf{T8} & \textbf{T9} & \textbf{T10} & \textbf{T11} & \textbf{T12} & \textbf{T13} \\
\taskmodeband{RVZeroBand}{Zero-Shot Evaluation}
LIV & {\scriptsize\mbox{0.00}} & {\scriptsize\mbox{13.33}} & {\scriptsize\mbox{6.67}} & {\scriptsize\mbox{80.00}} & {\scriptsize\mbox{0.00}} & {\scriptsize\mbox{0.00}} & {\scriptsize\mbox{0.00}} & {\scriptsize\mbox{0.00}} & {\scriptsize\mbox{6.67}} & {\scriptsize\mbox{6.67}} & {\scriptsize\mbox{0.00}} & {\scriptsize\mbox{26.67}} & {\scriptsize\mbox{53.33}} \\
ProcVLM-2B & {\scriptsize\mbox{0.00}} & {\scriptsize\mbox{0.00}} & {\scriptsize\mbox{13.33}} & {\scriptsize\mbox{0.00}} & {\scriptsize\mbox{0.00}} & {\scriptsize\mbox{33.33}} & {\scriptsize\mbox{33.33}} & {\scriptsize\mbox{13.33}} & {\scriptsize\mbox{26.67}} & {\scriptsize\mbox{0.00}} & {\scriptsize\mbox{0.00}} & {\scriptsize\mbox{20.00}} & {\scriptsize\mbox{46.67}} \\
RoboFAC-7B & {\scriptsize\mbox{0.00}} & {\scriptsize\mbox{0.00}} & {\scriptsize\mbox{0.00}} & {\scriptsize\mbox{0.00}} & {\scriptsize\mbox{0.00}} & {\scriptsize\mbox{0.00}} & {\scriptsize\mbox{33.33}} & {\scriptsize\mbox{0.00}} & {\scriptsize\mbox{0.00}} & {\scriptsize\mbox{73.33}} & {\scriptsize\mbox{0.00}} & {\scriptsize\mbox{0.00}} & {\scriptsize\mbox{0.00}} \\
RoboMeter-4B & {\scriptsize\mbox{\textbf{100.00}}} & {\scriptsize\mbox{0.00}} & {\scriptsize\mbox{\textbf{100.00}}} & {\scriptsize\mbox{66.67}} & {\scriptsize\mbox{13.33}} & {\scriptsize\mbox{26.67}} & {\scriptsize\mbox{\textbf{100.00}}} & {\scriptsize\mbox{\textbf{100.00}}} & {\scriptsize\mbox{20.00}} & {\scriptsize\mbox{86.67}} & {\scriptsize\mbox{93.33}} & {\scriptsize\mbox{\textbf{100.00}}} & {\scriptsize\mbox{93.33}} \\
\bottomrule
\end{tabularx}
\renewcommand{\scriptsize}{\fontsize{6}{7}\selectfont}
\par\vspace{1pt}
\parbox{\textwidth}{\scriptsize\raggedright \textbf{T1}: Arrange largest number.\enspace \textbf{T2}: Fill pen holder.\enspace \textbf{T3}: Fold clothes.\enspace \textbf{T4}: Hang mugs.\enspace \textbf{T5}: Insert tubes.\enspace \textbf{T6}: Make toast.\enspace \textbf{T7}: Organize table.\enspace \textbf{T8}: Pack objects into box.\enspace \textbf{T9}: Play stacking toy.\enspace \textbf{T10}: Put bottles into dustbin.\enspace \textbf{T11}: Stack blocks.\enspace \textbf{T12}: Store laptop and headphone.\enspace \textbf{T13}: Sweep blocks.}
\par\vspace{1pt}
\parbox{\textwidth}{\scriptsize\raggedright
$^{\dagger}$ 2.0 Preview checkpoints.\quad
$^{\ddagger}$ Qwen3-VL backbones.
}
\endgroup
\end{table}

\begin{table}[H]
\renewcommand{\scriptsize}{\fontsize{7}{8}\selectfont}
\renewcommand{\footnotesize}{\fontsize{9}{9}\selectfont}
\setlength{\belowcaptionskip}{4pt}
\centering
\caption{\textbf{TGA-CF Results on In-Domain Simulation Trajectories.}}
\label{tab:id_tga_cf_part1}
\label{tab:id_tga_cf}
\label{tab:id_tga_hard_part1}
\label{tab:id_tga_hard}
\begingroup
\footnotesize
\rmfamily
\setlength{\tabcolsep}{0.9pt}
\renewcommand{\arraystretch}{1.0}
\def\taskmodeband#1#2{%
  \specialrule{\lightrulewidth}{\aboverulesep}{2pt}
  \multicolumn{14}{@{}c@{}}{%
    \makebox[\textwidth][c]{%
      \rlap{\smash{\color{#1}%
        \rule[-\dp\csname @arstrutbox\endcsname]{\textwidth}%
        {\dimexpr\ht\csname @arstrutbox\endcsname
        +\dp\csname @arstrutbox\endcsname\relax}}}%
      \makebox[\textwidth][c]{\textit{#2}}}}\\%
  \specialrule{\lightrulewidth}{2pt}{\belowrulesep}
}
\begin{tabularx}{\textwidth}{
  @{} >{\raggedright\arraybackslash}p{0.24\textwidth}
  *{13}{>{\centering\arraybackslash}X} @{}
}
\toprule
\textbf{Model} & \textbf{T1} & \textbf{T2} & \textbf{T3} & \textbf{T4} & \textbf{T5} & \textbf{T6} & \textbf{T7} & \textbf{T8} & \textbf{T9} & \textbf{T10} & \textbf{T11} & \textbf{T12} & \textbf{T13} \\
\taskmodeband{RVZeroBand}{Zero-Shot Evaluation}
LIV & {\scriptsize\mbox{0.00}} & {\scriptsize\mbox{0.00}} & {\scriptsize\mbox{0.00}} & {\scriptsize\mbox{66.67}} & {\scriptsize\mbox{0.00}} & {\scriptsize\mbox{0.00}} & {\scriptsize\mbox{0.00}} & {\scriptsize\mbox{13.33}} & {\scriptsize\mbox{0.00}} & {\scriptsize\mbox{0.00}} & {\scriptsize\mbox{0.00}} & {\scriptsize\mbox{0.00}} & {\scriptsize\mbox{0.00}} \\
ProcVLM-2B & {\scriptsize\mbox{13.33}} & {\scriptsize\mbox{6.67}} & {\scriptsize\mbox{40.00}} & {\scriptsize\mbox{0.00}} & {\scriptsize\mbox{13.33}} & {\scriptsize\mbox{20.00}} & {\scriptsize\mbox{33.33}} & {\scriptsize\mbox{13.33}} & {\scriptsize\mbox{6.67}} & {\scriptsize\mbox{6.67}} & {\scriptsize\mbox{13.33}} & {\scriptsize\mbox{86.67}} & {\scriptsize\mbox{33.33}} \\
RoboFAC-7B & {\scriptsize\mbox{0.00}} & {\scriptsize\mbox{0.00}} & {\scriptsize\mbox{0.00}} & {\scriptsize\mbox{0.00}} & {\scriptsize\mbox{0.00}} & {\scriptsize\mbox{0.00}} & {\scriptsize\mbox{0.00}} & {\scriptsize\mbox{0.00}} & {\scriptsize\mbox{0.00}} & {\scriptsize\mbox{20.00}} & {\scriptsize\mbox{0.00}} & {\scriptsize\mbox{0.00}} & {\scriptsize\mbox{0.00}} \\
RoboMeter-4B & {\scriptsize\mbox{\textbf{80.00}}} & {\scriptsize\mbox{20.00}} & {\scriptsize\mbox{26.67}} & {\scriptsize\mbox{0.00}} & {\scriptsize\mbox{13.33}} & {\scriptsize\mbox{\textbf{53.33}}} & {\scriptsize\mbox{13.33}} & {\scriptsize\mbox{93.33}} & {\scriptsize\mbox{6.67}} & {\scriptsize\mbox{60.00}} & {\scriptsize\mbox{66.67}} & {\scriptsize\mbox{0.00}} & {\scriptsize\mbox{0.00}} \\
\bottomrule
\end{tabularx}
\renewcommand{\scriptsize}{\fontsize{6}{7}\selectfont}
\par\vspace{1pt}
\parbox{\textwidth}{\scriptsize\raggedright \textbf{T1}: Arrange largest number.\enspace \textbf{T2}: Fill pen holder.\enspace \textbf{T3}: Fold clothes.\enspace \textbf{T4}: Hang mugs.\enspace \textbf{T5}: Insert tubes.\enspace \textbf{T6}: Make toast.\enspace \textbf{T7}: Organize table.\enspace \textbf{T8}: Pack objects into box.\enspace \textbf{T9}: Play stacking toy.\enspace \textbf{T10}: Put bottles into dustbin.\enspace \textbf{T11}: Stack blocks.\enspace \textbf{T12}: Store laptop and headphone.\enspace \textbf{T13}: Sweep blocks.}
\par\vspace{1pt}
\parbox{\textwidth}{\scriptsize\raggedright
$^{\dagger}$ 2.0 Preview checkpoints.\quad
$^{\ddagger}$ Qwen3-VL backbones.
}
\endgroup
\end{table}

\begin{table}[H]
\renewcommand{\scriptsize}{\fontsize{7}{8}\selectfont}
\renewcommand{\footnotesize}{\fontsize{9}{9.5}\selectfont}
\setlength{\belowcaptionskip}{4pt}
\centering
\caption{\textbf{SIA Results on In-Domain Simulation Trajectories.}}
\label{tab:id_sia_part1}
\label{tab:id_sia}
\begingroup
\footnotesize
\rmfamily
\setlength{\tabcolsep}{0.9pt}
\renewcommand{\arraystretch}{1.0}
\def\taskmodeband#1#2{%
  \specialrule{\lightrulewidth}{\aboverulesep}{2pt}
  \multicolumn{14}{@{}c@{}}{%
    \makebox[\textwidth][c]{%
      \rlap{\smash{\color{#1}%
        \rule[-\dp\csname @arstrutbox\endcsname]{\textwidth}%
        {\dimexpr\ht\csname @arstrutbox\endcsname
        +\dp\csname @arstrutbox\endcsname\relax}}}%
      \makebox[\textwidth][c]{\textit{#2}}}}\\%
  \specialrule{\lightrulewidth}{2pt}{\belowrulesep}
}
\begin{tabularx}{\textwidth}{
  @{} >{\raggedright\arraybackslash}p{0.24\textwidth}
  *{13}{>{\centering\arraybackslash}X} @{}
}
\toprule
\textbf{Model} & \textbf{T1} & \textbf{T2} & \textbf{T3} & \textbf{T4} & \textbf{T5} & \textbf{T6} & \textbf{T7} & \textbf{T8} & \textbf{T9} & \textbf{T10} & \textbf{T11} & \textbf{T12} & \textbf{T13} \\
\taskmodeband{RVZeroBand}{Zero-Shot Evaluation}
ProcVLM-2B & {\scriptsize\mbox{10.65}} & {\scriptsize\mbox{\textbf{12.36}}} & {\scriptsize\mbox{1.75}} & {\scriptsize\mbox{18.25}} & {\scriptsize\mbox{8.21}} & {\scriptsize\mbox{\textbf{26.28}}} & {\scriptsize\mbox{\textbf{9.44}}} & {\scriptsize\mbox{6.47}} & {\scriptsize\mbox{5.39}} & {\scriptsize\mbox{16.30}} & {\scriptsize\mbox{21.39}} & {\scriptsize\mbox{7.72}} & {\scriptsize\mbox{\textbf{23.53}}} \\
RoboFAC-7B & {\scriptsize\mbox{\textbf{11.99}}} & {\scriptsize\mbox{8.82}} & {\scriptsize\mbox{6.66}} & {\scriptsize\mbox{\textbf{19.58}}} & {\scriptsize\mbox{16.25}} & {\scriptsize\mbox{12.82}} & {\scriptsize\mbox{6.62}} & {\scriptsize\mbox{4.83}} & {\scriptsize\mbox{8.80}} & {\scriptsize\mbox{15.29}} & {\scriptsize\mbox{\textbf{27.86}}} & {\scriptsize\mbox{1.64}} & {\scriptsize\mbox{1.65}} \\
FailSafe & {\scriptsize\mbox{9.76}} & {\scriptsize\mbox{3.17}} & {\scriptsize\mbox{\textbf{7.87}}} & {\scriptsize\mbox{19.30}} & {\scriptsize\mbox{\textbf{19.37}}} & {\scriptsize\mbox{23.76}} & {\scriptsize\mbox{2.63}} & {\scriptsize\mbox{\textbf{11.36}}} & {\scriptsize\mbox{\textbf{11.53}}} & {\scriptsize\mbox{\textbf{17.79}}} & {\scriptsize\mbox{18.83}} & {\scriptsize\mbox{\textbf{8.85}}} & {\scriptsize\mbox{11.08}} \\
\bottomrule
\end{tabularx}
\renewcommand{\scriptsize}{\fontsize{6}{7}\selectfont}
\par\vspace{1pt}
\parbox{\textwidth}{\scriptsize\raggedright \textbf{T1}: Arrange largest number.\enspace \textbf{T2}: Fill pen holder.\enspace \textbf{T3}: Fold clothes.\enspace \textbf{T4}: Hang mugs.\enspace \textbf{T5}: Insert tubes.\enspace \textbf{T6}: Make toast.\enspace \textbf{T7}: Organize table.\enspace \textbf{T8}: Pack objects into box.\enspace \textbf{T9}: Play stacking toy.\enspace \textbf{T10}: Put bottles into dustbin.\enspace \textbf{T11}: Stack blocks.\enspace \textbf{T12}: Store laptop and headphone.\enspace \textbf{T13}: Sweep blocks.}
\par\vspace{1pt}
\parbox{\textwidth}{\scriptsize\raggedright
$^{\dagger}$ 2.0 Preview checkpoints.\quad
$^{\ddagger}$ Qwen3-VL backbones.
}
\endgroup
\end{table}

\begin{table}[H]
\renewcommand{\scriptsize}{\fontsize{7}{8}\selectfont}
\renewcommand{\footnotesize}{\fontsize{9}{9.75}\selectfont}
\setlength{\belowcaptionskip}{4pt}
\centering
\caption{\textbf{VOC Results on In-Domain Simulation Trajectories.}}
\label{tab:id_voc_part1}
\label{tab:id_voc}
\begingroup
\footnotesize
\rmfamily
\setlength{\tabcolsep}{0.9pt}
\renewcommand{\arraystretch}{1.0}
\def\taskmodeband#1#2{%
  \specialrule{\lightrulewidth}{\aboverulesep}{2pt}
  \multicolumn{14}{@{}c@{}}{%
    \makebox[\textwidth][c]{%
      \rlap{\smash{\color{#1}%
        \rule[-\dp\csname @arstrutbox\endcsname]{\textwidth}%
        {\dimexpr\ht\csname @arstrutbox\endcsname
        +\dp\csname @arstrutbox\endcsname\relax}}}%
      \makebox[\textwidth][c]{\textit{#2}}}}\\%
  \specialrule{\lightrulewidth}{2pt}{\belowrulesep}
}
\begin{tabularx}{\textwidth}{
  @{} >{\raggedright\arraybackslash}p{0.24\textwidth}
  *{13}{>{\centering\arraybackslash}X} @{}
}
\toprule
\textbf{Model} & \textbf{T1} & \textbf{T2} & \textbf{T3} & \textbf{T4} & \textbf{T5} & \textbf{T6} & \textbf{T7} & \textbf{T8} & \textbf{T9} & \textbf{T10} & \textbf{T11} & \textbf{T12} & \textbf{T13} \\
\taskmodeband{RVZeroBand}{Zero-Shot Evaluation}
LIV & {\scriptsize\mbox{-33.16}} & {\scriptsize\mbox{42.81}} & {\scriptsize\mbox{20.17}} & {\scriptsize\mbox{76.13}} & {\scriptsize\mbox{-24.81}} & {\scriptsize\mbox{-37.39}} & {\scriptsize\mbox{-17.15}} & {\scriptsize\mbox{14.30}} & {\scriptsize\mbox{-4.29}} & {\scriptsize\mbox{43.03}} & {\scriptsize\mbox{-20.13}} & {\scriptsize\mbox{8.94}} & {\scriptsize\mbox{34.11}} \\
ProcVLM-2B & {\scriptsize\mbox{3.11}} & {\scriptsize\mbox{83.05}} & {\scriptsize\mbox{80.58}} & {\scriptsize\mbox{86.83}} & {\scriptsize\mbox{62.89}} & {\scriptsize\mbox{30.71}} & {\scriptsize\mbox{75.29}} & {\scriptsize\mbox{84.51}} & {\scriptsize\mbox{53.00}} & {\scriptsize\mbox{-5.36}} & {\scriptsize\mbox{53.80}} & {\scriptsize\mbox{71.93}} & {\scriptsize\mbox{49.79}} \\
RoboFAC-7B & {\scriptsize\mbox{34.97}} & {\scriptsize\mbox{73.63}} & {\scriptsize\mbox{-73.93}} & {\scriptsize\mbox{18.69}} & {\scriptsize\mbox{49.81}} & {\scriptsize\mbox{11.72}} & {\scriptsize\mbox{-4.88}} & {\scriptsize\mbox{52.89}} & {\scriptsize\mbox{49.10}} & {\scriptsize\mbox{40.92}} & {\scriptsize\mbox{46.29}} & {\scriptsize\mbox{-24.06}} & {\scriptsize\mbox{46.39}} \\
RoboMeter-4B & {\scriptsize\mbox{83.04}} & {\scriptsize\mbox{90.09}} & {\scriptsize\mbox{\textbf{94.49}}} & {\scriptsize\mbox{90.61}} & {\scriptsize\mbox{92.84}} & {\scriptsize\mbox{75.31}} & {\scriptsize\mbox{89.11}} & {\scriptsize\mbox{85.31}} & {\scriptsize\mbox{87.67}} & {\scriptsize\mbox{60.00}} & {\scriptsize\mbox{85.28}} & {\scriptsize\mbox{50.58}} & {\scriptsize\mbox{80.68}} \\
\bottomrule
\end{tabularx}
\renewcommand{\scriptsize}{\fontsize{6}{7}\selectfont}
\par\vspace{1pt}
\parbox{\textwidth}{\scriptsize\raggedright \textbf{T1}: Arrange largest number.\enspace \textbf{T2}: Fill pen holder.\enspace \textbf{T3}: Fold clothes.\enspace \textbf{T4}: Hang mugs.\enspace \textbf{T5}: Insert tubes.\enspace \textbf{T6}: Make toast.\enspace \textbf{T7}: Organize table.\enspace \textbf{T8}: Pack objects into box.\enspace \textbf{T9}: Play stacking toy.\enspace \textbf{T10}: Put bottles into dustbin.\enspace \textbf{T11}: Stack blocks.\enspace \textbf{T12}: Store laptop and headphone.\enspace \textbf{T13}: Sweep blocks.}
\par\vspace{1pt}
\parbox{\textwidth}{\scriptsize\raggedright
$^{\dagger}$ 2.0 Preview checkpoints.\quad
$^{\ddagger}$ Qwen3-VL backbones.
}
\endgroup
\end{table}

\begin{table}[H]
\renewcommand{\scriptsize}{\fontsize{7}{8}\selectfont}
\renewcommand{\footnotesize}{\fontsize{9}{11.5}\selectfont}
\setlength{\belowcaptionskip}{4pt}
\centering
\caption{\textbf{Cycle-VOC Results on In-Domain Simulation Trajectories.}}
\label{tab:id_cycle_voc_part1}
\label{tab:id_cycle_voc}
\begingroup
\footnotesize
\rmfamily
\setlength{\tabcolsep}{0.9pt}
\renewcommand{\arraystretch}{1.0}
\def\taskmodeband#1#2{%
  \specialrule{\lightrulewidth}{\aboverulesep}{2pt}
  \multicolumn{14}{@{}c@{}}{%
    \makebox[\textwidth][c]{%
      \rlap{\smash{\color{#1}%
        \rule[-\dp\csname @arstrutbox\endcsname]{\textwidth}%
        {\dimexpr\ht\csname @arstrutbox\endcsname
        +\dp\csname @arstrutbox\endcsname\relax}}}%
      \makebox[\textwidth][c]{\textit{#2}}}}\\%
  \specialrule{\lightrulewidth}{2pt}{\belowrulesep}
}
\begin{tabularx}{\textwidth}{
  @{} >{\raggedright\arraybackslash}p{0.24\textwidth}
  *{13}{>{\centering\arraybackslash}X} @{}
}
\toprule
\textbf{Model} & \textbf{T1} & \textbf{T2} & \textbf{T3} & \textbf{T4} & \textbf{T5} & \textbf{T6} & \textbf{T7} & \textbf{T8} & \textbf{T9} & \textbf{T10} & \textbf{T11} & \textbf{T12} & \textbf{T13} \\
\taskmodeband{RVZeroBand}{Zero-Shot Evaluation}
LIV & {\scriptsize\mbox{-33.16}} & {\scriptsize\mbox{42.81}} & {\scriptsize\mbox{20.17}} & {\scriptsize\mbox{76.13}} & {\scriptsize\mbox{-24.81}} & {\scriptsize\mbox{-37.39}} & {\scriptsize\mbox{-17.15}} & {\scriptsize\mbox{14.30}} & {\scriptsize\mbox{-4.29}} & {\scriptsize\mbox{43.03}} & {\scriptsize\mbox{-20.13}} & {\scriptsize\mbox{8.94}} & {\scriptsize\mbox{34.11}} \\
ProcVLM-2B & {\scriptsize\mbox{-1.46}} & {\scriptsize\mbox{77.63}} & {\scriptsize\mbox{63.08}} & {\scriptsize\mbox{86.13}} & {\scriptsize\mbox{53.03}} & {\scriptsize\mbox{41.10}} & {\scriptsize\mbox{71.38}} & {\scriptsize\mbox{84.94}} & {\scriptsize\mbox{48.11}} & {\scriptsize\mbox{-6.82}} & {\scriptsize\mbox{51.20}} & {\scriptsize\mbox{67.60}} & {\scriptsize\mbox{29.87}} \\
RoboFAC-7B & {\scriptsize\mbox{56.13}} & {\scriptsize\mbox{84.24}} & {\scriptsize\mbox{-65.11}} & {\scriptsize\mbox{35.37}} & {\scriptsize\mbox{66.55}} & {\scriptsize\mbox{15.84}} & {\scriptsize\mbox{11.29}} & {\scriptsize\mbox{71.14}} & {\scriptsize\mbox{62.44}} & {\scriptsize\mbox{48.77}} & {\scriptsize\mbox{65.56}} & {\scriptsize\mbox{17.58}} & {\scriptsize\mbox{66.56}} \\
RoboMeter-4B & {\scriptsize\mbox{\textbf{87.06}}} & {\scriptsize\mbox{\textbf{90.89}}} & {\scriptsize\mbox{\textbf{95.32}}} & {\scriptsize\mbox{92.82}} & {\scriptsize\mbox{\textbf{93.05}}} & {\scriptsize\mbox{\textbf{80.11}}} & {\scriptsize\mbox{\textbf{92.60}}} & {\scriptsize\mbox{90.53}} & {\scriptsize\mbox{89.35}} & {\scriptsize\mbox{73.78}} & {\scriptsize\mbox{\textbf{87.79}}} & {\scriptsize\mbox{70.31}} & {\scriptsize\mbox{\textbf{87.22}}} \\
\bottomrule
\end{tabularx}
\renewcommand{\scriptsize}{\fontsize{6}{7}\selectfont}
\par\vspace{1pt}
\parbox{\textwidth}{\scriptsize\raggedright \textbf{T1}: Arrange largest number.\enspace \textbf{T2}: Fill pen holder.\enspace \textbf{T3}: Fold clothes.\enspace \textbf{T4}: Hang mugs.\enspace \textbf{T5}: Insert tubes.\enspace \textbf{T6}: Make toast.\enspace \textbf{T7}: Organize table.\enspace \textbf{T8}: Pack objects into box.\enspace \textbf{T9}: Play stacking toy.\enspace \textbf{T10}: Put bottles into dustbin.\enspace \textbf{T11}: Stack blocks.\enspace \textbf{T12}: Store laptop and headphone.\enspace \textbf{T13}: Sweep blocks.}
\par\vspace{1pt}
\parbox{\textwidth}{\scriptsize\raggedright
$^{\dagger}$ 2.0 Preview checkpoints.\quad
$^{\ddagger}$ Qwen3-VL backbones.
}
\endgroup
\end{table}

\begin{table}[H]
\renewcommand{\scriptsize}{\fontsize{7}{8}\selectfont}
\renewcommand{\footnotesize}{\fontsize{9}{11.5}\selectfont}
\setlength{\belowcaptionskip}{4pt}
\centering
\caption{\textbf{Memory-VOC Results on In-Domain Simulation Trajectories.}}
\label{tab:id_memory_voc_part1}
\label{tab:id_memory_voc}
\label{tab:id_voc_mem_part1}
\label{tab:id_voc_mem}
\begingroup
\footnotesize
\rmfamily
\setlength{\tabcolsep}{0.9pt}
\renewcommand{\arraystretch}{1.0}
\def\taskmodeband#1#2{%
  \specialrule{\lightrulewidth}{\aboverulesep}{2pt}
  \multicolumn{3}{@{}c@{}}{%
    \makebox[\textwidth][c]{%
      \rlap{\smash{\color{#1}%
        \rule[-\dp\csname @arstrutbox\endcsname]{\textwidth}%
        {\dimexpr\ht\csname @arstrutbox\endcsname
        +\dp\csname @arstrutbox\endcsname\relax}}}%
      \makebox[\textwidth][c]{\textit{#2}}}}\\%
  \specialrule{\lightrulewidth}{2pt}{\belowrulesep}
}
\begin{tabularx}{\textwidth}{
  @{} >{\raggedright\arraybackslash}p{0.24\textwidth}
  *{2}{>{\centering\arraybackslash}X} @{}
}
\toprule
\textbf{Model} & \textbf{Press by number} & \textbf{Swap blocks} \\
\taskmodeband{RVZeroBand}{Zero-Shot Evaluation}
LIV & {\scriptsize\mbox{16.82}} & {\scriptsize\mbox{0.15}} \\
ProcVLM-2B & {\scriptsize\mbox{14.92}} & {\scriptsize\mbox{17.06}} \\
RoboFAC-7B & {\scriptsize\mbox{-25.54}} & {\scriptsize\mbox{-17.29}} \\
RoboMeter-4B & {\scriptsize\mbox{7.70}} & {\scriptsize\mbox{5.07}} \\
RoboReward-4B & {\scriptsize\mbox{89.25}} & {\scriptsize\mbox{75.06}} \\
RoboReward-8B & {\scriptsize\mbox{87.34}} & {\scriptsize\mbox{80.95}} \\
RynnValue-4B & {\scriptsize\mbox{1.84}} & {\scriptsize\mbox{-41.42}} \\
RynnValue-8B & {\scriptsize\mbox{9.64}} & {\scriptsize\mbox{-28.45}} \\
TOPReward-8B$^{\ddagger}$ & {\scriptsize\mbox{94.57}} & {\scriptsize\mbox{92.81}} \\
TOPReward-32B$^{\ddagger}$ & {\scriptsize\mbox{\textbf{95.33}}} & {\scriptsize\mbox{\textbf{97.60}}} \\
VLAC-2B & {\scriptsize\mbox{-1.90}} & {\scriptsize\mbox{-10.51}} \\
VLAC-8B & {\scriptsize\mbox{72.04}} & {\scriptsize\mbox{0.10}} \\
\taskmodeband{RVOneBand}{One-Shot Evaluation}
ProcVLM-2B & {\scriptsize\mbox{60.96}} & {\scriptsize\mbox{4.55}} \\
Robo-Dopamine-3B & {\scriptsize\mbox{18.64}} & {\scriptsize\mbox{9.46}} \\
Robo-Dopamine-4B$^{\dagger}$ & {\scriptsize\mbox{34.35}} & {\scriptsize\mbox{4.61}} \\
Robo-Dopamine-8B & {\scriptsize\mbox{4.37}} & {\scriptsize\mbox{1.88}} \\
Robo-Dopamine-8B$^{\dagger}$ & {\scriptsize\mbox{55.26}} & {\scriptsize\mbox{13.84}} \\
VLAC-2B & {\scriptsize\mbox{57.13}} & {\scriptsize\mbox{-31.31}} \\
VLAC-8B & {\scriptsize\mbox{\textbf{65.64}}} & {\scriptsize\mbox{\textbf{67.60}}} \\
\bottomrule
\end{tabularx}
\renewcommand{\scriptsize}{\fontsize{6}{7}\selectfont}
\par\vspace{1pt}
\parbox{\textwidth}{\scriptsize\raggedright
$^{\dagger}$ 2.0 Preview checkpoints.\quad
$^{\ddagger}$ Qwen3-VL backbones.
}
\endgroup
\end{table}

\begin{table}[H]
\renewcommand{\scriptsize}{\fontsize{7}{8}\selectfont}
\renewcommand{\footnotesize}{\fontsize{9}{11.5}\selectfont}
\setlength{\belowcaptionskip}{4pt}
\centering
\caption{\textbf{FPL Results on In-Domain Simulation Trajectories.}}
\label{tab:id_fpl_part1}
\label{tab:id_fpl}
\begingroup
\footnotesize
\rmfamily
\setlength{\tabcolsep}{0.9pt}
\renewcommand{\arraystretch}{1.0}
\def\taskmodeband#1#2{%
  \specialrule{\lightrulewidth}{\aboverulesep}{2pt}
  \multicolumn{14}{@{}c@{}}{%
    \makebox[\textwidth][c]{%
      \rlap{\smash{\color{#1}%
        \rule[-\dp\csname @arstrutbox\endcsname]{\textwidth}%
        {\dimexpr\ht\csname @arstrutbox\endcsname
        +\dp\csname @arstrutbox\endcsname\relax}}}%
      \makebox[\textwidth][c]{\textit{#2}}}}\\%
  \specialrule{\lightrulewidth}{2pt}{\belowrulesep}
}
\begin{tabularx}{\textwidth}{
  @{} >{\raggedright\arraybackslash}p{0.24\textwidth}
  *{13}{>{\centering\arraybackslash}X} @{}
}
\toprule
\textbf{Model} & \textbf{T1} & \textbf{T2} & \textbf{T3} & \textbf{T4} & \textbf{T5} & \textbf{T6} & \textbf{T7} & \textbf{T8} & \textbf{T9} & \textbf{T10} & \textbf{T11} & \textbf{T12} & \textbf{T13} \\
\taskmodeband{RVZeroBand}{Zero-Shot Evaluation}

LIV & {\scriptsize\mbox{35.56}} & {\scriptsize\mbox{32.54}} & {\scriptsize\mbox{45.44}} & {\scriptsize\mbox{29.66}} & {\scriptsize\mbox{28.69}} & {\scriptsize\mbox{36.83}} & {\scriptsize\mbox{44.40}} & {\scriptsize\mbox{27.90}} & {\scriptsize\mbox{\textbf{27.59}}} & {\scriptsize\mbox{30.57}} & {\scriptsize\mbox{41.72}} & {\scriptsize\mbox{45.96}} & {\scriptsize\mbox{28.75}} \\

ProcVLM-2B & {\scriptsize\mbox{38.83}} & {\scriptsize\mbox{35.64}} & {\scriptsize\mbox{39.17}} & {\scriptsize\mbox{24.40}} & {\scriptsize\mbox{43.91}} & {\scriptsize\mbox{40.18}} & {\scriptsize\mbox{30.18}} & {\scriptsize\mbox{42.42}} & {\scriptsize\mbox{42.83}} & {\scriptsize\mbox{40.17}} & {\scriptsize\mbox{39.74}} & {\scriptsize\mbox{38.73}} & {\scriptsize\mbox{34.14}} \\

RoboFAC-7B & {\scriptsize\mbox{22.66}} & {\scriptsize\mbox{41.74}} & {\scriptsize\mbox{37.14}} & {\scriptsize\mbox{28.89}} & {\scriptsize\mbox{41.85}} & {\scriptsize\mbox{44.95}} & {\scriptsize\mbox{37.25}} & {\scriptsize\mbox{50.16}} & {\scriptsize\mbox{41.96}} & {\scriptsize\mbox{36.58}} & {\scriptsize\mbox{39.05}} & {\scriptsize\mbox{39.74}} & {\scriptsize\mbox{43.89}} \\

RoboMeter-4B & {\scriptsize\mbox{30.05}} & {\scriptsize\mbox{\textbf{30.51}}} & {\scriptsize\mbox{\textbf{24.08}}} & {\scriptsize\mbox{36.12}} & {\scriptsize\mbox{27.90}} & {\scriptsize\mbox{42.20}} & {\scriptsize\mbox{29.06}} & {\scriptsize\mbox{37.37}} & {\scriptsize\mbox{34.16}} & {\scriptsize\mbox{34.31}} & {\scriptsize\mbox{28.69}} & {\scriptsize\mbox{30.52}} & {\scriptsize\mbox{21.96}} \\
\bottomrule
\end{tabularx}
\renewcommand{\scriptsize}{\fontsize{6}{7}\selectfont}
\par\vspace{1pt}
\parbox{\textwidth}{\scriptsize\raggedright \textbf{T1}: Arrange largest number.\enspace \textbf{T2}: Fill pen holder.\enspace \textbf{T3}: Fold clothes.\enspace \textbf{T4}: Hang mugs.\enspace \textbf{T5}: Insert tubes.\enspace \textbf{T6}: Make toast.\enspace \textbf{T7}: Organize table.\enspace \textbf{T8}: Pack objects into box.\enspace \textbf{T9}: Play stacking toy.\enspace \textbf{T10}: Put bottles into dustbin.\enspace \textbf{T11}: Stack blocks.\enspace \textbf{T12}: Store laptop and headphone.\enspace \textbf{T13}: Sweep blocks.}
\par\vspace{1pt}
\parbox{\textwidth}{\scriptsize\raggedright
$^{\dagger}$ 2.0 Preview checkpoints.\quad
$^{\ddagger}$ Qwen3-VL backbones.
}
\endgroup
\end{table}

\begin{table}[H]
\renewcommand{\scriptsize}{\fontsize{7}{8}\selectfont}
\renewcommand{\footnotesize}{\fontsize{9}{11.5}\selectfont}
\setlength{\belowcaptionskip}{4pt}
\centering
\caption{\textbf{TRR Results on In-Domain Simulation Trajectories.}}
\label{tab:id_trr_part1}
\label{tab:id_trr}
\begingroup
\footnotesize
\rmfamily
\setlength{\tabcolsep}{0.9pt}
\renewcommand{\arraystretch}{1.0}
\def\taskmodeband#1#2{%
  \specialrule{\lightrulewidth}{\aboverulesep}{2pt}
  \multicolumn{14}{@{}c@{}}{%
    \makebox[\textwidth][c]{%
      \rlap{\smash{\color{#1}%
        \rule[-\dp\csname @arstrutbox\endcsname]{\textwidth}%
        {\dimexpr\ht\csname @arstrutbox\endcsname
        +\dp\csname @arstrutbox\endcsname\relax}}}%
      \makebox[\textwidth][c]{\textit{#2}}}}\\%
  \specialrule{\lightrulewidth}{2pt}{\belowrulesep}
}
\begin{tabularx}{\textwidth}{
  @{} >{\raggedright\arraybackslash}p{0.24\textwidth}
  *{13}{>{\centering\arraybackslash}X} @{}
}
\toprule
\textbf{Model} & \textbf{T1} & \textbf{T2} & \textbf{T3} & \textbf{T4} & \textbf{T5} & \textbf{T6} & \textbf{T7} & \textbf{T8} & \textbf{T9} & \textbf{T10} & \textbf{T11} & \textbf{T12} & \textbf{T13} \\
\taskmodeband{RVZeroBand}{Zero-Shot Evaluation}
LIV & {\scriptsize\mbox{14.81}} & {\scriptsize\mbox{26.67}} & {\scriptsize\mbox{14.81}} & {\scriptsize\mbox{33.33}} & {\scriptsize\mbox{5.56}} & {\scriptsize\mbox{25.93}} & {\scriptsize\mbox{27.78}} & {\scriptsize\mbox{22.22}} & {\scriptsize\mbox{11.11}} & {\scriptsize\mbox{29.63}} & {\scriptsize\mbox{33.33}} & {\scriptsize\mbox{25.93}} & {\scriptsize\mbox{\textbf{40.74}}} \\
ProcVLM-2B & {\scriptsize\mbox{\textbf{25.93}}} & {\scriptsize\mbox{17.78}} & {\scriptsize\mbox{25.93}} & {\scriptsize\mbox{33.33}} & {\scriptsize\mbox{38.89}} & {\scriptsize\mbox{7.41}} & {\scriptsize\mbox{36.11}} & {\scriptsize\mbox{31.11}} & {\scriptsize\mbox{11.11}} & {\scriptsize\mbox{3.70}} & {\scriptsize\mbox{14.81}} & {\scriptsize\mbox{\textbf{37.04}}} & {\scriptsize\mbox{29.63}} \\
RoboFAC-7B & {\scriptsize\mbox{3.70}} & {\scriptsize\mbox{\textbf{40.00}}} & {\scriptsize\mbox{\textbf{44.44}}} & {\scriptsize\mbox{\textbf{44.44}}} & {\scriptsize\mbox{\textbf{61.11}}} & {\scriptsize\mbox{22.22}} & {\scriptsize\mbox{27.78}} & {\scriptsize\mbox{35.56}} & {\scriptsize\mbox{\textbf{22.22}}} & {\scriptsize\mbox{29.63}} & {\scriptsize\mbox{\textbf{48.15}}} & {\scriptsize\mbox{22.22}} & {\scriptsize\mbox{25.93}} \\
RoboMeter-4B & {\scriptsize\mbox{0.00}} & {\scriptsize\mbox{8.89}} & {\scriptsize\mbox{29.63}} & {\scriptsize\mbox{18.52}} & {\scriptsize\mbox{16.67}} & {\scriptsize\mbox{\textbf{29.63}}} & {\scriptsize\mbox{27.78}} & {\scriptsize\mbox{15.56}} & {\scriptsize\mbox{16.67}} & {\scriptsize\mbox{29.63}} & {\scriptsize\mbox{33.33}} & {\scriptsize\mbox{22.22}} & {\scriptsize\mbox{37.04}} \\
\bottomrule
\end{tabularx}
\renewcommand{\scriptsize}{\fontsize{6}{7}\selectfont}
\par\vspace{1pt}
\parbox{\textwidth}{\scriptsize\raggedright \textbf{T1}: Arrange largest number.\enspace \textbf{T2}: Fill pen holder.\enspace \textbf{T3}: Fold clothes.\enspace \textbf{T4}: Hang mugs.\enspace \textbf{T5}: Insert tubes.\enspace \textbf{T6}: Make toast.\enspace \textbf{T7}: Organize table.\enspace \textbf{T8}: Pack objects into box.\enspace \textbf{T9}: Play stacking toy.\enspace \textbf{T10}: Put bottles into dustbin.\enspace \textbf{T11}: Stack blocks.\enspace \textbf{T12}: Store laptop and headphone.\enspace \textbf{T13}: Sweep blocks.}
\par\vspace{1pt}
\parbox{\textwidth}{\scriptsize\raggedright
$^{\dagger}$ 2.0 Preview checkpoints.\quad
$^{\ddagger}$ Qwen3-VL backbones.
}
\endgroup
\end{table}

\begin{table}[H]
\renewcommand{\scriptsize}{\fontsize{7}{8}\selectfont}
\renewcommand{\footnotesize}{\fontsize{9}{11.5}\selectfont}
\setlength{\belowcaptionskip}{4pt}
\centering
\caption{\textbf{VS Results on In-Domain Simulation Trajectories.}}
\label{tab:id_vs_part1}
\label{tab:id_vs}
\begingroup
\footnotesize
\rmfamily
\setlength{\tabcolsep}{0.9pt}
\renewcommand{\arraystretch}{1.0}
\def\taskmodeband#1#2{%
  \specialrule{\lightrulewidth}{\aboverulesep}{2pt}
  \multicolumn{14}{@{}c@{}}{%
    \makebox[\textwidth][c]{%
      \rlap{\smash{\color{#1}%
        \rule[-\dp\csname @arstrutbox\endcsname]{\textwidth}%
        {\dimexpr\ht\csname @arstrutbox\endcsname
        +\dp\csname @arstrutbox\endcsname\relax}}}%
      \makebox[\textwidth][c]{\textit{#2}}}}\\%
  \specialrule{\lightrulewidth}{2pt}{\belowrulesep}
}
\begin{tabularx}{\textwidth}{
  @{} >{\raggedright\arraybackslash}p{0.24\textwidth}
  *{13}{>{\centering\arraybackslash}X} @{}
}
\toprule
\textbf{Model} & \textbf{T1} & \textbf{T2} & \textbf{T3} & \textbf{T4} & \textbf{T5} & \textbf{T6} & \textbf{T7} & \textbf{T8} & \textbf{T9} & \textbf{T10} & \textbf{T11} & \textbf{T12} & \textbf{T13} \\
\taskmodeband{RVZeroBand}{Zero-Shot Evaluation}
LIV & {\scriptsize\mbox{3.18}} & {\scriptsize\mbox{1.84}} & {\scriptsize\mbox{3.34}} & {\scriptsize\mbox{5.50}} & {\scriptsize\mbox{1.88}} & {\scriptsize\mbox{1.85}} & {\scriptsize\mbox{5.44}} & {\scriptsize\mbox{2.04}} & {\scriptsize\mbox{1.17}} & {\scriptsize\mbox{4.84}} & {\scriptsize\mbox{2.22}} & {\scriptsize\mbox{3.27}} & {\scriptsize\mbox{1.55}} \\
ProcVLM-2B & {\scriptsize\mbox{0.18}} & {\scriptsize\mbox{6.97}} & {\scriptsize\mbox{12.65}} & {\scriptsize\mbox{10.03}} & {\scriptsize\mbox{4.04}} & {\scriptsize\mbox{7.81}} & {\scriptsize\mbox{6.45}} & {\scriptsize\mbox{8.18}} & {\scriptsize\mbox{2.71}} & {\scriptsize\mbox{3.11}} & {\scriptsize\mbox{3.87}} & {\scriptsize\mbox{7.90}} & {\scriptsize\mbox{8.60}} \\
RoboFAC-7B & {\scriptsize\mbox{1.06}} & {\scriptsize\mbox{3.99}} & {\scriptsize\mbox{22.57}} & {\scriptsize\mbox{1.56}} & {\scriptsize\mbox{3.88}} & {\scriptsize\mbox{3.93}} & {\scriptsize\mbox{3.56}} & {\scriptsize\mbox{4.47}} & {\scriptsize\mbox{0.61}} & {\scriptsize\mbox{1.70}} & {\scriptsize\mbox{3.39}} & {\scriptsize\mbox{2.50}} & {\scriptsize\mbox{2.01}} \\
RoboMeter-4B & {\scriptsize\mbox{8.55}} & {\scriptsize\mbox{\textbf{24.82}}} & {\scriptsize\mbox{\textbf{38.79}}} & {\scriptsize\mbox{\textbf{22.87}}} & {\scriptsize\mbox{\textbf{24.98}}} & {\scriptsize\mbox{19.31}} & {\scriptsize\mbox{13.57}} & {\scriptsize\mbox{17.06}} & {\scriptsize\mbox{7.58}} & {\scriptsize\mbox{5.77}} & {\scriptsize\mbox{\textbf{19.55}}} & {\scriptsize\mbox{10.36}} & {\scriptsize\mbox{\textbf{20.43}}} \\
\bottomrule
\end{tabularx}
\renewcommand{\scriptsize}{\fontsize{6}{7}\selectfont}
\par\vspace{1pt}
\parbox{\textwidth}{\scriptsize\raggedright \textbf{T1}: Arrange largest number.\enspace \textbf{T2}: Fill pen holder.\enspace \textbf{T3}: Fold clothes.\enspace \textbf{T4}: Hang mugs.\enspace \textbf{T5}: Insert tubes.\enspace \textbf{T6}: Make toast.\enspace \textbf{T7}: Organize table.\enspace \textbf{T8}: Pack objects into box.\enspace \textbf{T9}: Play stacking toy.\enspace \textbf{T10}: Put bottles into dustbin.\enspace \textbf{T11}: Stack blocks.\enspace \textbf{T12}: Store laptop and headphone.\enspace \textbf{T13}: Sweep blocks.}
\par\vspace{1pt}
\parbox{\textwidth}{\scriptsize\raggedright
$^{\dagger}$ 2.0 Preview checkpoints.\quad
$^{\ddagger}$ Qwen3-VL backbones.
}
\endgroup
\end{table}

\begin{table}[H]
\renewcommand{\scriptsize}{\fontsize{7}{8}\selectfont}
\renewcommand{\footnotesize}{\fontsize{9}{11.5}\selectfont}
\setlength{\belowcaptionskip}{4pt}
\centering
\caption{\textbf{CSVC Results on In-Domain Simulation Trajectories.}}
\label{tab:id_csvc_part1}
\label{tab:id_csvc}
\label{tab:id_cspc_part1}
\label{tab:id_cspc}
\begingroup
\footnotesize
\rmfamily
\setlength{\tabcolsep}{0.9pt}
\renewcommand{\arraystretch}{1.0}
\def\taskmodeband#1#2{%
  \specialrule{\lightrulewidth}{\aboverulesep}{2pt}
  \multicolumn{5}{@{}c@{}}{%
    \makebox[\textwidth][c]{%
      \rlap{\smash{\color{#1}%
        \rule[-\dp\csname @arstrutbox\endcsname]{\textwidth}%
        {\dimexpr\ht\csname @arstrutbox\endcsname
        +\dp\csname @arstrutbox\endcsname\relax}}}%
      \makebox[\textwidth][c]{\textit{#2}}}}\\%
  \specialrule{\lightrulewidth}{2pt}{\belowrulesep}
}
\begin{tabularx}{\textwidth}{
  @{} >{\raggedright\arraybackslash}p{0.24\textwidth}
  *{4}{>{\centering\arraybackslash}X} @{}
}
\toprule
\textbf{Model} & \textbf{Fill pen holder} & \textbf{Hang mugs} & \textbf{Pack objects into box} & \textbf{Stack blocks} \\
\taskmodeband{RVZeroBand}{Zero-Shot Evaluation}
LIV & {\scriptsize\mbox{-17.92}} & {\scriptsize\mbox{-32.08}} & {\scriptsize\mbox{-4.68}} & {\scriptsize\mbox{8.37}} \\
ProcVLM-2B & {\scriptsize\mbox{5.10}} & {\scriptsize\mbox{39.47}} & {\scriptsize\mbox{21.21}} & {\scriptsize\mbox{-10.23}} \\
RoboFAC-7B & {\scriptsize\mbox{9.49}} & {\scriptsize\mbox{-11.83}} & {\scriptsize\mbox{11.67}} & {\scriptsize\mbox{20.13}} \\
RoboMeter-4B & {\scriptsize\mbox{38.45}} & {\scriptsize\mbox{54.87}} & {\scriptsize\mbox{-5.60}} & {\scriptsize\mbox{49.98}} \\
RoboReward-4B & \textemdash & \textemdash & \textemdash & \textemdash \\
RoboReward-8B & \textemdash & \textemdash & \textemdash & \textemdash \\
RynnValue-4B & {\scriptsize\mbox{21.94}} & {\scriptsize\mbox{55.24}} & {\scriptsize\mbox{16.48}} & {\scriptsize\mbox{24.26}} \\
\bottomrule
\end{tabularx}
\renewcommand{\scriptsize}{\fontsize{6}{7}\selectfont}
\par\vspace{1pt}
\parbox{\textwidth}{\scriptsize\raggedright
$^{\dagger}$ 2.0 Preview checkpoints.\quad
$^{\ddagger}$ Qwen3-VL backbones.
}
\endgroup
\end{table}

\FloatBarrier
\Needspace{34\baselineskip}
\subsection{Generalization Results}\label{app:simulation_ood_results}

We evaluate out-of-domain (OOD) performance under separate embodiment and environment shifts. Comparing each setting with the in-domain (ID) results reveals how the shift affects each metric under the shared evaluation protocol.

\subsubsection{Cross-Embodiment Generalization}\label{app:simulation_embodiment_results}

We replace ARX with a dual-arm UR5e platform while preserving the task goals and environment. This comparison tests whether value predictions transfer across robot morphologies and execution patterns.


\begin{table}[H]
\renewcommand{\scriptsize}{\fontsize{7}{8}\selectfont}
\renewcommand{\footnotesize}{\fontsize{9}{9.5}\selectfont}
\setlength{\belowcaptionskip}{4pt}
\centering
\caption{\textbf{SA Results on Cross-Embodiment Simulation Trajectories.}}
\label{tab:emb_sa_part1}
\label{tab:emb_sa}
\begingroup
\footnotesize
\rmfamily
\setlength{\tabcolsep}{0.9pt}
\renewcommand{\arraystretch}{1.0}
\def\taskmodeband#1#2{%
  \specialrule{\lightrulewidth}{\aboverulesep}{2pt}
  \multicolumn{14}{@{}c@{}}{%
    \makebox[\textwidth][c]{%
      \rlap{\smash{\color{#1}%
        \rule[-\dp\csname @arstrutbox\endcsname]{\textwidth}%
        {\dimexpr\ht\csname @arstrutbox\endcsname
        +\dp\csname @arstrutbox\endcsname\relax}}}%
      \makebox[\textwidth][c]{\textit{#2}}}}\\%
  \specialrule{\lightrulewidth}{2pt}{\belowrulesep}
}
\begin{tabularx}{\textwidth}{
  @{} >{\raggedright\arraybackslash}p{0.24\textwidth}
  *{13}{>{\centering\arraybackslash}X} @{}
}
\toprule
\textbf{Model} & \textbf{T1} & \textbf{T2} & \textbf{T3} & \textbf{T4} & \textbf{T5} & \textbf{T6} & \textbf{T7} & \textbf{T8} & \textbf{T9} & \textbf{T10} & \textbf{T11} & \textbf{T12} & \textbf{T13} \\
\taskmodeband{RVZeroBand}{Zero-Shot Evaluation}
LIV & {\scriptsize\mbox{35.00}} & {\scriptsize\mbox{45.00}} & {\scriptsize\mbox{38.33}} & {\scriptsize\mbox{26.67}} & {\scriptsize\mbox{45.00}} & {\scriptsize\mbox{36.67}} & {\scriptsize\mbox{35.00}} & {\scriptsize\mbox{63.00}} & {\scriptsize\mbox{57.50}} & {\scriptsize\mbox{83.33}} & {\scriptsize\mbox{70.00}} & {\scriptsize\mbox{43.33}} & {\scriptsize\mbox{91.67}} \\
ProcVLM-2B & {\scriptsize\mbox{66.67}} & {\scriptsize\mbox{\textbf{100.00}}} & {\scriptsize\mbox{85.00}} & {\scriptsize\mbox{83.33}} & {\scriptsize\mbox{\textbf{100.00}}} & {\scriptsize\mbox{83.33}} & {\scriptsize\mbox{53.75}} & {\scriptsize\mbox{90.00}} & {\scriptsize\mbox{47.50}} & {\scriptsize\mbox{55.00}} & {\scriptsize\mbox{33.33}} & {\scriptsize\mbox{66.67}} & {\scriptsize\mbox{81.67}} \\
RoboFAC-7B & {\scriptsize\mbox{48.33}} & {\scriptsize\mbox{\textbf{100.00}}} & {\scriptsize\mbox{86.67}} & {\scriptsize\mbox{85.00}} & {\scriptsize\mbox{\textbf{100.00}}} & {\scriptsize\mbox{51.67}} & {\scriptsize\mbox{90.00}} & {\scriptsize\mbox{91.00}} & {\scriptsize\mbox{65.00}} & {\scriptsize\mbox{\textbf{100.00}}} & {\scriptsize\mbox{96.67}} & {\scriptsize\mbox{83.33}} & {\scriptsize\mbox{\textbf{100.00}}} \\
RoboMeter-4B & {\scriptsize\mbox{38.33}} & {\scriptsize\mbox{99.00}} & {\scriptsize\mbox{78.33}} & {\scriptsize\mbox{55.00}} & {\scriptsize\mbox{\textbf{100.00}}} & {\scriptsize\mbox{70.00}} & {\scriptsize\mbox{\textbf{97.50}}} & {\scriptsize\mbox{\textbf{100.00}}} & {\scriptsize\mbox{\textbf{95.00}}} & {\scriptsize\mbox{85.00}} & {\scriptsize\mbox{95.00}} & {\scriptsize\mbox{85.00}} & {\scriptsize\mbox{\textbf{100.00}}} \\
\bottomrule
\end{tabularx}
\renewcommand{\scriptsize}{\fontsize{6}{7}\selectfont}
\par\vspace{1pt}
\parbox{\textwidth}{\scriptsize\raggedright \textbf{T1}: Arrange largest number.\enspace \textbf{T2}: Fill pen holder.\enspace \textbf{T3}: Fold clothes.\enspace \textbf{T4}: Hang mugs.\enspace \textbf{T5}: Insert tubes.\enspace \textbf{T6}: Make toast.\enspace \textbf{T7}: Organize table.\enspace \textbf{T8}: Pack objects into box.\enspace \textbf{T9}: Play stacking toy.\enspace \textbf{T10}: Put bottles into dustbin.\enspace \textbf{T11}: Stack blocks.\enspace \textbf{T12}: Store laptop and headphone.\enspace \textbf{T13}: Sweep blocks.}
\par\vspace{1pt}
\parbox{\textwidth}{\scriptsize\raggedright
$^{\dagger}$ 2.0 Preview checkpoints.\quad
$^{\ddagger}$ Qwen3-VL backbones.
}
\endgroup
\end{table}

\begin{table}[H]
\renewcommand{\scriptsize}{\fontsize{7}{8}\selectfont}
\renewcommand{\footnotesize}{\fontsize{9}{9.5}\selectfont}
\setlength{\belowcaptionskip}{4pt}
\centering
\caption{\textbf{TGA-CT Results on Cross-Embodiment Simulation Trajectories.}}
\label{tab:emb_tga_ct_part1}
\label{tab:emb_tga_ct}
\label{tab:emb_tga_easy_part1}
\label{tab:emb_tga_easy}
\begingroup
\footnotesize
\rmfamily
\setlength{\tabcolsep}{0.9pt}
\renewcommand{\arraystretch}{1.0}
\def\taskmodeband#1#2{%
  \specialrule{\lightrulewidth}{\aboverulesep}{2pt}
  \multicolumn{14}{@{}c@{}}{%
    \makebox[\textwidth][c]{%
      \rlap{\smash{\color{#1}%
        \rule[-\dp\csname @arstrutbox\endcsname]{\textwidth}%
        {\dimexpr\ht\csname @arstrutbox\endcsname
        +\dp\csname @arstrutbox\endcsname\relax}}}%
      \makebox[\textwidth][c]{\textit{#2}}}}\\%
  \specialrule{\lightrulewidth}{2pt}{\belowrulesep}
}
\begin{tabularx}{\textwidth}{
  @{} >{\raggedright\arraybackslash}p{0.24\textwidth}
  *{13}{>{\centering\arraybackslash}X} @{}
}
\toprule
\textbf{Model} & \textbf{T1} & \textbf{T2} & \textbf{T3} & \textbf{T4} & \textbf{T5} & \textbf{T6} & \textbf{T7} & \textbf{T8} & \textbf{T9} & \textbf{T10} & \textbf{T11} & \textbf{T12} & \textbf{T13} \\
\taskmodeband{RVZeroBand}{Zero-Shot Evaluation}
LIV & {\scriptsize\mbox{0.00}} & {\scriptsize\mbox{20.00}} & {\scriptsize\mbox{0.00}} & {\scriptsize\mbox{\textbf{100.00}}} & {\scriptsize\mbox{0.00}} & {\scriptsize\mbox{0.00}} & {\scriptsize\mbox{0.00}} & {\scriptsize\mbox{0.00}} & {\scriptsize\mbox{0.00}} & {\scriptsize\mbox{0.00}} & {\scriptsize\mbox{0.00}} & {\scriptsize\mbox{0.00}} & {\scriptsize\mbox{80.00}} \\
ProcVLM-2B & {\scriptsize\mbox{0.00}} & {\scriptsize\mbox{60.00}} & {\scriptsize\mbox{30.00}} & {\scriptsize\mbox{90.00}} & {\scriptsize\mbox{60.00}} & {\scriptsize\mbox{30.00}} & {\scriptsize\mbox{10.00}} & {\scriptsize\mbox{\textbf{100.00}}} & {\scriptsize\mbox{0.00}} & {\scriptsize\mbox{20.00}} & {\scriptsize\mbox{0.00}} & {\scriptsize\mbox{0.00}} & {\scriptsize\mbox{60.00}} \\
RoboFAC-7B & {\scriptsize\mbox{0.00}} & {\scriptsize\mbox{0.00}} & {\scriptsize\mbox{0.00}} & {\scriptsize\mbox{70.00}} & {\scriptsize\mbox{0.00}} & {\scriptsize\mbox{0.00}} & {\scriptsize\mbox{0.00}} & {\scriptsize\mbox{0.00}} & {\scriptsize\mbox{0.00}} & {\scriptsize\mbox{90.00}} & {\scriptsize\mbox{0.00}} & {\scriptsize\mbox{0.00}} & {\scriptsize\mbox{0.00}} \\
RoboMeter-4B & {\scriptsize\mbox{\textbf{100.00}}} & {\scriptsize\mbox{0.00}} & {\scriptsize\mbox{\textbf{100.00}}} & {\scriptsize\mbox{0.00}} & {\scriptsize\mbox{0.00}} & {\scriptsize\mbox{50.00}} & {\scriptsize\mbox{\textbf{100.00}}} & {\scriptsize\mbox{\textbf{100.00}}} & {\scriptsize\mbox{20.00}} & {\scriptsize\mbox{\textbf{100.00}}} & {\scriptsize\mbox{\textbf{100.00}}} & {\scriptsize\mbox{\textbf{100.00}}} & {\scriptsize\mbox{\textbf{100.00}}} \\
\bottomrule
\end{tabularx}
\renewcommand{\scriptsize}{\fontsize{6}{7}\selectfont}
\par\vspace{1pt}
\parbox{\textwidth}{\scriptsize\raggedright \textbf{T1}: Arrange largest number.\enspace \textbf{T2}: Fill pen holder.\enspace \textbf{T3}: Fold clothes.\enspace \textbf{T4}: Hang mugs.\enspace \textbf{T5}: Insert tubes.\enspace \textbf{T6}: Make toast.\enspace \textbf{T7}: Organize table.\enspace \textbf{T8}: Pack objects into box.\enspace \textbf{T9}: Play stacking toy.\enspace \textbf{T10}: Put bottles into dustbin.\enspace \textbf{T11}: Stack blocks.\enspace \textbf{T12}: Store laptop and headphone.\enspace \textbf{T13}: Sweep blocks.}
\par\vspace{1pt}
\parbox{\textwidth}{\scriptsize\raggedright
$^{\dagger}$ 2.0 Preview checkpoints.\quad
$^{\ddagger}$ Qwen3-VL backbones.
}
\endgroup
\end{table}

\begin{table}[H]
\renewcommand{\scriptsize}{\fontsize{7}{8}\selectfont}
\renewcommand{\footnotesize}{\fontsize{9}{9}\selectfont}
\setlength{\belowcaptionskip}{4pt}
\centering
\caption{\textbf{TGA-CF Results on Cross-Embodiment Simulation Trajectories.}}
\label{tab:emb_tga_cf_part1}
\label{tab:emb_tga_cf}
\label{tab:emb_tga_hard_part1}
\label{tab:emb_tga_hard}
\begingroup
\footnotesize
\rmfamily
\setlength{\tabcolsep}{0.9pt}
\renewcommand{\arraystretch}{1.0}
\def\taskmodeband#1#2{%
  \specialrule{\lightrulewidth}{\aboverulesep}{2pt}
  \multicolumn{14}{@{}c@{}}{%
    \makebox[\textwidth][c]{%
      \rlap{\smash{\color{#1}%
        \rule[-\dp\csname @arstrutbox\endcsname]{\textwidth}%
        {\dimexpr\ht\csname @arstrutbox\endcsname
        +\dp\csname @arstrutbox\endcsname\relax}}}%
      \makebox[\textwidth][c]{\textit{#2}}}}\\%
  \specialrule{\lightrulewidth}{2pt}{\belowrulesep}
}
\begin{tabularx}{\textwidth}{
  @{} >{\raggedright\arraybackslash}p{0.24\textwidth}
  *{13}{>{\centering\arraybackslash}X} @{}
}
\toprule
\textbf{Model} & \textbf{T1} & \textbf{T2} & \textbf{T3} & \textbf{T4} & \textbf{T5} & \textbf{T6} & \textbf{T7} & \textbf{T8} & \textbf{T9} & \textbf{T10} & \textbf{T11} & \textbf{T12} & \textbf{T13} \\
\taskmodeband{RVZeroBand}{Zero-Shot Evaluation}
LIV & {\scriptsize\mbox{0.00}} & {\scriptsize\mbox{0.00}} & {\scriptsize\mbox{0.00}} & {\scriptsize\mbox{30.00}} & {\scriptsize\mbox{0.00}} & {\scriptsize\mbox{0.00}} & {\scriptsize\mbox{0.00}} & {\scriptsize\mbox{10.00}} & {\scriptsize\mbox{0.00}} & {\scriptsize\mbox{0.00}} & {\scriptsize\mbox{0.00}} & {\scriptsize\mbox{0.00}} & {\scriptsize\mbox{0.00}} \\
ProcVLM-2B & {\scriptsize\mbox{40.00}} & {\scriptsize\mbox{40.00}} & {\scriptsize\mbox{20.00}} & {\scriptsize\mbox{30.00}} & {\scriptsize\mbox{20.00}} & {\scriptsize\mbox{0.00}} & {\scriptsize\mbox{10.00}} & {\scriptsize\mbox{60.00}} & {\scriptsize\mbox{10.00}} & {\scriptsize\mbox{0.00}} & {\scriptsize\mbox{0.00}} & {\scriptsize\mbox{60.00}} & {\scriptsize\mbox{40.00}} \\
RoboFAC-7B & {\scriptsize\mbox{0.00}} & {\scriptsize\mbox{0.00}} & {\scriptsize\mbox{0.00}} & {\scriptsize\mbox{80.00}} & {\scriptsize\mbox{0.00}} & {\scriptsize\mbox{0.00}} & {\scriptsize\mbox{0.00}} & {\scriptsize\mbox{0.00}} & {\scriptsize\mbox{0.00}} & {\scriptsize\mbox{70.00}} & {\scriptsize\mbox{0.00}} & {\scriptsize\mbox{0.00}} & {\scriptsize\mbox{0.00}} \\
RoboMeter-4B & {\scriptsize\mbox{20.00}} & {\scriptsize\mbox{30.00}} & {\scriptsize\mbox{10.00}} & {\scriptsize\mbox{10.00}} & {\scriptsize\mbox{10.00}} & {\scriptsize\mbox{40.00}} & {\scriptsize\mbox{\textbf{50.00}}} & {\scriptsize\mbox{\textbf{100.00}}} & {\scriptsize\mbox{0.00}} & {\scriptsize\mbox{30.00}} & {\scriptsize\mbox{\textbf{100.00}}} & {\scriptsize\mbox{10.00}} & {\scriptsize\mbox{0.00}} \\
\bottomrule
\end{tabularx}
\renewcommand{\scriptsize}{\fontsize{6}{7}\selectfont}
\par\vspace{1pt}
\parbox{\textwidth}{\scriptsize\raggedright \textbf{T1}: Arrange largest number.\enspace \textbf{T2}: Fill pen holder.\enspace \textbf{T3}: Fold clothes.\enspace \textbf{T4}: Hang mugs.\enspace \textbf{T5}: Insert tubes.\enspace \textbf{T6}: Make toast.\enspace \textbf{T7}: Organize table.\enspace \textbf{T8}: Pack objects into box.\enspace \textbf{T9}: Play stacking toy.\enspace \textbf{T10}: Put bottles into dustbin.\enspace \textbf{T11}: Stack blocks.\enspace \textbf{T12}: Store laptop and headphone.\enspace \textbf{T13}: Sweep blocks.}
\par\vspace{1pt}
\parbox{\textwidth}{\scriptsize\raggedright
$^{\dagger}$ 2.0 Preview checkpoints.\quad
$^{\ddagger}$ Qwen3-VL backbones.
}
\endgroup
\end{table}

\begin{table}[H]
\renewcommand{\scriptsize}{\fontsize{7}{8}\selectfont}
\renewcommand{\footnotesize}{\fontsize{9}{9.5}\selectfont}
\setlength{\belowcaptionskip}{4pt}
\centering
\caption{\textbf{SIA Results on Cross-Embodiment Simulation Trajectories.}}
\label{tab:emb_sia_part1}
\label{tab:emb_sia}
\begingroup
\footnotesize
\rmfamily
\setlength{\tabcolsep}{0.9pt}
\renewcommand{\arraystretch}{1.0}
\def\taskmodeband#1#2{%
  \specialrule{\lightrulewidth}{\aboverulesep}{2pt}
  \multicolumn{14}{@{}c@{}}{%
    \makebox[\textwidth][c]{%
      \rlap{\smash{\color{#1}%
        \rule[-\dp\csname @arstrutbox\endcsname]{\textwidth}%
        {\dimexpr\ht\csname @arstrutbox\endcsname
        +\dp\csname @arstrutbox\endcsname\relax}}}%
      \makebox[\textwidth][c]{\textit{#2}}}}\\%
  \specialrule{\lightrulewidth}{2pt}{\belowrulesep}
}
\begin{tabularx}{\textwidth}{
  @{} >{\raggedright\arraybackslash}p{0.24\textwidth}
  *{13}{>{\centering\arraybackslash}X} @{}
}
\toprule
\textbf{Model} & \textbf{T1} & \textbf{T2} & \textbf{T3} & \textbf{T4} & \textbf{T5} & \textbf{T6} & \textbf{T7} & \textbf{T8} & \textbf{T9} & \textbf{T10} & \textbf{T11} & \textbf{T12} & \textbf{T13} \\
\taskmodeband{RVZeroBand}{Zero-Shot Evaluation}
ProcVLM-2B & {\scriptsize\mbox{10.43}} & {\scriptsize\mbox{\textbf{13.38}}} & {\scriptsize\mbox{1.45}} & {\scriptsize\mbox{19.50}} & {\scriptsize\mbox{\textbf{21.38}}} & {\scriptsize\mbox{\textbf{32.11}}} & {\scriptsize\mbox{\textbf{17.91}}} & {\scriptsize\mbox{7.17}} & {\scriptsize\mbox{3.65}} & {\scriptsize\mbox{13.15}} & {\scriptsize\mbox{25.74}} & {\scriptsize\mbox{\textbf{23.69}}} & {\scriptsize\mbox{\textbf{12.00}}} \\
RoboFAC-7B & {\scriptsize\mbox{\textbf{11.69}}} & {\scriptsize\mbox{8.37}} & {\scriptsize\mbox{6.87}} & {\scriptsize\mbox{18.06}} & {\scriptsize\mbox{15.76}} & {\scriptsize\mbox{15.46}} & {\scriptsize\mbox{1.29}} & {\scriptsize\mbox{2.02}} & {\scriptsize\mbox{9.37}} & {\scriptsize\mbox{15.48}} & {\scriptsize\mbox{\textbf{29.85}}} & {\scriptsize\mbox{3.65}} & {\scriptsize\mbox{0.55}} \\
FailSafe & {\scriptsize\mbox{11.05}} & {\scriptsize\mbox{1.65}} & {\scriptsize\mbox{\textbf{10.23}}} & {\scriptsize\mbox{\textbf{19.73}}} & {\scriptsize\mbox{19.52}} & {\scriptsize\mbox{21.88}} & {\scriptsize\mbox{0.33}} & {\scriptsize\mbox{\textbf{11.00}}} & {\scriptsize\mbox{\textbf{13.22}}} & {\scriptsize\mbox{\textbf{17.43}}} & {\scriptsize\mbox{19.52}} & {\scriptsize\mbox{10.15}} & {\scriptsize\mbox{9.71}} \\
\bottomrule
\end{tabularx}
\renewcommand{\scriptsize}{\fontsize{6}{7}\selectfont}
\par\vspace{1pt}
\parbox{\textwidth}{\scriptsize\raggedright \textbf{T1}: Arrange largest number.\enspace \textbf{T2}: Fill pen holder.\enspace \textbf{T3}: Fold clothes.\enspace \textbf{T4}: Hang mugs.\enspace \textbf{T5}: Insert tubes.\enspace \textbf{T6}: Make toast.\enspace \textbf{T7}: Organize table.\enspace \textbf{T8}: Pack objects into box.\enspace \textbf{T9}: Play stacking toy.\enspace \textbf{T10}: Put bottles into dustbin.\enspace \textbf{T11}: Stack blocks.\enspace \textbf{T12}: Store laptop and headphone.\enspace \textbf{T13}: Sweep blocks.}
\par\vspace{1pt}
\parbox{\textwidth}{\scriptsize\raggedright
$^{\dagger}$ 2.0 Preview checkpoints.\quad
$^{\ddagger}$ Qwen3-VL backbones.
}
\endgroup
\end{table}

\begin{table}[H]
\renewcommand{\scriptsize}{\fontsize{7}{8}\selectfont}
\renewcommand{\footnotesize}{\fontsize{9}{9.75}\selectfont}
\setlength{\belowcaptionskip}{4pt}
\centering
\caption{\textbf{VOC Results on Cross-Embodiment Simulation Trajectories.}}
\label{tab:emb_voc_part1}
\label{tab:emb_voc}
\begingroup
\footnotesize
\rmfamily
\setlength{\tabcolsep}{0.9pt}
\renewcommand{\arraystretch}{1.0}
\def\taskmodeband#1#2{%
  \specialrule{\lightrulewidth}{\aboverulesep}{2pt}
  \multicolumn{14}{@{}c@{}}{%
    \makebox[\textwidth][c]{%
      \rlap{\smash{\color{#1}%
        \rule[-\dp\csname @arstrutbox\endcsname]{\textwidth}%
        {\dimexpr\ht\csname @arstrutbox\endcsname
        +\dp\csname @arstrutbox\endcsname\relax}}}%
      \makebox[\textwidth][c]{\textit{#2}}}}\\%
  \specialrule{\lightrulewidth}{2pt}{\belowrulesep}
}
\begin{tabularx}{\textwidth}{
  @{} >{\raggedright\arraybackslash}p{0.24\textwidth}
  *{13}{>{\centering\arraybackslash}X} @{}
}
\toprule
\textbf{Model} & \textbf{T1} & \textbf{T2} & \textbf{T3} & \textbf{T4} & \textbf{T5} & \textbf{T6} & \textbf{T7} & \textbf{T8} & \textbf{T9} & \textbf{T10} & \textbf{T11} & \textbf{T12} & \textbf{T13} \\
\taskmodeband{RVZeroBand}{Zero-Shot Evaluation}
LIV & {\scriptsize\mbox{-40.08}} & {\scriptsize\mbox{57.77}} & {\scriptsize\mbox{20.67}} & {\scriptsize\mbox{81.64}} & {\scriptsize\mbox{45.04}} & {\scriptsize\mbox{-28.38}} & {\scriptsize\mbox{45.16}} & {\scriptsize\mbox{6.65}} & {\scriptsize\mbox{18.60}} & {\scriptsize\mbox{28.28}} & {\scriptsize\mbox{-23.40}} & {\scriptsize\mbox{-31.61}} & {\scriptsize\mbox{40.16}} \\
ProcVLM-2B & {\scriptsize\mbox{10.66}} & {\scriptsize\mbox{82.24}} & {\scriptsize\mbox{59.18}} & {\scriptsize\mbox{91.14}} & {\scriptsize\mbox{77.68}} & {\scriptsize\mbox{22.00}} & {\scriptsize\mbox{69.97}} & {\scriptsize\mbox{92.14}} & {\scriptsize\mbox{49.10}} & {\scriptsize\mbox{8.42}} & {\scriptsize\mbox{50.78}} & {\scriptsize\mbox{75.88}} & {\scriptsize\mbox{48.57}} \\
RoboFAC-7B & {\scriptsize\mbox{32.28}} & {\scriptsize\mbox{69.23}} & {\scriptsize\mbox{-50.55}} & {\scriptsize\mbox{43.84}} & {\scriptsize\mbox{52.74}} & {\scriptsize\mbox{18.83}} & {\scriptsize\mbox{41.37}} & {\scriptsize\mbox{63.74}} & {\scriptsize\mbox{58.51}} & {\scriptsize\mbox{59.99}} & {\scriptsize\mbox{51.71}} & {\scriptsize\mbox{32.55}} & {\scriptsize\mbox{36.52}} \\
RoboMeter-4B & {\scriptsize\mbox{85.15}} & {\scriptsize\mbox{95.09}} & {\scriptsize\mbox{\textbf{95.67}}} & {\scriptsize\mbox{95.02}} & {\scriptsize\mbox{91.37}} & {\scriptsize\mbox{79.56}} & {\scriptsize\mbox{\textbf{95.76}}} & {\scriptsize\mbox{94.31}} & {\scriptsize\mbox{90.25}} & {\scriptsize\mbox{57.24}} & {\scriptsize\mbox{82.31}} & {\scriptsize\mbox{45.85}} & {\scriptsize\mbox{77.72}} \\
\bottomrule
\end{tabularx}
\renewcommand{\scriptsize}{\fontsize{6}{7}\selectfont}
\par\vspace{1pt}
\parbox{\textwidth}{\scriptsize\raggedright \textbf{T1}: Arrange largest number.\enspace \textbf{T2}: Fill pen holder.\enspace \textbf{T3}: Fold clothes.\enspace \textbf{T4}: Hang mugs.\enspace \textbf{T5}: Insert tubes.\enspace \textbf{T6}: Make toast.\enspace \textbf{T7}: Organize table.\enspace \textbf{T8}: Pack objects into box.\enspace \textbf{T9}: Play stacking toy.\enspace \textbf{T10}: Put bottles into dustbin.\enspace \textbf{T11}: Stack blocks.\enspace \textbf{T12}: Store laptop and headphone.\enspace \textbf{T13}: Sweep blocks.}
\par\vspace{1pt}
\parbox{\textwidth}{\scriptsize\raggedright
$^{\dagger}$ 2.0 Preview checkpoints.\quad
$^{\ddagger}$ Qwen3-VL backbones.
}
\endgroup
\end{table}

\begin{table}[H]
\renewcommand{\scriptsize}{\fontsize{7}{8}\selectfont}
\renewcommand{\footnotesize}{\fontsize{9}{11.5}\selectfont}
\setlength{\belowcaptionskip}{4pt}
\centering
\caption{\textbf{Cycle-VOC Results on Cross-Embodiment Simulation Trajectories.}}
\label{tab:emb_cycle_voc_part1}
\label{tab:emb_cycle_voc}
\begingroup
\footnotesize
\rmfamily
\setlength{\tabcolsep}{0.9pt}
\renewcommand{\arraystretch}{1.0}
\def\taskmodeband#1#2{%
  \specialrule{\lightrulewidth}{\aboverulesep}{2pt}
  \multicolumn{14}{@{}c@{}}{%
    \makebox[\textwidth][c]{%
      \rlap{\smash{\color{#1}%
        \rule[-\dp\csname @arstrutbox\endcsname]{\textwidth}%
        {\dimexpr\ht\csname @arstrutbox\endcsname
        +\dp\csname @arstrutbox\endcsname\relax}}}%
      \makebox[\textwidth][c]{\textit{#2}}}}\\%
  \specialrule{\lightrulewidth}{2pt}{\belowrulesep}
}
\begin{tabularx}{\textwidth}{
  @{} >{\raggedright\arraybackslash}p{0.24\textwidth}
  *{13}{>{\centering\arraybackslash}X} @{}
}
\toprule
\textbf{Model} & \textbf{T1} & \textbf{T2} & \textbf{T3} & \textbf{T4} & \textbf{T5} & \textbf{T6} & \textbf{T7} & \textbf{T8} & \textbf{T9} & \textbf{T10} & \textbf{T11} & \textbf{T12} & \textbf{T13} \\
\taskmodeband{RVZeroBand}{Zero-Shot Evaluation}
LIV & {\scriptsize\mbox{-40.08}} & {\scriptsize\mbox{57.77}} & {\scriptsize\mbox{20.67}} & {\scriptsize\mbox{81.64}} & {\scriptsize\mbox{45.04}} & {\scriptsize\mbox{-28.38}} & {\scriptsize\mbox{45.16}} & {\scriptsize\mbox{6.65}} & {\scriptsize\mbox{18.60}} & {\scriptsize\mbox{28.28}} & {\scriptsize\mbox{-23.40}} & {\scriptsize\mbox{-31.61}} & {\scriptsize\mbox{40.16}} \\
ProcVLM-2B & {\scriptsize\mbox{-0.23}} & {\scriptsize\mbox{71.79}} & {\scriptsize\mbox{38.67}} & {\scriptsize\mbox{88.81}} & {\scriptsize\mbox{69.10}} & {\scriptsize\mbox{22.52}} & {\scriptsize\mbox{61.85}} & {\scriptsize\mbox{89.83}} & {\scriptsize\mbox{42.79}} & {\scriptsize\mbox{-1.74}} & {\scriptsize\mbox{44.16}} & {\scriptsize\mbox{68.24}} & {\scriptsize\mbox{27.05}} \\
RoboFAC-7B & {\scriptsize\mbox{42.37}} & {\scriptsize\mbox{79.96}} & {\scriptsize\mbox{-40.39}} & {\scriptsize\mbox{55.21}} & {\scriptsize\mbox{67.60}} & {\scriptsize\mbox{15.48}} & {\scriptsize\mbox{58.20}} & {\scriptsize\mbox{77.53}} & {\scriptsize\mbox{61.72}} & {\scriptsize\mbox{66.36}} & {\scriptsize\mbox{63.48}} & {\scriptsize\mbox{55.55}} & {\scriptsize\mbox{59.48}} \\
RoboMeter-4B & {\scriptsize\mbox{\textbf{89.02}}} & {\scriptsize\mbox{\textbf{94.92}}} & {\scriptsize\mbox{\textbf{93.96}}} & {\scriptsize\mbox{96.07}} & {\scriptsize\mbox{\textbf{92.48}}} & {\scriptsize\mbox{\textbf{83.77}}} & {\scriptsize\mbox{\textbf{95.54}}} & {\scriptsize\mbox{93.39}} & {\scriptsize\mbox{\textbf{91.41}}} & {\scriptsize\mbox{73.45}} & {\scriptsize\mbox{86.28}} & {\scriptsize\mbox{67.89}} & {\scriptsize\mbox{\textbf{86.32}}} \\
\bottomrule
\end{tabularx}
\renewcommand{\scriptsize}{\fontsize{6}{7}\selectfont}
\par\vspace{1pt}
\parbox{\textwidth}{\scriptsize\raggedright \textbf{T1}: Arrange largest number.\enspace \textbf{T2}: Fill pen holder.\enspace \textbf{T3}: Fold clothes.\enspace \textbf{T4}: Hang mugs.\enspace \textbf{T5}: Insert tubes.\enspace \textbf{T6}: Make toast.\enspace \textbf{T7}: Organize table.\enspace \textbf{T8}: Pack objects into box.\enspace \textbf{T9}: Play stacking toy.\enspace \textbf{T10}: Put bottles into dustbin.\enspace \textbf{T11}: Stack blocks.\enspace \textbf{T12}: Store laptop and headphone.\enspace \textbf{T13}: Sweep blocks.}
\par\vspace{1pt}
\parbox{\textwidth}{\scriptsize\raggedright
$^{\dagger}$ 2.0 Preview checkpoints.\quad
$^{\ddagger}$ Qwen3-VL backbones.
}
\endgroup
\end{table}

\begin{table}[H]
\renewcommand{\scriptsize}{\fontsize{7}{8}\selectfont}
\renewcommand{\footnotesize}{\fontsize{9}{11.5}\selectfont}
\setlength{\belowcaptionskip}{4pt}
\centering
\caption{\textbf{Memory-VOC Results on Cross-Embodiment Simulation Trajectories.}}
\label{tab:emb_memory_voc_part1}
\label{tab:emb_memory_voc}
\label{tab:emb_voc_mem_part1}
\label{tab:emb_voc_mem}
\begingroup
\footnotesize
\rmfamily
\setlength{\tabcolsep}{0.9pt}
\renewcommand{\arraystretch}{1.0}
\def\taskmodeband#1#2{%
  \specialrule{\lightrulewidth}{\aboverulesep}{2pt}
  \multicolumn{3}{@{}c@{}}{%
    \makebox[\textwidth][c]{%
      \rlap{\smash{\color{#1}%
        \rule[-\dp\csname @arstrutbox\endcsname]{\textwidth}%
        {\dimexpr\ht\csname @arstrutbox\endcsname
        +\dp\csname @arstrutbox\endcsname\relax}}}%
      \makebox[\textwidth][c]{\textit{#2}}}}\\%
  \specialrule{\lightrulewidth}{2pt}{\belowrulesep}
}
\begin{tabularx}{\textwidth}{
  @{} >{\raggedright\arraybackslash}p{0.24\textwidth}
  *{2}{>{\centering\arraybackslash}X} @{}
}
\toprule
\textbf{Model} & \textbf{Press by number} & \textbf{Swap blocks} \\
\taskmodeband{RVZeroBand}{Zero-Shot Evaluation}
LIV & {\scriptsize\mbox{14.47}} & {\scriptsize\mbox{-20.30}} \\
ProcVLM-2B & {\scriptsize\mbox{20.81}} & {\scriptsize\mbox{18.43}} \\
RoboFAC-7B & {\scriptsize\mbox{9.04}} & {\scriptsize\mbox{5.83}} \\
RoboMeter-4B & {\scriptsize\mbox{10.74}} & {\scriptsize\mbox{-2.47}} \\
RoboReward-4B & {\scriptsize\mbox{83.49}} & {\scriptsize\mbox{75.73}} \\
RoboReward-8B & {\scriptsize\mbox{84.56}} & {\scriptsize\mbox{82.61}} \\
RynnValue-4B & {\scriptsize\mbox{-5.16}} & {\scriptsize\mbox{-30.25}} \\
RynnValue-8B & {\scriptsize\mbox{-18.97}} & {\scriptsize\mbox{-23.41}} \\
TOPReward-8B$^{\ddagger}$ & {\scriptsize\mbox{94.87}} & {\scriptsize\mbox{96.19}} \\
TOPReward-32B$^{\ddagger}$ & {\scriptsize\mbox{\textbf{97.26}}} & {\scriptsize\mbox{\textbf{98.40}}} \\
VLAC-2B & {\scriptsize\mbox{3.74}} & {\scriptsize\mbox{-0.26}} \\
VLAC-8B & {\scriptsize\mbox{37.91}} & {\scriptsize\mbox{16.73}} \\
\taskmodeband{RVOneBand}{One-Shot Evaluation}
ProcVLM-2B & {\scriptsize\mbox{37.51}} & {\scriptsize\mbox{29.58}} \\
Robo-Dopamine-3B & {\scriptsize\mbox{28.08}} & {\scriptsize\mbox{12.18}} \\
Robo-Dopamine-4B$^{\dagger}$ & {\scriptsize\mbox{7.99}} & {\scriptsize\mbox{21.27}} \\
Robo-Dopamine-8B & {\scriptsize\mbox{29.22}} & {\scriptsize\mbox{13.46}} \\
Robo-Dopamine-8B$^{\dagger}$ & {\scriptsize\mbox{32.91}} & {\scriptsize\mbox{-25.09}} \\
VLAC-2B & {\scriptsize\mbox{\textbf{81.28}}} & {\scriptsize\mbox{49.99}} \\
VLAC-8B & {\scriptsize\mbox{78.89}} & {\scriptsize\mbox{\textbf{78.31}}} \\
\bottomrule
\end{tabularx}
\renewcommand{\scriptsize}{\fontsize{6}{7}\selectfont}
\par\vspace{1pt}
\parbox{\textwidth}{\scriptsize\raggedright
$^{\dagger}$ 2.0 Preview checkpoints.\quad
$^{\ddagger}$ Qwen3-VL backbones.
}
\endgroup
\end{table}

\begin{table}[H]
\renewcommand{\scriptsize}{\fontsize{7}{8}\selectfont}
\renewcommand{\footnotesize}{\fontsize{9}{11.5}\selectfont}
\setlength{\belowcaptionskip}{4pt}
\centering
\caption{\textbf{FPL Results on Cross-Embodiment Simulation Trajectories.}}
\label{tab:emb_fpl_part1}
\label{tab:emb_fpl}
\begingroup
\footnotesize
\rmfamily
\setlength{\tabcolsep}{0.9pt}
\renewcommand{\arraystretch}{1.0}
\def\taskmodeband#1#2{%
  \specialrule{\lightrulewidth}{\aboverulesep}{2pt}
  \multicolumn{14}{@{}c@{}}{%
    \makebox[\textwidth][c]{%
      \rlap{\smash{\color{#1}%
        \rule[-\dp\csname @arstrutbox\endcsname]{\textwidth}%
        {\dimexpr\ht\csname @arstrutbox\endcsname
        +\dp\csname @arstrutbox\endcsname\relax}}}%
      \makebox[\textwidth][c]{\textit{#2}}}}\\%
  \specialrule{\lightrulewidth}{2pt}{\belowrulesep}
}
\begin{tabularx}{\textwidth}{
  @{} >{\raggedright\arraybackslash}p{0.24\textwidth}
  *{13}{>{\centering\arraybackslash}X} @{}
}
\toprule
\textbf{Model} & \textbf{T1} & \textbf{T2} & \textbf{T3} & \textbf{T4} & \textbf{T5} & \textbf{T6} & \textbf{T7} & \textbf{T8} & \textbf{T9} & \textbf{T10} & \textbf{T11} & \textbf{T12} & \textbf{T13} \\
\taskmodeband{RVZeroBand}{Zero-Shot Evaluation}
LIV & {\scriptsize\mbox{37.60}} & {\scriptsize\mbox{29.97}} & {\scriptsize\mbox{38.63}} & {\scriptsize\mbox{39.44}} & {\scriptsize\mbox{34.68}} & {\scriptsize\mbox{46.97}} & {\scriptsize\mbox{35.80}} & {\scriptsize\mbox{30.36}} & {\scriptsize\mbox{\textbf{10.66}}} & {\scriptsize\mbox{39.56}} & {\scriptsize\mbox{\textbf{22.07}}} & {\scriptsize\mbox{30.27}} & {\scriptsize\mbox{28.34}} \\
ProcVLM-2B & {\scriptsize\mbox{36.48}} & {\scriptsize\mbox{28.69}} & {\scriptsize\mbox{39.53}} & {\scriptsize\mbox{33.24}} & {\scriptsize\mbox{47.50}} & {\scriptsize\mbox{41.44}} & {\scriptsize\mbox{45.53}} & {\scriptsize\mbox{29.66}} & {\scriptsize\mbox{58.72}} & {\scriptsize\mbox{46.85}} & {\scriptsize\mbox{26.16}} & {\scriptsize\mbox{\textbf{12.15}}} & {\scriptsize\mbox{48.36}} \\
RoboFAC-7B & {\scriptsize\mbox{30.54}} & {\scriptsize\mbox{46.80}} & {\scriptsize\mbox{37.94}} & {\scriptsize\mbox{\textbf{13.02}}} & {\scriptsize\mbox{46.57}} & {\scriptsize\mbox{53.41}} & {\scriptsize\mbox{45.02}} & {\scriptsize\mbox{49.83}} & {\scriptsize\mbox{58.95}} & {\scriptsize\mbox{24.03}} & {\scriptsize\mbox{49.58}} & {\scriptsize\mbox{36.33}} & {\scriptsize\mbox{63.52}} \\
RoboMeter-4B & {\scriptsize\mbox{38.87}} & {\scriptsize\mbox{45.31}} & {\scriptsize\mbox{28.15}} & {\scriptsize\mbox{25.00}} & {\scriptsize\mbox{32.08}} & {\scriptsize\mbox{25.93}} & {\scriptsize\mbox{36.70}} & {\scriptsize\mbox{50.31}} & {\scriptsize\mbox{25.01}} & {\scriptsize\mbox{31.46}} & {\scriptsize\mbox{36.78}} & {\scriptsize\mbox{43.76}} & {\scriptsize\mbox{26.10}} \\
\bottomrule
\end{tabularx}
\renewcommand{\scriptsize}{\fontsize{6}{7}\selectfont}
\par\vspace{1pt}
\parbox{\textwidth}{\scriptsize\raggedright \textbf{T1}: Arrange largest number.\enspace \textbf{T2}: Fill pen holder.\enspace \textbf{T3}: Fold clothes.\enspace \textbf{T4}: Hang mugs.\enspace \textbf{T5}: Insert tubes.\enspace \textbf{T6}: Make toast.\enspace \textbf{T7}: Organize table.\enspace \textbf{T8}: Pack objects into box.\enspace \textbf{T9}: Play stacking toy.\enspace \textbf{T10}: Put bottles into dustbin.\enspace \textbf{T11}: Stack blocks.\enspace \textbf{T12}: Store laptop and headphone.\enspace \textbf{T13}: Sweep blocks.}
\par\vspace{1pt}
\parbox{\textwidth}{\scriptsize\raggedright
$^{\dagger}$ 2.0 Preview checkpoints.\quad
$^{\ddagger}$ Qwen3-VL backbones.
}
\endgroup
\end{table}

\begin{table}[H]
\renewcommand{\scriptsize}{\fontsize{7}{8}\selectfont}
\renewcommand{\footnotesize}{\fontsize{9}{11.5}\selectfont}
\setlength{\belowcaptionskip}{4pt}
\centering
\caption{\textbf{VS Results on Cross-Embodiment Simulation Trajectories.}}
\label{tab:emb_vs_part1}
\label{tab:emb_vs}
\begingroup
\footnotesize
\rmfamily
\setlength{\tabcolsep}{0.9pt}
\renewcommand{\arraystretch}{1.0}
\def\taskmodeband#1#2{%
  \specialrule{\lightrulewidth}{\aboverulesep}{2pt}
  \multicolumn{14}{@{}c@{}}{%
    \makebox[\textwidth][c]{%
      \rlap{\smash{\color{#1}%
        \rule[-\dp\csname @arstrutbox\endcsname]{\textwidth}%
        {\dimexpr\ht\csname @arstrutbox\endcsname
        +\dp\csname @arstrutbox\endcsname\relax}}}%
      \makebox[\textwidth][c]{\textit{#2}}}}\\%
  \specialrule{\lightrulewidth}{2pt}{\belowrulesep}
}
\begin{tabularx}{\textwidth}{
  @{} >{\raggedright\arraybackslash}p{0.24\textwidth}
  *{13}{>{\centering\arraybackslash}X} @{}
}
\toprule
\textbf{Model} & \textbf{T1} & \textbf{T2} & \textbf{T3} & \textbf{T4} & \textbf{T5} & \textbf{T6} & \textbf{T7} & \textbf{T8} & \textbf{T9} & \textbf{T10} & \textbf{T11} & \textbf{T12} & \textbf{T13} \\
\taskmodeband{RVZeroBand}{Zero-Shot Evaluation}
LIV & {\scriptsize\mbox{1.73}} & {\scriptsize\mbox{3.57}} & {\scriptsize\mbox{2.68}} & {\scriptsize\mbox{5.77}} & {\scriptsize\mbox{1.30}} & {\scriptsize\mbox{1.32}} & {\scriptsize\mbox{1.90}} & {\scriptsize\mbox{1.17}} & {\scriptsize\mbox{0.44}} & {\scriptsize\mbox{2.66}} & {\scriptsize\mbox{1.22}} & {\scriptsize\mbox{2.12}} & {\scriptsize\mbox{3.51}} \\
ProcVLM-2B & {\scriptsize\mbox{0.24}} & {\scriptsize\mbox{8.38}} & {\scriptsize\mbox{4.47}} & {\scriptsize\mbox{10.00}} & {\scriptsize\mbox{10.17}} & {\scriptsize\mbox{4.23}} & {\scriptsize\mbox{4.38}} & {\scriptsize\mbox{9.67}} & {\scriptsize\mbox{0.77}} & {\scriptsize\mbox{1.68}} & {\scriptsize\mbox{0.28}} & {\scriptsize\mbox{7.25}} & {\scriptsize\mbox{4.86}} \\
RoboFAC-7B & {\scriptsize\mbox{2.03}} & {\scriptsize\mbox{4.72}} & {\scriptsize\mbox{8.91}} & {\scriptsize\mbox{2.39}} & {\scriptsize\mbox{4.42}} & {\scriptsize\mbox{3.16}} & {\scriptsize\mbox{1.45}} & {\scriptsize\mbox{5.86}} & {\scriptsize\mbox{0.85}} & {\scriptsize\mbox{5.09}} & {\scriptsize\mbox{4.10}} & {\scriptsize\mbox{4.86}} & {\scriptsize\mbox{2.89}} \\
RoboMeter-4B & {\scriptsize\mbox{8.64}} & {\scriptsize\mbox{\textbf{23.85}}} & {\scriptsize\mbox{\textbf{36.59}}} & {\scriptsize\mbox{21.67}} & {\scriptsize\mbox{25.22}} & {\scriptsize\mbox{13.27}} & {\scriptsize\mbox{\textbf{19.20}}} & {\scriptsize\mbox{17.62}} & {\scriptsize\mbox{8.00}} & {\scriptsize\mbox{9.26}} & {\scriptsize\mbox{18.76}} & {\scriptsize\mbox{12.36}} & {\scriptsize\mbox{14.04}} \\
\bottomrule
\end{tabularx}
\renewcommand{\scriptsize}{\fontsize{6}{7}\selectfont}
\par\vspace{1pt}
\parbox{\textwidth}{\scriptsize\raggedright \textbf{T1}: Arrange largest number.\enspace \textbf{T2}: Fill pen holder.\enspace \textbf{T3}: Fold clothes.\enspace \textbf{T4}: Hang mugs.\enspace \textbf{T5}: Insert tubes.\enspace \textbf{T6}: Make toast.\enspace \textbf{T7}: Organize table.\enspace \textbf{T8}: Pack objects into box.\enspace \textbf{T9}: Play stacking toy.\enspace \textbf{T10}: Put bottles into dustbin.\enspace \textbf{T11}: Stack blocks.\enspace \textbf{T12}: Store laptop and headphone.\enspace \textbf{T13}: Sweep blocks.}
\par\vspace{1pt}
\parbox{\textwidth}{\scriptsize\raggedright
$^{\dagger}$ 2.0 Preview checkpoints.\quad
$^{\ddagger}$ Qwen3-VL backbones.
}
\endgroup
\end{table}

\FloatBarrier
\Needspace{29\baselineskip}
\subsubsection{Cross-Environment Generalization}\label{app:simulation_environment_results}

We retain ARX and vary surface appearance, lighting, and clutter while preserving the task goals. These results reveal the sensitivity of value predictions to changes in visual context.


\begin{table}[H]
\renewcommand{\scriptsize}{\fontsize{7}{8}\selectfont}
\renewcommand{\footnotesize}{\fontsize{9}{9.5}\selectfont}
\setlength{\belowcaptionskip}{4pt}
\centering
\caption{\textbf{SA Results on Cross-Environment Simulation Trajectories.}}
\label{tab:env_sa_part1}
\label{tab:env_sa}
\begingroup
\footnotesize
\rmfamily
\setlength{\tabcolsep}{0.9pt}
\renewcommand{\arraystretch}{1.0}
\def\taskmodeband#1#2{%
  \specialrule{\lightrulewidth}{\aboverulesep}{2pt}
  \multicolumn{14}{@{}c@{}}{%
    \makebox[\textwidth][c]{%
      \rlap{\smash{\color{#1}%
        \rule[-\dp\csname @arstrutbox\endcsname]{\textwidth}%
        {\dimexpr\ht\csname @arstrutbox\endcsname
        +\dp\csname @arstrutbox\endcsname\relax}}}%
      \makebox[\textwidth][c]{\textit{#2}}}}\\%
  \specialrule{\lightrulewidth}{2pt}{\belowrulesep}
}
\begin{tabularx}{\textwidth}{
  @{} >{\raggedright\arraybackslash}p{0.24\textwidth}
  *{13}{>{\centering\arraybackslash}X} @{}
}
\toprule
\textbf{Model} & \textbf{T1} & \textbf{T2} & \textbf{T3} & \textbf{T4} & \textbf{T5} & \textbf{T6} & \textbf{T7} & \textbf{T8} & \textbf{T9} & \textbf{T10} & \textbf{T11} & \textbf{T12} & \textbf{T13} \\
\taskmodeband{RVZeroBand}{Zero-Shot Evaluation}
LIV & {\scriptsize\mbox{\textbf{63.33}}} & {\scriptsize\mbox{60.00}} & {\scriptsize\mbox{48.33}} & {\scriptsize\mbox{\textbf{93.33}}} & {\scriptsize\mbox{70.00}} & {\scriptsize\mbox{45.00}} & {\scriptsize\mbox{51.25}} & {\scriptsize\mbox{37.00}} & {\scriptsize\mbox{67.50}} & {\scriptsize\mbox{46.67}} & {\scriptsize\mbox{76.67}} & {\scriptsize\mbox{43.33}} & {\scriptsize\mbox{43.33}} \\
ProcVLM-2B & {\scriptsize\mbox{46.67}} & {\scriptsize\mbox{51.00}} & {\scriptsize\mbox{71.67}} & {\scriptsize\mbox{70.00}} & {\scriptsize\mbox{45.00}} & {\scriptsize\mbox{26.67}} & {\scriptsize\mbox{51.25}} & {\scriptsize\mbox{\textbf{100.00}}} & {\scriptsize\mbox{\textbf{70.00}}} & {\scriptsize\mbox{0.00}} & {\scriptsize\mbox{68.33}} & {\scriptsize\mbox{88.33}} & {\scriptsize\mbox{80.00}} \\
RoboFAC-7B & {\scriptsize\mbox{45.00}} & {\scriptsize\mbox{92.00}} & {\scriptsize\mbox{68.33}} & {\scriptsize\mbox{43.33}} & {\scriptsize\mbox{82.50}} & {\scriptsize\mbox{30.00}} & {\scriptsize\mbox{67.50}} & {\scriptsize\mbox{\textbf{100.00}}} & {\scriptsize\mbox{57.50}} & {\scriptsize\mbox{23.33}} & {\scriptsize\mbox{91.67}} & {\scriptsize\mbox{90.00}} & {\scriptsize\mbox{\textbf{100.00}}} \\
RoboMeter-4B & {\scriptsize\mbox{56.67}} & {\scriptsize\mbox{\textbf{100.00}}} & {\scriptsize\mbox{\textbf{76.67}}} & {\scriptsize\mbox{73.33}} & {\scriptsize\mbox{\textbf{100.00}}} & {\scriptsize\mbox{60.00}} & {\scriptsize\mbox{\textbf{81.25}}} & {\scriptsize\mbox{96.00}} & {\scriptsize\mbox{47.50}} & {\scriptsize\mbox{66.67}} & {\scriptsize\mbox{75.00}} & {\scriptsize\mbox{60.00}} & {\scriptsize\mbox{\textbf{100.00}}} \\
\bottomrule
\end{tabularx}
\renewcommand{\scriptsize}{\fontsize{6}{7}\selectfont}
\par\vspace{1pt}
\parbox{\textwidth}{\scriptsize\raggedright \textbf{T1}: Arrange largest number.\enspace \textbf{T2}: Fill pen holder.\enspace \textbf{T3}: Fold clothes.\enspace \textbf{T4}: Hang mugs.\enspace \textbf{T5}: Insert tubes.\enspace \textbf{T6}: Make toast.\enspace \textbf{T7}: Organize table.\enspace \textbf{T8}: Pack objects into box.\enspace \textbf{T9}: Play stacking toy.\enspace \textbf{T10}: Put bottles into dustbin.\enspace \textbf{T11}: Stack blocks.\enspace \textbf{T12}: Store laptop and headphone.\enspace \textbf{T13}: Sweep blocks.}
\par\vspace{1pt}
\parbox{\textwidth}{\scriptsize\raggedright
$^{\dagger}$ 2.0 Preview checkpoints.\quad
$^{\ddagger}$ Qwen3-VL backbones.
}
\endgroup
\end{table}

\begin{table}[H]
\renewcommand{\scriptsize}{\fontsize{7}{8}\selectfont}
\renewcommand{\footnotesize}{\fontsize{9}{9.5}\selectfont}
\setlength{\belowcaptionskip}{4pt}
\centering
\caption{\textbf{TGA-CT Results on Cross-Environment Simulation Trajectories.}}
\label{tab:env_tga_ct_part1}
\label{tab:env_tga_ct}
\label{tab:env_tga_easy_part1}
\label{tab:env_tga_easy}
\begingroup
\footnotesize
\rmfamily
\setlength{\tabcolsep}{0.9pt}
\renewcommand{\arraystretch}{1.0}
\def\taskmodeband#1#2{%
  \specialrule{\lightrulewidth}{\aboverulesep}{2pt}
  \multicolumn{14}{@{}c@{}}{%
    \makebox[\textwidth][c]{%
      \rlap{\smash{\color{#1}%
        \rule[-\dp\csname @arstrutbox\endcsname]{\textwidth}%
        {\dimexpr\ht\csname @arstrutbox\endcsname
        +\dp\csname @arstrutbox\endcsname\relax}}}%
      \makebox[\textwidth][c]{\textit{#2}}}}\\%
  \specialrule{\lightrulewidth}{2pt}{\belowrulesep}
}
\begin{tabularx}{\textwidth}{
  @{} >{\raggedright\arraybackslash}p{0.24\textwidth}
  *{13}{>{\centering\arraybackslash}X} @{}
}
\toprule
\textbf{Model} & \textbf{T1} & \textbf{T2} & \textbf{T3} & \textbf{T4} & \textbf{T5} & \textbf{T6} & \textbf{T7} & \textbf{T8} & \textbf{T9} & \textbf{T10} & \textbf{T11} & \textbf{T12} & \textbf{T13} \\
\taskmodeband{RVZeroBand}{Zero-Shot Evaluation}
LIV & {\scriptsize\mbox{10.00}} & {\scriptsize\mbox{20.00}} & {\scriptsize\mbox{20.00}} & {\scriptsize\mbox{20.00}} & {\scriptsize\mbox{30.00}} & {\scriptsize\mbox{0.00}} & {\scriptsize\mbox{0.00}} & {\scriptsize\mbox{0.00}} & {\scriptsize\mbox{0.00}} & {\scriptsize\mbox{30.00}} & {\scriptsize\mbox{10.00}} & {\scriptsize\mbox{10.00}} & {\scriptsize\mbox{30.00}} \\
ProcVLM-2B & {\scriptsize\mbox{0.00}} & {\scriptsize\mbox{0.00}} & {\scriptsize\mbox{0.00}} & {\scriptsize\mbox{20.00}} & {\scriptsize\mbox{30.00}} & {\scriptsize\mbox{10.00}} & {\scriptsize\mbox{20.00}} & {\scriptsize\mbox{50.00}} & {\scriptsize\mbox{10.00}} & {\scriptsize\mbox{0.00}} & {\scriptsize\mbox{0.00}} & {\scriptsize\mbox{70.00}} & {\scriptsize\mbox{50.00}} \\
RoboFAC-7B & {\scriptsize\mbox{0.00}} & {\scriptsize\mbox{0.00}} & {\scriptsize\mbox{0.00}} & {\scriptsize\mbox{10.00}} & {\scriptsize\mbox{20.00}} & {\scriptsize\mbox{0.00}} & {\scriptsize\mbox{40.00}} & {\scriptsize\mbox{0.00}} & {\scriptsize\mbox{0.00}} & {\scriptsize\mbox{20.00}} & {\scriptsize\mbox{10.00}} & {\scriptsize\mbox{0.00}} & {\scriptsize\mbox{0.00}} \\
RoboMeter-4B & {\scriptsize\mbox{\textbf{100.00}}} & {\scriptsize\mbox{90.00}} & {\scriptsize\mbox{\textbf{100.00}}} & {\scriptsize\mbox{70.00}} & {\scriptsize\mbox{\textbf{100.00}}} & {\scriptsize\mbox{90.00}} & {\scriptsize\mbox{\textbf{100.00}}} & {\scriptsize\mbox{\textbf{100.00}}} & {\scriptsize\mbox{60.00}} & {\scriptsize\mbox{90.00}} & {\scriptsize\mbox{\textbf{90.00}}} & {\scriptsize\mbox{\textbf{100.00}}} & {\scriptsize\mbox{\textbf{100.00}}} \\
\bottomrule
\end{tabularx}
\renewcommand{\scriptsize}{\fontsize{6}{7}\selectfont}
\par\vspace{1pt}
\parbox{\textwidth}{\scriptsize\raggedright \textbf{T1}: Arrange largest number.\enspace \textbf{T2}: Fill pen holder.\enspace \textbf{T3}: Fold clothes.\enspace \textbf{T4}: Hang mugs.\enspace \textbf{T5}: Insert tubes.\enspace \textbf{T6}: Make toast.\enspace \textbf{T7}: Organize table.\enspace \textbf{T8}: Pack objects into box.\enspace \textbf{T9}: Play stacking toy.\enspace \textbf{T10}: Put bottles into dustbin.\enspace \textbf{T11}: Stack blocks.\enspace \textbf{T12}: Store laptop and headphone.\enspace \textbf{T13}: Sweep blocks.}
\par\vspace{1pt}
\parbox{\textwidth}{\scriptsize\raggedright
$^{\dagger}$ 2.0 Preview checkpoints.\quad
$^{\ddagger}$ Qwen3-VL backbones.
}
\endgroup
\end{table}

\begin{table}[H]
\renewcommand{\scriptsize}{\fontsize{7}{8}\selectfont}
\renewcommand{\footnotesize}{\fontsize{9}{9}\selectfont}
\setlength{\belowcaptionskip}{4pt}
\centering
\caption{\textbf{TGA-CF Results on Cross-Environment Simulation Trajectories.}}
\label{tab:env_tga_cf_part1}
\label{tab:env_tga_cf}
\label{tab:env_tga_hard_part1}
\label{tab:env_tga_hard}
\begingroup
\footnotesize
\rmfamily
\setlength{\tabcolsep}{0.9pt}
\renewcommand{\arraystretch}{1.0}
\def\taskmodeband#1#2{%
  \specialrule{\lightrulewidth}{\aboverulesep}{2pt}
  \multicolumn{14}{@{}c@{}}{%
    \makebox[\textwidth][c]{%
      \rlap{\smash{\color{#1}%
        \rule[-\dp\csname @arstrutbox\endcsname]{\textwidth}%
        {\dimexpr\ht\csname @arstrutbox\endcsname
        +\dp\csname @arstrutbox\endcsname\relax}}}%
      \makebox[\textwidth][c]{\textit{#2}}}}\\%
  \specialrule{\lightrulewidth}{2pt}{\belowrulesep}
}

\renewcommand{\scriptsize}{\fontsize{6}{7}\selectfont}
\par\vspace{1pt}
\parbox{\textwidth}{\scriptsize\raggedright \textbf{T1}: Arrange largest number.\enspace \textbf{T2}: Fill pen holder.\enspace \textbf{T3}: Fold clothes.\enspace \textbf{T4}: Hang mugs.\enspace \textbf{T5}: Insert tubes.\enspace \textbf{T6}: Make toast.\enspace \textbf{T7}: Organize table.\enspace \textbf{T8}: Pack objects into box.\enspace \textbf{T9}: Play stacking toy.\enspace \textbf{T10}: Put bottles into dustbin.\enspace \textbf{T11}: Stack blocks.\enspace \textbf{T12}: Store laptop and headphone.\enspace \textbf{T13}: Sweep blocks.}
\par\vspace{1pt}
\parbox{\textwidth}{\scriptsize\raggedright
$^{\dagger}$ 2.0 Preview checkpoints.\quad
$^{\ddagger}$ Qwen3-VL backbones.
}
\endgroup
\end{table}

\begin{table}[H]
\renewcommand{\scriptsize}{\fontsize{7}{8}\selectfont}
\renewcommand{\footnotesize}{\fontsize{9}{9.5}\selectfont}
\setlength{\belowcaptionskip}{4pt}
\centering
\caption{\textbf{SIA Results on Cross-Environment Simulation Trajectories.}}
\label{tab:env_sia_part1}
\label{tab:env_sia}
\begingroup
\footnotesize
\rmfamily
\setlength{\tabcolsep}{0.9pt}
\renewcommand{\arraystretch}{1.0}
\def\taskmodeband#1#2{%
  \specialrule{\lightrulewidth}{\aboverulesep}{2pt}
  \multicolumn{14}{@{}c@{}}{%
    \makebox[\textwidth][c]{%
      \rlap{\smash{\color{#1}%
        \rule[-\dp\csname @arstrutbox\endcsname]{\textwidth}%
        {\dimexpr\ht\csname @arstrutbox\endcsname
        +\dp\csname @arstrutbox\endcsname\relax}}}%
      \makebox[\textwidth][c]{\textit{#2}}}}\\%
  \specialrule{\lightrulewidth}{2pt}{\belowrulesep}
}

\renewcommand{\scriptsize}{\fontsize{6}{7}\selectfont}
\par\vspace{1pt}
\parbox{\textwidth}{\scriptsize\raggedright \textbf{T1}: Arrange largest number.\enspace \textbf{T2}: Fill pen holder.\enspace \textbf{T3}: Fold clothes.\enspace \textbf{T4}: Hang mugs.\enspace \textbf{T5}: Insert tubes.\enspace \textbf{T6}: Make toast.\enspace \textbf{T7}: Organize table.\enspace \textbf{T8}: Pack objects into box.\enspace \textbf{T9}: Play stacking toy.\enspace \textbf{T10}: Put bottles into dustbin.\enspace \textbf{T11}: Stack blocks.\enspace \textbf{T12}: Store laptop and headphone.\enspace \textbf{T13}: Sweep blocks.}
\par\vspace{1pt}
\parbox{\textwidth}{\scriptsize\raggedright
$^{\dagger}$ 2.0 Preview checkpoints.\quad
$^{\ddagger}$ Qwen3-VL backbones.
}
\endgroup
\end{table}

\begin{table}[H]
\renewcommand{\scriptsize}{\fontsize{7}{8}\selectfont}
\renewcommand{\footnotesize}{\fontsize{9}{9.75}\selectfont}
\setlength{\belowcaptionskip}{4pt}
\centering
\caption{\textbf{VOC Results on Cross-Environment Simulation Trajectories.}}
\label{tab:env_voc_part1}
\label{tab:env_voc}
\begingroup
\footnotesize
\rmfamily
\setlength{\tabcolsep}{0.9pt}
\renewcommand{\arraystretch}{1.0}
\def\taskmodeband#1#2{%
  \specialrule{\lightrulewidth}{\aboverulesep}{2pt}
  \multicolumn{14}{@{}c@{}}{%
    \makebox[\textwidth][c]{%
      \rlap{\smash{\color{#1}%
        \rule[-\dp\csname @arstrutbox\endcsname]{\textwidth}%
        {\dimexpr\ht\csname @arstrutbox\endcsname
        +\dp\csname @arstrutbox\endcsname\relax}}}%
      \makebox[\textwidth][c]{\textit{#2}}}}\\%
  \specialrule{\lightrulewidth}{2pt}{\belowrulesep}
}

\renewcommand{\scriptsize}{\fontsize{6}{7}\selectfont}
\par\vspace{1pt}
\parbox{\textwidth}{\scriptsize\raggedright \textbf{T1}: Arrange largest number.\enspace \textbf{T2}: Fill pen holder.\enspace \textbf{T3}: Fold clothes.\enspace \textbf{T4}: Hang mugs.\enspace \textbf{T5}: Insert tubes.\enspace \textbf{T6}: Make toast.\enspace \textbf{T7}: Organize table.\enspace \textbf{T8}: Pack objects into box.\enspace \textbf{T9}: Play stacking toy.\enspace \textbf{T10}: Put bottles into dustbin.\enspace \textbf{T11}: Stack blocks.\enspace \textbf{T12}: Store laptop and headphone.\enspace \textbf{T13}: Sweep blocks.}
\par\vspace{1pt}
\parbox{\textwidth}{\scriptsize\raggedright
$^{\dagger}$ 2.0 Preview checkpoints.\quad
$^{\ddagger}$ Qwen3-VL backbones.
}
\endgroup
\end{table}

\begin{table}[H]
\renewcommand{\scriptsize}{\fontsize{7}{8}\selectfont}
\renewcommand{\footnotesize}{\fontsize{9}{11.5}\selectfont}
\setlength{\belowcaptionskip}{4pt}
\centering
\caption{\textbf{Cycle-VOC Results on Cross-Environment Simulation Trajectories.}}
\label{tab:env_cycle_voc_part1}
\label{tab:env_cycle_voc}
\begingroup
\footnotesize
\rmfamily
\setlength{\tabcolsep}{0.9pt}
\renewcommand{\arraystretch}{1.0}
\def\taskmodeband#1#2{%
  \specialrule{\lightrulewidth}{\aboverulesep}{2pt}
  \multicolumn{14}{@{}c@{}}{%
    \makebox[\textwidth][c]{%
      \rlap{\smash{\color{#1}%
        \rule[-\dp\csname @arstrutbox\endcsname]{\textwidth}%
        {\dimexpr\ht\csname @arstrutbox\endcsname
        +\dp\csname @arstrutbox\endcsname\relax}}}%
      \makebox[\textwidth][c]{\textit{#2}}}}\\%
  \specialrule{\lightrulewidth}{2pt}{\belowrulesep}
}

\renewcommand{\scriptsize}{\fontsize{6}{7}\selectfont}
\par\vspace{1pt}
\parbox{\textwidth}{\scriptsize\raggedright \textbf{T1}: Arrange largest number.\enspace \textbf{T2}: Fill pen holder.\enspace \textbf{T3}: Fold clothes.\enspace \textbf{T4}: Hang mugs.\enspace \textbf{T5}: Insert tubes.\enspace \textbf{T6}: Make toast.\enspace \textbf{T7}: Organize table.\enspace \textbf{T8}: Pack objects into box.\enspace \textbf{T9}: Play stacking toy.\enspace \textbf{T10}: Put bottles into dustbin.\enspace \textbf{T11}: Stack blocks.\enspace \textbf{T12}: Store laptop and headphone.\enspace \textbf{T13}: Sweep blocks.}
\par\vspace{1pt}
\parbox{\textwidth}{\scriptsize\raggedright
$^{\dagger}$ 2.0 Preview checkpoints.\quad
$^{\ddagger}$ Qwen3-VL backbones.
}
\endgroup
\end{table}

\begin{table}[H]
\renewcommand{\scriptsize}{\fontsize{7}{8}\selectfont}
\renewcommand{\footnotesize}{\fontsize{9}{11.5}\selectfont}
\setlength{\belowcaptionskip}{4pt}
\centering
\caption{\textbf{Memory-VOC Results on Cross-Environment Simulation Trajectories.}}
\label{tab:env_memory_voc_part1}
\label{tab:env_memory_voc}
\label{tab:env_voc_mem_part1}
\label{tab:env_voc_mem}
\begingroup
\footnotesize
\rmfamily
\setlength{\tabcolsep}{0.9pt}
\renewcommand{\arraystretch}{1.0}
\def\taskmodeband#1#2{%
  \specialrule{\lightrulewidth}{\aboverulesep}{2pt}
  \multicolumn{3}{@{}c@{}}{%
    \makebox[\textwidth][c]{%
      \rlap{\smash{\color{#1}%
        \rule[-\dp\csname @arstrutbox\endcsname]{\textwidth}%
        {\dimexpr\ht\csname @arstrutbox\endcsname
        +\dp\csname @arstrutbox\endcsname\relax}}}%
      \makebox[\textwidth][c]{\textit{#2}}}}\\%
  \specialrule{\lightrulewidth}{2pt}{\belowrulesep}
}
\begin{tabularx}{\textwidth}{
  @{} >{\raggedright\arraybackslash}p{0.24\textwidth}
  *{2}{>{\centering\arraybackslash}X} @{}
}
\toprule
\textbf{Model} & \textbf{Press by number} & \textbf{Swap blocks} \\
\taskmodeband{RVZeroBand}{Zero-Shot Evaluation}
LIV & {\scriptsize\mbox{8.29}} & {\scriptsize\mbox{6.91}} \\
ProcVLM-2B & {\scriptsize\mbox{32.22}} & {\scriptsize\mbox{24.01}} \\
RoboFAC-7B & {\scriptsize\mbox{-56.49}} & {\scriptsize\mbox{-53.81}} \\
RoboMeter-4B & {\scriptsize\mbox{5.85}} & {\scriptsize\mbox{-23.78}} \\
RoboReward-4B & {\scriptsize\mbox{85.84}} & {\scriptsize\mbox{84.30}} \\
RoboReward-8B & {\scriptsize\mbox{75.31}} & {\scriptsize\mbox{88.46}} \\
RynnValue-4B & {\scriptsize\mbox{7.24}} & {\scriptsize\mbox{-22.44}} \\
RynnValue-8B & {\scriptsize\mbox{23.67}} & {\scriptsize\mbox{-22.13}} \\
TOPReward-8B$^{\ddagger}$ & {\scriptsize\mbox{93.06}} & {\scriptsize\mbox{90.56}} \\
TOPReward-32B$^{\ddagger}$ & {\scriptsize\mbox{\textbf{97.24}}} & {\scriptsize\mbox{\textbf{97.97}}} \\
VLAC-2B & {\scriptsize\mbox{-37.19}} & {\scriptsize\mbox{-29.60}} \\
VLAC-8B & {\scriptsize\mbox{2.16}} & {\scriptsize\mbox{-33.14}} \\
\taskmodeband{RVOneBand}{One-Shot Evaluation}
ProcVLM-2B & {\scriptsize\mbox{\textbf{56.50}}} & {\scriptsize\mbox{7.46}} \\
Robo-Dopamine-3B & {\scriptsize\mbox{35.30}} & {\scriptsize\mbox{8.26}} \\
Robo-Dopamine-4B$^{\dagger}$ & {\scriptsize\mbox{26.91}} & {\scriptsize\mbox{8.16}} \\
Robo-Dopamine-8B & {\scriptsize\mbox{26.93}} & {\scriptsize\mbox{-4.24}} \\
Robo-Dopamine-8B$^{\dagger}$ & {\scriptsize\mbox{40.12}} & {\scriptsize\mbox{\textbf{8.88}}} \\
VLAC-2B & {\scriptsize\mbox{26.70}} & {\scriptsize\mbox{-17.23}} \\
VLAC-8B & {\scriptsize\mbox{44.66}} & {\scriptsize\mbox{-32.45}} \\
\bottomrule
\end{tabularx}
\renewcommand{\scriptsize}{\fontsize{6}{7}\selectfont}
\par\vspace{1pt}
\parbox{\textwidth}{\scriptsize\raggedright
$^{\dagger}$ 2.0 Preview checkpoints.\quad
$^{\ddagger}$ Qwen3-VL backbones.
}
\endgroup
\end{table}

\begin{table}[H]
\renewcommand{\scriptsize}{\fontsize{7}{8}\selectfont}
\renewcommand{\footnotesize}{\fontsize{9}{10}\selectfont}
\setlength{\belowcaptionskip}{4pt}
\centering
\caption{\textbf{FPL Results on Cross-Environment Simulation Trajectories.}}
\label{tab:env_fpl_part1}
\label{tab:env_fpl}
\begingroup
\footnotesize
\rmfamily
\setlength{\tabcolsep}{0.9pt}
\renewcommand{\arraystretch}{1.0}
\def\taskmodeband#1#2{%
  \specialrule{\lightrulewidth}{\aboverulesep}{2pt}
  \multicolumn{14}{@{}c@{}}{%
    \makebox[\textwidth][c]{%
      \rlap{\smash{\color{#1}%
        \rule[-\dp\csname @arstrutbox\endcsname]{\textwidth}%
        {\dimexpr\ht\csname @arstrutbox\endcsname
        +\dp\csname @arstrutbox\endcsname\relax}}}%
      \makebox[\textwidth][c]{\textit{#2}}}}\\%
  \specialrule{\lightrulewidth}{2pt}{\belowrulesep}
}

\renewcommand{\scriptsize}{\fontsize{6}{7}\selectfont}
\par\vspace{1pt}
\parbox{\textwidth}{\scriptsize\raggedright \textbf{T1}: Arrange largest number.\enspace \textbf{T2}: Fill pen holder.\enspace \textbf{T3}: Fold clothes.\enspace \textbf{T4}: Hang mugs.\enspace \textbf{T5}: Insert tubes.\enspace \textbf{T6}: Make toast.\enspace \textbf{T7}: Organize table.\enspace \textbf{T8}: Pack objects into box.\enspace \textbf{T9}: Play stacking toy.\enspace \textbf{T10}: Put bottles into dustbin.\enspace \textbf{T11}: Stack blocks.\enspace \textbf{T12}: Store laptop and headphone.\enspace \textbf{T13}: Sweep blocks.}
\par\vspace{1pt}
\parbox{\textwidth}{\scriptsize\raggedright
$^{\dagger}$ 2.0 Preview checkpoints.\quad
$^{\ddagger}$ Qwen3-VL backbones.
}
\endgroup
\end{table}

\begin{table}[H]
\renewcommand{\scriptsize}{\fontsize{7}{8}\selectfont}
\renewcommand{\footnotesize}{\fontsize{9}{10}\selectfont}
\setlength{\belowcaptionskip}{4pt}
\centering
\caption{\textbf{VS Results on Cross-Environment Simulation Trajectories.}}
\label{tab:env_vs_part1}
\label{tab:env_vs}
\begingroup
\footnotesize
\rmfamily
\setlength{\tabcolsep}{0.9pt}
\renewcommand{\arraystretch}{1.0}
\def\taskmodeband#1#2{%
  \specialrule{\lightrulewidth}{\aboverulesep}{2pt}
  \multicolumn{14}{@{}c@{}}{%
    \makebox[\textwidth][c]{%
      \rlap{\smash{\color{#1}%
        \rule[-\dp\csname @arstrutbox\endcsname]{\textwidth}%
        {\dimexpr\ht\csname @arstrutbox\endcsname
        +\dp\csname @arstrutbox\endcsname\relax}}}%
      \makebox[\textwidth][c]{\textit{#2}}}}\\%
  \specialrule{\lightrulewidth}{2pt}{\belowrulesep}
}

\renewcommand{\scriptsize}{\fontsize{6}{7}\selectfont}
\par\vspace{1pt}
\parbox{\textwidth}{\scriptsize\raggedright \textbf{T1}: Arrange largest number.\enspace \textbf{T2}: Fill pen holder.\enspace \textbf{T3}: Fold clothes.\enspace \textbf{T4}: Hang mugs.\enspace \textbf{T5}: Insert tubes.\enspace \textbf{T6}: Make toast.\enspace \textbf{T7}: Organize table.\enspace \textbf{T8}: Pack objects into box.\enspace \textbf{T9}: Play stacking toy.\enspace \textbf{T10}: Put bottles into dustbin.\enspace \textbf{T11}: Stack blocks.\enspace \textbf{T12}: Store laptop and headphone.\enspace \textbf{T13}: Sweep blocks.}
\par\vspace{1pt}
\parbox{\textwidth}{\scriptsize\raggedright
$^{\dagger}$ 2.0 Preview checkpoints.\quad
$^{\ddagger}$ Qwen3-VL backbones.
}
\endgroup
\end{table}

\FloatBarrier
\subsection{Cross-Task Score Matrix Visualizations}
\label{app:simulation_tga_ct}

\begin{figure}[p]
  \centering
  \includegraphics[width=\linewidth,height=0.84\textheight,keepaspectratio]{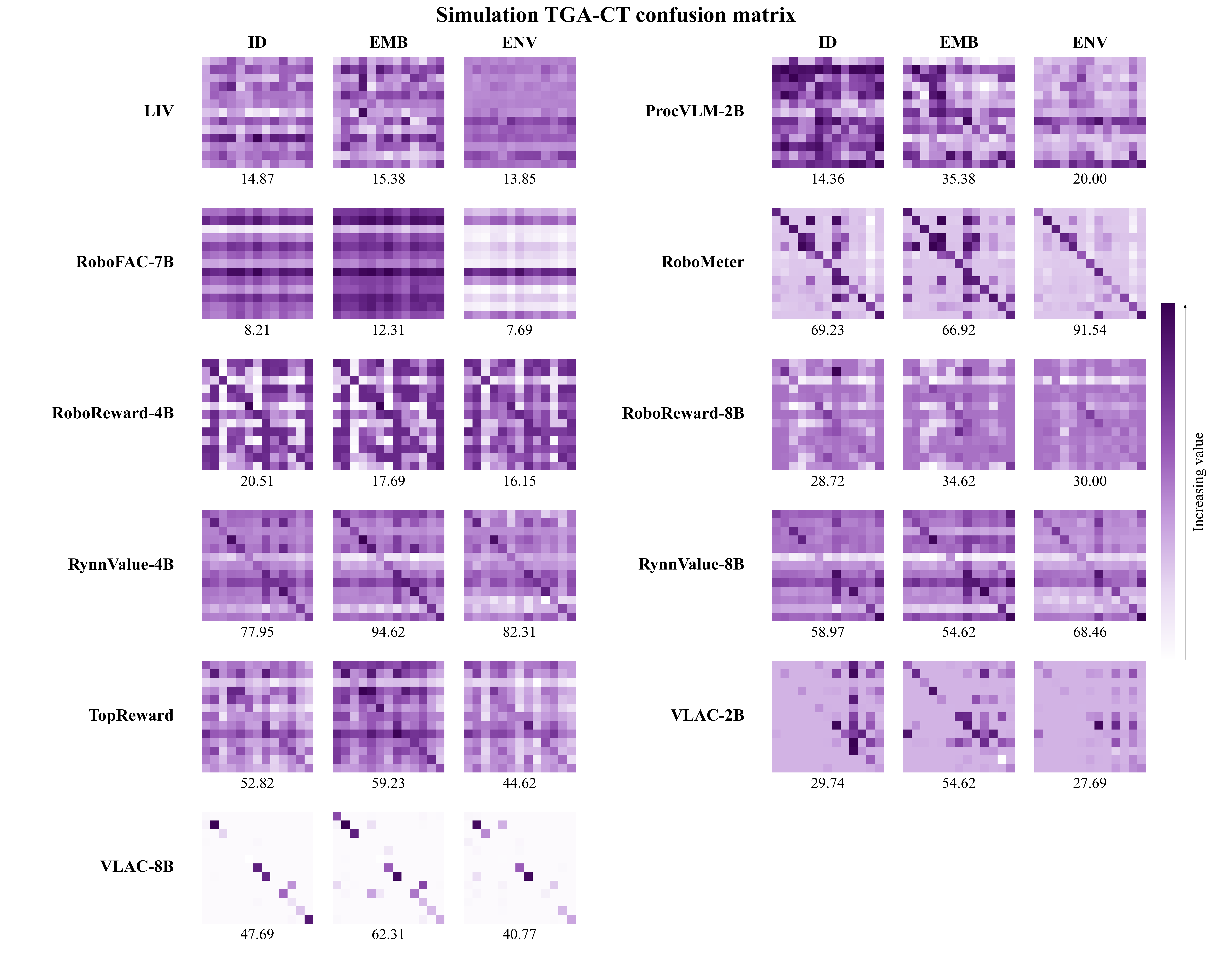}
  \caption{\textbf{Zero-Shot TGA-CT Score Matrices in Simulation.} Rows denote executed tasks and columns queried instructions; diagonal cells match the two. Darker cells indicate larger predicted gains within each matrix. Numbers below each matrix give TGA-CT accuracy (\%). TOPReward uses the 8B backbone.}
  \label{fig:tga_ct_sim_zero_shot}
  \par\vspace{10pt}
  \includegraphics[width=\linewidth,height=0.70\textheight,keepaspectratio]{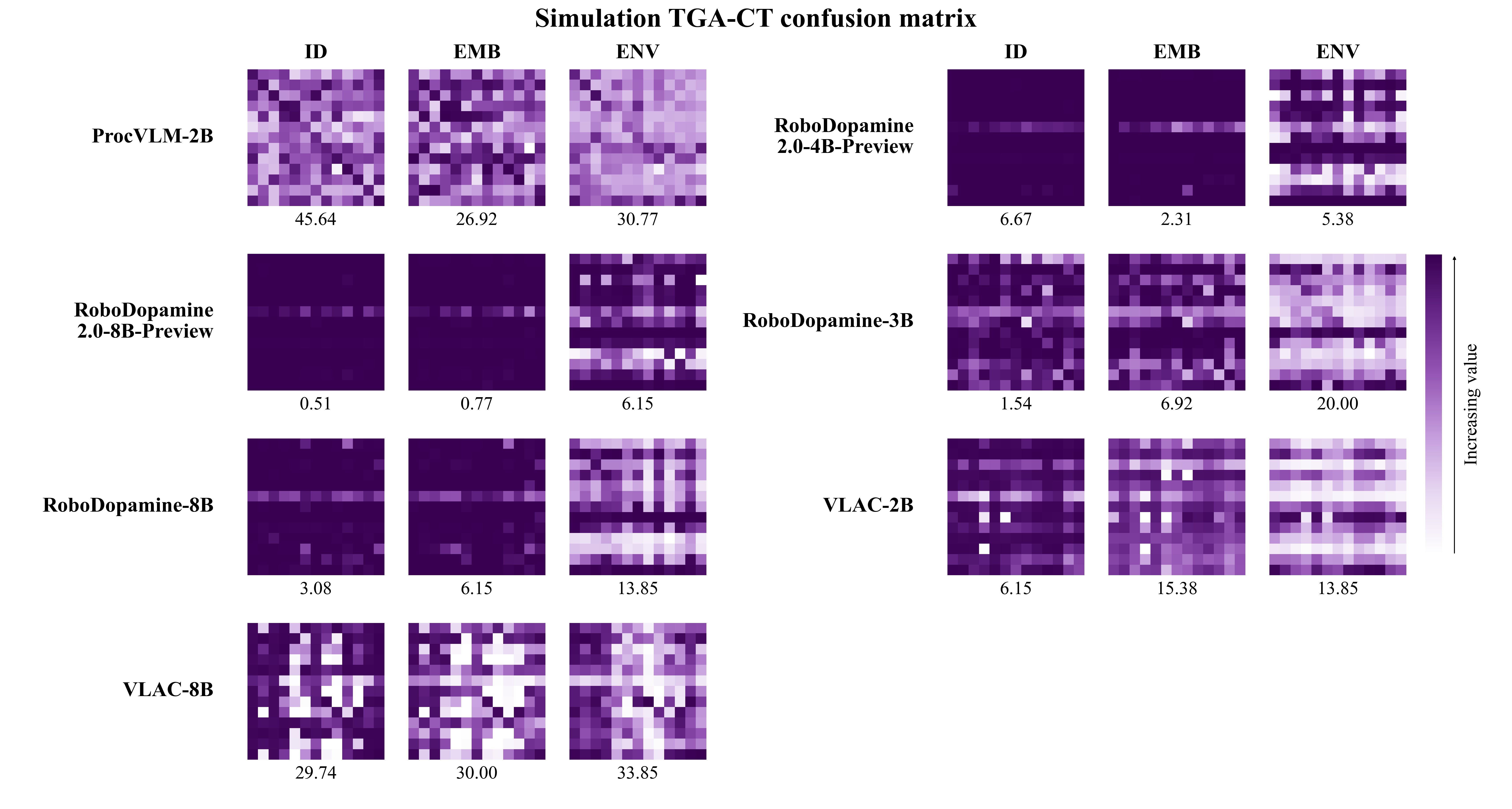}
  \caption{\textbf{One-Shot TGA-CT Score Matrices in Simulation.} Rows denote executed tasks and columns denote queried instructions. Darker cells indicate larger gains within each matrix; diagonal cells pair executions with matching instructions. Numbers below each matrix report TGA-CT accuracy (\%).}
  \label{fig:tga_ct_sim_one_shot}
\end{figure}

Figures~\ref{fig:tga_ct_sim_zero_shot} and~\ref{fig:tga_ct_sim_one_shot} show TGA-CT score matrices for 13 simulation tasks across the standard, cross-embodiment, and cross-environment settings. Each row compares the gain assigned to the matching instruction with gains assigned to alternative task instructions.

RoboMeter and RynnValue-4B show strong zero-shot diagonal selectivity. One-shot ProcVLM is more selective than the Robo-Dopamine variants, whose broad off-diagonal responses reveal gains assigned to unrelated instructions.

\FloatBarrier
\section{Full Real-World Benchmark Results}\label{app:realworld_results}
\subsection{Main Results}\label{app:realworld_id_results}

The following tables report in-domain (ID) results on real-world trajectories collected with the standard AgiBot Genie02 setup. These results provide the reference for the out-of-domain (OOD) comparisons below. Each table presents one metric, with task identifiers defined beneath it.


\begin{table}[H]
\renewcommand{\scriptsize}{\fontsize{7}{8}\selectfont}
\renewcommand{\footnotesize}{\fontsize{9}{9.5}\selectfont}
\setlength{\belowcaptionskip}{4pt}
\centering
\caption{\textbf{SA Results on In-Domain Real-World Trajectories.}}
\label{tab:real_id_sa_part1}
\label{tab:real_id_sa}
\begingroup
\footnotesize
\rmfamily
\setlength{\tabcolsep}{0.9pt}
\renewcommand{\arraystretch}{1.0}
\def\taskmodeband#1#2{%
  \specialrule{\lightrulewidth}{\aboverulesep}{2pt}
  \multicolumn{15}{@{}c@{}}{%
    \makebox[\textwidth][c]{%
      \rlap{\smash{\color{#1}%
        \rule[-\dp\csname @arstrutbox\endcsname]{\textwidth}%
        {\dimexpr\ht\csname @arstrutbox\endcsname
        +\dp\csname @arstrutbox\endcsname\relax}}}%
      \makebox[\textwidth][c]{\textit{#2}}}}\\%
  \specialrule{\lightrulewidth}{2pt}{\belowrulesep}
}

\renewcommand{\scriptsize}{\fontsize{6}{7}\selectfont}
\par\vspace{1pt}
\parbox{\textwidth}{\scriptsize\raggedright \textbf{T1}: Arrange family tea set.\enspace \textbf{T2}: Arrange flowers and reed diffuser.\enspace \textbf{T3}: Arrange seasoning jars.\enspace \textbf{T4}: Fill and place pen holder.\enspace \textbf{T5}: Load plastic plates into rack.\enspace \textbf{T6}: Organize books and pen.\enspace \textbf{T7}: Organize cosmetics storage box.\enspace \textbf{T8}: Organize desktop workstation.\enspace \textbf{T9}: Pack items for shipping.\enspace \textbf{T10}: Place pot on induction cooker.\enspace \textbf{T11}: Prepare breakfast.\enspace \textbf{T12}: Set the table.\enspace \textbf{T13}: Sort Blocks in Order.\enspace \textbf{T14}: Transfer Blocks Using Plate.}
\par\vspace{1pt}
\parbox{\textwidth}{\scriptsize\raggedright
$^{\dagger}$ 2.0 Preview checkpoints.\quad
$^{\ddagger}$ Qwen3-VL backbones.
}
\endgroup
\end{table}

\begin{table}[H]
\renewcommand{\scriptsize}{\fontsize{7}{8}\selectfont}
\renewcommand{\footnotesize}{\fontsize{9}{9.5}\selectfont}
\setlength{\belowcaptionskip}{4pt}
\centering
\caption{\textbf{TGA-CT Results on In-Domain Real-World Trajectories.}}
\label{tab:real_id_tga_ct_part1}
\label{tab:real_id_tga_ct}
\label{tab:real_id_tga_easy_part1}
\label{tab:real_id_tga_easy}
\begingroup
\footnotesize
\rmfamily
\setlength{\tabcolsep}{0.9pt}
\renewcommand{\arraystretch}{1.0}
\def\taskmodeband#1#2{%
  \specialrule{\lightrulewidth}{\aboverulesep}{2pt}
  \multicolumn{15}{@{}c@{}}{%
    \makebox[\textwidth][c]{%
      \rlap{\smash{\color{#1}%
        \rule[-\dp\csname @arstrutbox\endcsname]{\textwidth}%
        {\dimexpr\ht\csname @arstrutbox\endcsname
        +\dp\csname @arstrutbox\endcsname\relax}}}%
      \makebox[\textwidth][c]{\textit{#2}}}}\\%
  \specialrule{\lightrulewidth}{2pt}{\belowrulesep}
}

\renewcommand{\scriptsize}{\fontsize{6}{7}\selectfont}
\par\vspace{1pt}
\parbox{\textwidth}{\scriptsize\raggedright \textbf{T1}: Arrange family tea set.\enspace \textbf{T2}: Arrange flowers and reed diffuser.\enspace \textbf{T3}: Arrange seasoning jars.\enspace \textbf{T4}: Fill and place pen holder.\enspace \textbf{T5}: Load plastic plates into rack.\enspace \textbf{T6}: Organize books and pen.\enspace \textbf{T7}: Organize cosmetics storage box.\enspace \textbf{T8}: Organize desktop workstation.\enspace \textbf{T9}: Pack items for shipping.\enspace \textbf{T10}: Place pot on induction cooker.\enspace \textbf{T11}: Prepare breakfast.\enspace \textbf{T12}: Set the table.\enspace \textbf{T13}: Sort Blocks in Order.\enspace \textbf{T14}: Transfer Blocks Using Plate.}
\par\vspace{1pt}
\parbox{\textwidth}{\scriptsize\raggedright
$^{\dagger}$ 2.0 Preview checkpoints.\quad
$^{\ddagger}$ Qwen3-VL backbones.
}
\endgroup
\end{table}

\begin{table}[H]
\renewcommand{\scriptsize}{\fontsize{7}{8}\selectfont}
\renewcommand{\footnotesize}{\fontsize{9}{9}\selectfont}
\setlength{\belowcaptionskip}{4pt}
\centering
\caption{\textbf{TGA-CF Results on In-Domain Real-World Trajectories.}}
\label{tab:real_id_tga_cf_part1}
\label{tab:real_id_tga_cf}
\label{tab:real_id_tga_hard_part1}
\label{tab:real_id_tga_hard}
\begingroup
\footnotesize
\rmfamily
\setlength{\tabcolsep}{0.9pt}
\renewcommand{\arraystretch}{1.0}
\def\taskmodeband#1#2{%
  \specialrule{\lightrulewidth}{\aboverulesep}{2pt}
  \multicolumn{15}{@{}c@{}}{%
    \makebox[\textwidth][c]{%
      \rlap{\smash{\color{#1}%
        \rule[-\dp\csname @arstrutbox\endcsname]{\textwidth}%
        {\dimexpr\ht\csname @arstrutbox\endcsname
        +\dp\csname @arstrutbox\endcsname\relax}}}%
      \makebox[\textwidth][c]{\textit{#2}}}}\\%
  \specialrule{\lightrulewidth}{2pt}{\belowrulesep}
}

\renewcommand{\scriptsize}{\fontsize{6}{7}\selectfont}
\par\vspace{1pt}
\parbox{\textwidth}{\scriptsize\raggedright \textbf{T1}: Arrange family tea set.\enspace \textbf{T2}: Arrange flowers and reed diffuser.\enspace \textbf{T3}: Arrange seasoning jars.\enspace \textbf{T4}: Fill and place pen holder.\enspace \textbf{T5}: Load plastic plates into rack.\enspace \textbf{T6}: Organize books and pen.\enspace \textbf{T7}: Organize cosmetics storage box.\enspace \textbf{T8}: Organize desktop workstation.\enspace \textbf{T9}: Pack items for shipping.\enspace \textbf{T10}: Place pot on induction cooker.\enspace \textbf{T11}: Prepare breakfast.\enspace \textbf{T12}: Set the table.\enspace \textbf{T13}: Sort Blocks in Order.\enspace \textbf{T14}: Transfer Blocks Using Plate.}
\par\vspace{1pt}
\parbox{\textwidth}{\scriptsize\raggedright
$^{\dagger}$ 2.0 Preview checkpoints.\quad
$^{\ddagger}$ Qwen3-VL backbones.
}
\endgroup
\end{table}

\begin{table}[H]
\renewcommand{\scriptsize}{\fontsize{7}{8}\selectfont}
\renewcommand{\footnotesize}{\fontsize{9}{9.5}\selectfont}
\setlength{\belowcaptionskip}{4pt}
\centering
\caption{\textbf{SIA Results on In-Domain Real-World Trajectories.}}
\label{tab:real_id_sia_part1}
\label{tab:real_id_sia}
\begingroup
\footnotesize
\rmfamily
\setlength{\tabcolsep}{0.9pt}
\renewcommand{\arraystretch}{1.0}
\def\taskmodeband#1#2{%
  \specialrule{\lightrulewidth}{\aboverulesep}{2pt}
  \multicolumn{15}{@{}c@{}}{%
    \makebox[\textwidth][c]{%
      \rlap{\smash{\color{#1}%
        \rule[-\dp\csname @arstrutbox\endcsname]{\textwidth}%
        {\dimexpr\ht\csname @arstrutbox\endcsname
        +\dp\csname @arstrutbox\endcsname\relax}}}%
      \makebox[\textwidth][c]{\textit{#2}}}}\\%
  \specialrule{\lightrulewidth}{2pt}{\belowrulesep}
}

\renewcommand{\scriptsize}{\fontsize{6}{7}\selectfont}
\par\vspace{1pt}
\parbox{\textwidth}{\scriptsize\raggedright \textbf{T1}: Arrange family tea set.\enspace \textbf{T2}: Arrange flowers and reed diffuser.\enspace \textbf{T3}: Arrange seasoning jars.\enspace \textbf{T4}: Fill and place pen holder.\enspace \textbf{T5}: Load plastic plates into rack.\enspace \textbf{T6}: Organize books and pen.\enspace \textbf{T7}: Organize cosmetics storage box.\enspace \textbf{T8}: Organize desktop workstation.\enspace \textbf{T9}: Pack items for shipping.\enspace \textbf{T10}: Place pot on induction cooker.\enspace \textbf{T11}: Prepare breakfast.\enspace \textbf{T12}: Set the table.\enspace \textbf{T13}: Sort Blocks in Order.\enspace \textbf{T14}: Transfer Blocks Using Plate.}
\par\vspace{1pt}
\parbox{\textwidth}{\scriptsize\raggedright
$^{\dagger}$ 2.0 Preview checkpoints.\quad
$^{\ddagger}$ Qwen3-VL backbones.
}
\endgroup
\end{table}

\begin{table}[H]
\renewcommand{\scriptsize}{\fontsize{7}{8}\selectfont}
\renewcommand{\footnotesize}{\fontsize{9}{9}\selectfont}
\setlength{\belowcaptionskip}{4pt}
\centering
\caption{\textbf{VOC Results on In-Domain Real-World Trajectories.}}
\label{tab:real_id_voc_part1}
\label{tab:real_id_voc}
\begingroup
\footnotesize
\rmfamily
\setlength{\tabcolsep}{0.9pt}
\renewcommand{\arraystretch}{1.0}
\def\taskmodeband#1#2{%
  \specialrule{\lightrulewidth}{\aboverulesep}{2pt}
  \multicolumn{15}{@{}c@{}}{%
    \makebox[\textwidth][c]{%
      \rlap{\smash{\color{#1}%
        \rule[-\dp\csname @arstrutbox\endcsname]{\textwidth}%
        {\dimexpr\ht\csname @arstrutbox\endcsname
        +\dp\csname @arstrutbox\endcsname\relax}}}%
      \makebox[\textwidth][c]{\textit{#2}}}}\\%
  \specialrule{\lightrulewidth}{2pt}{\belowrulesep}
}

\renewcommand{\scriptsize}{\fontsize{6}{7}\selectfont}
\par\vspace{1pt}
\parbox{\textwidth}{\scriptsize\raggedright \textbf{T1}: Arrange family tea set.\enspace \textbf{T2}: Arrange flowers and reed diffuser.\enspace \textbf{T3}: Arrange seasoning jars.\enspace \textbf{T4}: Fill and place pen holder.\enspace \textbf{T5}: Load plastic plates into rack.\enspace \textbf{T6}: Organize books and pen.\enspace \textbf{T7}: Organize cosmetics storage box.\enspace \textbf{T8}: Organize desktop workstation.\enspace \textbf{T9}: Pack items for shipping.\enspace \textbf{T10}: Place pot on induction cooker.\enspace \textbf{T11}: Prepare breakfast.\enspace \textbf{T12}: Set the table.\enspace \textbf{T13}: Sort Blocks in Order.\enspace \textbf{T14}: Transfer Blocks Using Plate.}
\par\vspace{1pt}
\parbox{\textwidth}{\scriptsize\raggedright
$^{\dagger}$ 2.0 Preview checkpoints.\quad
$^{\ddagger}$ Qwen3-VL backbones.
}
\endgroup
\end{table}

\begin{table}[H]
\renewcommand{\scriptsize}{\fontsize{7}{8}\selectfont}
\renewcommand{\footnotesize}{\fontsize{9}{11.5}\selectfont}
\setlength{\belowcaptionskip}{4pt}
\centering
\caption{\textbf{Cycle-VOC Results on In-Domain Real-World Trajectories.}}
\label{tab:real_id_cycle_voc_part1}
\label{tab:real_id_cycle_voc}
\begingroup
\footnotesize
\rmfamily
\setlength{\tabcolsep}{0.9pt}
\renewcommand{\arraystretch}{1.0}
\def\taskmodeband#1#2{%
  \specialrule{\lightrulewidth}{\aboverulesep}{2pt}
  \multicolumn{15}{@{}c@{}}{%
    \makebox[\textwidth][c]{%
      \rlap{\smash{\color{#1}%
        \rule[-\dp\csname @arstrutbox\endcsname]{\textwidth}%
        {\dimexpr\ht\csname @arstrutbox\endcsname
        +\dp\csname @arstrutbox\endcsname\relax}}}%
      \makebox[\textwidth][c]{\textit{#2}}}}\\%
  \specialrule{\lightrulewidth}{2pt}{\belowrulesep}
}

\renewcommand{\scriptsize}{\fontsize{6}{7}\selectfont}
\par\vspace{1pt}
\parbox{\textwidth}{\scriptsize\raggedright \textbf{T1}: Arrange family tea set.\enspace \textbf{T2}: Arrange flowers and reed diffuser.\enspace \textbf{T3}: Arrange seasoning jars.\enspace \textbf{T4}: Fill and place pen holder.\enspace \textbf{T5}: Load plastic plates into rack.\enspace \textbf{T6}: Organize books and pen.\enspace \textbf{T7}: Organize cosmetics storage box.\enspace \textbf{T8}: Organize desktop workstation.\enspace \textbf{T9}: Pack items for shipping.\enspace \textbf{T10}: Place pot on induction cooker.\enspace \textbf{T11}: Prepare breakfast.\enspace \textbf{T12}: Set the table.\enspace \textbf{T13}: Sort Blocks in Order.\enspace \textbf{T14}: Transfer Blocks Using Plate.}
\par\vspace{1pt}
\parbox{\textwidth}{\scriptsize\raggedright
$^{\dagger}$ 2.0 Preview checkpoints.\quad
$^{\ddagger}$ Qwen3-VL backbones.
}
\endgroup
\end{table}

\begin{table}[H]
\renewcommand{\scriptsize}{\fontsize{7}{8}\selectfont}
\renewcommand{\footnotesize}{\fontsize{9}{11.5}\selectfont}
\setlength{\belowcaptionskip}{4pt}
\centering
\caption{\textbf{Memory-VOC Results on In-Domain Real-World Trajectories.}}
\label{tab:real_id_memory_voc_part1}
\label{tab:real_id_memory_voc}
\label{tab:real_id_voc_mem_part1}
\label{tab:real_id_voc_mem}
\begingroup
\footnotesize
\rmfamily
\setlength{\tabcolsep}{0.9pt}
\renewcommand{\arraystretch}{1.0}
\def\taskmodeband#1#2{%
  \specialrule{\lightrulewidth}{\aboverulesep}{2pt}
  \multicolumn{7}{@{}c@{}}{%
    \makebox[\textwidth][c]{%
      \rlap{\smash{\color{#1}%
        \rule[-\dp\csname @arstrutbox\endcsname]{\textwidth}%
        {\dimexpr\ht\csname @arstrutbox\endcsname
        +\dp\csname @arstrutbox\endcsname\relax}}}%
      \makebox[\textwidth][c]{\textit{#2}}}}\\%
  \specialrule{\lightrulewidth}{2pt}{\belowrulesep}
}

\renewcommand{\scriptsize}{\fontsize{6}{7}\selectfont}
\par\vspace{1pt}
\parbox{\textwidth}{\scriptsize\raggedright \textbf{T1}: Apply dish soap repeatedly.\enspace \textbf{T2}: Enter keypad number sequence.\enspace \textbf{T3}: Operate drawers in sequence.\enspace \textbf{T4}: Press button five times.\enspace \textbf{T5}: Season meat skewers repeatedly.\enspace \textbf{T6}: Strike drum five times.}
\par\vspace{1pt}
\parbox{\textwidth}{\scriptsize\raggedright
$^{\dagger}$ 2.0 Preview checkpoints.\quad
$^{\ddagger}$ Qwen3-VL backbones.
}
\endgroup
\end{table}

\begin{table}[H]
\renewcommand{\scriptsize}{\fontsize{7}{8}\selectfont}
\renewcommand{\footnotesize}{\fontsize{9}{11.5}\selectfont}
\setlength{\belowcaptionskip}{4pt}
\centering
\caption{\textbf{FPL Results on In-Domain Real-World Trajectories.}}
\label{tab:real_id_fpl_part1}
\label{tab:real_id_fpl}
\begingroup
\footnotesize
\rmfamily
\setlength{\tabcolsep}{0.9pt}
\renewcommand{\arraystretch}{1.0}
\def\taskmodeband#1#2{%
  \specialrule{\lightrulewidth}{\aboverulesep}{2pt}
  \multicolumn{15}{@{}c@{}}{%
    \makebox[\textwidth][c]{%
      \rlap{\smash{\color{#1}%
        \rule[-\dp\csname @arstrutbox\endcsname]{\textwidth}%
        {\dimexpr\ht\csname @arstrutbox\endcsname
        +\dp\csname @arstrutbox\endcsname\relax}}}%
      \makebox[\textwidth][c]{\textit{#2}}}}\\%
  \specialrule{\lightrulewidth}{2pt}{\belowrulesep}
}

\renewcommand{\scriptsize}{\fontsize{6}{7}\selectfont}
\par\vspace{1pt}
\parbox{\textwidth}{\scriptsize\raggedright \textbf{T1}: Arrange family tea set.\enspace \textbf{T2}: Arrange flowers and reed diffuser.\enspace \textbf{T3}: Arrange seasoning jars.\enspace \textbf{T4}: Fill and place pen holder.\enspace \textbf{T5}: Load plastic plates into rack.\enspace \textbf{T6}: Organize books and pen.\enspace \textbf{T7}: Organize cosmetics storage box.\enspace \textbf{T8}: Organize desktop workstation.\enspace \textbf{T9}: Pack items for shipping.\enspace \textbf{T10}: Place pot on induction cooker.\enspace \textbf{T11}: Prepare breakfast.\enspace \textbf{T12}: Set the table.\enspace \textbf{T13}: Sort Blocks in Order.\enspace \textbf{T14}: Transfer Blocks Using Plate.}
\par\vspace{1pt}
\parbox{\textwidth}{\scriptsize\raggedright
$^{\dagger}$ 2.0 Preview checkpoints.\quad
$^{\ddagger}$ Qwen3-VL backbones.
}
\endgroup
\end{table}

\begin{table}[H]
\renewcommand{\scriptsize}{\fontsize{7}{8}\selectfont}
\renewcommand{\footnotesize}{\fontsize{9}{11.5}\selectfont}
\setlength{\belowcaptionskip}{4pt}
\centering
\caption{\textbf{TRR Results on In-Domain Real-World Trajectories.}}
\label{tab:real_id_trr_part1}
\label{tab:real_id_trr}
\begingroup
\footnotesize
\rmfamily
\setlength{\tabcolsep}{0.9pt}
\renewcommand{\arraystretch}{1.0}
\def\taskmodeband#1#2{%
  \specialrule{\lightrulewidth}{\aboverulesep}{2pt}
  \multicolumn{15}{@{}c@{}}{%
    \makebox[\textwidth][c]{%
      \rlap{\smash{\color{#1}%
        \rule[-\dp\csname @arstrutbox\endcsname]{\textwidth}%
        {\dimexpr\ht\csname @arstrutbox\endcsname
        +\dp\csname @arstrutbox\endcsname\relax}}}%
      \makebox[\textwidth][c]{\textit{#2}}}}\\%
  \specialrule{\lightrulewidth}{2pt}{\belowrulesep}
}

\renewcommand{\scriptsize}{\fontsize{6}{7}\selectfont}
\par\vspace{1pt}
\parbox{\textwidth}{\scriptsize\raggedright \textbf{T1}: Arrange family tea set.\enspace \textbf{T2}: Arrange flowers and reed diffuser.\enspace \textbf{T3}: Arrange seasoning jars.\enspace \textbf{T4}: Fill and place pen holder.\enspace \textbf{T5}: Load plastic plates into rack.\enspace \textbf{T6}: Organize books and pen.\enspace \textbf{T7}: Organize cosmetics storage box.\enspace \textbf{T8}: Organize desktop workstation.\enspace \textbf{T9}: Pack items for shipping.\enspace \textbf{T10}: Place pot on induction cooker.\enspace \textbf{T11}: Prepare breakfast.\enspace \textbf{T12}: Set the table.\enspace \textbf{T13}: Sort Blocks in Order.\enspace \textbf{T14}: Transfer Blocks Using Plate.}
\par\vspace{1pt}
\parbox{\textwidth}{\scriptsize\raggedright
$^{\dagger}$ 2.0 Preview checkpoints.\quad
$^{\ddagger}$ Qwen3-VL backbones.
}
\endgroup
\end{table}

\begin{table}[H]
\renewcommand{\scriptsize}{\fontsize{7}{8}\selectfont}
\renewcommand{\footnotesize}{\fontsize{9}{11.5}\selectfont}
\setlength{\belowcaptionskip}{4pt}
\centering
\caption{\textbf{VS Results on In-Domain Real-World Trajectories.}}
\label{tab:real_id_vs_part1}
\label{tab:real_id_vs}
\begingroup
\footnotesize
\rmfamily
\setlength{\tabcolsep}{0.9pt}
\renewcommand{\arraystretch}{1.0}
\def\taskmodeband#1#2{%
  \specialrule{\lightrulewidth}{\aboverulesep}{2pt}
  \multicolumn{15}{@{}c@{}}{%
    \makebox[\textwidth][c]{%
      \rlap{\smash{\color{#1}%
        \rule[-\dp\csname @arstrutbox\endcsname]{\textwidth}%
        {\dimexpr\ht\csname @arstrutbox\endcsname
        +\dp\csname @arstrutbox\endcsname\relax}}}%
      \makebox[\textwidth][c]{\textit{#2}}}}\\%
  \specialrule{\lightrulewidth}{2pt}{\belowrulesep}
}

\renewcommand{\scriptsize}{\fontsize{6}{7}\selectfont}
\par\vspace{1pt}
\parbox{\textwidth}{\scriptsize\raggedright \textbf{T1}: Arrange family tea set.\enspace \textbf{T2}: Arrange flowers and reed diffuser.\enspace \textbf{T3}: Arrange seasoning jars.\enspace \textbf{T4}: Fill and place pen holder.\enspace \textbf{T5}: Load plastic plates into rack.\enspace \textbf{T6}: Organize books and pen.\enspace \textbf{T7}: Organize cosmetics storage box.\enspace \textbf{T8}: Organize desktop workstation.\enspace \textbf{T9}: Pack items for shipping.\enspace \textbf{T10}: Place pot on induction cooker.\enspace \textbf{T11}: Prepare breakfast.\enspace \textbf{T12}: Set the table.\enspace \textbf{T13}: Sort Blocks in Order.\enspace \textbf{T14}: Transfer Blocks Using Plate.}
\par\vspace{1pt}
\parbox{\textwidth}{\scriptsize\raggedright
$^{\dagger}$ 2.0 Preview checkpoints.\quad
$^{\ddagger}$ Qwen3-VL backbones.
}
\endgroup
\end{table}

\begin{table}[H]
\renewcommand{\scriptsize}{\fontsize{7}{8}\selectfont}
\renewcommand{\footnotesize}{\fontsize{9}{11.5}\selectfont}
\setlength{\belowcaptionskip}{4pt}
\centering
\caption{\textbf{CSVC Results on In-Domain Real-World Trajectories.}}
\label{tab:real_id_csvc_part1}
\label{tab:real_id_csvc}
\label{tab:real_id_cspc_part1}
\label{tab:real_id_cspc}
\begingroup
\footnotesize
\rmfamily
\setlength{\tabcolsep}{0.9pt}
\renewcommand{\arraystretch}{1.0}
\def\taskmodeband#1#2{%
  \specialrule{\lightrulewidth}{\aboverulesep}{2pt}
  \multicolumn{9}{@{}c@{}}{%
    \makebox[\textwidth][c]{%
      \rlap{\smash{\color{#1}%
        \rule[-\dp\csname @arstrutbox\endcsname]{\textwidth}%
        {\dimexpr\ht\csname @arstrutbox\endcsname
        +\dp\csname @arstrutbox\endcsname\relax}}}%
      \makebox[\textwidth][c]{\textit{#2}}}}\\%
  \specialrule{\lightrulewidth}{2pt}{\belowrulesep}
}

\renewcommand{\scriptsize}{\fontsize{6}{7}\selectfont}
\par\vspace{1pt}
\parbox{\textwidth}{\scriptsize\raggedright \textbf{T1}: Arrange family tea set.\enspace \textbf{T2}: Arrange flowers and reed diffuser.\enspace \textbf{T3}: Arrange seasoning jars.\enspace \textbf{T4}: Fill and place pen holder.\enspace \textbf{T5}: Load plastic plates into rack.\enspace \textbf{T6}: Organize cosmetics storage box.\enspace \textbf{T7}: Organize desktop workstation.\enspace \textbf{T8}: Set the table.}
\par\vspace{1pt}
\parbox{\textwidth}{\scriptsize\raggedright
$^{\dagger}$ 2.0 Preview checkpoints.\quad
$^{\ddagger}$ Qwen3-VL backbones.
}
\endgroup
\end{table}

\FloatBarrier
\Needspace{34\baselineskip}
\subsection{Generalization Results}\label{app:realworld_ood_results}

We evaluate out-of-domain (OOD) performance under separate embodiment and environment shifts. Comparing each setting with the in-domain (ID) results reveals how the shift affects each metric under the shared evaluation protocol.

\subsubsection{Cross-Embodiment Generalization}\label{app:realworld_embodiment_results}

We replace AgiBot Genie02 with a dual-arm ARX setup while preserving the task goals. This setting tests whether value predictions remain informative across changes in the physical robot and its execution patterns.


\begin{table}[H]
\renewcommand{\scriptsize}{\fontsize{7}{8}\selectfont}
\renewcommand{\footnotesize}{\fontsize{9}{9.5}\selectfont}
\setlength{\belowcaptionskip}{4pt}
\centering
\caption{\textbf{SA Results on Cross-Embodiment Real-World Trajectories.}}
\label{tab:real_emb_sa_part1}
\label{tab:real_emb_sa}
\begingroup
\footnotesize
\rmfamily
\setlength{\tabcolsep}{0.9pt}
\renewcommand{\arraystretch}{1.0}
\def\taskmodeband#1#2{%
  \specialrule{\lightrulewidth}{\aboverulesep}{2pt}
  \multicolumn{15}{@{}c@{}}{%
    \makebox[\textwidth][c]{%
      \rlap{\smash{\color{#1}%
        \rule[-\dp\csname @arstrutbox\endcsname]{\textwidth}%
        {\dimexpr\ht\csname @arstrutbox\endcsname
        +\dp\csname @arstrutbox\endcsname\relax}}}%
      \makebox[\textwidth][c]{\textit{#2}}}}\\%
  \specialrule{\lightrulewidth}{2pt}{\belowrulesep}
}

\renewcommand{\scriptsize}{\fontsize{6}{7}\selectfont}
\par\vspace{1pt}
\parbox{\textwidth}{\scriptsize\raggedright \textbf{T1}: Arrange family tea set.\enspace \textbf{T2}: Arrange flowers and reed diffuser.\enspace \textbf{T3}: Arrange seasoning jars.\enspace \textbf{T4}: Fill and place pen holder.\enspace \textbf{T5}: Load plastic plates into rack.\enspace \textbf{T6}: Organize books and pen.\enspace \textbf{T7}: Organize cosmetics storage box.\enspace \textbf{T8}: Organize desktop workstation.\enspace \textbf{T9}: Pack items for shipping.\enspace \textbf{T10}: Place pot on induction cooker.\enspace \textbf{T11}: Prepare breakfast.\enspace \textbf{T12}: Set the table.\enspace \textbf{T13}: Sort Blocks in Order.\enspace \textbf{T14}: Transfer Blocks Using Plate.}
\par\vspace{1pt}
\parbox{\textwidth}{\scriptsize\raggedright
$^{\dagger}$ 2.0 Preview checkpoints.\quad
$^{\ddagger}$ Qwen3-VL backbones.
}
\endgroup
\end{table}

\begin{table}[H]
\renewcommand{\scriptsize}{\fontsize{7}{8}\selectfont}
\renewcommand{\footnotesize}{\fontsize{9}{9.5}\selectfont}
\setlength{\belowcaptionskip}{4pt}
\centering
\caption{\textbf{TGA-CT Results on Cross-Embodiment Real-World Trajectories.}}
\label{tab:real_emb_tga_ct_part1}
\label{tab:real_emb_tga_ct}
\label{tab:real_emb_tga_easy_part1}
\label{tab:real_emb_tga_easy}
\begingroup
\footnotesize
\rmfamily
\setlength{\tabcolsep}{0.9pt}
\renewcommand{\arraystretch}{1.0}
\def\taskmodeband#1#2{%
  \specialrule{\lightrulewidth}{\aboverulesep}{2pt}
  \multicolumn{15}{@{}c@{}}{%
    \makebox[\textwidth][c]{%
      \rlap{\smash{\color{#1}%
        \rule[-\dp\csname @arstrutbox\endcsname]{\textwidth}%
        {\dimexpr\ht\csname @arstrutbox\endcsname
        +\dp\csname @arstrutbox\endcsname\relax}}}%
      \makebox[\textwidth][c]{\textit{#2}}}}\\%
  \specialrule{\lightrulewidth}{2pt}{\belowrulesep}
}

\renewcommand{\scriptsize}{\fontsize{6}{7}\selectfont}
\par\vspace{1pt}
\parbox{\textwidth}{\scriptsize\raggedright \textbf{T1}: Arrange family tea set.\enspace \textbf{T2}: Arrange flowers and reed diffuser.\enspace \textbf{T3}: Arrange seasoning jars.\enspace \textbf{T4}: Fill and place pen holder.\enspace \textbf{T5}: Load plastic plates into rack.\enspace \textbf{T6}: Organize books and pen.\enspace \textbf{T7}: Organize cosmetics storage box.\enspace \textbf{T8}: Organize desktop workstation.\enspace \textbf{T9}: Pack items for shipping.\enspace \textbf{T10}: Place pot on induction cooker.\enspace \textbf{T11}: Prepare breakfast.\enspace \textbf{T12}: Set the table.\enspace \textbf{T13}: Sort Blocks in Order.\enspace \textbf{T14}: Transfer Blocks Using Plate.}
\par\vspace{1pt}
\parbox{\textwidth}{\scriptsize\raggedright
$^{\dagger}$ 2.0 Preview checkpoints.\quad
$^{\ddagger}$ Qwen3-VL backbones.
}
\endgroup
\end{table}

\begin{table}[H]
\renewcommand{\scriptsize}{\fontsize{7}{8}\selectfont}
\renewcommand{\footnotesize}{\fontsize{9}{9}\selectfont}
\setlength{\belowcaptionskip}{4pt}
\centering
\caption{\textbf{TGA-CF Results on Cross-Embodiment Real-World Trajectories.}}
\label{tab:real_emb_tga_cf_part1}
\label{tab:real_emb_tga_cf}
\label{tab:real_emb_tga_hard_part1}
\label{tab:real_emb_tga_hard}
\begingroup
\footnotesize
\rmfamily
\setlength{\tabcolsep}{0.9pt}
\renewcommand{\arraystretch}{1.0}
\def\taskmodeband#1#2{%
  \specialrule{\lightrulewidth}{\aboverulesep}{2pt}
  \multicolumn{15}{@{}c@{}}{%
    \makebox[\textwidth][c]{%
      \rlap{\smash{\color{#1}%
        \rule[-\dp\csname @arstrutbox\endcsname]{\textwidth}%
        {\dimexpr\ht\csname @arstrutbox\endcsname
        +\dp\csname @arstrutbox\endcsname\relax}}}%
      \makebox[\textwidth][c]{\textit{#2}}}}\\%
  \specialrule{\lightrulewidth}{2pt}{\belowrulesep}
}

\renewcommand{\scriptsize}{\fontsize{6}{7}\selectfont}
\par\vspace{1pt}
\parbox{\textwidth}{\scriptsize\raggedright \textbf{T1}: Arrange family tea set.\enspace \textbf{T2}: Arrange flowers and reed diffuser.\enspace \textbf{T3}: Arrange seasoning jars.\enspace \textbf{T4}: Fill and place pen holder.\enspace \textbf{T5}: Load plastic plates into rack.\enspace \textbf{T6}: Organize books and pen.\enspace \textbf{T7}: Organize cosmetics storage box.\enspace \textbf{T8}: Organize desktop workstation.\enspace \textbf{T9}: Pack items for shipping.\enspace \textbf{T10}: Place pot on induction cooker.\enspace \textbf{T11}: Prepare breakfast.\enspace \textbf{T12}: Set the table.\enspace \textbf{T13}: Sort Blocks in Order.\enspace \textbf{T14}: Transfer Blocks Using Plate.}
\par\vspace{1pt}
\parbox{\textwidth}{\scriptsize\raggedright
$^{\dagger}$ 2.0 Preview checkpoints.\quad
$^{\ddagger}$ Qwen3-VL backbones.
}
\endgroup
\end{table}

\begin{table}[H]
\renewcommand{\scriptsize}{\fontsize{7}{8}\selectfont}
\renewcommand{\footnotesize}{\fontsize{9}{9.5}\selectfont}
\setlength{\belowcaptionskip}{4pt}
\centering
\caption{\textbf{SIA Results on Cross-Embodiment Real-World Trajectories.}}
\label{tab:real_emb_sia_part1}
\label{tab:real_emb_sia}
\begingroup
\footnotesize
\rmfamily
\setlength{\tabcolsep}{0.9pt}
\renewcommand{\arraystretch}{1.0}
\def\taskmodeband#1#2{%
  \specialrule{\lightrulewidth}{\aboverulesep}{2pt}
  \multicolumn{15}{@{}c@{}}{%
    \makebox[\textwidth][c]{%
      \rlap{\smash{\color{#1}%
        \rule[-\dp\csname @arstrutbox\endcsname]{\textwidth}%
        {\dimexpr\ht\csname @arstrutbox\endcsname
        +\dp\csname @arstrutbox\endcsname\relax}}}%
      \makebox[\textwidth][c]{\textit{#2}}}}\\%
  \specialrule{\lightrulewidth}{2pt}{\belowrulesep}
}

\renewcommand{\scriptsize}{\fontsize{6}{7}\selectfont}
\par\vspace{1pt}
\parbox{\textwidth}{\scriptsize\raggedright \textbf{T1}: Arrange family tea set.\enspace \textbf{T2}: Arrange flowers and reed diffuser.\enspace \textbf{T3}: Arrange seasoning jars.\enspace \textbf{T4}: Fill and place pen holder.\enspace \textbf{T5}: Load plastic plates into rack.\enspace \textbf{T6}: Organize books and pen.\enspace \textbf{T7}: Organize cosmetics storage box.\enspace \textbf{T8}: Organize desktop workstation.\enspace \textbf{T9}: Pack items for shipping.\enspace \textbf{T10}: Place pot on induction cooker.\enspace \textbf{T11}: Prepare breakfast.\enspace \textbf{T12}: Set the table.\enspace \textbf{T13}: Sort Blocks in Order.\enspace \textbf{T14}: Transfer Blocks Using Plate.}
\par\vspace{1pt}
\parbox{\textwidth}{\scriptsize\raggedright
$^{\dagger}$ 2.0 Preview checkpoints.\quad
$^{\ddagger}$ Qwen3-VL backbones.
}
\endgroup
\end{table}

\begin{table}[H]
\renewcommand{\scriptsize}{\fontsize{7}{8}\selectfont}
\renewcommand{\footnotesize}{\fontsize{9}{9}\selectfont}
\setlength{\belowcaptionskip}{4pt}
\centering
\caption{\textbf{VOC Results on Cross-Embodiment Real-World Trajectories.}}
\label{tab:real_emb_voc_part1}
\label{tab:real_emb_voc}
\begingroup
\footnotesize
\rmfamily
\setlength{\tabcolsep}{0.9pt}
\renewcommand{\arraystretch}{1.0}
\def\taskmodeband#1#2{%
  \specialrule{\lightrulewidth}{\aboverulesep}{2pt}
  \multicolumn{15}{@{}c@{}}{%
    \makebox[\textwidth][c]{%
      \rlap{\smash{\color{#1}%
        \rule[-\dp\csname @arstrutbox\endcsname]{\textwidth}%
        {\dimexpr\ht\csname @arstrutbox\endcsname
        +\dp\csname @arstrutbox\endcsname\relax}}}%
      \makebox[\textwidth][c]{\textit{#2}}}}\\%
  \specialrule{\lightrulewidth}{2pt}{\belowrulesep}
}

\renewcommand{\scriptsize}{\fontsize{6}{7}\selectfont}
\par\vspace{1pt}
\parbox{\textwidth}{\scriptsize\raggedright \textbf{T1}: Arrange family tea set.\enspace \textbf{T2}: Arrange flowers and reed diffuser.\enspace \textbf{T3}: Arrange seasoning jars.\enspace \textbf{T4}: Fill and place pen holder.\enspace \textbf{T5}: Load plastic plates into rack.\enspace \textbf{T6}: Organize books and pen.\enspace \textbf{T7}: Organize cosmetics storage box.\enspace \textbf{T8}: Organize desktop workstation.\enspace \textbf{T9}: Pack items for shipping.\enspace \textbf{T10}: Place pot on induction cooker.\enspace \textbf{T11}: Prepare breakfast.\enspace \textbf{T12}: Set the table.\enspace \textbf{T13}: Sort Blocks in Order.\enspace \textbf{T14}: Transfer Blocks Using Plate.}
\par\vspace{1pt}
\parbox{\textwidth}{\scriptsize\raggedright
$^{\dagger}$ 2.0 Preview checkpoints.\quad
$^{\ddagger}$ Qwen3-VL backbones.
}
\endgroup
\end{table}

\begin{table}[H]
\renewcommand{\scriptsize}{\fontsize{7}{8}\selectfont}
\renewcommand{\footnotesize}{\fontsize{9}{11.5}\selectfont}
\setlength{\belowcaptionskip}{4pt}
\centering
\caption{\textbf{Cycle-VOC Results on Cross-Embodiment Real-World Trajectories.}}
\label{tab:real_emb_cycle_voc_part1}
\label{tab:real_emb_cycle_voc}
\begingroup
\footnotesize
\rmfamily
\setlength{\tabcolsep}{0.9pt}
\renewcommand{\arraystretch}{1.0}
\def\taskmodeband#1#2{%
  \specialrule{\lightrulewidth}{\aboverulesep}{2pt}
  \multicolumn{15}{@{}c@{}}{%
    \makebox[\textwidth][c]{%
      \rlap{\smash{\color{#1}%
        \rule[-\dp\csname @arstrutbox\endcsname]{\textwidth}%
        {\dimexpr\ht\csname @arstrutbox\endcsname
        +\dp\csname @arstrutbox\endcsname\relax}}}%
      \makebox[\textwidth][c]{\textit{#2}}}}\\%
  \specialrule{\lightrulewidth}{2pt}{\belowrulesep}
}

\renewcommand{\scriptsize}{\fontsize{6}{7}\selectfont}
\par\vspace{1pt}
\parbox{\textwidth}{\scriptsize\raggedright \textbf{T1}: Arrange family tea set.\enspace \textbf{T2}: Arrange flowers and reed diffuser.\enspace \textbf{T3}: Arrange seasoning jars.\enspace \textbf{T4}: Fill and place pen holder.\enspace \textbf{T5}: Load plastic plates into rack.\enspace \textbf{T6}: Organize books and pen.\enspace \textbf{T7}: Organize cosmetics storage box.\enspace \textbf{T8}: Organize desktop workstation.\enspace \textbf{T9}: Pack items for shipping.\enspace \textbf{T10}: Place pot on induction cooker.\enspace \textbf{T11}: Prepare breakfast.\enspace \textbf{T12}: Set the table.\enspace \textbf{T13}: Sort Blocks in Order.\enspace \textbf{T14}: Transfer Blocks Using Plate.}
\par\vspace{1pt}
\parbox{\textwidth}{\scriptsize\raggedright
$^{\dagger}$ 2.0 Preview checkpoints.\quad
$^{\ddagger}$ Qwen3-VL backbones.
}
\endgroup
\end{table}

\begin{table}[H]
\renewcommand{\scriptsize}{\fontsize{7}{8}\selectfont}
\renewcommand{\footnotesize}{\fontsize{9}{11.5}\selectfont}
\setlength{\belowcaptionskip}{4pt}
\centering
\caption{\textbf{Memory-VOC Results on Cross-Embodiment Real-World Trajectories.}}
\label{tab:real_emb_memory_voc_part1}
\label{tab:real_emb_memory_voc}
\label{tab:real_emb_voc_mem_part1}
\label{tab:real_emb_voc_mem}
\begingroup
\footnotesize
\rmfamily
\setlength{\tabcolsep}{0.9pt}
\renewcommand{\arraystretch}{1.0}
\def\taskmodeband#1#2{%
  \specialrule{\lightrulewidth}{\aboverulesep}{2pt}
  \multicolumn{7}{@{}c@{}}{%
    \makebox[\textwidth][c]{%
      \rlap{\smash{\color{#1}%
        \rule[-\dp\csname @arstrutbox\endcsname]{\textwidth}%
        {\dimexpr\ht\csname @arstrutbox\endcsname
        +\dp\csname @arstrutbox\endcsname\relax}}}%
      \makebox[\textwidth][c]{\textit{#2}}}}\\%
  \specialrule{\lightrulewidth}{2pt}{\belowrulesep}
}

\renewcommand{\scriptsize}{\fontsize{6}{7}\selectfont}
\par\vspace{1pt}
\parbox{\textwidth}{\scriptsize\raggedright \textbf{T1}: Apply dish soap repeatedly.\enspace \textbf{T2}: Enter keypad number sequence.\enspace \textbf{T3}: Operate drawers in sequence.\enspace \textbf{T4}: Press button five times.\enspace \textbf{T5}: Season meat skewers repeatedly.\enspace \textbf{T6}: Strike drum five times.}
\par\vspace{1pt}
\parbox{\textwidth}{\scriptsize\raggedright
$^{\dagger}$ 2.0 Preview checkpoints.\quad
$^{\ddagger}$ Qwen3-VL backbones.
}
\endgroup
\end{table}

\begin{table}[H]
\renewcommand{\scriptsize}{\fontsize{7}{8}\selectfont}
\renewcommand{\footnotesize}{\fontsize{9}{11.5}\selectfont}
\setlength{\belowcaptionskip}{4pt}
\centering
\caption{\textbf{FPL Results on Cross-Embodiment Real-World Trajectories.}}
\label{tab:real_emb_fpl_part1}
\label{tab:real_emb_fpl}
\begingroup
\footnotesize
\rmfamily
\setlength{\tabcolsep}{0.9pt}
\renewcommand{\arraystretch}{1.0}
\def\taskmodeband#1#2{%
  \specialrule{\lightrulewidth}{\aboverulesep}{2pt}
  \multicolumn{15}{@{}c@{}}{%
    \makebox[\textwidth][c]{%
      \rlap{\smash{\color{#1}%
        \rule[-\dp\csname @arstrutbox\endcsname]{\textwidth}%
        {\dimexpr\ht\csname @arstrutbox\endcsname
        +\dp\csname @arstrutbox\endcsname\relax}}}%
      \makebox[\textwidth][c]{\textit{#2}}}}\\%
  \specialrule{\lightrulewidth}{2pt}{\belowrulesep}
}

\renewcommand{\scriptsize}{\fontsize{6}{7}\selectfont}
\par\vspace{1pt}
\parbox{\textwidth}{\scriptsize\raggedright \textbf{T1}: Arrange family tea set.\enspace \textbf{T2}: Arrange flowers and reed diffuser.\enspace \textbf{T3}: Arrange seasoning jars.\enspace \textbf{T4}: Fill and place pen holder.\enspace \textbf{T5}: Load plastic plates into rack.\enspace \textbf{T6}: Organize books and pen.\enspace \textbf{T7}: Organize cosmetics storage box.\enspace \textbf{T8}: Organize desktop workstation.\enspace \textbf{T9}: Pack items for shipping.\enspace \textbf{T10}: Place pot on induction cooker.\enspace \textbf{T11}: Prepare breakfast.\enspace \textbf{T12}: Set the table.\enspace \textbf{T13}: Sort Blocks in Order.\enspace \textbf{T14}: Transfer Blocks Using Plate.}
\par\vspace{1pt}
\parbox{\textwidth}{\scriptsize\raggedright
$^{\dagger}$ 2.0 Preview checkpoints.\quad
$^{\ddagger}$ Qwen3-VL backbones.
}
\endgroup
\end{table}

\begin{table}[H]
\renewcommand{\scriptsize}{\fontsize{7}{8}\selectfont}
\renewcommand{\footnotesize}{\fontsize{9}{11.5}\selectfont}
\setlength{\belowcaptionskip}{4pt}
\centering
\caption{\textbf{VS Results on Cross-Embodiment Real-World Trajectories.}}
\label{tab:real_emb_vs_part1}
\label{tab:real_emb_vs}
\begingroup
\footnotesize
\rmfamily
\setlength{\tabcolsep}{0.9pt}
\renewcommand{\arraystretch}{1.0}
\def\taskmodeband#1#2{%
  \specialrule{\lightrulewidth}{\aboverulesep}{2pt}
  \multicolumn{15}{@{}c@{}}{%
    \makebox[\textwidth][c]{%
      \rlap{\smash{\color{#1}%
        \rule[-\dp\csname @arstrutbox\endcsname]{\textwidth}%
        {\dimexpr\ht\csname @arstrutbox\endcsname
        +\dp\csname @arstrutbox\endcsname\relax}}}%
      \makebox[\textwidth][c]{\textit{#2}}}}\\%
  \specialrule{\lightrulewidth}{2pt}{\belowrulesep}
}

\renewcommand{\scriptsize}{\fontsize{6}{7}\selectfont}
\par\vspace{1pt}
\parbox{\textwidth}{\scriptsize\raggedright \textbf{T1}: Arrange family tea set.\enspace \textbf{T2}: Arrange flowers and reed diffuser.\enspace \textbf{T3}: Arrange seasoning jars.\enspace \textbf{T4}: Fill and place pen holder.\enspace \textbf{T5}: Load plastic plates into rack.\enspace \textbf{T6}: Organize books and pen.\enspace \textbf{T7}: Organize cosmetics storage box.\enspace \textbf{T8}: Organize desktop workstation.\enspace \textbf{T9}: Pack items for shipping.\enspace \textbf{T10}: Place pot on induction cooker.\enspace \textbf{T11}: Prepare breakfast.\enspace \textbf{T12}: Set the table.\enspace \textbf{T13}: Sort Blocks in Order.\enspace \textbf{T14}: Transfer Blocks Using Plate.}
\par\vspace{1pt}
\parbox{\textwidth}{\scriptsize\raggedright
$^{\dagger}$ 2.0 Preview checkpoints.\quad
$^{\ddagger}$ Qwen3-VL backbones.
}
\endgroup
\end{table}

\FloatBarrier
\Needspace{29\baselineskip}
\subsubsection{Cross-Environment Generalization}\label{app:realworld_environment_results}

We retain AgiBot Genie02 and vary tablecloth appearance, lighting, and tabletop clutter. The results measure robustness to these scene changes while the robot embodiment and task requirements remain fixed.


\begin{table}[H]
\renewcommand{\scriptsize}{\fontsize{7}{8}\selectfont}
\renewcommand{\footnotesize}{\fontsize{9}{9.5}\selectfont}
\setlength{\belowcaptionskip}{4pt}
\centering
\caption{\textbf{SA Results on Cross-Environment Real-World Trajectories.}}
\label{tab:real_env_sa_part1}
\label{tab:real_env_sa}
\begingroup
\footnotesize
\rmfamily
\setlength{\tabcolsep}{0.9pt}
\renewcommand{\arraystretch}{1.0}
\def\taskmodeband#1#2{%
  \specialrule{\lightrulewidth}{\aboverulesep}{2pt}
  \multicolumn{15}{@{}c@{}}{%
    \makebox[\textwidth][c]{%
      \rlap{\smash{\color{#1}%
        \rule[-\dp\csname @arstrutbox\endcsname]{\textwidth}%
        {\dimexpr\ht\csname @arstrutbox\endcsname
        +\dp\csname @arstrutbox\endcsname\relax}}}%
      \makebox[\textwidth][c]{\textit{#2}}}}\\%
  \specialrule{\lightrulewidth}{2pt}{\belowrulesep}
}

\renewcommand{\scriptsize}{\fontsize{6}{7}\selectfont}
\par\vspace{1pt}
\parbox{\textwidth}{\scriptsize\raggedright \textbf{T1}: Arrange family tea set.\enspace \textbf{T2}: Arrange flowers and reed diffuser.\enspace \textbf{T3}: Arrange seasoning jars.\enspace \textbf{T4}: Fill and place pen holder.\enspace \textbf{T5}: Load plastic plates into rack.\enspace \textbf{T6}: Organize books and pen.\enspace \textbf{T7}: Organize cosmetics storage box.\enspace \textbf{T8}: Organize desktop workstation.\enspace \textbf{T9}: Pack items for shipping.\enspace \textbf{T10}: Place pot on induction cooker.\enspace \textbf{T11}: Prepare breakfast.\enspace \textbf{T12}: Set the table.\enspace \textbf{T13}: Sort Blocks in Order.\enspace \textbf{T14}: Transfer Blocks Using Plate.}
\par\vspace{1pt}
\parbox{\textwidth}{\scriptsize\raggedright
$^{\dagger}$ 2.0 Preview checkpoints.\quad
$^{\ddagger}$ Qwen3-VL backbones.
}
\endgroup
\end{table}

\begin{table}[H]
\renewcommand{\scriptsize}{\fontsize{7}{8}\selectfont}
\renewcommand{\footnotesize}{\fontsize{9}{9.5}\selectfont}
\setlength{\belowcaptionskip}{4pt}
\centering
\caption{\textbf{TGA-CT Results on Cross-Environment Real-World Trajectories.}}
\label{tab:real_env_tga_ct_part1}
\label{tab:real_env_tga_ct}
\label{tab:real_env_tga_easy_part1}
\label{tab:real_env_tga_easy}
\begingroup
\footnotesize
\rmfamily
\setlength{\tabcolsep}{0.9pt}
\renewcommand{\arraystretch}{1.0}
\def\taskmodeband#1#2{%
  \specialrule{\lightrulewidth}{\aboverulesep}{2pt}
  \multicolumn{15}{@{}c@{}}{%
    \makebox[\textwidth][c]{%
      \rlap{\smash{\color{#1}%
        \rule[-\dp\csname @arstrutbox\endcsname]{\textwidth}%
        {\dimexpr\ht\csname @arstrutbox\endcsname
        +\dp\csname @arstrutbox\endcsname\relax}}}%
      \makebox[\textwidth][c]{\textit{#2}}}}\\%
  \specialrule{\lightrulewidth}{2pt}{\belowrulesep}
}

\renewcommand{\scriptsize}{\fontsize{6}{7}\selectfont}
\par\vspace{1pt}
\parbox{\textwidth}{\scriptsize\raggedright \textbf{T1}: Arrange family tea set.\enspace \textbf{T2}: Arrange flowers and reed diffuser.\enspace \textbf{T3}: Arrange seasoning jars.\enspace \textbf{T4}: Fill and place pen holder.\enspace \textbf{T5}: Load plastic plates into rack.\enspace \textbf{T6}: Organize books and pen.\enspace \textbf{T7}: Organize cosmetics storage box.\enspace \textbf{T8}: Organize desktop workstation.\enspace \textbf{T9}: Pack items for shipping.\enspace \textbf{T10}: Place pot on induction cooker.\enspace \textbf{T11}: Prepare breakfast.\enspace \textbf{T12}: Set the table.\enspace \textbf{T13}: Sort Blocks in Order.\enspace \textbf{T14}: Transfer Blocks Using Plate.}
\par\vspace{1pt}
\parbox{\textwidth}{\scriptsize\raggedright
$^{\dagger}$ 2.0 Preview checkpoints.\quad
$^{\ddagger}$ Qwen3-VL backbones.
}
\endgroup
\end{table}

\begin{table}[H]
\renewcommand{\scriptsize}{\fontsize{7}{8}\selectfont}
\renewcommand{\footnotesize}{\fontsize{9}{9}\selectfont}
\setlength{\belowcaptionskip}{4pt}
\centering
\caption{\textbf{TGA-CF Results on Cross-Environment Real-World Trajectories.}}
\label{tab:real_env_tga_cf_part1}
\label{tab:real_env_tga_cf}
\label{tab:real_env_tga_hard_part1}
\label{tab:real_env_tga_hard}
\begingroup
\footnotesize
\rmfamily
\setlength{\tabcolsep}{0.9pt}
\renewcommand{\arraystretch}{1.0}
\def\taskmodeband#1#2{%
  \specialrule{\lightrulewidth}{\aboverulesep}{2pt}
  \multicolumn{15}{@{}c@{}}{%
    \makebox[\textwidth][c]{%
      \rlap{\smash{\color{#1}%
        \rule[-\dp\csname @arstrutbox\endcsname]{\textwidth}%
        {\dimexpr\ht\csname @arstrutbox\endcsname
        +\dp\csname @arstrutbox\endcsname\relax}}}%
      \makebox[\textwidth][c]{\textit{#2}}}}\\%
  \specialrule{\lightrulewidth}{2pt}{\belowrulesep}
}

\renewcommand{\scriptsize}{\fontsize{6}{7}\selectfont}
\par\vspace{1pt}
\parbox{\textwidth}{\scriptsize\raggedright \textbf{T1}: Arrange family tea set.\enspace \textbf{T2}: Arrange flowers and reed diffuser.\enspace \textbf{T3}: Arrange seasoning jars.\enspace \textbf{T4}: Fill and place pen holder.\enspace \textbf{T5}: Load plastic plates into rack.\enspace \textbf{T6}: Organize books and pen.\enspace \textbf{T7}: Organize cosmetics storage box.\enspace \textbf{T8}: Organize desktop workstation.\enspace \textbf{T9}: Pack items for shipping.\enspace \textbf{T10}: Place pot on induction cooker.\enspace \textbf{T11}: Prepare breakfast.\enspace \textbf{T12}: Set the table.\enspace \textbf{T13}: Sort Blocks in Order.\enspace \textbf{T14}: Transfer Blocks Using Plate.}
\par\vspace{1pt}
\parbox{\textwidth}{\scriptsize\raggedright
$^{\dagger}$ 2.0 Preview checkpoints.\quad
$^{\ddagger}$ Qwen3-VL backbones.
}
\endgroup
\end{table}

\begin{table}[H]
\renewcommand{\scriptsize}{\fontsize{7}{8}\selectfont}
\renewcommand{\footnotesize}{\fontsize{9}{9.5}\selectfont}
\setlength{\belowcaptionskip}{4pt}
\centering
\caption{\textbf{SIA Results on Cross-Environment Real-World Trajectories.}}
\label{tab:real_env_sia_part1}
\label{tab:real_env_sia}
\begingroup
\footnotesize
\rmfamily
\setlength{\tabcolsep}{0.9pt}
\renewcommand{\arraystretch}{1.0}
\def\taskmodeband#1#2{%
  \specialrule{\lightrulewidth}{\aboverulesep}{2pt}
  \multicolumn{15}{@{}c@{}}{%
    \makebox[\textwidth][c]{%
      \rlap{\smash{\color{#1}%
        \rule[-\dp\csname @arstrutbox\endcsname]{\textwidth}%
        {\dimexpr\ht\csname @arstrutbox\endcsname
        +\dp\csname @arstrutbox\endcsname\relax}}}%
      \makebox[\textwidth][c]{\textit{#2}}}}\\%
  \specialrule{\lightrulewidth}{2pt}{\belowrulesep}
}

\renewcommand{\scriptsize}{\fontsize{6}{7}\selectfont}
\par\vspace{1pt}
\parbox{\textwidth}{\scriptsize\raggedright \textbf{T1}: Arrange family tea set.\enspace \textbf{T2}: Arrange flowers and reed diffuser.\enspace \textbf{T3}: Arrange seasoning jars.\enspace \textbf{T4}: Fill and place pen holder.\enspace \textbf{T5}: Load plastic plates into rack.\enspace \textbf{T6}: Organize books and pen.\enspace \textbf{T7}: Organize cosmetics storage box.\enspace \textbf{T8}: Organize desktop workstation.\enspace \textbf{T9}: Pack items for shipping.\enspace \textbf{T10}: Place pot on induction cooker.\enspace \textbf{T11}: Prepare breakfast.\enspace \textbf{T12}: Set the table.\enspace \textbf{T13}: Sort Blocks in Order.\enspace \textbf{T14}: Transfer Blocks Using Plate.}
\par\vspace{1pt}
\parbox{\textwidth}{\scriptsize\raggedright
$^{\dagger}$ 2.0 Preview checkpoints.\quad
$^{\ddagger}$ Qwen3-VL backbones.
}
\endgroup
\end{table}

\begin{table}[H]
\renewcommand{\scriptsize}{\fontsize{7}{8}\selectfont}
\renewcommand{\footnotesize}{\fontsize{9}{9}\selectfont}
\setlength{\belowcaptionskip}{4pt}
\centering
\caption{\textbf{VOC Results on Cross-Environment Real-World Trajectories.}}
\label{tab:real_env_voc_part1}
\label{tab:real_env_voc}
\begingroup
\footnotesize
\rmfamily
\setlength{\tabcolsep}{0.9pt}
\renewcommand{\arraystretch}{1.0}
\def\taskmodeband#1#2{%
  \specialrule{\lightrulewidth}{\aboverulesep}{2pt}
  \multicolumn{15}{@{}c@{}}{%
    \makebox[\textwidth][c]{%
      \rlap{\smash{\color{#1}%
        \rule[-\dp\csname @arstrutbox\endcsname]{\textwidth}%
        {\dimexpr\ht\csname @arstrutbox\endcsname
        +\dp\csname @arstrutbox\endcsname\relax}}}%
      \makebox[\textwidth][c]{\textit{#2}}}}\\%
  \specialrule{\lightrulewidth}{2pt}{\belowrulesep}
}

\renewcommand{\scriptsize}{\fontsize{6}{7}\selectfont}
\par\vspace{1pt}
\parbox{\textwidth}{\scriptsize\raggedright \textbf{T1}: Arrange family tea set.\enspace \textbf{T2}: Arrange flowers and reed diffuser.\enspace \textbf{T3}: Arrange seasoning jars.\enspace \textbf{T4}: Fill and place pen holder.\enspace \textbf{T5}: Load plastic plates into rack.\enspace \textbf{T6}: Organize books and pen.\enspace \textbf{T7}: Organize cosmetics storage box.\enspace \textbf{T8}: Organize desktop workstation.\enspace \textbf{T9}: Pack items for shipping.\enspace \textbf{T10}: Place pot on induction cooker.\enspace \textbf{T11}: Prepare breakfast.\enspace \textbf{T12}: Set the table.\enspace \textbf{T13}: Sort Blocks in Order.\enspace \textbf{T14}: Transfer Blocks Using Plate.}
\par\vspace{1pt}
\parbox{\textwidth}{\scriptsize\raggedright
$^{\dagger}$ 2.0 Preview checkpoints.\quad
$^{\ddagger}$ Qwen3-VL backbones.
}
\endgroup
\end{table}

\begin{table}[H]
\renewcommand{\scriptsize}{\fontsize{7}{8}\selectfont}
\renewcommand{\footnotesize}{\fontsize{9}{11.5}\selectfont}
\setlength{\belowcaptionskip}{4pt}
\centering
\caption{\textbf{Cycle-VOC Results on Cross-Environment Real-World Trajectories.}}
\label{tab:real_env_cycle_voc_part1}
\label{tab:real_env_cycle_voc}
\begingroup
\footnotesize
\rmfamily
\setlength{\tabcolsep}{0.9pt}
\renewcommand{\arraystretch}{1.0}
\def\taskmodeband#1#2{%
  \specialrule{\lightrulewidth}{\aboverulesep}{2pt}
  \multicolumn{15}{@{}c@{}}{%
    \makebox[\textwidth][c]{%
      \rlap{\smash{\color{#1}%
        \rule[-\dp\csname @arstrutbox\endcsname]{\textwidth}%
        {\dimexpr\ht\csname @arstrutbox\endcsname
        +\dp\csname @arstrutbox\endcsname\relax}}}%
      \makebox[\textwidth][c]{\textit{#2}}}}\\%
  \specialrule{\lightrulewidth}{2pt}{\belowrulesep}
}

\renewcommand{\scriptsize}{\fontsize{6}{7}\selectfont}
\par\vspace{1pt}
\parbox{\textwidth}{\scriptsize\raggedright \textbf{T1}: Arrange family tea set.\enspace \textbf{T2}: Arrange flowers and reed diffuser.\enspace \textbf{T3}: Arrange seasoning jars.\enspace \textbf{T4}: Fill and place pen holder.\enspace \textbf{T5}: Load plastic plates into rack.\enspace \textbf{T6}: Organize books and pen.\enspace \textbf{T7}: Organize cosmetics storage box.\enspace \textbf{T8}: Organize desktop workstation.\enspace \textbf{T9}: Pack items for shipping.\enspace \textbf{T10}: Place pot on induction cooker.\enspace \textbf{T11}: Prepare breakfast.\enspace \textbf{T12}: Set the table.\enspace \textbf{T13}: Sort Blocks in Order.\enspace \textbf{T14}: Transfer Blocks Using Plate.}
\par\vspace{1pt}
\parbox{\textwidth}{\scriptsize\raggedright
$^{\dagger}$ 2.0 Preview checkpoints.\quad
$^{\ddagger}$ Qwen3-VL backbones.
}
\endgroup
\end{table}

\begin{table}[H]
\renewcommand{\scriptsize}{\fontsize{7}{8}\selectfont}
\renewcommand{\footnotesize}{\fontsize{9}{11.5}\selectfont}
\setlength{\belowcaptionskip}{4pt}
\centering
\caption{\textbf{Memory-VOC Results on Cross-Environment Real-World Trajectories.}}
\label{tab:real_env_memory_voc_part1}
\label{tab:real_env_memory_voc}
\label{tab:real_env_voc_mem_part1}
\label{tab:real_env_voc_mem}
\begingroup
\footnotesize
\rmfamily
\setlength{\tabcolsep}{0.9pt}
\renewcommand{\arraystretch}{1.0}
\def\taskmodeband#1#2{%
  \specialrule{\lightrulewidth}{\aboverulesep}{2pt}
  \multicolumn{7}{@{}c@{}}{%
    \makebox[\textwidth][c]{%
      \rlap{\smash{\color{#1}%
        \rule[-\dp\csname @arstrutbox\endcsname]{\textwidth}%
        {\dimexpr\ht\csname @arstrutbox\endcsname
        +\dp\csname @arstrutbox\endcsname\relax}}}%
      \makebox[\textwidth][c]{\textit{#2}}}}\\%
  \specialrule{\lightrulewidth}{2pt}{\belowrulesep}
}

\renewcommand{\scriptsize}{\fontsize{6}{7}\selectfont}
\par\vspace{1pt}
\parbox{\textwidth}{\scriptsize\raggedright \textbf{T1}: Apply dish soap repeatedly.\enspace \textbf{T2}: Enter keypad number sequence.\enspace \textbf{T3}: Operate drawers in sequence.\enspace \textbf{T4}: Press button five times.\enspace \textbf{T5}: Season meat skewers repeatedly.\enspace \textbf{T6}: Strike drum five times.}
\par\vspace{1pt}
\parbox{\textwidth}{\scriptsize\raggedright
$^{\dagger}$ 2.0 Preview checkpoints.\quad
$^{\ddagger}$ Qwen3-VL backbones.
}
\endgroup
\end{table}

\begin{table}[H]
\renewcommand{\scriptsize}{\fontsize{7}{8}\selectfont}
\renewcommand{\footnotesize}{\fontsize{9}{9.5}\selectfont}
\setlength{\belowcaptionskip}{4pt}
\centering
\caption{\textbf{FPL Results on Cross-Environment Real-World Trajectories.}}
\label{tab:real_env_fpl_part1}
\label{tab:real_env_fpl}
\begingroup
\footnotesize
\rmfamily
\setlength{\tabcolsep}{0.9pt}
\renewcommand{\arraystretch}{1.0}
\def\taskmodeband#1#2{%
  \specialrule{\lightrulewidth}{\aboverulesep}{2pt}
  \multicolumn{15}{@{}c@{}}{%
    \makebox[\textwidth][c]{%
      \rlap{\smash{\color{#1}%
        \rule[-\dp\csname @arstrutbox\endcsname]{\textwidth}%
        {\dimexpr\ht\csname @arstrutbox\endcsname
        +\dp\csname @arstrutbox\endcsname\relax}}}%
      \makebox[\textwidth][c]{\textit{#2}}}}\\%
  \specialrule{\lightrulewidth}{2pt}{\belowrulesep}
}

\renewcommand{\scriptsize}{\fontsize{6}{7}\selectfont}
\par\vspace{1pt}
\parbox{\textwidth}{\scriptsize\raggedright \textbf{T1}: Arrange family tea set.\enspace \textbf{T2}: Arrange flowers and reed diffuser.\enspace \textbf{T3}: Arrange seasoning jars.\enspace \textbf{T4}: Fill and place pen holder.\enspace \textbf{T5}: Load plastic plates into rack.\enspace \textbf{T6}: Organize books and pen.\enspace \textbf{T7}: Organize cosmetics storage box.\enspace \textbf{T8}: Organize desktop workstation.\enspace \textbf{T9}: Pack items for shipping.\enspace \textbf{T10}: Place pot on induction cooker.\enspace \textbf{T11}: Prepare breakfast.\enspace \textbf{T12}: Set the table.\enspace \textbf{T13}: Sort Blocks in Order.\enspace \textbf{T14}: Transfer Blocks Using Plate.}
\par\vspace{1pt}
\parbox{\textwidth}{\scriptsize\raggedright
$^{\dagger}$ 2.0 Preview checkpoints.\quad
$^{\ddagger}$ Qwen3-VL backbones.
}
\endgroup
\end{table}

\begin{table}[H]
\renewcommand{\scriptsize}{\fontsize{7}{8}\selectfont}
\renewcommand{\footnotesize}{\fontsize{9}{9.5}\selectfont}
\setlength{\belowcaptionskip}{4pt}
\centering
\caption{\textbf{VS Results on Cross-Environment Real-World Trajectories.}}
\label{tab:real_env_vs_part1}
\label{tab:real_env_vs}
\begingroup
\footnotesize
\rmfamily
\setlength{\tabcolsep}{0.9pt}
\renewcommand{\arraystretch}{1.0}
\def\taskmodeband#1#2{%
  \specialrule{\lightrulewidth}{\aboverulesep}{2pt}
  \multicolumn{15}{@{}c@{}}{%
    \makebox[\textwidth][c]{%
      \rlap{\smash{\color{#1}%
        \rule[-\dp\csname @arstrutbox\endcsname]{\textwidth}%
        {\dimexpr\ht\csname @arstrutbox\endcsname
        +\dp\csname @arstrutbox\endcsname\relax}}}%
      \makebox[\textwidth][c]{\textit{#2}}}}\\%
  \specialrule{\lightrulewidth}{2pt}{\belowrulesep}
}

\renewcommand{\scriptsize}{\fontsize{6}{7}\selectfont}
\par\vspace{1pt}
\parbox{\textwidth}{\scriptsize\raggedright \textbf{T1}: Arrange family tea set.\enspace \textbf{T2}: Arrange flowers and reed diffuser.\enspace \textbf{T3}: Arrange seasoning jars.\enspace \textbf{T4}: Fill and place pen holder.\enspace \textbf{T5}: Load plastic plates into rack.\enspace \textbf{T6}: Organize books and pen.\enspace \textbf{T7}: Organize cosmetics storage box.\enspace \textbf{T8}: Organize desktop workstation.\enspace \textbf{T9}: Pack items for shipping.\enspace \textbf{T10}: Place pot on induction cooker.\enspace \textbf{T11}: Prepare breakfast.\enspace \textbf{T12}: Set the table.\enspace \textbf{T13}: Sort Blocks in Order.\enspace \textbf{T14}: Transfer Blocks Using Plate.}
\par\vspace{1pt}
\parbox{\textwidth}{\scriptsize\raggedright
$^{\dagger}$ 2.0 Preview checkpoints.\quad
$^{\ddagger}$ Qwen3-VL backbones.
}
\endgroup
\end{table}

\FloatBarrier
\subsection{Cross-Task Score Matrix Visualizations}
\label{app:realworld_tga_ct}

\begin{figure}[p]
  \centering
  \includegraphics[width=\linewidth,height=0.84\textheight,keepaspectratio]{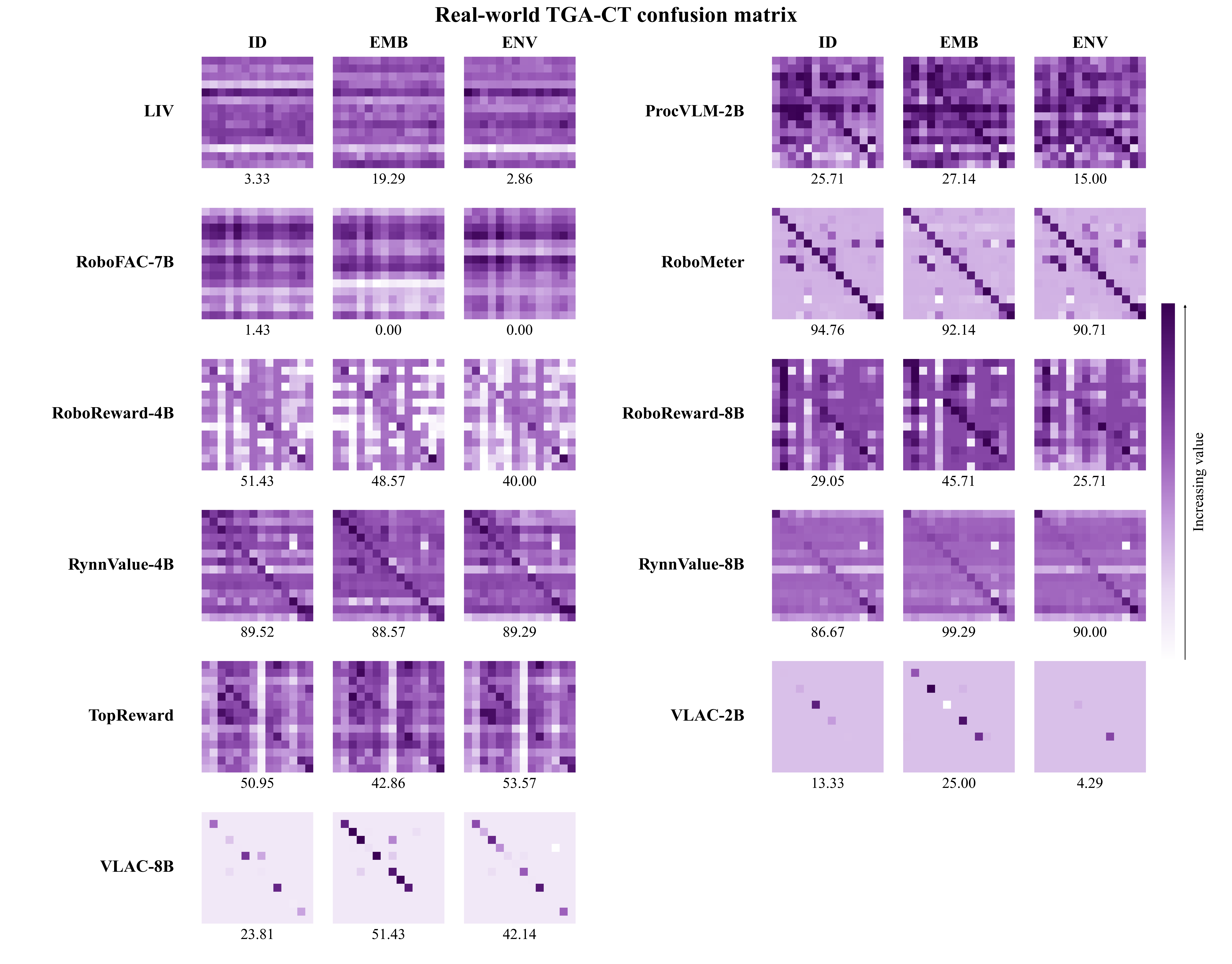}
  \caption{\textbf{Zero-Shot TGA-CT Score Matrices in the Real World.} Rows denote executed tasks and columns queried instructions; diagonal cells match the two. Darker cells indicate larger predicted gains within each matrix. Numbers below each matrix give TGA-CT accuracy (\%). TOPReward uses the 8B backbone.}
  \label{fig:tga_ct_real_zero_shot}
  \par\vspace{10pt}
  \includegraphics[width=\linewidth,height=0.70\textheight,keepaspectratio]{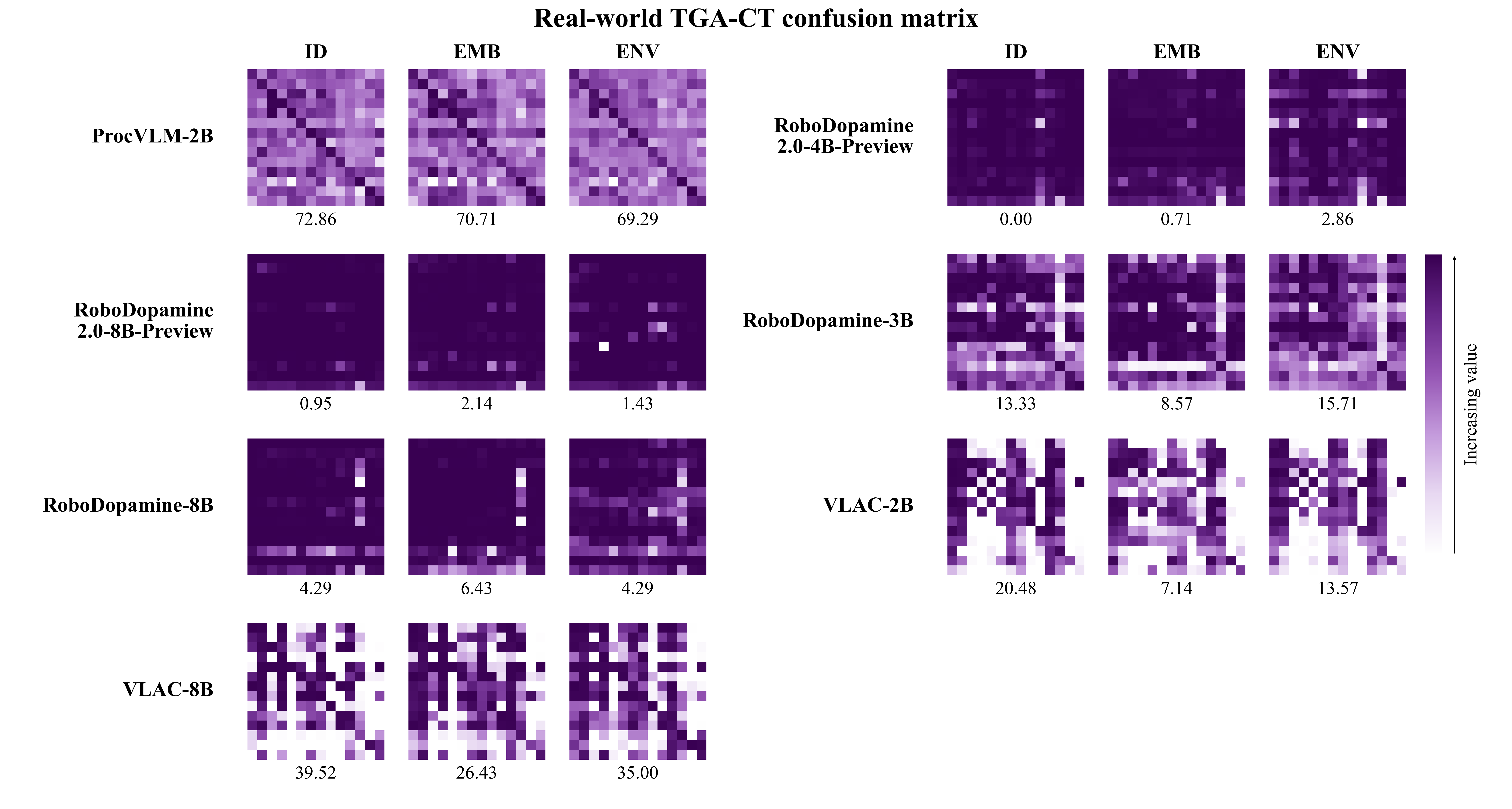}
  \caption{\textbf{One-Shot TGA-CT Score Matrices in the Real World.} Rows denote executed tasks and columns denote queried instructions. Darker cells indicate larger gains within each matrix; diagonal cells pair executions with matching instructions. Numbers below each matrix report TGA-CT accuracy (\%).}
  \label{fig:tga_ct_real_one_shot}
\end{figure}

Figures~\ref{fig:tga_ct_real_zero_shot} and~\ref{fig:tga_ct_real_one_shot} show TGA-CT score matrices for 14 real-world tasks across the standard, cross-embodiment, and cross-environment settings. Each row compares the gain assigned to the matching instruction with gains assigned to alternative task instructions.

RoboMeter and the RynnValue variants retain strong zero-shot diagonal selectivity across settings. In one-shot evaluation, ProcVLM is more selective than the other demonstration-based configurations, while Robo-Dopamine often assigns large gains to both matching and mismatched instructions.

\clearpage

\end{document}